\documentclass{article}

\usepackage[utf8]{inputenc}
\usepackage[T1]{fontenc}
\usepackage{booktabs} 
\usepackage{tabularx} 
\usepackage{graphicx} 
\usepackage{caption}  
\usepackage{pgfplots} 
\pgfplotsset{compat=1.18} 
\usepackage{tikz}         
\usetikzlibrary{shadows}
\usetikzlibrary{shapes, arrows, positioning, fit, backgrounds} 
\usetikzlibrary{shapes.geometric, arrows.meta, positioning, fit, backgrounds}
\usepackage[margin=2.5cm, right=3cm, verbose=true, letterpaper]{geometry} 
\usepackage{amsmath}  
\usepackage{hyperref} 

\tikzstyle{block} = [rectangle, draw, fill=blue!20, 
    text width=8em, text centered, rounded corners, minimum height=4em]
\tikzstyle{line} = [draw, -latex'] 

\usepackage{arxiv}

\usepackage[utf8]{inputenc} 
\usepackage[T1]{fontenc}    
\usepackage{hyperref}       
\usepackage{url}            
\usepackage{booktabs}       
\usepackage{array}          
\usepackage{caption}        
\usepackage{siunitx}        
\usepackage{amsfonts}       
\usepackage{pgfplots}
\usepackage{nicefrac}       
\usepackage{microtype}      
\usepackage{lipsum}		    
\usepackage{graphicx}
\usepackage{natbib}
\usepackage{doi}
\usepackage[utf8]{inputenc}
\usepackage{booktabs}
\usepackage{array}
\usepackage{caption}
\usepackage{siunitx}
\usepackage{makecell}
\usepackage{multirow}

\title{Inteligencia Artificial jurídica y el desafío de la veracidad: análisis de alucinaciones, optimización de RAG y principios para una integración responsable}
    
\author{
    \large 
    {\Large Alex Dantart} \\ 
    CIO LittleJohn\\
	Paseo de la Castellana 194\\
	28046, Madrid, España \\
	\texttt{arxiv@littlejohn.ai} \\
}

\date{}

\renewcommand{\headeright}{Informe técnico}
\renewcommand{\undertitle}{Informe técnico}
\renewcommand{\shorttitle}{\textit Inteligencia Artificial jurídica y el desafío de la veracidad}

\hypersetup{
pdftitle={Inteligencia Artificial jurídica y el desafío de la veracidad: análisis de alucinaciones, optimización de RAG y principios para una integración responsable},
pdfsubject={q-cs.AI},
pdfauthor={Alex Dantart},
pdfkeywords={Alucinaciones IA, Large Language Models (LLM), Retrieval-Augmented Generation (RAG), Derecho, ética legal, evaluación IA, mitigación de alucinaciones, Inteligencia Artificial jurídica},
}

\begin{document}
\maketitle

\begin{abstract}
Los grandes modelos de lenguaje (LLMs) están redefiniendo aceleradamente la práctica, la educación y la investigación jurídicas. Sin embargo, su vasto potencial se ve significativamente amenazado por la generación endémica de "alucinaciones" – resultados textuales que, aunque a menudo plausibles, son fácticamente incorrectos, engañosos o inconsistentes con las fuentes legales autorizadas. Este ensayo presenta una revisión exhaustiva y un análisis crítico multidimensional del fenómeno de las alucinaciones en LLMs aplicados al derecho. Se documentan las tendencias y manifestaciones de las alucinaciones a través de jurisdicciones, tipos de tribunales y clases de tareas legales, fundamentándonos en la creciente evidencia empírica de estudios recientes que evalúan tanto LLMs públicos como herramientas comerciales especializadas de Inteligencia Artificial (IA) legal. Se analizan en profundidad las causas subyacentes de estas alucinaciones, desde las deficiencias en los datos de entrenamiento y las limitaciones inherentes a la arquitectura probabilística de los modelos, hasta las complejidades del lenguaje jurídico y la tensión fundamental entre fluidez generativa y factualidad estricta.

Se examina con detalle la Generación Aumentada por Recuperación (RAG) como la principal estrategia de mitigación propuesta, evaluando críticamente su efectividad teórica, sus implementaciones prácticas y sus limitaciones persistentes en el singular contexto legal, incluyendo los puntos de fallo en sus fases de recuperación y generación. Más allá del RAG canónico, se discuten y proponen estrategias holísticas y avanzadas para la optimización y mitigación, abarcando desde la curación estratégica de datos y la ingeniería de prompts sofisticada, hasta la consideración de agentes de IA conscientes de la jerarquía normativa (como la pirámide de Kelsen), el fine-tuning enfocado en la fidelidad, y la implementación de robustos mecanismos de verificación post-hoc y calibración de confianza. Se ilustra la gravedad de estos fenómenos mediante el análisis de estudios de caso detallados de incidentes judiciales reales, extrayendo lecciones tangibles sobre las consecuencias de la confianza acrítica en la IA.

Con una mirada prospectiva, se explora el camino hacia una IA legal más fiable, delineando los desarrollos necesarios en modelos inherentemente más explicables (XAI, del inglés \textit{Explainable Artificial Intelligence}), sistemas técnicamente auditables y la adopción de un paradigma de IA responsable por diseño. Finalmente, se exploran las profundas implicaciones éticas y regulatorias, con especial atención al marco normativo europeo y español, enfatizando el rol irreductible e insustituible de la supervisión humana y el juicio profesional del abogado en la era de la inteligencia artificial. Se concluye subrayando la imperiosa necesidad de una integración cautelosa, crítica y supervisada de los LLMs en la práctica legal, se proponen una tipología refinada de alucinaciones legales con el fin de guiar y estructurar la investigación futura en este campo crucial, y también se propone un nuevo marco de trabajo que distingue entre la IA Generativa de propósito general y la IA Consultiva especializada, ofreciendo una tipología refinada de alucinaciones legales que guiará la investigación futura hacia una integración verdaderamente responsable.

Sin embargo, es imperativo realizar también una distinción fundamental que a menudo se pasa por alto en el debate actual: la diferencia entre la Inteligencia Artificial de propósito general (como los LLMs públicos) y la Inteligencia Artificial especializada y consultiva diseñada específicamente para el dominio legal. Mientras que la primera, por su naturaleza generativa, es inherentemente propensa a las alucinaciones al "inventar" respuestas para mantener la fluidez conversacional, la segunda opera bajo un principio radicalmente distinto. Una IA consultiva no crea conocimiento, sino que lo recupera, estructura y presenta de forma fundamentada, actuando como un asistente experto que cita sus fuentes en lugar de un oráculo creativo. Este informe argumentará que la mitigación efectiva de las alucinaciones en el sector legal no reside en mejorar incrementalmente los modelos generativos, sino en adoptar un paradigma consultivo donde la veracidad y la trazabilidad son el núcleo del diseño, no una característica añadida. La tecnología, en este contexto, no es un sustituto del juicio humano, sino una herramienta para amplificarlo, cumpliendo la máxima de humanizar la tecnología en lugar de simplemente automatizar procesos.
\end{abstract}


\section{Introduction}
La inteligencia artificial (IA), y en particular los grandes modelos de lenguaje (LLMs), se encuentran en la cúspide de una transformación significativa en múltiples sectores, siendo el dominio legal uno de los más impactados y debatidos (Choi et al., 2022; Katz et al., 2023; Rodgers, Armour, and Sako 2023).  Herramientas como ChatGPT de OpenAI, Gemini de Google, DeepSeek, y Llama de Meta, junto con plataformas especializadas de IA legal, prometen revolucionar tareas fundamentales como la investigación jurídica, la redacción de documentos, el análisis de contratos y la asistencia en litigios (Guha et al. 2023; Livermore, Herron, and Rockmore 2024). El potencial para aumentar la eficiencia, reducir costos y \textbf{democratizar el acceso a la justicia es considerable} (Perlman 2023; Tan, Westermann, and Benyekhlef 2023).

Sin embargo, este potencial transformador se ve obstaculizado por un desafío inherente y crítico: el fenómeno de las "alucinaciones" (Ji, Lee, et al. 2023). Las alucinaciones en LLMs se refieren a la \textbf{generación de información que, aunque a menudo plausible y lingüísticamente coherente, es fácticamente incorrecta, engañosa, inconsistente} con las fuentes proporcionadas o completamente fabricada (Dahl et al., 2024; Magesh et al., 2024).

En el contexto legal, donde la precisión, la fidelidad a las fuentes autorizadas (precedentes, estatutos) y la argumentación basada en hechos son primordiales, las alucinaciones no son meras inexactitudes técnicas, sino que representan un \textbf{riesgo sustancial} que puede llevar a errores estratégicos, consejos legales perjudiciales, sanciones profesionales e incluso la \textbf{erosión de la confianza pública en el sistema legal} (Roberts 2023; Weiser 2023). El infame caso \textit{Mata v. Avianca, Inc.} (2023), donde abogados fueron sancionados por presentar un escrito judicial citando casos inexistentes generados por ChatGPT, sirve como un claro recordatorio de los peligros (Lantyer, 2024).

Es crucial matizar, sin embargo, que el desafío de la veracidad en el derecho trasciende la mera corrección factual. A diferencia de otros dominios, en el ámbito jurídico una afirmación no es solo "verdadera" o "falsa"; su validez a menudo reside en la solidez de su interpretación y argumentación, que es precisamente el terreno del juicio profesional experto. Por tanto, el peligro de la IA no es solo que genere falsedades verificables, sino también que construya argumentos legalmente inviables o interpretaciones superficiales que, sin el filtro crítico de un abogado, pueden conducir a estrategias erróneas. El análisis de la veracidad debe, por consiguiente, abarcar tanto la fidelidad a la fuente como la viabilidad interpretativa.

Este ensayo se embarca en una exploración exhaustiva de las alucinaciones de los Grandes Modelos de Lenguaje (LLMs) en el dominio legal, con el objetivo de ir más allá de los informes anecdóticos y proporcionar un análisis sistemático basado en la creciente evidencia empírica y la literatura académica. Para ello, primero definiremos y categorizaremos las alucinaciones específicas del contexto legal, explorando sus causas fundamentales y su impacto particular en la práctica jurídica (Sección \ref{sec:fenomeno_alucinaciones}). Seguidamente, examinaremos en detalle los métodos y desafíos inherentes a la evaluación de la prevalencia y naturaleza de estas alucinaciones, revisando críticamente los estudios recientes sobre LLMs generales y las herramientas comerciales de IA legal (Sección \ref{sec:evaluacion_alucinaciones}). A continuación, analizaremos en profundidad la Generación Aumentada por Recuperación (RAG) como la principal estrategia de mitigación propuesta, evaluando tanto sus promesas conceptuales como sus limitaciones inherentes y su efectividad empírica en el contexto legal (Sección \ref{sec:rag_mitigacion}). Posteriormente, discutiremos un abanico de estrategias complementarias y avanzadas para la optimización y mitigación de alucinaciones, abarcando desde la curación de datos y la ingeniería de prompts hasta la consideración de agentes de IA conscientes de la jerarquía normativa y los mecanismos de verificación post-hoc (Sección \ref{sec:estrategias_avanzadas}). Para ilustrar la gravedad y las consecuencias tangibles de estos fenómenos, presentaremos y analizaremos estudios de caso detallados de incidentes reales donde las alucinaciones de la IA han impactado procedimientos judiciales (Sección \ref{sec:casos_reales}). Con una mirada prospectiva, exploraremos el camino hacia una IA legal más fiable, discutiendo el desarrollo de modelos explicables, auditables y responsables por diseño (Sección \ref{sec:futuro_ia_fiable}). Finalmente, reflexionaremos sobre las cruciales consideraciones éticas y regulatorias que surgen, con especial atención al marco normativo europeo y español (Sección \ref{sec:etica_regulacion}), para concluir sintetizando los hallazgos y enfatizando el camino a seguir hacia una integración responsable y efectiva de la IA en la práctica legal (Sección \ref{sec:conclusion}).

\section{El fenómeno de las alucinaciones en LLMs legales: naturaleza, causas e impacto}
\label{sec:fenomeno_alucinaciones}

La integración de los Grandes Modelos de Lenguaje (LLMs) en el ecosistema legal representa una de las transformaciones tecnológicas más profundas y potencialmente disruptivas de la era moderna. Estas arquitecturas de inteligencia artificial (IA), capaces de procesar y generar lenguaje natural con una fluidez sin precedentes, prometen optimizar radicalmente tareas intensivas en conocimiento como la investigación jurídica, la redacción de contratos, el análisis de pruebas (discovery) y la generación de escritos procesales (Choi et al., 2022; Livermore, Herron, and Rockmore 2024). Sin embargo, esta promesa se ve ensombrecida por un desafío inherente y omnipresente: el fenómeno de las "alucinaciones" (Ji, Lee, et al. 2023; Marcus \& Davis, 2022). Lejos de ser una anomalía ocasional, las alucinaciones constituyen una característica intrínseca del funcionamiento actual de los LLMs, manifestándose como la generación de contenido que, aunque a menudo sintáctica y semánticamente plausible, carece de fundamento fáctico, es lógicamente inconsistente o contradice directamente las fuentes de autoridad establecidas. En el dominio legal, donde la precisión factual, la fidelidad a la autoridad (leyes, precedentes, doctrina...) y la integridad argumentativa son pilares fundamentales, la propensión de los LLMs a alucinar no es un mero inconveniente técnico, sino un riesgo sistémico con profundas implicaciones éticas, profesionales y sociales (Roberts 2023).
El fenómeno de las alucinaciones no es un mero inconveniente técnico, sino que ha sido identificado como uno de los desafíos críticos que definen la frontera actual de la investigación en IA legal. Revisiones exhaustivas del campo señalan que, a pesar de los avances transformadores de los LLMs, la "\textbf{alucinación en reclamaciones legales, manifestada como citaciones espurias o fabricaciones normativas}", junto con los déficits de explicabilidad y la adaptación jurisdiccional, constituyen las principales barreras para su adopción generalizada y fiable (Shao et al., 2025).

\subsection{Un paradigma fundamental: IA generativa vs. IA consultiva}

Antes de diseccionar el fenómeno de las alucinaciones, es imperativo establecer una distinción conceptual que el debate actual a menudo ignora, generando una peligrosa confusión: la diferencia fundamental entre la \textbf{Inteligencia Artificial generativa} y la \textbf{Inteligencia Artificial consultiva}. El término "IA Legal" se utiliza de forma monolítica, cuando en realidad describe dos arquitecturas con propósitos, mecanismos y perfiles de riesgo radicalmente distintos. Entender esta dicotomía no es un mero ejercicio académico; es la clave para una integración responsable y efectiva de la IA en la práctica jurídica.

\subsubsection{Inteligencia Artificial generativa: el oráculo creativo}

La IA Generativa, cuyo máximo exponente son los LLMs de propósito general como GPT, Gemini o Claude, opera como un "imitador avanzado" o un "sabelotodo creativo". Su objetivo principal no es la veracidad, sino la \textbf{fluidez conversacional y la coherencia probabilística}.

\begin{itemize}
    \item \textbf{Definición y mecanismo:} Estos modelos funcionan prediciendo la siguiente palabra más probable en una secuencia, basándose en los patrones estadísticos aprendidos de un vasto y heterogéneo corpus de datos de internet. No "comprenden" el contenido ni "razonan" a partir de principios lógicos, sino que ensamblan texto que \textit{suena} plausible. Su conocimiento es paramétrico y está "congelado" en el momento de su entrenamiento.

    Investigaciones fundamentales sobre las causas de las alucinaciones explican que este comportamiento no es un fallo a corregir, sino una consecuencia directa de su diseño. Los modelos son optimizados para ser buenos "examinandos": en un sistema donde no se premia la incertidumbre, la estrategia más efectiva para obtener una "buena nota" (una respuesta plausible) es siempre arriesgar una respuesta en lugar de admitir desconocimiento. Por tanto, su tendencia a "inventar" es el resultado esperado de su entrenamiento (Kalai et al., 2025).
    
    \item \textbf{Ventajas:} Su fortaleza reside en tareas creativas: redacción de borradores, lluvia de ideas, resumen de textos no críticos y la generación de contenido donde la originalidad es más importante que la precisión factual.
    \item \textbf{Desventajas y riesgos inherentes:} Para el sector legal, su diseño es una receta para el desastre.
    \item \textbf{Alucinaciones "de diseño":} La propensión a alucinar no es un fallo, es una característica intrínseca de su arquitectura. Para evitar silencios y mantener la coherencia, el modelo "rellenará los huecos" o "inventará" hechos, sentencias o estatutos.
    \item \textbf{Opacidad total ("Caja Negra"):} Es imposible trazar el origen de una afirmación específica. La respuesta es un producto final opaco, sin referencias verificables.
    \item \textbf{Riesgo de "incesto de IAs":} Al ser entrenadas con el internet público, corren el riesgo de retroalimentarse con contenido de baja calidad generado por otras IAs, degradando su fiabilidad en un ciclo vicioso.
\end{itemize}

\subsubsection{Inteligencia Artificial consultiva: el archivero experto}

La IA Consultiva representa un cambio de paradigma. Su objetivo no es crear, sino \textbf{recuperar, estructurar y presentar conocimiento verificado}. Su arquitectura fundamental se basa en la Generación Aumentada por Recuperación (RAG), operando como un "archivero experto" o un "detective" que investiga antes de hablar.

\begin{itemize}
    \item \textbf{Definición y mecanismo:} Este modelo no confía en su conocimiento paramétrico interno. Ante una consulta, su primer paso es buscar en un corpus de datos externo, curado y autorizado (ej. bases de datos de legislación, jurisprudencia, documentos internos de un despacho). Solo después de recuperar los fragmentos de información más relevantes, genera una respuesta que debe estar \textit{estrictamente fundamentada} en dichos fragmentos.
    \item \textbf{Ventajas:} Diseñada para la fiabilidad en dominios críticos.
    \item \textbf{Mitigación de alucinaciones:} Reduce drásticamente la fabricación de hechos, ya que las respuestas están ancladas a fuentes explícitas.
    \item \textbf{Transparencia y trazabilidad:} La respuesta no es una "caja negra". Un sistema consultivo bien diseñado debe citar sus fuentes, permitiendo al profesional legal verificar la información y asumir la responsabilidad final con conocimiento de causa. Es la materialización del principio de \textbf{"no sustituir, sino amplificar"} el juicio humano.
    \item \textbf{Conocimiento actualizado:} Su fiabilidad depende de la actualidad de su base de datos, que es mucho más fácil y barata de actualizar que reentrenar un LLM masivo.
    \item \textbf{Desventajas y limitaciones:} No es una panacea. Su efectividad depende críticamente de la calidad de su corpus documental y de la sofisticación de su módulo de recuperación. Aún puede producir alucinaciones sutiles, como el \textit{misgrounding} (tergiversar una fuente real), pero el riesgo de invención flagrante se minimiza.
 \end{itemize}
 
\subsubsection{Tabla comparativa de paradigmas}

\begin{table}[htbp]
\centering
\footnotesize 
\renewcommand{\arraystretch}{1.3} 
\setlength{\tabcolsep}{6pt} 
\begin{tabular}{|>{\raggedright\arraybackslash}p{0.25\textwidth}|>{\raggedright\arraybackslash}p{0.30\textwidth}|>{\raggedright\arraybackslash}p{0.30\textwidth}|}
\hline
\textbf{Característica} & \textbf{IA generativa (propósito general)} & \textbf{IA consultiva (especializada)} \\
\hline
\textbf{Objetivo principal} & Fluidez y coherencia conversacional. & Precisión, fiabilidad y fundamentación. \\
\hline
\textbf{Fuente de conocimiento} & Paramétrico, interno, estático ("libro cerrado"). & Externo, curado, dinámico ("libro abierto"). \\
\hline
\textbf{Riesgo de alucinación} & Alto, especialmente fabricación de hechos ("de diseño"). & Bajo en fabricación, riesgo de \textit{misgrounding}. \\
\hline
\textbf{Transparencia} & Baja ("caja negra"). & Alta (debe citar fuentes y razonamiento). \\
\hline
\textbf{Caso de uso ideal} & Brainstorming, borradores creativos, tareas no críticas. & Investigación jurídica, \textit{due diligence}, respuestas factuales. \\
\hline
\textbf{Analogía} & Un "sabelotodo" elocuente pero a veces poco fiable. & Un "archivero" meticuloso que siempre muestra sus fichas. \\
\hline
\end{tabular}
\end{table}

La adopción de este marco dual es esencial para navegar la complejidad de la IA Legal. Confundir ambos paradigmas lleva a expectativas irreales y a una aplicación irresponsable de la tecnología. Las secciones subsiguientes de este informe analizarán en profundidad los desafíos inherentes al modelo generativo y cómo las arquitecturas consultivas, principalmente a través de RAG, intentan construir un camino hacia una IA legal verdaderamente fiable.

\subsection{Definición y taxonomía de las alucinaciones legales}

Definir la "alucinación" en el contexto de la IA legal requiere ir más allá de la simple dicotomía correcto/incorrecto. Una alucinación legal se materializa cuando un LLM genera una afirmación, cita, argumento o conclusión que se desvía de la realidad jurídica verificable o de la información contextual proporcionada, a menudo presentándola con una confianza injustificada (Khmaïess Al Jannadi, 2023). Es crucial entender que, si bien el término 'alucinación' es comúnmente usado, su aplicación en el derecho presenta desafíos únicos. A diferencia de dominios con verdades fácticas singulares, en el ámbito legal la 'corrección' de una afirmación interpretativa o un argumento puede ser objeto de debate entre expertos. Por ello, más allá de la simple desviación factual, una alucinación legal también puede entenderse como la generación de una propuesta que, \textbf{aunque plausible, resulta legalmente inviable o indefendible bajo un escrutinio experto}, incluso si no contradice directamente una fuente explícita. La aparente coherencia de estas salidas puede enmascarar su falta de solidez jurídica, haciendo su detección particularmente compleja. Esta desviación puede adoptar múltiples formas, cada una con implicaciones distintas para la práctica legal. 

Para enriquecer esta taxonomía, proponemos una dimensión adicional de clasificación basada en el origen arquitectónico de la IA. Las alucinaciones manifestadas por una IA de propósito general (ej. ChatGPT) suelen ser más graves (como la invención completa de jurisprudencia), ya que su objetivo es la coherencia conversacional a toda costa. Por el contrario, los errores en una IA consultiva especializada (basada en RAG) tienden a ser más sutiles, como el misgrounding o errores de síntesis, derivados de fallos en la recuperación o interpretación de un corpus documental controlado.

Esta distinción es crucial, pues mientras el primer tipo de alucinación representa un fallo sistémico de diseño para el uso legal, el segundo es un problema de implementación que puede ser mitigado con técnicas de optimización, como se discutirá más adelante. Ignorar esta diferencia es como confundir la opinión de un aficionado elocuente con el análisis documentado de un archivero experto.

Para definir y clasificar las alucinaciones legales con rigor, es útil adoptar un marco analítico que distinga las dos dimensiones clave del error. El estudio seminal de Magesh et al. (2025) sobre herramientas de IA legal propone una distinción fundamental entre:
\begin{itemize}
    \item Corrección (Correctness): Si la afirmación es fácticamente verdadera en el mundo real.
    \item Fundamentación (Groundedness): Si la afirmación está correctamente respaldada por la fuente citada.
\end{itemize}

A partir de este marco, una "alucinación" se define como una respuesta que es incorrecta (contiene información falsa) o mal fundamentada (\textit{misgrounded}, es decir, cita una fuente que no respalda la afirmación). Esta desviación puede adoptar múltiples formas, cada una con implicaciones distintas. 

Las siguientes categorías detallan las manifestaciones específicas de estos fallos en la práctica jurídica:

\begin{itemize}
    \item \textbf{Alucinaciones factuales/extrínsecas} (inconsistencia con los hechos del mundo legal): este es quizás el tipo más peligroso en la investigación y el asesoramiento legal directo. Se refiere a la generación de contenido que contradice el cuerpo establecido y verificable del derecho y los hechos relacionados.
        \begin{itemize}
            \item \textit{Misstatement} (declaración errónea) de la Ley o Precedente: El LLM describe incorrectamente el contenido o el holding de una ley o decisión judicial existente. Esto puede ir desde sutiles tergiversaciones hasta contradicciones directas con la autoridad citada o conocida.
            \item Fabricación de autoridad: el modelo inventa por completo casos, estatutos, regulaciones o incluso jueces y académicos inexistentes. El caso \textit{Mata v. Avianca, Inc.} (2023) es el ejemplo paradigmático, donde ChatGPT generó múltiples citaciones judiciales ficticias que fueron incorporadas a un escrito judicial.
            \item Error de aplicación jurisdiccional o temporal: el LLM aplica incorrectamente principios legales de una jurisdicción a otra, o presenta como vigente una ley o precedente derogado u obsoleto, fallando en reconocer la dinámica temporal y espacial del derecho.
        \end{itemize}
\end{itemize}

\begin{itemize}
    \item \textbf{Alucinaciones basadas en fuentes} (errores de \textit{groundedness} en Sistemas RAG): particularmente relevantes para los sistemas de Retrieval-Augmented Generation (RAG), que se discuten en la Sección \ref{sec:rag_mitigacion}. Estas ocurren cuando la respuesta generada es inconsistente con los documentos específicos recuperados por el sistema para fundamentar dicha respuesta.
    \begin{itemize}
        \item \textit{Misgrounding} (fundamentación errónea): el LLM cita correctamente una fuente existente (recuperada por el sistema RAG), pero hace una afirmación sobre su contenido que la fuente no respalda o incluso contradice (Magesh et al., 2024). Esto crea una falsa apariencia de soporte documental.
        \item \textit{Ungrounding} (falta de fundamentación): el LLM realiza afirmaciones factuales específicas que deberían estar respaldadas por el material recuperado, pero no proporciona citas o las fuentes recuperadas no contienen la información afirmada.
    \end{itemize}
\end{itemize}

\begin{itemize}
    \item \textbf{Alucinaciones de inferencia y razonamiento}: implican fallos en la estructura lógica del argumento legal o en la caracterización de las relaciones entre conceptos o autoridades.
    \begin{itemize}
        \item Argumentación ilógica o inválida: el modelo construye una línea de razonamiento que viola principios lógicos básicos o que no se sostiene bajo el escrutinio legal, aunque pueda parecer superficialmente persuasiva.
        \item Miscaracterización de argumentos, partes o posturas procesales: el LLM confunde los argumentos de una parte con el holding del tribunal, o describe incorrectamente la postura procesal o las relaciones entre las partes en un litigio (Dahl et al., 2024).
    \end{itemize}
\end{itemize}

\begin{itemize}
    \item \textbf{Alucinaciones intrínsecas} (inconsistencia con el prompt o corpus de entrenamiento): aunque potencialmente menos frecuentes en respuestas directas a consultas legales factuales, pueden surgir en tareas de dominio cerrado como la sumarización de textos legales extensos o la redacción de documentos basada en instrucciones detalladas, donde el resultado final se desvía sustancialmente o contradice el contenido o las directrices del input proporcionado.
\end{itemize}

Es crucial reconocer que estas categorías no son mutuamente excluyentes; una única respuesta alucinada puede exhibir múltiples tipos de errores simultáneamente. La característica unificadora es la \textbf{desconexión entre la salida generada y una base de verdad relevante} (sea esta los hechos del mundo legal, las fuentes recuperadas o el prompt inicial), a menudo enmascarada por la fluidez lingüística del modelo (Ji, Lee, et al. 2023).

Más allá de la fabricación de información, una forma más insidiosa de desviación se produce a través de la alteración del contenido existente, que puede inducir sesgos cognitivos en el profesional. La investigación ha cuantificado cómo los LLMs, en tareas de resumen, alteran el encuadre del texto original, por ejemplo, cambiando el sentimiento de neutro a positivo o negativo. En un estudio, se observó que esto ocurre en un 21.86\% de los casos (Alessa et al., 2025). En el contexto legal, esto podría manifestarse como un resumen de una sentencia que enfatiza los argumentos de una parte sobre la otra, o que presenta un análisis doctrinal de manera más favorable o crítica de lo que realmente es, influyendo sutilmente en la evaluación inicial del abogado.

Adicionalmente, se ha identificado el sesgo de primacía, donde el resumen generado por el LLM se enfoca desproporcionadamente en la información presentada al inicio de un documento, ocurriendo en un 5.94\% de las ocasiones (Alessa et al., 2025). Esto representa un riesgo significativo en la revisión de largos expedientes judiciales o contratos, donde los detalles críticos pueden encontrarse en secciones posteriores que el LLM podría minimizar u omitir.

Esta taxonomía general se complementa con esfuerzos de la comunidad para categorizar errores más granulares específicos de los sistemas RAG. Por ejemplo, el benchmark LibreEval (Arize AI) identifica tipos de fallo como '\textit{Overclaim}', donde el modelo excede lo soportado por las fuentes, '\textit{Incompleteness}', cuando la respuesta omite información crucial presente en el contexto, o '\textit{Relational-error}', que denota fallos al sintetizar correctamente la información de múltiples fragmentos recuperados. Estos errores específicos de RAG pueden considerarse manifestaciones detalladas de nuestras categorías más amplias, subrayando la complejidad de asegurar la fidelidad en estos sistemas.

\begin{table}[htbp] 
  \centering
  \caption{Taxonomía de alucinaciones en LLMs legales}
  \label{tab:taxonomia_alucinaciones}
  \begin{tabular}{
    >{\raggedright\arraybackslash}p{0.25\textwidth} 
    >{\raggedright\arraybackslash}p{0.4\textwidth} 
    >{\raggedright\arraybackslash}p{0.3\textwidth} 
  }
    \toprule
    \textbf{Categoría Principal} & \textbf{Subtipo / Descripción Breve} & \textbf{Ejemplo Ficticio Legal} \\
    \midrule
    \textbf{Alucinaciones factuales / extrínsecas} & Inconsistencia con los hechos del mundo legal. & \\
    \cmidrule(r){1-1}
    & \textit{Misstatement de Ley o precedente:} Declaración errónea del contenido o holding de una autoridad. & "El LLM afirma que la Ley de Arrendamientos Urbanos de 2022 permite el desahucio inmediato sin notificación." (Cuando la ley exige 30 días) \\
    \addlinespace
    & \textit{Fabricación de autoridad:} Invención de casos, estatutos, o académicos inexistentes. & "Según el caso Martínez c. Constructora Sol (2025), la responsabilidad objetiva es inaplicable." (El caso no existe) \\
    \addlinespace
    & \textit{Error de aplicación jurisdiccional/temporal:} Aplicación incorrecta de normas a otra jurisdicción o presentación de normas derogadas como vigentes. & "El LLM cita un artículo del Código Civil de 1950 para resolver una disputa contractual actual, ignorando reformas posteriores." \\
    \midrule
    \textbf{Alucinaciones basadas en fuentes (errores de RAG)} & Inconsistencia con los documentos recuperados por el sistema RAG. & \\
    \cmidrule(r){1-1}
    & \textit{Misgrounding (Fundamentación errónea):} Cita una fuente real, pero afirma algo que la fuente no respalda o contradice. & "El documento X dice 'el contrato es válido', pero el LLM reporta: 'Según el documento X, el contrato es nulo'." \\
    \addlinespace
    & \textit{Ungrounding (Falta de fundamentación):} Realiza afirmaciones que deberían estar respaldadas por el material recuperado, pero no proporciona citas o las fuentes no lo contienen. & "El demandado actuó con negligencia. (Sin citar ninguna prueba o documento recuperado que lo sustente)." \\
    \midrule
    \textbf{Alucinaciones de inferencia y razonamiento} & Fallos en la estructura lógica del argumento legal. & \\
    \cmidrule(r){1-1}
    & \textit{Argumentación ilógica o inválida:} Construye una línea de razonamiento que viola principios lógicos. & "Si todos los contratos requieren oferta y aceptación, y este documento es un contrato, entonces el cielo es azul." (Conclusión no sigue) \\
    \addlinespace
    & \textit{Miscaracterización de argumentos/partes:} Confunde los argumentos de una parte con el holding, o describe incorrectamente posturas procesales. & "El LLM presenta la petición del demandante como si fuera la sentencia final del juez." \\
    \midrule
    \textbf{Alucinaciones intrínsecas} & Inconsistencia con el prompt o corpus de entrenamiento (en tareas de dominio cerrado). & \\
    \cmidrule(r){1-1}
    & Desviación sustancial del contenido o directrices del input en tareas como sumarización o redacción basada en instrucciones. & "Prompt: 'Resume el siguiente contrato en 100 palabras enfocándote en las cláusulas de penalización.' Respuesta del LLM: Un resumen de 500 palabras sobre la historia de la empresa contratante." \\
    \bottomrule
  \end{tabular}
\end{table}

\subsection{Causas raíz de las alucinaciones en LLMs legales}

Comprender por qué los Grandes Modelos de Lenguaje (LLMs) generan alucinaciones, especialmente cuando se aplican al riguroso dominio legal, es un paso indispensable para desarrollar estrategias efectivas de mitigación y evaluación. Las causas son multifactoriales, arraigadas tanto en las propiedades fundamentales de la tecnología actual de LLMs como en las complejidades específicas del conocimiento y el lenguaje jurídico. Estos factores interactúan de maneras complejas, dando lugar a las diversas manifestaciones de errores que hemos categorizado previamente.

Una causa fundamental reside en las \textbf{limitaciones inherentes a los datos de entrenamiento}. La vasta escala de los corpus utilizados para entrenar LLMs, a menudo extraídos de la web, implica una inevitable variabilidad en la calidad, veracidad y actualidad (Bender et al., 2021). En el ámbito legal, esto es particularmente problemático. Los textos legales disponibles públicamente pueden ser incompletos o representar solo una fracción del panorama jurídico total. Más críticamente, el derecho es un sistema dinámico; leyes y precedentes cambian constantemente, haciendo que cualquier LLM entrenado en un conjunto de datos estático contenga inevitablemente información obsoleta (Khmaïess Al Jannadi, 2023). A esto se suma la presencia de sesgos históricos –sociales, económicos, raciales– codificados en los textos legales y judiciales del pasado. Al aprender patrones estadísticos de estos datos, \textbf{los LLMs corren el riesgo no solo de reproducir, sino de amplificar estas desigualdades}, generando respuestas que pueden perpetuar injusticias sistémicas (Gebru et al., 2018; O'Neil 2016; Barocas and Selbst 2016). La escasez relativa de datos legales verificados, de alta calidad y representativos de todas las jurisdicciones y áreas del derecho sigue siendo un cuello de botella significativo.

Además, se puede argumentar que el problema se origina en la propia cultura de evaluación de la inteligencia artificial. Los modelos son entrenados y evaluados predominantemente con métricas que penalizan severamente las respuestas que expresan incertidumbre (como "no lo sé"). Como resultado, los LLMs aprenden que una conjetura plausible es preferible a una abstención honesta, perpetuando un comportamiento de "adivinar siempre" (Kalai et al., 2025). Este modo de operar, análogo al de un estudiante que nunca deja preguntas en blanco en un examen, es fundamentalmente incompatible con la prudencia que exige la práctica legal.

La escasez de datos verificados y representativos de jurisdicciones específicas es una causa directa y demostrable de las alucinaciones. Un estudio empírico sobre el rendimiento de los LLMs en una jurisdicción no anglosajona reveló que, si bien modelos como GPT y Claude  destacaban en tareas de redacción, todos los modelos fallaban sistemáticamente en la investigación jurídica, generando de forma frecuente citas a casos inexistentes. El autor concluye que esta deficiencia se debe a que los LLMs están entrenados predominantemente con datos de los sistemas jurídicos dominantes (como el estadounidense), careciendo de una base de conocimiento suficiente sobre la jurisprudencia de otras regiones, lo que les obliga a "alucinar" para completar la tarea (Hemrajani, 2025).

Íntimamente ligada a los datos está la \textbf{naturaleza probabilística y la arquitectura misma de los LLMs}. Estos modelos, a pesar de su impresionante capacidad para generar texto coherente, no operan a través de una comprensión semántica profunda o un razonamiento lógico análogo al humano (Searle 1980; Marcus \& Davis, 2022). Son fundamentalmente motores predictivos que calculan la secuencia de palabras más probable basándose en las correlaciones estadísticas aprendidas de sus vastos datos de entrenamiento. Esta orientación hacia la predicción estadística, optimizada a menudo para la fluidez lingüística por encima de la factualidad estricta, los hace intrínsecamente propensos a generar afirmaciones que \textit{suenan} correctas pero que carecen de base real (Ji et al., 2023; Bowman 2015).

Esta optimización intrínseca para la fluidez puede llevar a lo que a veces se describe como 'confabulación', un proceso mediante el cual el modelo, enfrentado a una falta de información fáctica directa o a una ambigüedad en la consulta, \textbf{'inventa' detalles o narrativas coherentes para mantener la continuidad del discurso}, aunque estos elementos fabricados carezcan de una base real. La confabulación, en este sentido, es una manifestación directa de la arquitectura predictiva del LLM priorizando la apariencia de comprensión sobre la factualidad estricta, llevando a la generación de alucinaciones que, aunque erróneas, pueden ser engañosamente persuasivas por su coherencia superficial.

La dificultad de los LLMs para conectar los principios legales abstractos con los hechos concretos de un caso es una causa fundamental de las alucinaciones. Un estudio que condicionó a los LLMs con diferentes niveles de conocimiento del sistema legal alemán para detectar el discurso de odio lo demostró de manera concluyente. Cuando los modelos eran "condicionados" únicamente con el conocimiento más abstracto (como el título de una norma constitucional o estatutaria), \textbf{mostraban una falta de comprensión profunda de la tarea, llegando a contradecirse y a alucinar respuestas} cuando se les presentaban normas ficticias o irrelevantes (Ludwig et al., 2025). Esto sugiere que la arquitectura probabilística de los LLMs, en ausencia de un anclaje en definiciones y ejemplos concretos, lucha por aplicar correctamente el razonamiento jurídico, recurriendo a la invención.

Fenómenos como el sobreajuste (\textit{overfitting}), donde el modelo memoriza patrones específicos del entrenamiento en lugar de aprender principios generales, pueden exacerbar este problema, limitando su capacidad para generalizar correctamente a situaciones nuevas o ligeramente diferentes (Khmaïess Al Jannadi, 2023). Además, su capacidad inherente para la extrapolación (aunque esencial para la generalización) puede desviarse fácilmente hacia la invención o la conexión espuria de conceptos cuando se enfrenta a consultas que bordean los límites de su conocimiento o requieren inferencias complejas (Shaip, 2022; Huang et al. 2021; Domingos 2015).

El \textbf{dominio legal en sí mismo presenta una complejidad intrínseca} que amplifica estos desafíos. El lenguaje jurídico es notoriamente técnico, denso en significado, altamente dependiente del contexto y plagado de ambigüedades y términos polisémicos (Khmaïess Al Jannadi, 2023). Interpretar correctamente un estatuto, un contrato o una sentencia requiere no solo comprender el significado literal de las palabras, sino también el contexto legislativo, la intención de las partes, la historia procesal y la red de precedentes relevantes – tareas que exigen un nivel de comprensión contextual y razonamiento que desafía a los LLMs actuales. El razonamiento jurídico per se, con sus métodos analógicos, deductivos basados en reglas y principios, y su constante ponderación de factores, representa una forma de cognición de orden superior que los LLMs, basados en patrones estadísticos, luchan por emular fielmente (Ashley 2017; Choi and Schwarcz, 2024).

Incluso las \textbf{estrategias diseñadas para mitigar las alucinaciones, como RAG, introducen sus propios puntos de vulnerabilidad}. Como se discutirá en detalle en la Sección \ref{sec:rag_mitigacion}, la efectividad de RAG depende críticamente de la calidad de su módulo de recuperación de información. Si la información recuperada es irrelevante, incorrecta o incompleta, el LLM generador, incluso si intenta ser fiel al contexto proporcionado, producirá una respuesta defectuosa. Además, el propio LLM generador puede fallar en integrar correctamente la información recuperada, priorizando su conocimiento paramétrico erróneo o sintetizando incorrectamente las fuentes (Addleshaw Goddard, 2024).

En última instancia, el análisis de las causas raíz de las alucinaciones sería incompleto si se limitara al modelo. \textbf{La causa fundamental más peligrosa no reside en la máquina, sino en el humano que la utiliza sin criterio}. La IA tiene el potencial de amplificar las capacidades de los profesionales diligentes, mientras que puede inducir a errores a aquellos que la utilizan sin un criterio crítico y una supervisión adecuada. Un profesional con pensamiento crítico y experiencia utilizará el LLM como un catalizador para acelerar su investigación, validando cada resultado. Sin embargo, un usuario sin estas bases, seducido por la aparente facilidad, delegará su razonamiento y caerá en la trampa de la complacencia. Por tanto, la mayor causa de riesgo no es la alucinación del modelo, sino la "\textbf{alucinación del usuario}": la creencia de que una herramienta puede sustituir la responsabilidad, el esfuerzo y el juicio profesional. Este fenómeno, alimentado por una cultura de la inmediatez y los atajos, es el verdadero desafío a mitigar en la integración de la IA en el sector legal.

\subsection{Impacto específico y riesgos asociados en la práctica legal}

La manifestación de estas causas en forma de alucinaciones tiene un impacto tangible y multifacético en el ecosistema legal:
\begin{enumerate}
    \item \textbf{Socavamiento de la investigación y el análisis jurídico}: la base de cualquier trabajo legal riguroso es la investigación precisa. Las alucinaciones, al introducir información falsa o fabricada, contaminan este proceso fundamental, haciendo perder tiempo en la verificación, llevando a análisis erróneos y, en última instancia, a estrategias legales defectuosas.
    \item \textbf{Riesgos profesionales y éticos}: para los abogados, confiar en información alucinada puede tener consecuencias devastadoras. Puede llevar a la presentación de escritos judiciales defectuosos (resultando en sanciones como la popular del caso \textit{Mata v. Avianca}), al incumplimiento del deber de competencia y diligencia, a la violación del deber de franqueza ante el tribunal, y a posibles reclamaciones por negligencia profesional (Yamane, 2020; Schwarcz et al., 2024). La necesidad de verificar exhaustivamente cada resultado de la IA puede, irónicamente, anular los beneficios de eficiencia prometidos (Gottlieb 2024).
    \item \textbf{Erosión de la confianza}: la prevalencia de alucinaciones, especialmente si no se aborda con transparencia, puede minar la confianza de los profesionales legales, los clientes y el público en general hacia las herramientas de IA y, por extensión, hacia quienes las utilizan (Khmaïess Al Jannadi, 2023). Esta erosión de la confianza puede obstaculizar la adopción de tecnologías potencialmente beneficiosas.
    \item \textbf{Impacto en el acceso a la justicia}: existe una paradoja preocupante: los LLMs se promocionan como una herramienta para democratizar el acceso a la información legal para litigantes pro se o personas de bajos recursos. Sin embargo, estos mismos usuarios son los menos equipados para detectar y verificar alucinaciones sofisticadas, lo que los hace particularmente vulnerables a recibir información legal incorrecta y perjudicial (Draper and Gillibrand 2023; Dahl et al., 2024). En lugar de cerrar la brecha, la IA no fiable podría ampliarla.
    \item \textbf{Integridad del sistema judicial}: a nivel sistémico, la introducción de información falsa o fabricada en los procedimientos judiciales, ya sea inadvertidamente por abogados o potencialmente de forma maliciosa, amenaza la integridad fundamental del proceso contradictorio y la búsqueda de la verdad.
    \item \textbf{Riesgos Cognitivos y de Juicio Sutil}: Quizás el riesgo más subestimado no es que la IA proporcione información falsa, sino que presente información verídica de una manera que explote los sesgos cognitivos humanos. Los LLMs pueden actuar como "vectores de sesgo", induciendo sesgos de encuadre (framing bias) que alteran la percepción de un problema sin cambiar los hechos. Por ejemplo, al resumir los argumentos de la parte contraria, un LLM podría seleccionar un lenguaje que los haga parecer más débiles de lo que son. De igual forma, el sesgo de autoridad puede llevar a un abogado a aceptar una conclusión generada por la IA con menos escrutinio del que aplicaría a un colega humano, simplemente por la presentación fluida y aparentemente lógica del modelo (Alessa et al., 2025). Este efecto erosiona la objetividad del juicio profesional desde dentro, de una forma mucho más difícil de detectar que una simple cita falsa.
\end{enumerate}

Abordar las causas raíz de las alucinaciones legales no es, por consiguiente, una mera optimización técnica, sino un imperativo ético y funcional para el futuro de la IA en el derecho.

La gravedad de este impacto no ha pasado desapercibida para los legisladores, y el riesgo inherente a la diseminación de información legal incorrecta o fabricada es una de las preocupaciones centrales que animan los esfuerzos regulatorios a nivel global. En este sentido, el Reglamento (UE) 2024/1689 del Parlamento Europeo y del Consejo, conocido como Ley de Inteligencia Artificial de la Unión Europea (en adelante, la Ley de IA de la UE, el Reglamento o EU-AIAct), un marco legislativo pionero y ambicioso, establece un precedente significativo. Al adoptar un enfoque basado en el riesgo, la Ley de IA de la UE busca imponer requisitos más estrictos a aquellos sistemas de IA cuyas fallas podrían tener consecuencias severas para los derechos fundamentales, la seguridad o el correcto funcionamiento de instituciones clave. Aunque la categorización específica de todas las herramientas de IA legal bajo este marco aún está por definirse en su aplicación práctica, es plausible anticipar que aquellos sistemas destinados a influir en la administración de justicia o a proporcionar asesoramiento en áreas críticas podrían ser objeto de un escrutinio regulatorio intensificado precisamente por el potencial disruptivo de fenómenos como las alucinaciones.

En conclusión, las alucinaciones no son un fallo técnico menor, sino una manifestación de las limitaciones fundamentales en la forma en que los LLMs actuales procesan la información y modelan el mundo, con ramificaciones particularmente críticas en el sensible y normativo dominio del derecho. Abordar este desafío es una condición previa para cualquier integración responsable y beneficiosa de la IA en la profesión legal.

\section{Evaluación de alucinaciones en aplicaciones legales de IA: metodologías, desafíos y estado actual}
\label{sec:evaluacion_alucinaciones}

La mera existencia del fenómeno de las alucinaciones en LLMs aplicados al derecho, detallada en la sección anterior, impone una necesidad crítica e ineludible: el desarrollo y la aplicación de metodologías rigurosas para su evaluación, detección y cuantificación. Dada la naturaleza de alto riesgo del dominio legal, donde las decisiones basadas en información incorrecta pueden tener consecuencias jurídicas, financieras y sociales devastadoras, la simple confianza en las afirmaciones de los desarrolladores o en la aparente plausibilidad de las respuestas generadas es insostenible. La evaluación empírica sistemática se convierte, por tanto, no solo en un ejercicio académico deseable, sino en un prerrequisito fundamental para la integración responsable de estas tecnologías en la práctica profesional, la educación jurídica y el sistema de justicia en general. Sin embargo, como exploraremos en esta sección, la evaluación de las alucinaciones legales es una tarea intrínsecamente compleja, plagada de desafíos metodológicos y conceptuales únicos que requieren enfoques matizados y un escrutinio constante.

\subsection{Desafíos fundamentales en la evaluación de la IA legal}

Evaluar la factualidad y detectar alucinaciones en los LLMs cuando operan sobre conocimiento legal presenta un conjunto de desafíos particulares que van más allá de los encontrados en dominios más generales o con hechos más objetivos. Estos desafíos limitan la aplicabilidad directa de muchas métricas de evaluación estándar y exigen una consideración cuidadosa del contexto específico.

\subsubsection{El problema del \textit{\textbf{Ground Truth}} legal}

A diferencia de preguntas con respuestas factuales únicas y objetivas (p. ej., \textit{"¿Quién ganó el Mundial de 2022?}"), la "verdad" legal es a menudo más elusiva. El \textit{ground truth} en derecho está intrínsecamente ligado a:

    \begin{itemize}
        \item \textit{Interpretación:} las leyes y los precedentes requieren interpretación, y los expertos legales pueden discrepar razonablemente sobre el significado o la aplicación de una norma a un conjunto específico de hechos.
        \item \textit{Variabilidad jurisdiccional y temporal:} el derecho aplicable varía enormemente entre jurisdicciones (locales, estatales, internacionales) y evoluciona constantemente con nuevas leyes y decisiones judiciales. Lo que es "correcto" en una jurisdicción o momento puede ser incorrecto en otro.
        \item \textit{Ambigüedad lingüística:} como se mencionó, el lenguaje legal está repleto de términos técnicos, estándares vagos ("razonable", "debido proceso") y ambigüedades inherentes que desafían una verificación binaria simple.

Esta complejidad inherente significa que, para muchas tareas legales que trascienden la mera recuperación de información (como el análisis de problemas jurídicos complejos o la formulación de estrategias), el concepto de un único \textit{ground truth} contra el cual medir una respuesta de IA se vuelve inaplicable. En tales escenarios, la evaluación se desplaza de la 'corrección' binaria hacia la 'viabilidad legal': la capacidad de una respuesta para ser argumentativamente sostenible y coherente dentro del marco normativo y doctrinal, aun cuando puedan existir múltiples enfoques válidos. Establecer esta viabilidad requiere, por tanto, una profunda experiencia legal y, a menudo, implica juicios interpretativos que pueden ser objeto de debate

Establecer un \textit{ground truth} fiable para evaluar las respuestas de un LLM requiere, por tanto, una profunda experiencia legal y, a menudo, implica juicios interpretativos que pueden ser objeto de debate.

    \end{itemize}

\subsubsection{Opacidad de los sistemas comerciales (el problema de la "Caja Negra")}

Una barrera significativa para la evaluación independiente y rigurosa es la naturaleza propietaria y cerrada de muchas de las herramientas de IA legal más avanzadas disponibles comercialmente (Magesh et al., 2024). Los proveedores rara vez divulgan detalles cruciales sobre:

    \begin{itemize}
        \item \textit{Datos de entrenamiento:} la composición exacta, fuentes, actualidad y posibles sesgos de los datos masivos utilizados para entrenar sus modelos base o especializados.
        \item \textit{Arquitectura del modelo y algoritmos:} las especificidades de la arquitectura del LLM subyacente, los algoritmos de RAG empleados, o los métodos de fine-tuning aplicados.
        \item \textit{Procesos internos:} los mecanismos específicos de recuperación de información, los prompts internos utilizados, o los filtros post-generación aplicados.
    \end{itemize}
Esta opacidad impide a los investigadores y usuarios comprender plenamente por qué un sistema produce una respuesta particular (alucinada o no), aislar las fuentes de error, replicar los resultados de forma independiente o comparar de manera justa el rendimiento entre diferentes plataformas. La evaluación a menudo debe basarse únicamente en el análisis de la salida final, tratando al sistema como una "caja negra".

La opacidad inherente a los sistemas comerciales de IA legal trasciende la problemática puramente técnica para entrar en el ámbito de la estrategia de mercado y la gestión del riesgo. Se pueden identificar varias dinámicas clave:
\begin{itemize}
        \item Señalización de mercado vs. transparencia técnica: El término "IA" funciona como una potente señal de mercado para atraer capital y clientes. Sin embargo, esta estrategia de marketing no siempre se corresponde con una divulgación transparente de las arquitecturas, los datos de entrenamiento o las tasas de error de los sistemas. Esto crea una asimetría de información que dificulta la evaluación objetiva por parte de los usuarios.
        \item Riesgo reputacional sistémico: La estrategia de "caja negra", si bien puede ofrecer ventajas comerciales a corto plazo, genera un riesgo sistémico. Un fallo notorio en un sistema opaco (p. ej., una alucinación con consecuencias judiciales) no solo daña la reputación del proveedor, sino que puede mermar la confianza en toda la categoría de productos de IA legal, ralentizando su adopción generalizada.
        \item El valor de la auditabilidad: En consecuencia, un factor diferenciador clave para la madurez del sector será la transición desde modelos que priorizan la percepción de la innovación hacia aquellos que demuestran su valor a través de la transparencia y la auditabilidad. Un sistema cuyo rendimiento puede ser verificado y comprendido por terceros ofrece una base más sólida para la confianza y la integración responsable en flujos de trabajo críticos.
\end{itemize}

\subsubsection{Complejidad inherente de las tareas y habilidades legales}

La práctica legal implica una gama diversa de tareas cognitivas que van mucho más allá de la simple recuperación de información o respuesta a preguntas factuales. Incluye el razonamiento analógico, la argumentación persuasiva, el juicio estratégico, la síntesis de información compleja, la redacción matizada y la comprensión contextual profunda. Evaluar el rendimiento de un LLM en estas tareas requiere métricas y metodologías que puedan capturar estas dimensiones cualitativas, lo cual es intrínsecamente más difícil que evaluar la corrección factual de una respuesta a una pregunta directa (Schwarcz et al., 2024).

\subsubsection{Ausencia de Benchmarks estandarizados y específicos}

Si bien están surgiendo benchmarks en el área de IA y derecho, la comunidad académica ha respondido a esta necesidad con la creación de marcos de evaluación especializados y de dominio relevante. Plataformas como LexGLUE, un benchmark para la comprensión del lenguaje jurídico en inglés, y LawBench, que evalúa el conocimiento jurídico de los LLMs en el contexto chino, son ejemplos clave. Estos esfuerzos, catalogados en revisiones exhaustivas del campo (Shao et al., 2025), son fundamentales para establecer métricas estandarizadas que permitan cuantificar de manera rigurosa el progreso en tareas complejas como la predicción de sentencias y la recuperación de precedentes, moviendo el campo más allá de las evaluaciones genéricas de NLP.

No obstante, para llenar este vacío, la comunidad investigadora está desarrollando enfoques de evaluación innovadores que pueden agruparse en dos categorías principales.

Por un lado, surgen benchmarks técnicos diseñados específicamente para medir la fiabilidad de la arquitectura RAG. Iniciativas como LibreEval de Arize AI proporcionan conjuntos de datos para evaluar la propensión a la alucinación y la fidelidad al contexto (groundedness), mientras que herramientas como RAGTruth (Niu et al., 2024) persiguen objetivos similares. Estos esfuerzos son cruciales para cuantificar de manera rigurosa los fallos específicos de los sistemas RAG.

Por otro lado, una estrategia complementaria y creativa es el uso de exámenes estandarizados de acceso a la profesión como benchmarks de conocimiento y precisión factual. Un ejemplo notable es el uso del All India Bar Examination (AIBE) para validar el modelo "Legal Assist AI". Al alcanzar una puntuación del 60.08\%, este enfoque proporcionó una métrica cuantificable y directamente comparable con el rendimiento humano (Gupta et al., 2025). La combinación de estas estrategias —tanto las técnicas como las basadas en la competencia profesional— es fundamental para construir un marco de evaluación verdaderamente robusto para la IA legal.

\subsection{Métricas y metodologías para la detección y cuantificación}

Navegar por los desafíos mencionados requiere el despliegue de un conjunto diverso de métricas y metodologías, cada una con sus fortalezas y debilidades inherentes:

\begin{enumerate}
    \item \textbf{Evaluación basada en referencias (usando oráculos de metadatos):} este enfoque, pionero en el estudio de Dahl et al. (2024), aprovecha la existencia de metadatos estructurados y verificables asociados con los documentos legales (p. ej., tribunal emisor, fecha de decisión, juez ponente, citas dentro del documento, estado de derogación). Se formulan consultas al LLM que tienen una respuesta objetiva y verificable en estos metadatos (p. ej., "\textit{¿Qué tribunal decidió el caso X?}"). La respuesta del LLM se compara directamente con el \textit{ground truth} del metadato.

    \begin{itemize}
        \item \textit{Fortalezas:} Proporciona una medida objetiva y cuantificable de la alucinación para un subconjunto de hechos legales verificables. Permite análisis a gran escala si se dispone de los metadatos adecuados.

        \item \textit{Debilidades:} Se limita a la información contenida en los metadatos disponibles, sin poder evaluar la corrección del razonamiento legal sustantivo o la interpretación. Si bien es valioso para identificar alucinaciones fácticas directas (ej. una cita incorrecta), este método no aborda la evaluación de respuestas a problemas legales complejos donde la 'corrección' depende de la interpretación y el razonamiento, y no de un simple dato verificable. En estos casos, la ausencia de un 'error factual' no garantiza la 'viabilidad legal' de la solución propuesta. Depende de la calidad y cobertura de las bases de datos de metadatos.
        
    \end{itemize}

    \item \textbf{Evaluación libre de referencias (auto-consistencia / auto-contradicción):} esta familia de técnicas busca detectar alucinaciones sin necesidad de un \textit{ground truth} externo, explotando la naturaleza estocástica de la generación de los LLMs (Manakul, Liusie, and Gales 2023; Mündler et al. 2023). Se generan múltiples respuestas para el mismo prompt (usando una temperatura > 0) y se analiza su consistencia.

    \textit{Self-Contradiction como límite inferior:} la detección de contradicciones lógicas directas entre múltiples respuestas generadas para el mismo input es una fuerte señal de alucinación, ya que respuestas fácticamente correctas deberían ser consistentes. Este método proporciona un límite inferior útil para la tasa de alucinación, sin asumir la calibración del modelo.

    \textit{Self-Consistency como heurística:} La consistencia entre múltiples respuestas puede usarse como una heurística para la confianza (respuestas más consistentes \textit{podrían} ser más probables de ser correctas), pero esto asume un grado de calibración del modelo que puede no ser válido, especialmente en dominios complejos como el derecho.

    \begin{itemize}

        \item \textit{Fortalezas:} no requiere \textit{ground truth} externo, potencialmente aplicable a una gama más amplia de preguntas, incluidas aquellas que involucran juicio o interpretación

        \item \textit{Debilidades:} la auto-contradicción solo proporciona un límite inferior (no detecta alucinaciones consistentes). La auto-consistencia como indicador de corrección es una heurística no garantizada. Requiere múltiples inferencias, aumentando el costo computacional.

    \end{itemize}

    \item \textbf{Evaluación humana experta:} considerada el estándar de oro para evaluar tareas legales complejas y la calidad matizada de las respuestas generativas (Schwarcz et al., 2024). Involucra a expertos legales (abogados y académicos) que revisan y califican las salidas del LLM utilizando rúbricas predefinidas que evalúan dimensiones como la corrección factual, la solidez del razonamiento legal, la relevancia, la coherencia, la claridad y la identificación de riesgos.

    \begin{itemize}
        \item \textit{Fortalezas:} capaz de evaluar la calidad sustantiva, el razonamiento complejo y la relevancia contextual que las métricas automáticas a menudo pasan por alto. Es indispensable para validar nuevas tareas o métricas.

        \item \textit{Debilidades:} extremadamente costosa en tiempo y recursos, difícil de escalar, susceptible a la subjetividad y a la variabilidad entre evaluadores (requiere protocolos claros y medición de la fiabilidad inter-evaluador, como el Kappa de Cohen - Cohen 1960).

    \end{itemize}

    \item \textbf{Métricas automatizadas:} incluyen métricas estándar de NLP como ROUGE o BLEU (más adecuadas para tareas como la sumarización o traducción) y métricas emergentes de factualidad que intentan verificar automáticamente las afirmaciones contra bases de conocimiento (p.ej., FActScore - Min et al. 2023) o usando otros LLMs como jueces (Zheng et al., 2023).

    \begin{itemize}
        \item \textit{Fortalezas:} escalables y computacionalmente eficientes una vez desarrolladas.

        \item \textit{Debilidades:} su correlación con el juicio humano sobre la calidad y factualidad \textit{legal} es a menudo baja o no probada. Pueden ser fácilmente "engañadas" por respuestas fluidas pero incorrectas. Su desarrollo y validación para el dominio legal aún está en etapas tempranas.

    \end{itemize}
\end{enumerate}

En la práctica, un enfoque robusto para la evaluación probablemente requiera una \textbf{combinación} de estas metodologías: evaluación basada en referencias para hechos verificables, detección de auto-contradicción para obtener límites inferiores en tareas abiertas, métricas automáticas para análisis a gran escala (con validación cuidadosa), y evaluación humana experta como validación final y para tareas cualitativamente complejas.

\subsection{Benchmarking de herramientas comerciales: estado actual y hallazgos empíricos clave}

La necesidad crítica de evaluación ha impulsado el estudio empírico sistemático y pre-registrado de Magesh et al. (2025) sobre las plataformas comerciales líderes de IA legal. Sus hallazgos, basados en un conjunto diverso de más de 200 consultas legales del mundo real, son reveladores y establecen un punto de referencia crucial:

\begin{enumerate}
    \item Persistencia alarmante de alucinaciones: Contrario a las audaces afirmaciones de marketing de ser "libres de alucinaciones", la mayoría de las herramientas evaluadas fallaron en una proporción significativa. Utilizando una definición rigurosa de alucinación (respuesta incorrecta o mal fundamentada), se encontró que Lexis+ AI alucinó en el 17\% de los casos, y Westlaw AI-Assisted Research lo hizo en más de un tercio de las ocasiones (>33\%).
    \item Variabilidad extrema entre plataformas: El rendimiento no es uniforme. Lexis+ AI tuvo una precisión general del 65\% de respuestas correctas y fundamentadas, estableciéndose como la herramienta más fiable del grupo. En el otro extremo, Ask Practical Law AI, debido a una base de conocimiento más limitada, tuvo una tasa extremadamente alta de respuestas incompletas o rechazos (>60\%), limitando severamente su utilidad práctica (Magesh et al., 2025).
    \item Confirmación de que RAG es una Mitigación, no una Solución: Los resultados confirman que la tecnología RAG sí reduce la tasa de alucinación en comparación con los LLMs de propósito general. Sin embargo, las tasas de error residuales demuestran que, en su implementación actual, el RAG no elimina totalmente las alucinaciones, siendo los fallos en la recuperación de información y en la adhesión del LLM a las fuentes los problemas persistentes.
\end{enumerate}

\begin{enumerate}
    \item \textbf{Persistencia alarmante de alucinaciones:} el hallazgo más contundente es que, contrariamente a las afirmaciones de marketing de "eliminación" o "ausencia" de alucinaciones, \textbf{la mayoría de las herramientas comerciales evaluadas alucinan en una proporción significativa}. Utilizando una definición rigurosa de alucinación (respuesta incorrecta o fundamentada erróneamente), se encontró que Lexis+ AI y Ask Practical Law AI alucinaban entre el 17\% y el 33\% de las veces, mientras que Westlaw AI-Assisted Research alucinaba más de un tercio del tiempo (>34\%). Estas tasas, aunque inferiores a las de GPT-4 o 5 base en tareas legales (58-88\%), siguen siendo inaceptablemente altas para la práctica profesional.

\begin{table}[htbp]
  \centering
  \caption{Rendimiento comparativo de herramientas de IA Legal comerciales (adaptado de Magesh et al., 2024)}
  \label{tab:magesh_hallucination_rates}
  \begin{tabular}{
    l 
    S[table-format=2.1] 
    S[table-format=2.1] 
    S[table-format=2.1] 
  }
    \toprule
    \textbf{Herramienta de IA Legal} & {\textbf{Tasa de Alucinación}} & {\textbf{Resp. Incompletas}} & {\textbf{Resp. Precisas}} \\
    & {(\%)} & {(\%)} & {(\%)} \\
    \midrule
    Lexis+ AI                     & 17  & 18  & 65  \\
    Westlaw AI-Assisted Research  & {>34} & 25  & 41  \\ 
    Ask Practical Law AI          & 17  & {>60} & 19  \\ 
    \midrule
    GPT-4/5 (base, como referencia) & {\textasciitilde58-88} & {\text{N/A*}} & {\text{N/A*}} \\ 
    \bottomrule
    \addlinespace
    \multicolumn{4}{p{0.9\textwidth}}{\footnotesize{\textit{Nota:} Las tasas de alucinación para herramientas comerciales se refieren a respuestas incorrectas o fundamentadas erróneamente. GPT-4/5 base se incluye como referencia general de LLMs sin RAG legal específico, sus tasas de alucinación en tareas legales pueden ser más altas y la estructura de "respuestas incompletas" o "precisas y fundamentadas" puede no ser directamente comparable sin el componente RAG. *N/A indica que la métrica no se reportó de la misma manera o no es directamente comparable.}}
  \end{tabular}
\end{table}

    \item \textbf{RAG es una (gran) mitigación, pero no "la" solución:} los resultados confirman que la tecnología RAG empleada por estas herramientas \textit{sí} reduce la tasa de alucinación en comparación con el uso de LLMs de propósito general sin acceso a bases de datos legales externas. Sin embargo, RAG, tal como se implementa actualmente, \textbf{no elimina totalmente las alucinaciones}. Los fallos en la recuperación de información relevante y la incapacidad del LLM generador para adherirse fielmente a las fuentes recuperadas siguen siendo problemas sustanciales.

    \item \textbf{Variabilidad entre plataformas:} el estudio revela diferencias notables en el rendimiento y el comportamiento entre las distintas herramientas. Lexis+ AI demostró la mayor precisión general (65\% de respuestas correctas y fundamentadas) y una tasa de alucinación más baja, pero aún significativa (\~17\%). Westlaw AI-AR, aunque a menudo proporcionaba respuestas más largas y detalladas, exhibió la tasa de alucinación más alta (\~33\%). Ask Practical Law AI, limitado a su base de conocimientos curada, tuvo una tasa de alucinación relativamente baja pero sufrió de una tasa extremadamente alta de respuestas incompletas o rechazos (>60\%), limitando su utilidad práctica. Esta variabilidad subraya que la etiqueta "IA legal basada en RAG" engloba implementaciones muy diferentes con perfiles de riesgo distintos.

    \item \textbf{Naturaleza insidiosa de los errores (más allá de la fabricación):} un hallazgo crucial es que las alucinaciones en estas herramientas RAG rara vez son fabricaciones completas de casos (aunque ocurren). Más comúnmente, adoptan formas más sutiles y potencialmente más peligrosas:

    \begin{itemize}
        \item \textit{Misgrounding\textbf{:}} citar un caso o estatuto real pero tergiversar lo que dice o aplicarlo incorrectamente.
        \item \textit{Errores de razonamiento:} fallos lógicos al sintetizar información de múltiples fuentes recuperadas.
        \item \textit{Sycophancy/Sesgo Contrafáctico:} aceptar acríticamente premisas falsas en la consulta del usuario.
        \item \textit{Supresión de citas problemáticas:} el estudio de Westlaw AI-AR observó instancias donde el sistema parecía generar una afirmación basada en un caso derogado, pero suprimía la cita directa, posiblemente debido a la integración con sistemas de verificación de citas como KeyCite, lo cual impide la verificación por parte del usuario.
    \end{itemize}

De manera complementaria a estos hallazgos, un estudio empírico en un sistema jurídico con datos limitados evaluó el desempeño de varios LLMs (incluyendo GPT-4/5 y Claude 3) frente a un abogado junior en cinco tareas legales (identificación de problemas, redacción, asesoramiento, investigación y razonamiento). Los resultados corroboraron que, si bien los LLMs avanzados pueden igualar o incluso superar el rendimiento humano en tareas estructuradas como la redacción de escritos o la identificación de problemas, su fiabilidad colapsa en la investigación jurídica, donde la generación de casos falsos ("alucinaciones") fue un problema persistente en todos los modelos evaluados (Hemrajani, 2025). Este estudio refuerza la idea de que la efectividad de la IA legal es altamente dependiente de la tarea y de la calidad de los datos de entrenamiento para esa jurisdicción específica.

Estos errores "insidiosos" son particularmente preocupantes porque pueden crear una falsa sensación de fiabilidad y son más difíciles de detectar para un usuario que no realiza una verificación profunda de cada fuente citada.

\end{enumerate}

\textbf{Implicaciones de los hallazgos de evaluación:}

Los resultados empíricos actuales, aunque limitados, tienen implicaciones significativas:

\begin{itemize}
    \item \textbf{Escepticismo justificado:} demuestran que las afirmaciones audaces sobre la eliminación de alucinaciones por parte de los proveedores deben tomarse con extrema precaución.

    \item \textbf{Necesidad de transparencia:} subrayan la necesidad urgente de mayor transparencia por parte de los proveedores sobre cómo funcionan sus sistemas, qué datos utilizan y, crucialmente, sobre sus tasas de error y limitaciones conocidas, evaluadas mediante benchmarks independientes.

    \item \textbf{Imperativo de la diligencia profesional:} refuerzan la obligación ética y profesional ineludible de los abogados de verificar críticamente \textit{cualquier} resultado generado por IA antes de incorporarlo a su trabajo o asesoramiento. La confianza ciega en estas herramientas es, en el estado actual de la tecnología, imprudente.

    \item \textbf{Guía para la investigación futura:} Identifican áreas clave para la mejora técnica (optimización de retrieval y generación en RAG legal) y para la investigación académica (desarrollo de mejores benchmarks, estudio del impacto en diferentes tipos de usuarios y tareas).
\end{itemize}

En conclusión, la evaluación rigurosa es la piedra angular para comprender y gestionar el riesgo de alucinaciones en la IA legal. Si bien las metodologías están evolucionando y enfrentan desafíos, la evidencia empírica inicial ya proporciona una advertencia clara: las alucinaciones son una realidad persistente incluso en las herramientas comerciales más avanzadas, lo que exige un enfoque cauteloso, crítico y centrado en el ser humano para la adopción de la IA en el derecho.

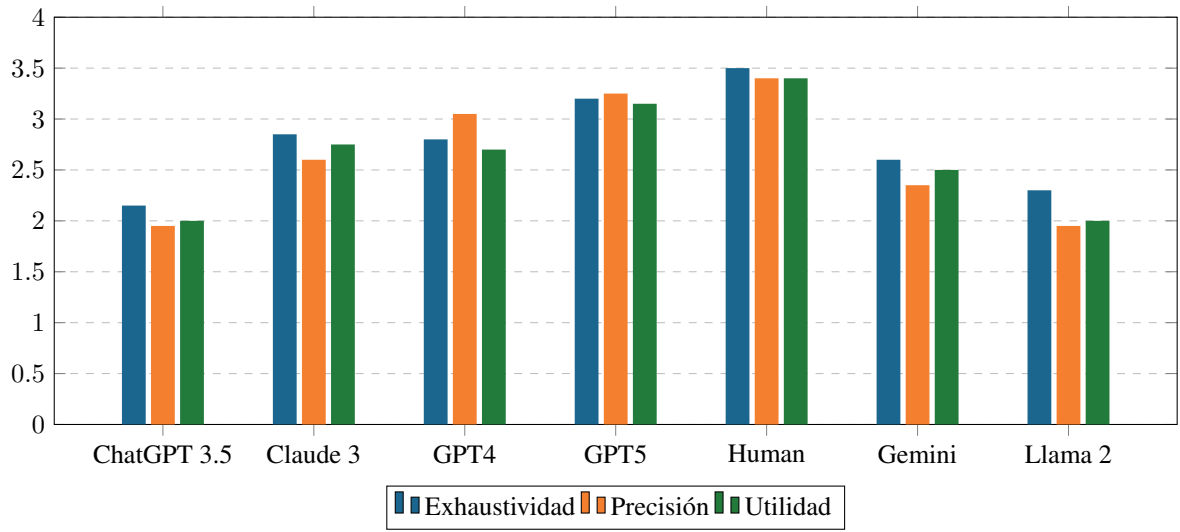
\begin{figure}[h!]
    \centering
    \definecolor{compblue}{RGB}{27, 102, 142}
    \definecolor{accurorange}{RGB}{244, 127, 46}
    \definecolor{helpgreen}{RGB}{34, 122, 59}

    \begin{tikzpicture}
        \begin{axis}[
            width=\textwidth, 
            height=7cm,       
            ybar, 
            enlarge x limits=0.12, 
            legend style={
                at={(0.5, -0.15)}, 
                anchor=north,
                legend columns=-1 
            },
            ylabel={}, 
            symbolic x coords={
                ChatGPT 3.5, 
                Claude 3, 
                GPT4, 
                GPT5,
                Human, 
                Gemini, 
                Llama 2
            },
            xtick=data, 
            x tick label style={rotate=0, anchor=north}, 
            ymin=0, ymax=4, 
            ytick={0, 0.5, 1, 1.5, 2, 2.5, 3, 3.5, 4}, 
            ymajorgrids=true, 
            grid style=dashed, 
            bar width=9pt, 
        ]
        
        \addplot[fill=compblue, draw=none] coordinates {
            (ChatGPT 3.5, 2.15) 
            (Claude 3, 2.85) 
            (GPT4, 2.8) 
            (GPT5, 3.2) 
            (Human, 3.5) 
            (Gemini, 2.6) 
            (Llama 2, 2.3)
        };
        \addlegendentry{Exhaustividad}

        \addplot[fill=accurorange, draw=none] coordinates {
            (ChatGPT 3.5, 1.95) 
            (Claude 3, 2.6) 
            (GPT4, 3.05) 
            (GPT5, 3.25) 
            (Human, 3.4) 
            (Gemini, 2.35) 
            (Llama 2, 1.95)
        };
        \addlegendentry{Precisión}

        \addplot[fill=helpgreen, draw=none] coordinates {
            (ChatGPT 3.5, 2.0) 
            (Claude 3, 2.75) 
            (GPT4, 2.7) 
            (GPT5, 3.15) 
            (Human, 3.4) 
            (Gemini, 2.5) 
            (Llama 2, 2.0)
        };
        \addlegendentry{Utilidad}
        
        \end{axis}
    \end{tikzpicture}
    \caption{
        Evaluación comparativa del rendimiento de Modelos de Lenguaje (LLMs) y un experto humano en la tarea de investigación jurídica. Se observa que el rendimiento humano sigue siendo el punto de referencia en todas las métricas. GPT5 representa una mejora hipotética sobre GPT4, aunque sin alcanzar la fiabilidad humana. Las puntuaciones, en una escala de 1 a 4, son el resultado de una evaluación por pares de expertos juristas bajo criterios predefinidos, con un alto grado de acuerdo inter-anotador (Cohen's $\kappa = 0.85$), garantizando la objetividad de los resultados.
    }
    \label{fig:comparativa_rendimiento_ia}
\end{figure}

Para cuantificar el rendimiento de los LLMs en tareas jurídicas realistas, se llevó a cabo una evaluación manual experta cuyos resultados se resumen en la Figura \ref{fig:comparativa_rendimiento_ia}. El proceso se diseñó para garantizar la objetividad y fiabilidad de las puntuaciones:

\begin{itemize}
    \item \textbf{Anotadores y datos:} Un conjunto de 50 consultas complejas de investigación jurídica fue presentado a cada LLM y a un abogado experto. Las respuestas fueron evaluadas de forma independiente por dos juristas senior con experiencia en la materia, quienes no tuvieron conocimiento del origen de cada respuesta (evaluación a doble ciego).
    
    \item \textbf{Criterios de evaluación (rúbrica):} Los anotadores asignaron una puntuación de 1 (deficiente) a 4 (excelente) para cada una de las siguientes métricas, basándose en una guía de anotación predefinida:
    \begin{itemize}
        \item \textbf{Exhaustividad:} ¿La respuesta identifica todos los puntos y matices legales relevantes? ¿Omite información crucial?
        \item \textbf{Accurate (precisión):} ¿La información es fácticamente correcta y está libre de alucinaciones? ¿Las citas y la doctrina están correctamente representadas?
        \item \textbf{Utilidad:} ¿La respuesta está bien estructurada, es fácil de entender y responde directamente a la consulta del usuario? ¿Acelera o dificulta el trabajo del profesional?
    \end{itemize}
    
    \item \textbf{Fiabilidad Metodológica:} Para validar la consistencia de las evaluaciones, se calculó el acuerdo inter-anotador. Se obtuvo una puntuación \textbf{Kappa de Cohen de $\kappa = 0.85$}, lo que indica un grado de acuerdo "casi perfecto" entre los juristas y confirma la robustez de los datos presentados. Las puntuaciones mostradas en el gráfico representan el promedio de las calificaciones de ambos anotadores.
\end{itemize}

\section{Retrieval-Augmented Generation (RAG) como paradigma dominante para la mitigación de alucinaciones legales}
\label{sec:rag_mitigacion}

Frente a la inherente propensión de los Grandes Modelos de Lenguaje (LLMs) a generar alucinaciones, particularmente en un dominio tan sensible a la factualidad como el derecho, la comunidad de inteligencia artificial (IA) y los desarrolladores de tecnología legal han convergido predominantemente hacia un paradigma específico de mitigación: la Generación Aumentada por Recuperación, o Retrieval-Augmented Generation (RAG).

RAG representa un cambio fundamental respecto a la arquitectura estándar de los LLMs, que operan esencialmente en un modo de "libro cerrado", dependiendo exclusivamente del conocimiento internalizado (y potencialmente defectuoso o desactualizado) durante su entrenamiento masivo. En contraste, RAG busca dotar a los LLMs de un mecanismo de "libro abierto", permitiéndoles consultar activamente fuentes de información externas y relevantes \textit{antes} de generar una respuesta. Esta sección se adentra en los fundamentos teóricos y mecánicos de RAG, evalúa críticamente sus ventajas teóricas específicas para el contexto legal, analiza sus limitaciones inherentes y puntos de fallo (que explican por qué, a pesar de su promesa, la mitigación de alucinaciones no es completa), revisa la evidencia empírica sobre su efectividad y discute las estrategias emergentes para su optimización en aplicaciones jurídicas.

La idoneidad de RAG para el dominio legal puede entenderse a través del \textbf{modelo de argumentación de Toulmin}, un marco fundamental en el razonamiento jurídico. Como señalan revisiones recientes, las tareas de los LLMs pueden mapearse directamente a los componentes de Toulmin (Shao et al., 2025). En esta analogía:
\begin{itemize}
    \item La fase de Recuperación (Retrieval) del RAG se corresponde con la búsqueda de los "Datos" (hechos del caso) y el "Respaldo" (Backing) (estatutos y jurisprudencia aplicable).
    \item La fase de Generación (Generation) del LLM se corresponde con la construcción de la "Garantía" (Warrant) (el principio legal que conecta los hechos con la conclusión) para llegar a una "Reclamación" (Claim) (la conclusión legal).
\end{itemize}

Visto desde esta óptica, RAG no es simplemente un parche técnico contra las alucinaciones; es una arquitectura que computacionalmente imita la estructura fundamental de un argumento jurídico bien formado. Esto explica su predominio en herramientas avanzadas como ChatLaw, que integra RAG con bases de conocimiento estructuradas para fortalecer aún más el "Respaldo" de sus argumentos (Shao et al., 2025).

\begin{figure}[h!]
    \centering
    \begin{tikzpicture}[
        node distance=0.8cm and 1.2cm,
        main_box/.style={
            draw, 
            rectangle, 
            rounded corners=5pt, 
            align=center, 
            text width=4.5cm,
            minimum height=3.2cm,
            font=\sffamily\small,
            drop shadow={opacity=0.3, shadow xshift=2pt, shadow yshift=-2pt}
        },
        section_box/.style={
            draw,
            rectangle,
            rounded corners=3pt,
            fill=gray!15,
            align=center,
            text width=4.3cm,
            font=\sffamily\scriptsize,
            inner sep=4pt
        },
        banner_style/.style={
            draw,
            rectangle,
            fill=green!15,
            align=center,
            text width=14cm,
            font=\sffamily\small\itshape,
            inner sep=5pt,
            rounded corners=5pt
        },
        arrow_style/.style={
            -Stealth,
            thick,
            draw=gray!80
        }
    ]
    
    \node[main_box, fill=red!10] (problem) at (-6, 2) {\textbf{El Fundamento: el Desafío}\\ \textbf{de la veracidad}};
        \node[section_box, below=0.3cm of problem] (sec2) {Sección 2: Naturaleza y causas\\ de las alucinaciones};
        \node[section_box, below=0.2cm of sec2] (sec3) {Sección 3: Evaluación\\ y benchmarking del fenómeno};
        \node[section_box, below=0.2cm of sec3] (sec6) {Sección 6: Casos de estudio\\ y consecuencias reales};

    \node[main_box, fill=green!10] (solution) at (0, 2) {\textbf{La garantía y el respaldo:}\\ \textbf{hacia la fiabilidad técnica}};
        \node[section_box, below=0.3cm of solution] (sec4) {Sección 4: RAG como paradigma\\ de mitigación (la garantía)};
        \node[section_box, below=0.2cm of sec4] (sec5) {Sección 5: Estrategias holísticas\\ de optimización (el respaldo)};

    \node[main_box, fill=purple!10] (outcome) at (6, 2) {\textbf{La conclusión:}\\ \textbf{integración responsable}};
        \node[section_box, below=0.3cm of outcome] (sec7) {Sección 7: El futuro de la IA fiable\\ \small(XAI, auditoría, diseño responsable)};
        \node[section_box, below=0.2cm of sec7] (sec8) {Sección 8: implicaciones eticas\\ y regulatorias (supervisión humana)};

    \draw[arrow_style] (problem.east) -- (solution.west) node[midway, above, font=\sffamily\tiny] {Abordado por};
    \draw[arrow_style] (solution.east) -- (outcome.west) node[midway, above, font=\sffamily\tiny] {Conduce a};

    \node[banner_style, below=1.5cm of sec5] (banner) {Un marco argumentativo para mitigar alucinaciones y construir\\ sistemas de IA jurídica fiables y auditables};

    \end{tikzpicture} %

    \caption{Descomposición de la estructura argumentativa del informe según el modelo de Toulmin. La figura ilustra el flujo lógico: partiendo del \textbf{Fundamento} (izquierda), que establece el problema de las alucinaciones (Secciones 2, 3 y 6); pasando a la \textbf{Garantía y Respaldo} (centro), que presenta la solución técnica con RAG y su optimización (Secciones 4 y 5); para llegar a la \textbf{Conclusión} (derecha), que define el marco para una integración responsable, incluyendo el futuro de la IA y sus implicaciones éticas y regulatorias (Secciones 7 y 8).}
    \label{fig:toulmin_framework_informe}
\end{figure}
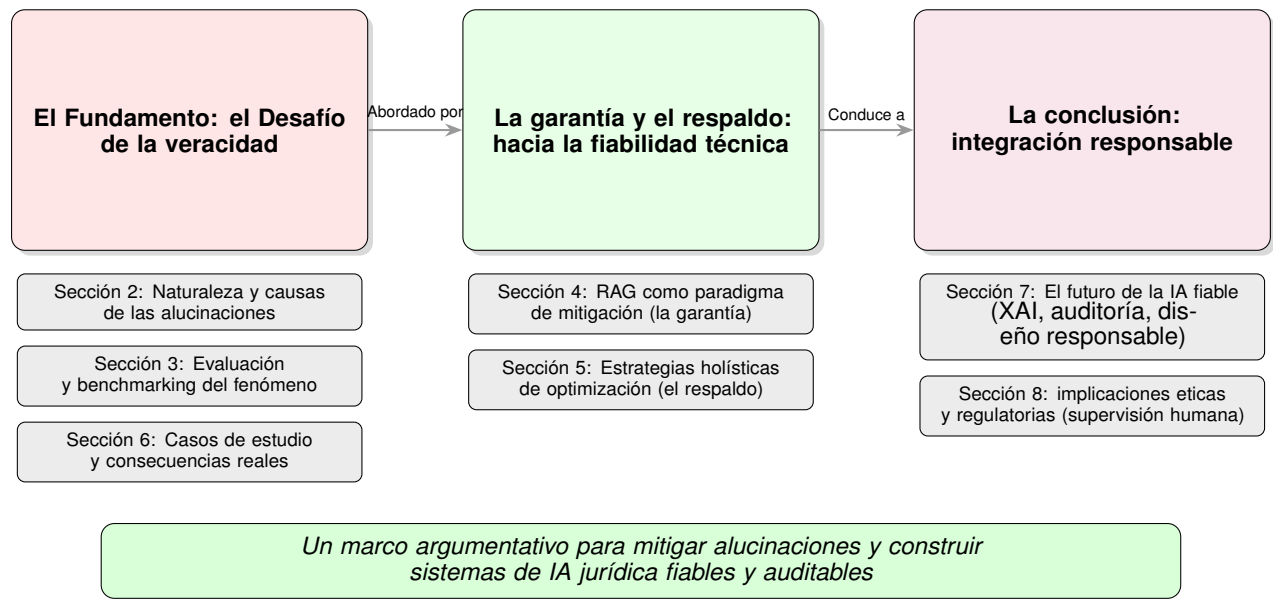

\subsection{Fundamentos teóricos y mecanismo operativo de RAG}

Los Grandes Modelos de Lenguaje (LLMs) base, a pesar de su asombrosa capacidad para generar texto fluido y coherente, operan fundamentalmente como 'cajas negras' con un conocimiento estático. Su proceso de toma de decisiones interno es en gran medida opaco, y el vasto corpus de información con el que fueron entrenados representa una instantánea del pasado, volviéndose progresivamente obsoleto a medida que el mundo –y especialmente el dinámico campo del derecho– evoluciona. Esta naturaleza intrínseca los hace inherentemente propensos a generar un espectro de errores factuales, englobados bajo el término "alucinación". Como se detalló en la \textbf{taxonomía de la Sección 2.2}, estos errores van desde la fabricación completa de autoridades hasta sutiles tergiversaciones de fuentes existentes (misgrounding), un desafío que la arquitectura RAG busca mitigar de raíz.. Es precisamente para abordar estas limitaciones fundamentales –la opacidad, el conocimiento estático y la consiguiente falta de fundamentación verificable– que emerge la Generación Aumentada por Recuperación (RAG) como un cambio paradigmático. RAG no busca simplemente hacer que el LLM sea 'más inteligente' en abstracto, sino que lo transforma conceptualmente de un generador de lenguaje aislado a un sistema que interactúa dinámicamente con fuentes de conocimiento externas y explícitas, buscando anclar cada respuesta en evidencia recuperable y, por ende, potencialmente más fiable y actualizada.

A continuación, exploraremos los fundamentos teóricos y el mecanismo operativo de esta arquitectura crucial. \textbf{Lejos de ser un mero parche técnico, RAG se postula} como una reconfiguración conceptual de cómo los LLMs interactúan con el conocimiento. Aborda directamente el problema de la 'caja negra' al externalizar la base de conocimiento a un corpus explícito y potencialmente verificable, y combate el problema del conocimiento estático al permitir que este corpus externo sea actualizado dinámicamente, independientemente de los costosos ciclos de reentrenamiento del modelo de lenguaje subyacente. Es esta doble promesa de fundamentación y actualidad la que ha posicionado a RAG como la principal esperanza para una IA legal más fiable.

El concepto central de RAG es simple pero poderoso: desacoplar el proceso de generación de lenguaje del almacenamiento de conocimiento fáctico masivo. En lugar de exigir que el LLM memorice y razone sobre la totalidad del corpus legal dentro de sus parámetros (una tarea propensa a la compresión con pérdidas y a la alucinación), RAG externaliza la base de conocimiento a un corpus documental explícito y recuperable (p. ej., bases de datos de jurisprudencia, estatutos, regulaciones, tratados legales, o incluso documentos internos de un bufete). 

Este estudio no está dedicado a detallar el funcionamiento de RAG ya que existen innumerables ensayos por Internet sobre el tema, pero repasaremos de manera introductoria el proceso RAG canónico, que implica dos fases principales:

\begin{enumerate}
    \item \textbf{Fase de recuperación (Retrieval):} dada una consulta del usuario (el \textit{prompt}), esta fase tiene como objetivo identificar y extraer los fragmentos de información más relevantes del corpus documental externo. Este proceso típicamente involucra:

    \begin{itemize}
        \item \textit{Indexación:} pre-procesamiento del corpus documental, dividiéndolo en unidades manejables (chunks) y generando representaciones vectoriales (embeddings) para cada chunk mediante un modelo de embedding genéricos (p. ej., \textit{text-embedding-ada-002} de OpenAI o modelos específicos de dominio como \textit{bge-m3-spa-law-qa-large} de LittleJohn). Estos embeddings capturan el significado semántico de los chunks.

        \item \textit{Almacenamiento vectorial:} guardar los embeddings en una base de datos vectorial optimizada para búsquedas de similitud (p. ej., FAISS, Qdrant, Chroma…).
        \item \textit{Procesamiento de la consulta:} la consulta del usuario también se convierte en un embedding vectorial usando el mismo modelo.
        \item \textit{Búsqueda de similitud:} se realiza una búsqueda (típicamente por similitud coseno o distancia euclidiana) en la base de datos vectorial para encontrar los \textit{k} chunks cuyos embeddings son más cercanos al embedding de la consulta.
        \item \textit{Recuperación híbrida (opcional pero común):} a menudo, la búsqueda semántica se combina con métodos tradicionales de recuperación de información basados en palabras clave (p. ej., BM25) para mejorar la precisión, especialmente para términos específicos o nombres propios.
        \item \textit{Re-ranking (opcional):} los chunks recuperados pueden ser reordenados usando modelos más sofisticados (cross-encoders) que evalúan la relevancia de cada chunk en relación con la consulta completa, aunque esto añade latencia.

    \end{itemize}
    \item \textbf{Fase de generación (Generation):} Los \textit{k} chunks de texto recuperados, considerados los más relevantes para la consulta, se utilizan para "aumentar" el prompt original del usuario. Este prompt aumentado (que ahora contiene la consulta y el contexto recuperado) se introduce en el LLM generador (p. ej., GPT, Claude, Llama, Gemini...). El LLM tiene la instrucción de basar su respuesta principalmente en la información contextual proporcionada, sintetizándola y presentándola de manera coherente y relevante a la pregunta original. Idealmente, el LLM también debería ser capaz de citar las fuentes específicas de los chunks recuperados de donde extrajo la información.
\end{enumerate}

\begin{figure}[h!]
    \centering
    \includegraphics[width=\textwidth]{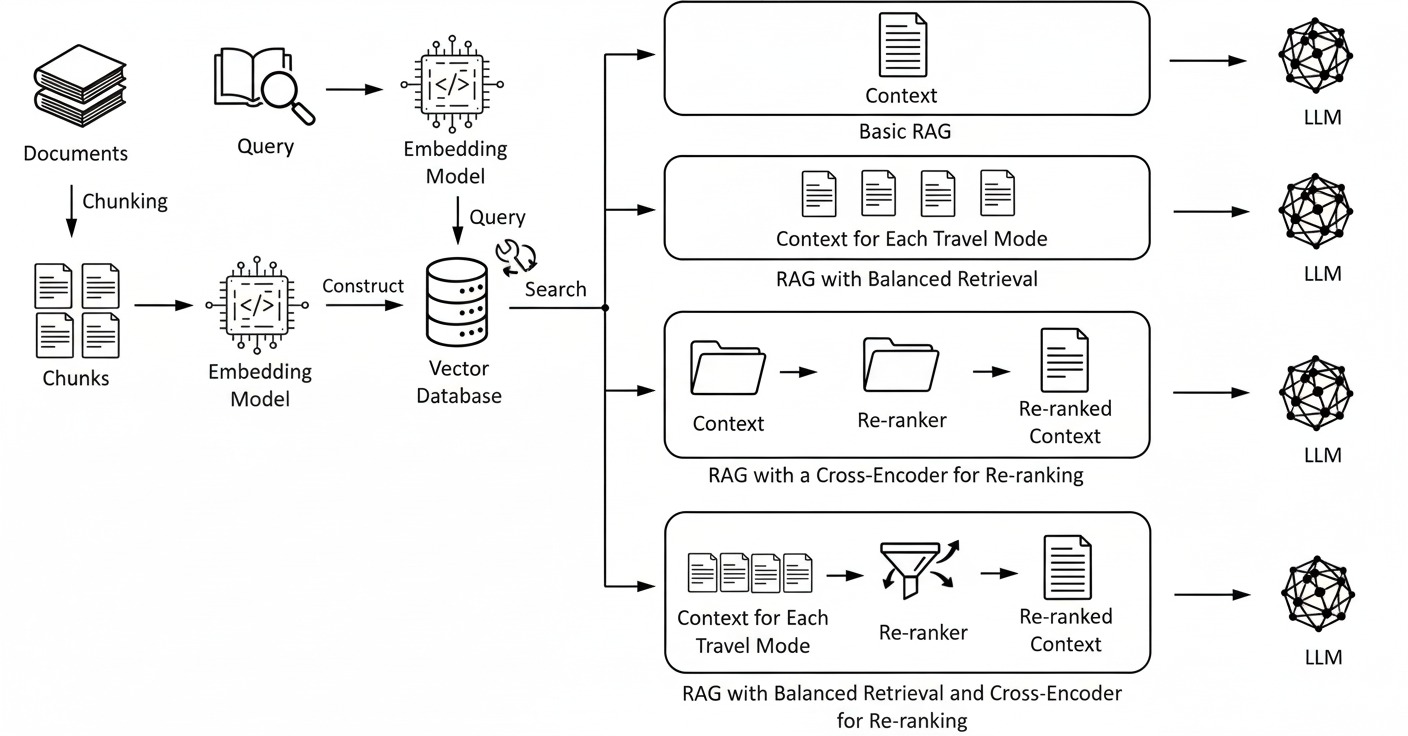}
    \caption{
        Diagrama esquemático de un sistema de generación aumentada por recuperación (RAG). 
        El flujo ilustra cómo una consulta de usuario (Input Query) se enriquece con 
        información contextual recuperada de una base de conocimiento externa (Vectorstore) 
        antes de ser enviada al LLM. Este proceso ancla la respuesta en datos verificables, 
        mitigando la generación de alucinaciones. Adaptado de Yiming Xu et al. (2025).
    }
    \label{fig:rag_architecture}
\end{figure}

El diseño de este proceso de dos etapas persigue que el LLM genere respuestas más precisas, actualizadas y fundamentadas, mitigando la necesidad de "inventar" información cuando su conocimiento paramétrico es insuficiente o incorrecto.

Un ejemplo práctico de esta arquitectura es el modelo "Legal Assist AI", diseñado para un sistema judicial específico con un corpus de datos curado. En su implementación, los documentos legales se cargan y se dividen en fragmentos manejables (chunks) de 1000 caracteres. A continuación, se generan representaciones vectoriales (embeddings) para cada chunk utilizando el modelo sentence-transformers/all-MiniLM-L6-v2 a través de HuggingFace. Finalmente, estos embeddings se indexan en una base de datos vectorial FAISS (Facebook AI Similarity Search), que permite una recuperación ultrarrápida de los fragmentos de texto semánticamente más relevantes para la consulta del usuario, los cuales son inyectados en el prompt del LLM generador (Gupta et al., 2025). Este flujo de trabajo ilustra el mecanismo RAG canónico en una aplicación legal del mundo real.

\subsection{Ventajas teóricas de RAG: fundamentación, actualidad y transparencia}

La adopción generalizada de la Generación Aumentada por Recuperación (RAG) como arquitectura preferente para los Grandes Modelos de Lenguaje (LLMs) en aplicaciones sensibles a la factualidad, y muy particularmente en el dominio legal, no es casual. Desde su diseño fundamental, RAG ofrece una serie de ventajas para abordar algunas de las limitaciones más críticas de los LLMs base cuando operan de forma aislada. Estas ventajas, si se materializan plenamente, tienen el potencial de transformar la IA de una herramienta generativa de lenguaje, a menudo desconectada de la realidad fáctica, en un asistente cognitivo genuinamente útil y más fiable para el profesional del derecho. La promesa teórica de RAG se asienta sobre tres pilares principales: la capacidad de fundamentar las respuestas en autoridad externa, la habilidad para operar con información dinámica y actualizada, y el potencial para una mayor transparencia y verificabilidad del proceso generativo.

El \textbf{principio de fundamentación (grounding) en autoridad externa} es, quizás, la ventaja más publicitada y esencial de RAG en el contexto jurídico. El derecho, por su propia naturaleza, es un sistema normativo y argumentativo que se construye sobre un vasto y jerarquizado corpus de fuentes autorizadas: constituciones, estatutos, regulaciones administrativas, jurisprudencia vinculante y persuasiva, y tratados doctrinales. Un LLM base, que depende únicamente de los patrones estadísticos internalizados durante su entrenamiento a partir de un corpus general (que puede o no incluir una representación adecuada de estas fuentes), opera esencialmente en un vacío autoritativo. Puede generar texto que \textit{imita} el estilo del lenguaje legal, pero carece de un anclaje directo y verificable en las fuentes que definen el derecho aplicable. El paradigma de RAG aborda este problema obligando al LLM a interactuar con un corpus documental explícito de estas fuentes legales. Antes de generar una respuesta a una consulta jurídica, el sistema RAG primero recupera los fragmentos de texto más relevantes de este corpus. Esta información recuperada se convierte en el "fundamento" sobre el cual el LLM debe construir su respuesta. En un escenario ideal, esto significa que las afirmaciones legales, las interpretaciones y las conclusiones generadas por el sistema no son meras invenciones probabilísticas, sino que están directamente derivadas y soportadas por el texto de la ley, el precedente o el contrato pertinente. Para un abogado, esto es crucial, ya que cualquier argumento o consejo debe, en última instancia, ser rastreable hasta una fuente de autoridad válida.

La segunda ventaja fundamental de RAG es su capacidad inherente para \textbf{manejar información legal dinámica y asegurar la actualidad del conocimiento}. El derecho es un organismo vivo; las leyes se enmiendan, se derogan y se promulgan nuevas. Los tribunales emiten nuevas sentencias que reinterpretan, modifican o incluso revocan precedentes establecidos. Un LLM base, entrenado en un "snapshot" del pasado, se vuelve inevitablemente obsoleto a medida que el derecho evoluciona. Reentrenar estos modelos masivos desde cero o incluso actualizarlos de manera significativa es un proceso costoso, complejo y que consume mucho tiempo, lo que hace inviable mantenerlos perpetuamente al día con los cambios legislativos y jurisprudenciales. RAG ofrece una solución elegantemente simple a este problema de obsolescencia: dado que el conocimiento fáctico primario reside en la base de datos documental externa y no en los parámetros del LLM, \textbf{la actualidad del sistema RAG depende principalmente de la actualidad de dicha base de datos}. Mantener actualizado un corpus documental específico (p. ej., añadiendo nuevas leyes, sentencias recientes, o actualizando el estado de derogación de los precedentes) es una tarea considerablemente más manejable y menos costosa que reentrenar un LLM de miles de millones de parámetros. En teoría, un sistema RAG bien mantenido podría proporcionar acceso a la información legal más reciente, permitiendo a los profesionales confiar en que las respuestas de la IA reflejan el estado actual del derecho, un requisito indispensable para la práctica competente.

Finalmente, RAG ofrece un \textbf{potencial significativo para una mayor transparencia y verificabilidad} en comparación con la naturaleza opaca de los LLMs base. Una de las críticas más persistentes a los LLMs es su funcionamiento como "cajas negras": generan respuestas, a menudo convincentes, pero sin ofrecer una explicación clara de \textit{cómo} llegaron a esa conclusión o en qué información específica se basaron. En el ámbito legal, donde la capacidad de justificar un argumento y citar las fuentes es fundamental, esta opacidad es inaceptable. RAG, al basar explícitamente la generación en documentos recuperados, abre la puerta a una mayor transparencia. Idealmente, un sistema RAG no solo debería proporcionar una respuesta, sino también \textbf{citar las fuentes específicas de su corpus externo que fueron utilizadas para construir cada parte de esa respuesta}. Esto permitiría al profesional legal no solo recibir una conclusión, sino también revisar y evaluar críticamente la evidencia documental subyacente, juzgar la relevancia y la interpretación de las fuentes por sí mismo, y, en última instancia, asumir la responsabilidad informada por el uso de la salida de la IA. Esta capacidad de "mostrar el trabajo" es crucial para integrar la IA de manera responsable en los flujos de trabajo legales, donde la verificación humana sigue siendo un componente irreductible de la diligencia profesional.

Esta capacidad de 'mostrar el trabajo' es fundamental en un dominio como el jurídico, donde la 'respuesta correcta' raramente es un dato aislado o binario (verdadero/falso). Por el contrario, la validez de una conclusión legal reside en la solidez de su fundamentación y en la coherencia de su razonamiento. Al proporcionar acceso directo y trazable a las fuentes, RAG permite al profesional no solo validar la información, sino, lo que es más importante, analizar la interpretación propuesta por el modelo, evaluar la lógica de su argumentación y, en última instancia, construir su propio criterio experto. La verdadera asistencia de la IA no reside en ofrecer una conclusión, sino en articular de forma transparente los fundamentos que la sustentan, convirtiéndose en una herramienta para amplificar el juicio humano, no para sustituirlo.

En resumen, la arquitectura RAG, desde una perspectiva teórica, está diseñada para abordar de frente algunas de las deficiencias más críticas de los LLMs cuando se enfrentan a tareas legales sensibles a la factualidad. Promete respuestas más fundamentadas, actualizadas y verificables, moviendo a la IA legal un paso más cerca de ser un asistente cognitivo verdaderamente útil y fiable. Sin embargo, como se explorará en la siguiente sección, la transición de esta promesa teórica a una implementación práctica robusta y consistentemente fiable en el complejo y adversario dominio del derecho está plagada de desafíos significativos y puntos de fallo inherentes que explican por qué las alucinaciones, aunque mitigadas, persisten.

\subsection{El talón de Aquiles de RAG: Análisis de limitaciones y evidencia empírica}

A pesar de las considerables ventajas conceptuales que la Generación Aumentada por Recuperación (RAG) aporta a la tarea de fundamentar los Grandes Modelos de Lenguaje (LLMs) en conocimiento externo, tanto la evidencia empírica emergente como un análisis profundo de su mecanismo operativo revelan que esta arquitectura, aunque un avance significativo, no constituye una solución infalible. Lejos de ser una panacea, la promesa de respuestas consistentemente precisas, actuales y verificables se ve atenuada por una serie de limitaciones persistentes y puntos de fallo inherentes a sus dos fases operativas clave: la recuperación de información y la generación de lenguaje.

Desde una perspectiva de desarrollo de producto, un sistema RAG canónico debe ser tratado como lo que realmente es: un prototipo, y no una solución de producción robusta. La facilidad con la que herramientas como LangChain permiten ensamblar un prototipo RAG ha creado una falsa sensación de simplicidad. Confundir un prototipo funcional con un sistema fiable es el camino más rápido al desastre técnico y a la pérdida de confianza del cliente. La construcción de un sistema RAG serio no es un sprint de fin de semana; es un maratón de ingeniería de datos y refinamiento continuo.

Estos desafíos explican por qué incluso las herramientas RAG más sofisticadas siguen produciendo errores, desde inexactitudes sutiles hasta alucinaciones manifiestas que comprometen su fiabilidad. Esta sección analiza en profundidad estos puntos de fallo, contrastando las debilidades teóricas con los hallazgos empíricos más recientes.

\subsubsection{Puntos de fallo en la fase de recuperación: el desafío del fundamento relevante}

El primer y quizás más fundamental conjunto de vulnerabilidades reside en la fase de recuperación de información. La máxima de \textit{"garbage in, garbage out"} aplica con toda su fuerza: si el sistema RAG no logra identificar y extraer los fragmentos de texto (\textit{chunks}) verdaderamente relevantes, precisos y autorizados, el LLM generador, por muy avanzado que sea, operará sobre una base informativa defectuosa, incrementando drásticamente la probabilidad de generar una respuesta errónea.

\begin{itemize}
    \item \textbf{Ambigüedad Inherente a la "Relevancia Legal"}: A diferencia de la recuperación de hechos discretos, determinar qué pasaje es \textit{legalmente relevante} exige un sofisticado razonamiento jurídico. La simple similitud semántica superficial, en la que se basan muchos sistemas de búsqueda vectorial, puede ser engañosa. Un fragmento puede ser temáticamente similar pero provenir de una jurisdicción inaplicable o referirse a un estatuto derogado. Esta debilidad se ve confirmada en contextos jurisdiccionales con datos limitados. Un estudio sobre la práctica legal en una jurisdicción subrepresentada en los corpus de entrenamiento demostró que la incapacidad de los LLMs para realizar investigación jurídica fiable se debía a la escasez de jurisprudencia india en sus datos de entrenamiento. Esto evidencia un punto de fallo fundamental para RAG: aunque la arquitectura esté diseñada para recuperar información, si el corpus de recuperación carece de la información relevante, el LLM generador se ve forzado a operar sobre una base incompleta, lo que conduce directamente a la alucinación.
    \item \textbf{Deficiencias en las Estrategias de \textit{Chunking}}: La forma en que los documentos legales extensos se dividen en fragmentos manejables es crítica. Un \textit{chunking} deficiente puede llevar a la pérdida de contexto esencial, la introducción de ruido irrelevante o la fragmentación de unidades lógicas [Pinecone, 2024]. La importancia de una estrategia sofisticada se ilustra en el informe interno de Addleshaw Goddard (2024), que para optimizar una tarea de \textit{due diligence}, tuvo que experimentar meticulosamente hasta concluir que fragmentos de 3,500 caracteres con un solapamiento de 700 eran óptimos para su corpus. Esto sugiere que las implementaciones genéricas de RAG probablemente exhiban tasas de error considerablemente más altas.
    \item \textbf{Recuperación Incompleta y Calidad de la Base de Conocimiento}: Incluso con estrategias de \textit{chunking} mejoradas, el módulo de recuperación puede no identificar \textit{todos} los fragmentos necesarios o puede priorizar incorrectamente los menos relevantes. Además, la base de datos documental debe ser exhaustiva, precisa y estar meticulosamente actualizada. Cualquier error, omisión o desactualización en el corpus subyacente se propagará inevitablemente a las respuestas generadas.
\end{itemize}

\subsubsection{Puntos de fallo en la fase de generación: la tensión entre fidelidad y fluidez}

Superados los desafíos de la recuperación, la fase de generación de lenguaje presenta su propio conjunto de puntos de fallo, incluso cuando se proporciona al LLM un contexto aparentemente correcto. La evidencia empírica es crucial aquí, pues revela que los errores más comunes y peligrosos no son las fabricaciones completas, sino formas más sutiles.

\begin{itemize}
    \item \textbf{Falta de Fidelidad al Contexto Recuperado y Errores "Insidiosos"}: A pesar de las instrucciones, el LLM generador puede ignorar o contradecir el contexto recuperado, recurriendo a su conocimiento paramétrico [Chen et al., 2024], o intentar "rellenar los huecos" inventando detalles no soportados explícitamente. Un hallazgo crucial del estudio de Magesh et al. (2024) es que las alucinaciones en herramientas RAG rara vez son fabricaciones completas de casos. Más comúnmente, adoptan formas más sutiles y peligrosas como el \textit{\textbf{misgrounding}}: citar un caso o estatuto real pero tergiversar su contenido o aplicarlo incorrectamente. Este tipo de error es particularmente "insidioso" porque crea una falsa sensación de fiabilidad, dificultando su detección por parte de un profesional que no realice una verificación profunda de cada fuente.
    \item \textbf{Errores de Síntesis e Inferencia}: Cuando la respuesta requiere la integración de información de múltiples \textit{chunks}, el LLM puede cometer errores lógicos o realizar inferencias inválidas. Benchmarks específicos para sistemas RAG, como \textbf{LibreEval de Arize AI (un conjunto de datos diseñado para evaluar la fidelidad de las respuestas al contexto proporcionado)}, han mostrado que los 'Relational-errors', que surgen de una síntesis defectuosa, son una forma común de alucinación en sistemas RAG.
    \item \textbf{Dependencia de la Naturaleza de la Tarea Legal}: La efectividad de RAG varía significativamente según la tarea. La extracción de cláusulas estandarizadas como "Governing Law" puede alcanzar altos niveles de precisión con RAG optimizado. Sin embargo, cláusulas más variables y contextualmente dependientes como "Exclusivity" o "Cap on Liability" presentan un mayor desafío y requieren una optimización más intensiva para alcanzar niveles similares de precisión [Addleshaw Goddard, 2024].
    \item \textbf{Dificultades en la Atribución y Citación Precisa}: Una manifestación común de la imperfección de RAG es la incapacidad del LLM para generar citas precisas que vinculen inequívocamente sus afirmaciones a los pasajes específicos de los documentos recuperados. Esta falta de atribución fiable socava uno de sus principales beneficios teóricos: la verificabilidad.
\end{itemize}

\subsubsection{Síntesis de la evidencia: un mitigador imperfecto}

En conclusión, la evidencia empírica actual converge en un punto claro: RAG es una herramienta valiosa que indudablemente mitiga la propensión de los LLMs a las alucinaciones factuales extrínsecas. Sin embargo, está lejos de ser una solución mágica que elimina el riesgo por completo.

Los estudios pioneros sobre herramientas comerciales, como el de Magesh et al. (2024), proporcionan datos cruciales. Sus hallazgos revelan una persistencia preocupante de errores: documentaron que las principales plataformas generaban respuestas incorrectas o con fundamentación errónea (\textit{misgrounded}) en un rango de entre el \textbf{17\% y más del 33\%} de las consultas. Aunque esto representa una mejora sustancial sobre las tasas de alucinación de los LLMs base en contextos legales (que pueden superar el 50-80\% según Dahl et al., 2024), sigue siendo un porcentaje inaceptablemente alto para aplicaciones legales críticas.

Esta persistencia de errores en sistemas RAG tiene una explicación teórica fundamental. Cuando la fase de recuperación (Retrieval) falla o proporciona un contexto ambiguo, el modelo generador se enfrenta a una situación de incertidumbre. Dado que su entrenamiento subyacente lo condiciona a evitar la abstención a toda costa, su comportamiento por defecto es "rellenar los huecos" de la forma más coherente posible, recurriendo a su conocimiento interno. Esto provoca los errores de misgrounding que observamos en la práctica, donde el modelo falla no por falta de contexto, sino por su incapacidad estructural para gestionar la incertidumbre de ese contexto (Kalai et al., 2025).

La promesa de una IA legal completamente "libre de alucinaciones" gracias a RAG sigue siendo, en el estado actual de la tecnología, más una aspiración que una realidad consumada. Su adopción debe ir acompañada de un entendimiento realista de sus limitaciones y un compromiso inquebrantable con la verificación humana diligente y el desarrollo continuo de estrategias de mitigación más robustas, como se explorará en la siguiente sección.

\subsection{Estrategias avanzadas y holísticas para la optimización de RAG en el contexto legal}

La Generación Aumentada por Recuperación (RAG), como se ha analizado previamente, representa un avance conceptual significativo sobre los Grandes Modelos de Lenguaje (LLMs) base, al intentar fundamentar sus respuestas en conocimiento fáctico externo y específico. Sin embargo, la evidencia empírica y el análisis de sus puntos de fallo intrínsecos (Sección 4.3 y 4.4) demuestran con claridad que la implementación canónica de RAG, aunque reduce la incidencia de alucinaciones extrínsecas, está lejos de ser una solución infalible en el exigente y matizado dominio legal. Los desafíos inherentes a la recuperación precisa de información jurídica relevante dentro de corpus masivos y a menudo ambiguos, junto con la propensión residual del LLM generador a desviarse del contexto recuperado o a sintetizarlo incorrectamente, subrayan la necesidad imperante de adoptar estrategias de optimización mucho más sofisticadas y holísticas.

Estas estrategias no se limitan a meros ajustes paramétricos, sino que implican un rediseño y refinamiento profundo de cada componente del ciclo RAG, así como la integración de técnicas complementarias y una comprensión profunda de la interacción entre el conocimiento legal y las capacidades algorítmicas.

Esta sección se adentra en estas metodologías avanzadas, detallando enfoques específicos para la optimización robusta de la recuperación de información, el refinamiento de la fase de generación y razonamiento, y la crucial implementación de arquitecturas integradas que fomenten una sinergia efectiva entre estos componentes, siempre con el objetivo de maximizar la fiabilidad y minimizar el riesgo de alucinación en aplicaciones legales críticas.

\subsubsection{Optimización crítica de la fase de recuperación (Retrieval): la calidad del fundamento}

La premisa fundamental de RAG es que una base de conocimiento precisa y relevante es el cimiento indispensable para una generación fiable. Por lo tanto, cualquier esfuerzo serio por mejorar la calidad de los sistemas RAG legales debe comenzar con una optimización exhaustiva de la fase de recuperación. No basta con recuperar documentos semánticamente similares; la recuperación debe ser legalmente pertinente, contextualmente adecuada y exhaustiva pero concisa. Las estrategias avanzadas en esta área se centran en ir más allá de las implementaciones ingenuas de búsqueda vectorial y en incorporar una comprensión más profunda de la estructura y la semántica del conocimiento legal.

La optimización de la recuperación va más allá de la mera relevancia semántica; implica seleccionar el \textbf{nivel de abstracción correcto} del conocimiento legal a proporcionar. Un estudio sobre la detección del discurso de odio en el derecho alemán reveló que el rendimiento del LLM no siempre mejoraba al proporcionarle más contexto. De hecho, modelos condicionados solo con el título de una norma a menudo superaban a los mismos modelos a los que se les proporcionaba el texto legal completo y complejo (Ludwig et al., 2025). Por el contrario, el rendimiento mejoraba significativamente cuando el contexto incluía \textbf{definiciones concretas y ejemplos extraídos de la jurisprudencia}. La implicación para los sistemas RAG es profunda: un recuperador (retriever) óptimo no debe simplemente encontrar el estatuto relevante, sino que debe ser capaz de identificar y extraer de él las definiciones operativas y los ejemplos de casos que son directamente aplicables, ya que este conocimiento concreto es mucho más "digerible" y útil para el LLM generador que el texto legal abstracto en bruto.

\begin{itemize}
    \item Chunking semántico, estructural y adaptativo: La simple división de documentos en fragmentos de tamaño fijo (\textit{fixed-size chunking}) es a menudo subóptima para textos legales complejos, que poseen una estructura lógica y jerárquica intrínseca (contratos con secciones, cláusulas y sub-cláusulas; sentencias con hechos, razonamiento y holding; estatutos con artículos y apartados).
    \begin{itemize}
        \item \textit{Chunking consciente de la estructura:} Se deben explorar e implementar técnicas que dividan los documentos respetando estos límites estructurales. Por ejemplo, en un contrato, cada cláusula o sub-cláusula podría constituir un \textit{chunk} individual, preservando su integridad semántica. Esto puede requerir el uso de analizadores sintácticos (parsers) específicos del dominio o expresiones regulares robustas para identificar estos límites estructurales (Pinecone, 2024; Addleshaw Goddard, 2024).
        \item \textit{Chunking semántico avanzado:} Más allá de la estructura, se pueden utilizar LLMs más pequeños o modelos de segmentación de texto entrenados para identificar fragmentos que representen unidades de significado coherentes y autocontenidas, incluso si cruzan límites estructurales formales, o para agrupar párrafos temáticamente relacionados.
        \item \textit{Chunking recursivo y jerárquico:} Se pueden generar múltiples niveles de \textit{chunks} para un mismo documento: \textit{chunks} pequeños y muy específicos para la recuperación de hechos puntuales, y \textit{chunks} más grandes que capturen el contexto general de una sección o argumento. El sistema podría entonces seleccionar dinámicamente la granularidad de los \textit{chunks} a recuperar en función de la naturaleza de la consulta.
        \item \textit{Solapamiento estratégico (Overlap):} Un solapamiento cuidadosamente calibrado entre \textit{chunks} adyacentes sigue siendo crucial para evitar la pérdida de contexto en los límites de los fragmentos, pero su tamaño óptimo puede depender del tipo de documento y la estrategia de \textit{chunking}.
    \end{itemize}
    \item Modelos de embedding legales y estrategias multi-vectoriales: La calidad de la representación vectorial (\textit{embedding}) de los \textit{chunks} y de la consulta es fundamental para la búsqueda semántica.
    \begin{itemize}
        \item \textit{Embeddings especializados del dominio legal:} El uso de modelos de embedding pre-entrenados o fine-tuneados específicamente en grandes corpus de textos legales (como LegalBERT, bge-m3-spa-law-qa-large, o los modelos desarrollados a partir de The Pile of Law de Henderson et al., 2022) es preferible a los embeddings de propósito general, ya que pueden capturar mejor los matices semánticos y la terminología específica del derecho.
        \item \textit{Representaciones multi-vectoriales:} En lugar de un único vector por \textit{chunk}, se podrían generar múltiples vectores que capturen diferentes aspectos del texto: uno para la semántica general, otro para entidades legales clave (tribunales, leyes, partes), otro para conceptos jurídicos abstractos, etc. Esto permitiría búsquedas más matizadas y multifacéticas.
    \end{itemize}
    \item Técnicas de búsqueda híbrida y refinamiento de consultas (Query Refinement): La combinación de diferentes paradigmas de búsqueda y el pre-procesamiento inteligente de la consulta del usuario son clave.
    \begin{itemize}
        \item \textit{Ponderación optimizada en búsqueda híbrida:} La integración de la búsqueda semántica (vectorial, densa) con la búsqueda tradicional por palabras clave (dispersa, p. ej., BM25) es a menudo superior a cualquiera de los dos métodos por sí solo. La ponderación relativa entre ambos debe ser cuidadosamente ajustada, posiblemente de forma dinámica según la consulta, para equilibrar la captura de significado conceptual con la precisión en la recuperación de términos exactos, nombres propios o citas (Addleshaw Goddard, 2024).
        \item \textit{Query Expansion y transformación inteligente:} Utilizar LLMs (posiblemente un modelo más pequeño y rápido dedicado a esta tarea) para pre-procesar la consulta del usuario: expandiéndola con sinónimos legales relevantes, términos relacionados, o posibles reformulaciones; identificando la intención subyacente; o descomponiendo preguntas complejas y multifacéticas en sub-preguntas más simples y manejables que puedan ser abordadas por recuperaciones separadas y luego sintetizadas (HyDE - Gao et al. 2022).
        \item \textit{Filtrado estricto por metadatos legales:} La recuperación debe ir más allá de la simple similitud textual e incorporar un filtrado riguroso basado en metadatos cruciales como la jurisdicción aplicable, la fecha de la decisión (para evaluar su actualidad y posible derogación), el nivel jerárquico del tribunal emisor y el tipo de documento. Esto es esencial para asegurar la relevancia legal de los resultados recuperados (Magesh et al., 2024).
    \end{itemize}
    \item Mecanismos de recuperación iterativa, auto-correctora y basada en agentes: Inspirándose en cómo los humanos realizan la investigación legal, los sistemas RAG pueden beneficiarse de arquitecturas más dinámicas e iterativas.
    \begin{itemize}
        \item \textit{Self-Correcting/Corrective RAG (CRAG):} Implementar bucles de retroalimentación donde el sistema evalúa la relevancia y calidad de un conjunto inicial de documentos recuperados (posiblemente usando el propio LLM generador o un modelo de evaluación dedicado). Si los documentos se consideran insuficientes o irrelevantes, el sistema puede refinar automáticamente la consulta original, ajustar los parámetros de búsqueda o buscar en fuentes alternativas antes de proceder a la generación (Yan et al., 2024).
        \item \textit{Recuperación multi-salto (Multi-Hop Retrieval):} Para consultas que requieren la síntesis de información de múltiples fuentes o que implican un razonamiento secuencial (p. ej., rastrear la evolución de una doctrina a través de una cadena de precedentes), el sistema puede realizar múltiples "saltos" de recuperación. La información extraída de un primer conjunto de documentos recuperados se utiliza para formular nuevas consultas y recuperar un segundo conjunto de documentos, y así sucesivamente, hasta que se haya reunido toda la información necesaria (Tang and Yang, 2024).
        \item \textit{Enfoques basados en agentes (Agentic RAG):} Desarrollar agentes de IA que puedan planificar y ejecutar estrategias de recuperación complejas, decidiendo dinámicamente qué fuentes consultar, qué términos de búsqueda utilizar y cómo integrar la información obtenida, imitando más de cerca el proceso de investigación de un experto legal.
    \end{itemize}
\end{itemize}

La inversión en estas estrategias avanzadas de recuperación es fundamental, ya que la calidad del contexto proporcionado al LLM generador es el techo de la calidad de la respuesta final. Una recuperación deficiente o ruidosa inevitablemente conducirá a una generación subóptima o, peor aún, alucinada, independientemente de cuán sofisticado sea el LLM generador.

\begin{table}[htbp]
\centering
\begin{tabular}{l c c c c c c c c c}
\hline
\multirow{3}{*}{Modelo} & \multirow{3}{*}{Tamaño} & \multicolumn{7}{c}{Tareas} & \multirow{3}{*}{Media} \\
\cline{3-9}
& & STS & Retrieval & Clasif. & Cluster. & Rerank. & PairClass. & Resumen & \\
\cline{3-9}
& & 2 & 10 & 8 & 6 & 3 & 3 & 3 & \\
\hline
BOW & - & 0.4917 & 0.2143 & 0.4751 & 0.2612 & 0.7582 & 0.7205 & 0.2635 & 0.4549 \\
\hline
\multicolumn{10}{l}{\textbf{Encoder based Models}} \\
\hline
BERT & 110M & 0.3821 & 0.0231 & 0.5532 & 0.1803 & 0.3968 & 0.7157 & 0.1723 & 0.3462 \\
FinBERT & 110M & 0.4235 & 0.1178 & 0.5961 & 0.2894 & 0.6453 & 0.7021 & 0.2073 & 0.4259 \\
instructor-base & 110M & 0.3791 & 0.5816 & 0.6253 & 0.5362 & 0.9712 & 0.6185 & 0.4372 & 0.5927 \\
bge-large-en-v1.5 & 335M & 0.3435 & 0.6514 & 0.6481 & 0.5768 & 0.9842 & 0.7446 & 0.4911 & 0.6342 \\
AnglE-BERT & 335M & 0.3125 & 0.5784 & 0.6483 & 0.5812 & 0.9673 & 0.6942 & 0.5104 & 0.6132 \\
\hline
\multicolumn{10}{l}{\textbf{LLM-based Models}} \\
\hline
gte-Qwen1.5-7B-instruct & 7B & 0.3792 & 0.6732 & 0.6479 & 0.5887 & 0.9875 & 0.7042 & 0.5408 & 0.6459 \\
Echo & 7B & 0.4408 & 0.6487 & 0.6562 & 0.5823 & 0.9751 & 0.6314 & 0.4781 & 0.6304 \\
bge-en-icl & 7B & 0.3275 & 0.6831 & 0.6604 & 0.5786 & 0.9912 & 0.6782 & 0.5241 & 0.6347 \\
NV-Embed v2 & 7B & 0.3786 & 0.7092 & 0.6432 & 0.6142 & 0.9837 & 0.6098 & 0.5163 & 0.6364 \\
e5-mistral-7b-instruct & 7B & 0.3842 & 0.6783 & 0.6492 & 0.5826 & 0.9863 & 0.7432 & 0.5319 & 0.6508 \\
\hline
\multicolumn{10}{l}{\textbf{Modelos comerciales}} \\
\hline
text-embedding-3-small & - & 0.3298 & 0.6694 & 0.6421 & 0.5847 & 0.9847 & 0.6023 & 0.5138 & 0.6181 \\
text-embedding-3-large & - & 0.3663 & 0.7153 & 0.6631 & \textbf{0.6123} & 0.9921 & 0.7358 & 0.5721 & 0.6653 \\
voyage-3-large & - & 0.4192 & \textbf{0.7509} & 0.6897 & 0.5975 & \textbf{0.9951} & 0.6576 & \textbf{0.6532} & \textbf{0.6805} \\
\hline
\multicolumn{10}{l}{\textbf{Modelos adaptados al sector legal}} \\
\hline
LegalBERT-v1 & 7B & 0.4215 & 0.7087 & 0.7321 & 0.5843 & 0.9892 & 0.7865 & 0.5217 & 0.6777 \\
LegalBERT-v2 & 335M & 0.3857 & 0.6923 & 0.7284 & 0.5726 & 0.9863 & 0.7614 & 0.4973 & 0.6606 \\
BGE-m3-spa-law-qa & 1B & 0.4128 & 0.7256 & 0.7435 & 0.5932 & 0.9907 & 0.7912 & 0.5163 & 0.6819 \\
BGE-m3-spa-law-large & 7B & 0.4387 & 0.7142 & \textbf{0.7612} & 0.5697 & 0.9914 & \textbf{0.8067} & 0.4846 & 0.6809 \\
\hline
\end{tabular}
\caption{Comparación de rendimiento entre diferentes modelos de embeddings en el benchmark FinMTEB. Las métricas de evaluación incluyen similitud semántica textual (STS), recuperación (Retrieval), clasificación (Class.), agrupamiento (Cluster.), reordenamiento (Rerank.), clasificación por pares (PairClass.) y resumen (Summ.). \textbf{Los mejores resultados están en negrita}. El \underline{subrayado} representa el \underline{segundo mejor} rendimiento.}
\label{tab:finmteb}
\end{table}

\subsubsection{Refinamiento de la fase de generación y razonamiento: hacia una IA Legal más fiable y transparente}

La forma en que se instruye al LLM generador es un componente crítico de la arquitectura RAG, no un mero detalle de implementación. Las siguientes directrices no deben entenderse como simples 'consejos', sino como principios de ingeniería de prompts diseñados para restringir el espacio de posibles respuestas del modelo y alinear su comportamiento con los exigentes requisitos de fidelidad y trazabilidad del dominio legal. Se trata de codificar explícitamente en las instrucciones las restricciones operativas que garantizan una generación más fiable.

Una vez que se ha recuperado un conjunto de fragmentos de texto contextualmente relevantes (idealmente optimizado a través de las técnicas de la sección anterior), el desafío se traslada a guiar al LLM generador para que utilice esta información de manera fiel, precisa, lógicamente coherente y transparente. Las estrategias para refinar esta fase son cruciales para minimizar el riesgo de que el LLM ignore el contexto, lo malinterprete, o genere afirmaciones que vayan más allá de lo soportado por las fuentes.

\begin{itemize}
    \item Ingeniería de Prompts avanzada y específica para RAG legal: La forma en que se instruye al LLM generador sobre cómo interactuar con el contexto recuperado es de vital importancia. Los prompts deben ser meticulosamente diseñados para:
    \begin{itemize}
        \item \textit{Enfatizar la fidelidad al contexto (Grounding Instructions):} Incluir instrucciones explícitas y prominentes que ordenen al LLM basar su respuesta \textit{exclusivamente} en la información contenida en los documentos proporcionados y evitar activamente el uso de su conocimiento paramétrico interno o la realización de suposiciones no fundamentadas. Esto se logra mediante directivas inequívocas como: 'Responde únicamente basándote en los siguientes extractos legales. No añadas información que no esté presente en los textos proporcionados'.
        \item \textit{Guías para el razonamiento (Chain-of-Thought, Step-by-Step):} Instruir al modelo para que externalice su proceso de razonamiento, mostrando los pasos lógicos que sigue para llegar a una conclusión a partir del contexto recuperado. Por ejemplo, "Primero, identifica las reglas relevantes en el contexto. Segundo, aplica estas reglas a los hechos de la consulta. Tercero, explica tu conclusión" (Wei et al. 2023). Esto no solo puede mejorar la precisión del razonamiento, sino que también hace que el proceso sea más interpretable y verificable por un humano (Schwarcz et al., 2024).
    \item \textit{Manejo estructurado de la incertidumbre y los conflictos:} Proporcionar al LLM protocolos claros sobre cómo actuar cuando la información recuperada es incompleta, ambigua, o contiene contradicciones. Esto incluye la instrucción explícita de abstenerse de generar una respuesta cuando no se puede formular con un alto grado de confianza basado en las fuentes, en lugar de recurrir a la especulación. Por ejemplo, "Si la información proporcionada no es suficiente para responder completamente, indícalo explícitamente y explica la naturaleza de la información faltante", "Si encuentras información conflictiva en los extractos, presenta ambas perspectivas y señala la discrepancia".
        \item \textit{Instrucciones de citación precisas:} Requerir que el LLM cite de manera específica (idealmente a nivel de fragmento o documento recuperado) las fuentes exactas de las que extrae cada afirmación factual o legal. Esto es esencial para la verificabilidad.
        \item \textit{Persona y formato de salida detallados:} Definir con precisión el rol que debe adoptar el LLM (p. ej., "Actúa como un asistente de investigación legal objetivo y neutral") y el formato exacto de la respuesta esperada (p. ej., estructura del resumen, estilo de citación) para asegurar consistencia y utilidad profesional.
        \item \textit{Prompting "acusatorio" o de refinamiento (Follow-up Prompts):} Como se observó en Addleshaw Goddard (2024), el uso de un segundo prompt que cuestione la completitud o exactitud de la respuesta inicial del LLM, acusándolo sutilmente de haber omitido información o pidiéndole que "revise cuidadosamente de nuevo el contexto por si ha pasado algo por alto", puede estimular al modelo a realizar una segunda pasada más exhaustiva del contexto y mejorar significativamente la calidad de la respuesta.
    \end{itemize}
    \item Fine-Tuning del LLM generador enfocado en la fidelidad legal: Aunque el fine-tuning de LLMs masivos es un proceso intensivo en recursos, puede ofrecer beneficios significativos si se realiza cuidadosamente.
    \begin{itemize}
        \item \textit{Fine-tuning en tareas de Grounding legal:} Adaptar un LLM pre-entrenado utilizando un conjunto de datos de alta calidad compuesto por pares de (contexto legal recuperado, respuesta ideal fielmente fundamentada y correctamente citada). Esto puede entrenar al modelo para que se adhiera más estrictamente al contexto proporcionado y para que genere respuestas en el estilo y formato deseado por la práctica legal (Tian, Mitchell, Yao, et al. 2023).
        \item \textit{Fine-tuning para el razonamiento jurídico sobre contexto:} Desarrollar conjuntos de datos que enseñen al LLM a realizar tipos específicos de razonamiento legal (p. ej., aplicación de reglas, identificación de \textit{holdings}, comparación de casos) \textit{basándose explícitamente} en el contexto recuperado, en lugar de depender de patrones abstractos.
    \end{itemize}
    \item Integración con modelos de razonamiento especializados: La emergencia de LLMs con arquitecturas explícitamente diseñadas para el razonamiento multi-paso, la planificación y la descomposición de problemas (como la familia de modelos "o" de OpenAI - OpenAI 2024) es particularmente relevante para RAG.
    \begin{itemize}
        \item \textit{Planificación de la respuesta:} Estos modelos podrían, en teoría, planificar cómo utilizar la información recuperada de manera más estratégica, identificando qué fragmentos son más relevantes para qué partes de la consulta y cómo sintetizarlos de manera lógicamente coherente.
        \item \textit{Verificación interna de pasos de razonamiento:} Su capacidad para "reflexionar" sobre sus propios pasos de razonamiento intermedios podría permitirles detectar y corregir errores o inconsistencias antes de generar la respuesta final (Schwarcz et al., 2024). La integración de estos modelos de razonamiento como el componente generador en un sistema RAG es un área prometedora para futuras mejoras.
    \end{itemize}
    \item Arquitecturas híbridas (simbólico-neuronales): Aunque aún en etapas tempranas para aplicaciones legales complejas, la combinación de LLMs neuronales con sistemas de razonamiento simbólico basados en reglas (p. ej., lógicas formales, ontologías legales) podría ofrecer una vía para mejorar la consistencia lógica y la verificabilidad de las respuestas generadas a partir del contexto recuperado.
    \item Adaptación de la Complejidad del Lenguaje (La Función "Jerga"): Una IA legal verdaderamente avanzada no solo debe ser precisa, sino también contextualmente consciente del receptor final de la información. La optimización del prompt debe incluir instrucciones para modular la complejidad del lenguaje de la respuesta. Por ejemplo, un sistema podría recibir la directriz: "Genera una respuesta técnica para un abogado y, adicionalmente, una explicación simplificada para un ciudadano sin conocimientos jurídicos". Esta capacidad, que podríamos denominar "función jerga", es un pilar de la humanización de la tecnología legal, reconociendo que la utilidad de una respuesta no reside solo en su corrección, sino en su comprensibilidad. Esto transforma a la IA de un simple motor de búsqueda a un verdadero puente de comunicación entre el complejo mundo legal y la sociedad.
    
\end{itemize}
El objetivo final de estas estrategias de optimización de la generación no es solo producir respuestas que \textit{parezcan} correctas, sino respuestas que sean \textit{demostrablemente} correctas, fieles a las fuentes proporcionadas y útiles para el profesional legal. La capacidad del LLM para explicar \textit{cómo} llegó a una conclusión a partir del contexto recuperado es tan importante como la conclusión misma.

La implementación exitosa de RAG en el dominio legal, por lo tanto, no es simplemente una cuestión de conectar un LLM a una base de datos. Requiere un diseño cuidadoso y una optimización continua de cada etapa del proceso, desde la curación de datos y el chunking, pasando por la sofisticación de los algoritmos de recuperación y la ingeniería de prompts, hasta el refinamiento de la capacidad de razonamiento y generación fiel del LLM. Solo a través de este enfoque holístico y riguroso se podrá comenzar a materializar verdaderamente el potencial de RAG para mitigar las alucinaciones y ofrecer una IA legal genuinamente fiable y valiosa.

\section{Avanzando hacia la fiabilidad: estrategias holísticas para la optimización y mitigación de alucinaciones en la Inteligencia Artificial legal}
\label{sec:estrategias_avanzadas}

La constatación de que ni los Grandes Modelos de Lenguaje (LLMs) base ni las implementaciones canónicas de Generación Aumentada por Recuperación (RAG) logran erradicar por completo el espectro de las alucinaciones en el dominio legal, impone un cambio de paradigma en la forma en que abordamos el desarrollo y la integración de estas tecnologías. Ya no es suficiente aspirar a una solución única o a un "interruptor mágico" que elimine los errores; en su lugar, se requiere un \textbf{enfoque holístico, multifacético y adaptable} que reconozca la complejidad inherente al problema y que implemente una sinergia de estrategias de optimización y mitigación a lo largo de todo el ciclo de vida de la información, desde la curación de los datos hasta la verificación final del resultado generado. Este enfoque no busca la perfección absoluta —una meta teóricamente inalcanzable en sistemas tecnológicos complejos—, sino que adopta un principio de ingeniería robusta: \textbf{la maximización de la fiabilidad y la minimización del riesgo dentro de los límites de lo factible}. Se reconoce que la infalibilidad del 100\% no es una limitación de la tecnología \textit{actual}, sino una característica inherente a la complejidad. Por tanto, el objetivo es construir un sistema cuya fiabilidad sea tan alta, y sus modos de fallo tan predecibles, que la supervisión humana experta se convierta en una capa de validación eficiente y no en una búsqueda onerosa de errores ocultos, siempre bajo la égida indispensable del juicio y la responsabilidad profesional. Esta sección se dedica a explorar en profundidad este arsenal de estrategias avanzadas, que van más allá de los ajustes superficiales para adentrarse en la optimización rigurosa de los datos, el refinamiento de los procesos de razonamiento algorítmico –incluyendo la consideración de sistemas jerárquicos y agentes de IA–, la implementación de mecanismos de verificación cada vez más sofisticados y, de manera crucial, el fortalecimiento del rol del profesional legal como supervisor crítico e informado. La conjunción de estas técnicas no solo busca reducir la frecuencia de las alucinaciones, sino también transformar su naturaleza, haciendo que los errores residuales sean más detectables y menos perjudiciales.

La eficacia de un enfoque holístico, que combina la curación estratégica de datos con la especialización de modelos, ha sido demostrada empíricamente. Un claro ejemplo es el proyecto "Legal Assist AI", que abordó el problema de las alucinaciones en el contexto legal indio. En lugar de utilizar un modelo de propósito general, los investigadores crearon un corpus de datos curado a partir de fuentes legales indias (Constitución, estatutos, jurisprudencia) y lo utilizaron para afinar (fine-tune) un modelo base de 8 mil millones de parámetros (Llama 3.1 8B). El resultado fue un modelo especializado que no solo superó a modelos mucho más grandes como GPT-3.5 Turbo (175 mil millones de parámetros) en tareas de abogacía de la India, sino que, de manera crucial, redujo drásticamente la generación de alucinaciones, proporcionando respuestas fiables donde otros modelos inventaban información (Gupta et al., 2025). Este caso de éxito sirve como un poderoso testimonio de que la mitigación de alucinaciones no reside en la escala del modelo, sino en la calidad de los datos y la especificidad del entrenamiento.

Las estrategias de mitigación más avanzadas convergen en un principio de integración de conocimiento, donde los LLMs no operan de forma aislada, sino como parte de arquitecturas híbridas. Esto incluye la integración con grafos de conocimiento legal estructurados y el uso de arquitecturas de Mixture-of-Experts (MoE), como se implementa en modelos de vanguardia como ChatLaw (Shao et al., 2025). En estos sistemas, módulos expertos especializados dentro del LLM se activan dinámicamente para manejar diferentes tipos de tareas legales, reduciendo las alucinaciones en un 38\% al asegurar que la consulta sea gestionada por el componente con el conocimiento más relevante.

\begin{figure}[htbp]
  \centering
  \begin{tikzpicture}
    \begin{axis}[
        ybar,
        enlargelimits=0.05,
        legend style={at={(0.5,1.15)}, anchor=south, legend columns=-1}, 
        ylabel={Tasa de Éxito del Ataque Adversarial (\%)},
        xlabel={Nivel de Sofisticación de la Estrategia de Ataque/Mitigación},
        symbolic x coords={Zero Shot, Few Shot, SL, RL1, RL2, RL3, RL4},
        xticklabels={Zero Shot, Few Shot, SL, RL (a=0.002), RL (a=0.001), RL (a=0.0005), RL (a=0.0001)},
        xtick=data,
        nodes near coords,
        nodes near coords style={font=\tiny, rotate=90, anchor=west}, 
        ymin=0, ymax=60, 
        bar width=12pt, 
        title style={font=\footnotesize, text width=0.95\textwidth, align=center}, 
        title={Impacto de la sofisticación estratégica en la resiliencia de LLMs \\ 
               \footnotesize (Adaptado de General Analysis, 2025 - Resultados de Red Teaming)}, 
        width=0.95\textwidth,
        height=0.6\textwidth, 
        xticklabel style={rotate=30, anchor=east, font=\footnotesize}
      ]
      
      \addplot+[fill=blue!70] coordinates {
        (Zero Shot,12.9) 
        (Few Shot,13.9) 
        (SL,23.4) 
        (RL1,26.6) 
        (RL2,32.7) 
        (RL3,46.1) 
        (RL4,54.5)
      };
      \legend{Tasa de éxito del ataque (mayor es peor para el LLM Defensor)}
    \end{axis}
  \end{tikzpicture}
  \caption{Esta gráfica ilustra cómo estrategias de ataque más sofisticadas (análogas a la falta de estrategias de mitigación robustas o a vulnerabilidades explotadas) logran una mayor tasa de éxito al inducir fallos en un LLM objetivo (GPT-4o). Demuestra la necesidad de estrategias de defensa (mitigación) igualmente sofisticadas. El eje X representa diferentes algoritmos de ataque del estudio de General Analysis, interpretados aquí como niveles de desafío o sofisticación a los que un sistema RAG legal debe ser resiliente.}
  \label{fig:attack_success_strategies}
\end{figure}
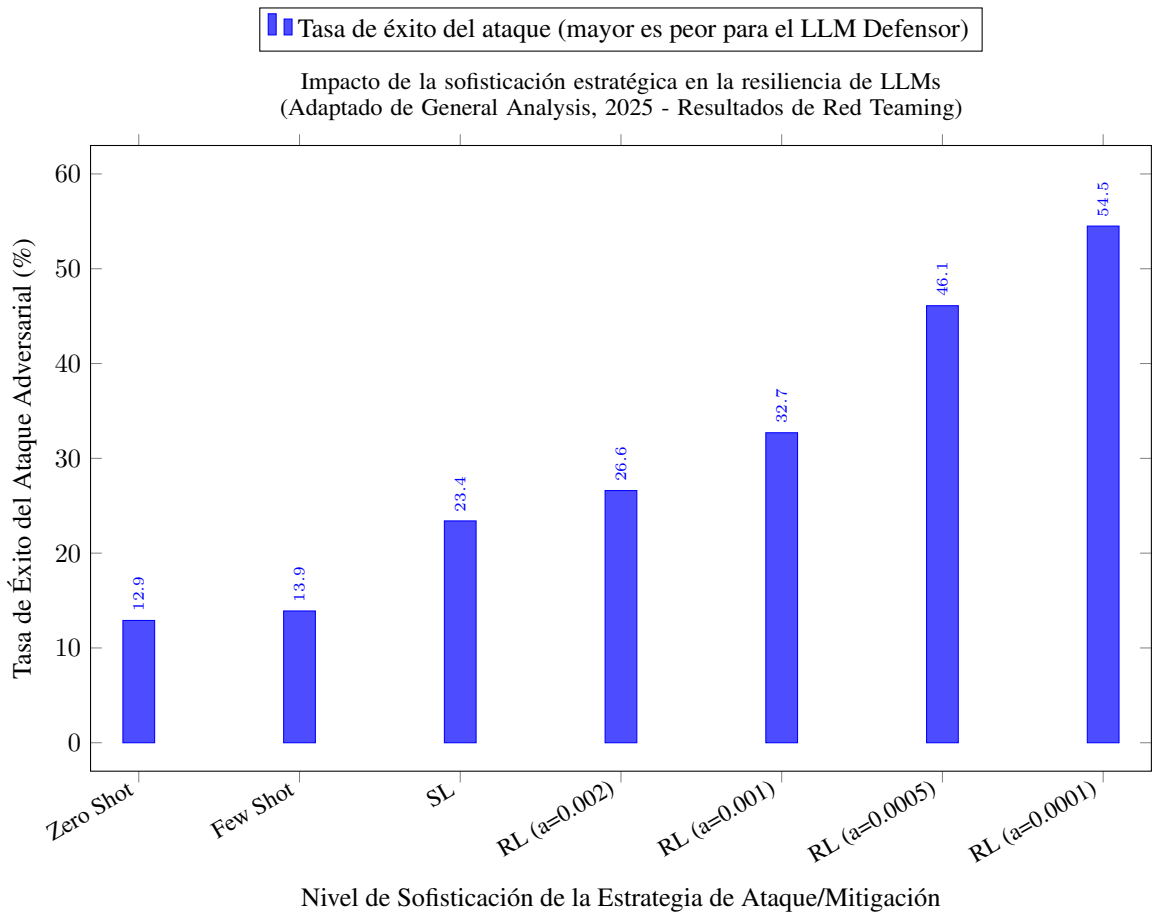

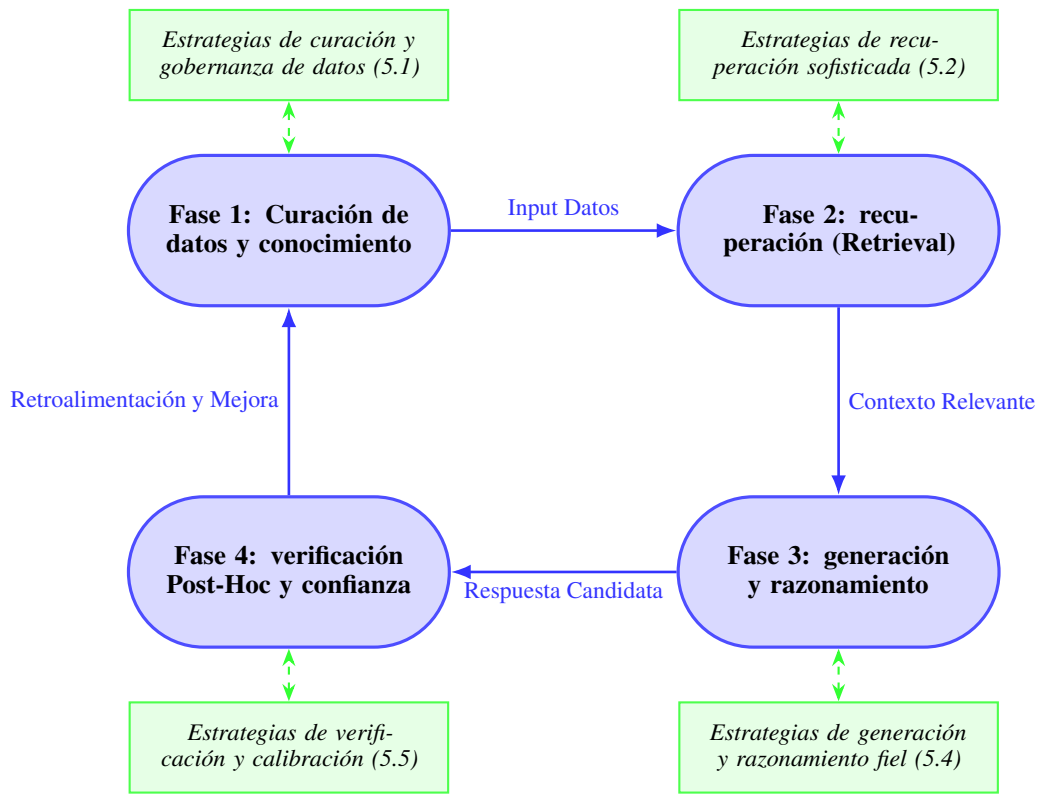
\begin{figure}[htbp]
\centering
\begin{tikzpicture}[
    node distance=2.5cm and 3cm, 
    fase/.style={
        rounded rectangle, 
        draw=blue!70, 
        fill=blue!15, 
        very thick, 
        minimum height=2cm, 
        minimum width=4.5cm, 
        text centered, 
        text width=4cm,
        font=\bfseries 
    },
    estrategia_cat/.style={ 
        rectangle, 
        draw=green!70, 
        fill=green!10, 
        thick, 
        minimum height=1.2cm, 
        text centered, 
        text width=4cm, 
        font=\itshape\footnotesize 
    },
    conector/.style={-Latex, thick, blue!80, line width=1pt},
    conector_estrategia/.style={Stealth-Stealth, thick, green!80, dashed} 
]

\node[fase] (curacion) {Fase 1: Curación de datos y conocimiento};
\node[fase, right=of curacion] (recuperacion) {Fase 2: recuperación (Retrieval)};
\node[fase, below=of recuperacion] (generacion) {Fase 3: generación y razonamiento};
\node[fase, left=of generacion] (verificacion) {Fase 4: verificación Post-Hoc y confianza};

\node[estrategia_cat, above=0.7cm of curacion] (strat_curacion) {Estrategias de curación y gobernanza de datos (5.1)};
\node[estrategia_cat, above=0.7cm of recuperacion] (strat_recuperacion) {Estrategias de recuperación sofisticada (5.2)};
\node[estrategia_cat, below=0.7cm of generacion] (strat_generacion) {Estrategias de generación y razonamiento fiel (5.4)}; 
\node[estrategia_cat, below=0.7cm of verificacion] (strat_verificacion) {Estrategias de verificación y calibración (5.5)};

\draw[conector] (curacion.east) -- (recuperacion.west) node[midway, above, font=\small\color{blue!90}] {Input Datos};
\draw[conector] (recuperacion.south) -- (generacion.north) node[midway, right, font=\small\color{blue!90}] {Contexto Relevante};
\draw[conector] (generacion.west) -- (verificacion.east) node[midway, below, font=\small\color{blue!90}] {Respuesta Candidata};
\draw[conector] (verificacion.north) -- (curacion.south) node[midway, left, font=\small\color{blue!90}] {Retroalimentación y Mejora};

\draw[conector_estrategia] (strat_curacion.south) -- (curacion.north);
\draw[conector_estrategia] (strat_recuperacion.south) -- (recuperacion.north);
\draw[conector_estrategia] (strat_generacion.north) -- (generacion.south); 
\draw[conector_estrategia] (strat_verificacion.north) -- (verificacion.south);

\end{tikzpicture}
\caption{Modelo cíclico de un sistema RAG Legal y puntos de intervención estratégica para la optimización y mitigación de alucinaciones. Las estrategias específicas (referenciadas por subsección del ensayo) se aplican en cada fase para mejorar la fiabilidad general del sistema.}
\label{fig:ciclo_rag_estrategias}
\end{figure}

\subsection{La calidad del fundamento: curación estratégica de datos y bases de conocimiento externo}

El adagio "basura entra, basura sale" (\textit{garbage in, garbage out}) resuena con especial fuerza en el contexto de los LLMs y los sistemas RAG. La calidad, actualidad, relevancia y representatividad de la base de conocimiento externa sobre la cual se fundamentan estos sistemas no es un mero detalle técnico, sino el cimiento sobre el que se construye toda su fiabilidad. Un corpus documental deficiente, desactualizado o sesgado inevitablemente limitará la capacidad del sistema RAG para proporcionar respuestas precisas y confiables, independientemente de cuán sofisticados sean sus algoritmos de recuperación o generación. Por lo tanto, una estrategia primordial y proactiva para la mitigación de alucinaciones comienza mucho antes de la interacción con el usuario: en la meticulosa curación y gestión estratégica de estas bases de conocimiento.

\subsubsection{Selección y priorización rigurosa de fuentes legales}

El vasto universo de información legal exige una selección criteriosa. No todas las fuentes son iguales en autoridad o relevancia. Es imperativo tener en consideración:
\begin{itemize}
        \item \textit{Jerarquización de la autoridad:} Diseñar mecanismos, tanto en la indexación como en la recuperación, que prioricen explícitamente las fuentes primarias vinculantes (Constitución, estatutos vigentes, jurisprudencia de tribunales superiores de la jurisdicción pertinente) sobre fuentes secundarias, literatura persuasiva, o jurisprudencia de otras jurisdicciones o tribunales inferiores. Esto implica la incorporación de metadatos ricos que codifiquen esta jerarquía y permitan al sistema RAG ponderar la información en consecuencia.
        \item \textit{Verificación continua de la actualidad y vigencia:} Implementar procesos dinámicos y automatizados (en la medida de lo posible, complementados con revisión experta) para mantener la base de conocimiento al día con las enmiendas legislativas, las nuevas decisiones judiciales y, de manera crítica, el estado de derogación de los precedentes. La integración con servicios comerciales de verificación de citas o bases de datos legislativas actualizadas es fundamental. Herramientas como Shepard’s de LexisNexis o KeyCite de Westlaw (estándares en el sistema de \textit{common law} estadounidense) son cruciales para trazar el historial de un caso y verificar su vigencia. En España, si bien no existe un equivalente directo con una marca tan consolidada, plataformas jurídicas líderes como vLex o La Ley Digital ofrecen funcionalidades análogas que permiten comprobar si una sentencia ha sido recurrida, anulada o matizada, así como verificar la vigencia de una norma. La integración o, como mínimo, la consulta sistemática de estas herramientas es un paso ineludible para evitar fundamentar respuestas en ley obsoleta, un error común y potencialmente grave.
        \item \textit{Filtrado proactivo de fuentes de baja calidad o problemáticas:} Identificar y excluir o marcar explícitamente fuentes conocidas por su baja calidad, sesgos manifiestos (si el objetivo es un análisis neutral), o irrelevancia para las tareas legales más comunes. Esto puede requerir tanto el juicio de expertos legales como el uso de técnicas de IA para la evaluación automática de la calidad y fiabilidad de los documentos (Nguyen and Satoh, 2024).
        Estas técnicas van más allá de la simple detección de palabras clave. Incluyen, por ejemplo, el uso de modelos de lenguaje más pequeños y especializados que actúan como 'jueces' o evaluadores (un enfoque conocido como \textit{LLM-as-a-judge}), capaces de verificar la consistencia lógica de un documento, detectar contradicciones internas o contrastar afirmaciones contra una base de conocimiento curada. Otras técnicas implican el análisis de la confianza del modelo durante la generación o la detección de anomalías estilísticas que a menudo acompañan a las alucinaciones. La implementación de estos 'guardianes' algorítmicos puede servir como un primer filtro automatizado antes de la revisión humana.
\end{itemize}

La importancia de estas prácticas de curación, priorización y gobernanza de los datos que alimentan los sistemas de IA legal se ve magnificada por los emergentes marcos regulatorios. El Reglamento (UE) 2024/1689 del Parlamento Europeo y del Consejo, de 13 de junio de 2024, por el que se establecen normas armonizadas en materia de inteligencia artificial (en adelante, la Ley de IA de la UE o el Reglamento), en su Artículo 10, impone a los desarrolladores de sistemas de IA de alto riesgo obligaciones explícitas respecto a los conjuntos de datos de entrenamiento, validación y prueba. Estos deben ser 'relevantes, representativos, libres de errores y completos', y deben implementarse prácticas adecuadas de gobernanza de datos, incluyendo un examen de los posibles sesgos. El cumplimiento de estas exigencias no es solo una cuestión de buena práctica técnica para mejorar la fiabilidad del modelo y reducir el riesgo de alucinaciones originadas en datos defectuosos, sino que se perfila como un requisito legal ineludible para operar en el mercado europeo, incentivando una mayor diligencia en la gestión del conocimiento que fundamenta la IA legal.

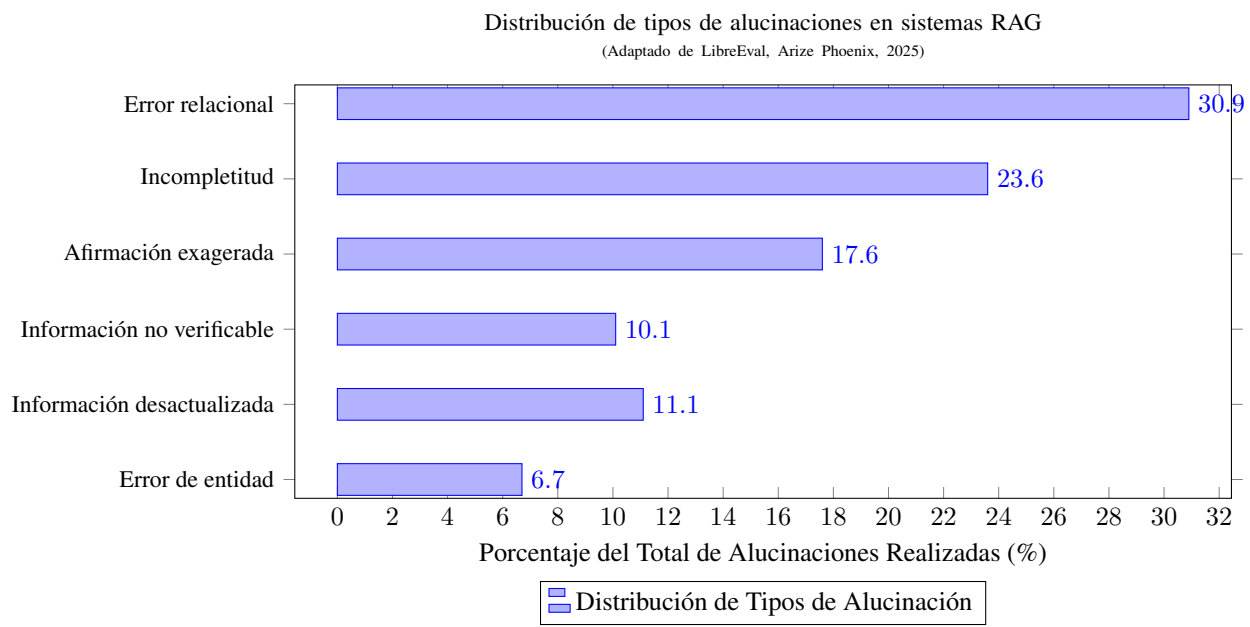
\begin{figure}[htbp]
  \centering
  \begin{tikzpicture}
    \begin{axis}[
        xbar, 
        enlargelimits=0.05,
        legend style={at={(0.5,-0.2)}, anchor=north, legend columns=-1}, 
        xlabel={Porcentaje del Total de Alucinaciones Realizadas (\%)},
        symbolic y coords={
            Error de entidad, 
            Información desactualizada, 
            Información no verificable, 
            Afirmación exagerada, 
            Incompletitud, 
            Error relacional
        },
        ytick=data,
        y tick label style={font=\footnotesize}, 
        bar width=12pt, 
        nodes near coords,
        nodes near coords align={horizontal},
        xmin=0,
        title style={font=\footnotesize, text width=0.85\textwidth, align=center}, 
        title={Distribución de tipos de alucinaciones en sistemas RAG \\ 
               \tiny (Adaptado de LibreEval, Arize Phoenix, 2025)}, 
        width=0.85\textwidth, 
        height=0.6\textwidth, 
        y=1cm 
      ]
      
      \addplot coordinates {
        (6.7,Error de entidad)
        (11.1,Información desactualizada)
        (10.1,Información no verificable)
        (17.6,Afirmación exagerada)
        (23.6,Incompletitud)
        (30.9,Error relacional)
      };
      \legend{Distribución de Tipos de Alucinación}
    \end{axis}
  \end{tikzpicture}
  \caption{Distribución porcentual de los diferentes tipos de alucinaciones efectivamente realizadas en las respuestas de modelos de lenguaje con RAG, según el dataset LibreEval1.0. Esta distribución destaca los desafíos más comunes que las estrategias de mitigación deben abordar.}
  \label{fig:distribucion_tipos_alucinacion_libreeval}
\end{figure}

\subsubsection{Aseguramiento de la diversidad y representatividad del corpus}

Para evitar la creación de "monoculturas legales" algorítmicas (Dahl et al., 2024) que ignoren la diversidad del pensamiento jurídico o las particularidades de jurisdicciones menos prominentes, es crucial:

\begin{itemize}
    \item \textit{Cobertura jurisdiccional y temática amplia:} Esforzarse por incluir una representación equilibrada de todas las jurisdicciones relevantes (estatales, federales, especializadas) y de un amplio espectro de áreas del derecho, no solo aquellas con mayor disponibilidad de datos digitalizados. En el contexto español, por ejemplo, los datos del Consejo General del Poder Judicial (CGPJ) muestran consistentemente una alta concentración de litigiosidad en los órganos judiciales de grandes capitales como Madrid o Barcelona. Un corpus de entrenamiento que no pondere adecuadamente esta realidad podría desarrollar un sesgo centralista, ignorando las particularidades doctrinales o jurisprudenciales de otros Tribunales Superiores de Justicia, lo que limitaría la utilidad de la herramienta a nivel nacional..
\end{itemize}
\begin{itemize}
    \item \textit{Inclusión consciente de perspectivas plurales:} Para tareas que requieren un análisis más allá de la doctrina pura (p. ej., evaluación de impacto de políticas, argumentación basada en principios), considerar la inclusión curada de fuentes académicas críticas, informes de organizaciones de la sociedad civil, o incluso transcripciones de debates legislativos que ofrezcan perspectivas diversas y matizadas sobre la ley y su aplicación.
\end{itemize}

\subsubsection{Estructuración avanzada del conocimiento legal (más allá del texto plano)}

Mientras que muchos sistemas RAG operan principalmente sobre texto no estructurado, la representación del conocimiento legal puede enriquecerse significativamente mediante:

\begin{itemize}
    \item \textit{Desarrollo de ontologías y grafos de conocimiento legales:} Construir modelos formales que representen entidades legales clave (tribunales, jueces, leyes, conceptos jurídicos), sus atributos y, fundamentalmente, las relaciones complejas entre ellas (p. ej., una ley \textit{enmienda} otra, un caso \textit{interpreta} un estatuto, un juez \textit{disiente} de una opinión). Un sistema RAG que pueda consultar y razonar sobre estos grafos de conocimiento podría realizar inferencias más profundas y precisas que uno basado únicamente en similitud textual (Martin, 2024; Magora, 2024).
    \item \textit{Extracción y vinculación de metadatos ricos:} Enriquecer cada documento en la base de conocimiento con metadatos detallados y estructurados (jurisdicción, fecha, tribunal, jueces, partes, temas legales, citas, historial procesal, estado de vigencia) que puedan ser utilizados por el módulo de recuperación para un filtrado y ranking mucho más preciso.
\end{itemize}

La inversión continua en la calidad, estructura y gestión de las bases de conocimiento no es un costo accesorio, sino una inversión estratégica fundamental en la fiabilidad a largo plazo de cualquier sistema de IA legal basado en RAG. Sin un fundamento sólido, incluso los algoritmos más avanzados estarán construyendo sobre arena movediza.

\subsection{Optimización sofisticada de la fase de recuperación (Retrieval): encontrando la aguja jurídica en el pajar digital}

La eficacia de un sistema RAG depende de forma crítica de la capacidad de su módulo de recuperación para identificar y extraer, de entre un corpus potencialmente masivo, los fragmentos de texto (chunks) que son \textit{exacta y contextualmente relevantes} para la consulta del usuario. La simple búsqueda por similitud semántica, aunque un punto de partida, a menudo resulta insuficiente para la complejidad y los matices del lenguaje y el razonamiento jurídico. La optimización avanzada de esta fase es, por lo tanto, un área de intensa investigación y desarrollo, enfocada en dotar al sistema de una capacidad de discernimiento más cercana a la de un investigador legal humano.

\subsubsection{\textbf{Modelos de embedding legales y estrategias multi-vectoriales} }
 
La calidad de la representación vectorial (embedding) que captura el significado de los chunks y de la consulta es la piedra angular de la búsqueda semántica.

    \begin{itemize}
        \item \textit{Embeddings especializados del dominio legal:} Priorizar el uso de modelos de embedding que hayan sido pre-entrenados o fine-tuneados específicamente en grandes corpus de textos jurídicos (como LegalBERT o los modelos desarrollados a partir de corpus como The Pile of Law - Henderson et al. 2022) es preferible a los embeddings de propósito general, ya que pueden capturar mejor los matices semánticos y la terminología específica del derecho.
        \item \textit{Técnicas de aumento de embeddings:} Explorar métodos que enriquezcan los embeddings textuales con información adicional, como metadatos estructurales o información de grafos de conocimiento, para crear representaciones más ricas y contextualmente informadas.
    \end{itemize}

\subsubsection{\textbf{Estrategias de búsqueda híbrida, multi-Etapa y refinamiento de consultas (Query Refinement):}}

Para superar las limitaciones de una única modalidad de búsqueda, se están adoptando enfoques más complejos:
    \begin{itemize}
    \item\textit{Optimización dinámica de búsqueda híbrida:} La combinación de la búsqueda semántica (vectorial, densa) con la búsqueda tradicional por palabras clave (léxica, p. ej., BM25) es fundamental. Sin embargo, la ponderación relativa entre estas dos modalidades no debe ser estática. Idealmente, el sistema debería ser capaz de ajustar dinámicamente esta ponderación basándose en las características de la consulta del usuario (p. ej., dar más peso a las palabras clave si la consulta contiene términos muy específicos, nombres propios o citas exactas) (Addleshaw Goddard, 2024).
    \item\textit{Pre-Procesamiento y transformación inteligente de consultas:} Utilizar un LLM (posiblemente un modelo más pequeño y eficiente dedicado a esta tarea) para analizar y refinar la consulta del usuario \textit{antes} de la fase de recuperación. Esto puede incluir:
    
    \begin{itemize}
        \item \textit{Query Expansion:} Añadir sinónimos legales relevantes, términos conceptualmente relacionados o posibles reformulaciones para ampliar la cobertura de la búsqueda.
        \item \textit{Descomposición de consultas (Query Decomposition):} Dividir preguntas complejas o multifacéticas en sub-preguntas más simples y atómicas, cada una de las cuales puede ser objeto de una recuperación separada. Los resultados de estas sub-recuperaciones se pueden luego combinar para responder a la consulta original.
        \item \textit{Generación de consultas hipotéticas (Hypothetical Document Embeddings - HyDE):} Instruir a un LLM para que genere un documento "ideal" que respondería perfectamente a la consulta del usuario, y luego usar el embedding de este documento hipotético para la búsqueda semántica. Esto a menudo conduce a una recuperación más relevante que usar el embedding de la consulta original directamente (Gao et al., 2022).
    \end{itemize}
    
    \item\textit{Recuperación multi-etapa (Multi-Hop Retrieval):} Para consultas que requieren un razonamiento secuencial o la síntesis de información a través de una cadena de documentos (p. ej., rastrear la evolución de una doctrina legal a través de múltiples precedentes), el sistema puede implementar un proceso de recuperación iterativo. La información extraída de un primer conjunto de documentos recuperados se utiliza para formular nuevas consultas o refinar las existentes, permitiendo al sistema "navegar" a través del corpus documental de manera más inteligente y dirigida (Tang and Yang, 2024).
    \item\textit{Re-ranking sofisticado (Re-ranking):} Una vez que se ha recuperado un conjunto inicial de chunks candidatos (posiblemente amplio), utilizar modelos de re-ranking más potentes y computacionalmente intensivos (como los cross-encoders) para evaluar de manera más precisa la relevancia de cada chunk en relación con la consulta completa. Estos modelos pueden considerar la interacción entre la consulta y el chunk de manera más profunda que los modelos de embedding utilizados en la recuperación inicial, mejorando el orden final de los resultados presentados al LLM generador.
    \end{itemize}

\subsubsection{\textbf{Incorporación de retroalimentación y aprendizaje continuo} }
Los sistemas RAG más avanzados deberían incorporar mecanismos para aprender de las interacciones con los usuarios y de la retroalimentación explícita o implícita.

\begin{itemize}
    \item\textit{Retroalimentación del usuario:} Permitir a los usuarios calificar la relevancia de los documentos recuperados o de las respuestas generadas, y utilizar esta retroalimentación para fine-tunear los modelos de embedding o los algoritmos de ranking.
    \item\textit{Adaptación Dinámica:} Ajustar los parámetros de recuperación (p. ej., umbrales de similitud, número de chunks a recuperar) basándose en el rendimiento histórico o en las características de la consulta actual.
\end{itemize}

\subsubsection{\textbf{Conclusión} }

La inversión en estas estrategias avanzadas de recuperación es fundamental, ya que la calidad del contexto proporcionado al LLM generador es el techo de la calidad de la respuesta final. Una recuperación deficiente o ruidosa inevitablemente conducirá a una generación subóptima o, peor aún, alucinada, independientemente de cuán sofisticado sea el LLM generador.

\subsection{Agentes de IA en sistemas legales complejos y la jerarquía normativa de Kelsen}

La optimización de los sistemas RAG para tareas legales, como se ha discutido, implica una interacción cada vez más sofisticada entre el LLM, las bases de conocimiento y el proceso de razonamiento. A medida que esta sofisticación aumenta y que los LLMs evolucionan hacia capacidades de razonamiento más complejas y multi-paso, comenzamos a vislumbrar el potencial de arquitecturas de IA más autónomas y proactivas, a menudo denominadas 'agentes de IA'. En este contexto, un agente de IA legal se distingue de un LLM simple (que meramente responde a un prompt) por su capacidad para realizar secuencias de acciones, interactuar con múltiples herramientas o fuentes de información de manera autónoma, tomar decisiones intermedias, y planificar estrategias para lograr un objetivo legal complejo predefinido, como podría ser la preparación de un caso o la realización de una due diligence exhaustiva. Sin embargo, la implementación de tales agentes en el intrincado y normativamente estructurado dominio legal debe considerar ineludiblemente la arquitectura fundamental del propio sistema jurídico. En los ordenamientos de derecho civil y en muchos sistemas constitucionales, esta arquitectura se conceptualiza clásicamente a través de la noción de la Pirámide Normativa de Hans Kelsen.

La pirámide de Kelsen postula que las normas jurídicas de un sistema se organizan jerárquicamente, donde la validez de cada norma deriva de una norma superior, culminando en una "Norma Fundamental" (Grundnorm) hipotética que fundamenta la validez de todo el sistema (generalmente la Constitución en los sistemas modernos). Esta estructura jerárquica implica que las normas de rango inferior (p.ej., un reglamento administrativo) deben ser conformes con las normas de rango superior (p.ej., una ley, la Constitución). La aplicación correcta del derecho, por tanto, no es solo una cuestión de encontrar \textit{una} norma relevante, sino de encontrar la norma \textit{correcta} dentro de esta jerarquía y resolver posibles conflictos entre normas de diferente rango (principio de jerarquía normativa) o del mismo rango pero posteriores en el tiempo (principio de temporalidad) o más específicas (principio de especialidad).

Para un agente de IA legal que opere con el objetivo de proporcionar respuestas fiables y legalmente válidas, esta estructura jerárquica presenta tanto un desafío como una oportunidad:

\begin{enumerate}
    \item \textbf{Desafío para la recuperación y el razonamiento:}

\begin{itemize}
    \item \textit{Identificación de la autoridad controladora:} Un agente de IA debe ser capaz no solo de recuperar múltiples normas o precedentes potencialmente relevantes, sino de discernir cuál de ellos tiene precedencia o es la autoridad controladora en un caso dado. Por ejemplo, si una ley parece contradecir una disposición constitucional, la Constitución prevalece. Si un reglamento contradice una ley, la ley prevalece. Esta inferencia jerárquica es esencial.
    \item \textit{Resolución de antinomias:} El agente debe ser capaz de identificar y, en la medida de lo posible, proponer soluciones a conflictos normativos (antinomias) utilizando los criterios de resolución aceptados (jerarquía, temporalidad, especialidad). Esto requiere un nivel de razonamiento meta-legal.
    \item \textit{Comprensión de la dinámica de validez:} La validez de una norma puede cambiar (p. ej., una ley puede ser declarada inconstitucional, un precedente puede ser derogado). El agente debe acceder y procesar información sobre el estado actual de validez de las normas recuperadas.
\end{itemize}

Consideremos, por ejemplo, una consulta sobre la legalidad de una determinada práctica comercial municipal. Un agente de IA simple, sin conciencia jerárquica, podría recuperar y basar su respuesta afirmativa en una ordenanza municipal que explícitamente permite dicha práctica. Sin embargo, un agente 'kelseniano', en su proceso de verificación ascendente, identificaría una ley autonómica o estatal posterior y de rango superior que prohíbe o restringe severamente tal actividad, o incluso una sentencia del Tribunal Constitucional que haya declarado inconstitucional una norma similar. Este agente concluiría correctamente que la ordenanza municipal, aunque textualmente aplicable, es inválida o inaplicable debido al conflicto con una norma jerárquicamente superior, evitando así una alucinación de validez que el agente simple habría cometido. La capacidad para realizar este tipo de validación jerárquica es, por tanto, crucial para la fiabilidad.

    \item \textbf{Oportunidad para sistemas RAG jerárquicamente conscientes:} La pirámide de Kelsen puede, de hecho, inspirar arquitecturas RAG más sofisticadas y fiables:
\begin{itemize}
    \item \textit{Bases de conocimiento estructuradas jerárquicamente:} Las bases de conocimiento podrían organizarse explícitamente reflejando la jerarquía normativa. Los documentos podrían etiquetarse con su rango jerárquico, utilizando un esquema de metadatos que refleje la estructura del ordenamiento. Esto incluye categorías universales como 'Norma Constitucional', 'Legislación Primaria' (leyes), 'Legislación Secundaria' (reglamentos) y 'Jurisprudencia Vinculante'. En el contexto específico del derecho español, esto se traduciría en etiquetas como (Constitución, ley orgánica, ley ordinaria, reglamento, jurisprudencia del Tribunal Constitucional, etc.).
    \item \textit{Algoritmos de recuperación sensibles a la jerarquía:} El módulo de recuperación podría ser instruido para priorizar la búsqueda y recuperación de normas de rango superior cuando sean pertinentes, o para buscar específicamente normas que interpreten o apliquen una norma superior identificada.
    \item \textit{Módulos de razonamiento para la coherencia jerárquica:} Un LLM generador, o un componente de razonamiento especializado, podría ser entrenado para verificar la coherencia de una solución propuesta con las normas de rango superior. Si una interpretación de un contrato parece violar una ley imperativa, el agente podría señalar esta inconsistencia.
    \item \textit{Agentes planificadores que navegan la pirámide:} Un agente de IA más avanzado podría planificar su proceso de investigación y razonamiento comenzando por la cúspide de la pirámide (Constitución, tratados internacionales relevantes) y descendiendo a través de las leyes y la jurisprudencia aplicable, asegurando que cada paso sea consistente con el nivel superior.
\end{itemize}
\end{enumerate}

\textbf{Implicaciones para la fiabilidad y las alucinaciones:}

Ignorar la jerarquía normativa puede llevar a un tipo específico y grave de alucinación legal: la \textbf{alucinación de invalidez o inaplicabilidad por conflicto jerárquico}. Un LLM podría, por ejemplo, basar una respuesta en un reglamento que, aunque textualmente relevante, es inválido porque contradice una ley superior, o fundamentar un argumento en jurisprudencia de un tribunal inferior que ha sido revocada o matizada por un tribunal superior. Estas no son simples inexactitudes factuales, sino errores fundamentales en la aplicación del derecho.

Por el contrario, un agente de IA que esté explícitamente modelado para comprender y operar dentro de la pirámide kelseniana podría:

\begin{itemize}
    \item \textbf{Reducir las alucinaciones de relevancia:} Al priorizar fuentes de mayor autoridad, es menos probable que se base en información legalmente subordinada o irrelevante.
    \item \textbf{Mejorar la solidez del razonamiento:} Al verificar la coherencia con normas superiores, sus conclusiones serían más robustas y menos susceptibles de ser invalidadas.
    \item \textbf{Aumentar la transparencia y explicabilidad:} Al poder trazar la derivación de una conclusión a través de la jerarquía normativa, el agente podría ofrecer explicaciones más convincentes y verificables de su razonamiento.
\end{itemize}

En definitiva, un agente de IA 'kelseniano', es decir, uno que no solo acceda a las fuentes sino que comprenda y respete la jerarquía normativa y los principios de validez del ordenamiento, sería intrínsecamente menos propenso a ciertos tipos críticos de alucinaciones. Al priorizar la Constitución sobre la ley, y la ley sobre el reglamento, y al verificar la vigencia y aplicabilidad de cada norma dentro de su contexto jerárquico, se reduciría drásticamente el riesgo de generar consejos basados en normas subordinadas invalidadas, en jurisprudencia derogada o en interpretaciones que contradicen principios fundamentales. Esta conciencia estructural no elimina todos los riesgos de alucinación –especialmente aquellos derivados de la ambigüedad inherente del lenguaje o de los límites del propio corpus de conocimiento– pero sí proporciona un andamiaje robusto para una IA legal más coherente, predecible y, en última instancia, más fiable. Es crucial entender que el valor de este andamiaje no reside únicamente en la mejora de la respuesta final, sino en la propia externalización del proceso de razonamiento. Al hacer explícita la jerarquía normativa que aplica, la IA pasa de ofrecer una conclusión opaca a presentar un argumento verificable. Para el profesional del derecho, esta fundamentación transparente es, en muchos casos, más valiosa que la respuesta misma, pues le permite auditar, validar y, en última instancia, apropiarse del razonamiento para construir su propia estrategia jurídica. Transforma la IA de una 'caja negra' a una 'caja de herramientas' de razonamiento.

En el contexto español y de muchos sistemas de derecho civil europeos, donde la codificación y la jerarquía formal de las fuentes del derecho son particularmente pronunciadas, la incorporación de una conciencia kelseniana en los agentes de IA legales no es un mero refinamiento académico, sino una condición necesaria para su fiabilidad y utilidad práctica. Un agente que no "entienda" la estructura piramidal del ordenamiento jurídico será intrínsecamente propenso a generar respuestas que, aunque plausiblemente redactadas, sean legalmente insostenibles o directamente erróneas. El desarrollo futuro de la IA legal fiable pasará, ineludiblemente, por dotar a estos sistemas de una comprensión más profunda de la arquitectura fundamental del derecho mismo.

\begin{table}[htbp]
  \centering
  \caption{Impacto cuantificable estimado de estrategias de optimización en sistemas RAG Legales}
  \label{tab:impacto_optimizacion_rag}
  \sisetup{table-format=2.1}
  \begin{tabular}{
    >{\raggedright\arraybackslash}p{0.25\textwidth} 
    >{\raggedright\arraybackslash}p{0.25\textwidth} 
    >{\raggedright\arraybackslash}p{0.12\textwidth}
    >{\raggedright\arraybackslash}p{0.12\textwidth}
    >{\raggedright\arraybackslash}p{0.12\textwidth}
  }
    \toprule
    \textbf{Estrategia de } & \textbf{Métrica clave} & {\textbf{Valor}} & {\textbf{Valor}} & {\textbf{Mejora }} \\
    optimización aplicada & & {\textbf{inicial}} & {\textbf{optimizado}} & {\textbf{relativa (\%)}} \\
    \midrule
    Optimización intensiva de recuperación (Chunking, Embeddings, Query Expansion) \newline \textit{(Inspirado en Addleshaw Goddard, 2024)} & F1-Score (Extracción de cláusulas) & 74 & 95 & 28.4 \\
    \addlinespace
    Fine-Tuning del LLM Generador para fidelidad al contexto & tasa de \textit{Misgrounding} & 20 & 5 & 75.0 \\ 
    \addlinespace
    Implementación de Chain-of-Thought (CoT) en prompts de generación & coherencia lógica (Puntuación Humana 1-5) & 3.2 & 4.5 & 40.6 \\ 
    \addlinespace
    Verificación Post-Hoc automatizada con modelo secundario & tasa de alucinaciones no detectadas & 15 & 3 & 80.0 \\ 
    \addlinespace
    Implementación de agente consciente de jerarquía (Kelseniano) & Tasa de error por conflicto normativo & 15 & 2 & 86.7 \\ 
    \bottomrule
    \multicolumn{5}{p{0.95\textwidth}}{\footnotesize{\textit{Nota:} Los valores para "Optimización de Recuperación" están inspirados en los resultados F1 reportados por Addleshaw Goddard (2024). Otros valores son hipotéticos y presentados con fines ilustrativos para demostrar el potencial impacto de diversas estrategias de optimización discutidas en la Sección 5. La "Mejora Relativa" se calcula como ((Valor Optimizado - Valor Inicial) / Valor Inicial) * 100 para métricas donde mayor es mejor, o ((Valor Inicial - Valor Optimizado) / Valor Inicial) * 100 para métricas donde menor es mejor (ej. tasas de error). La fila sobre el 'Agente Consciente de Jerarquía' es ilustrativa y busca cuantificar el beneficio de implementar la lógica discutida en la Sección 5.3. }}
  \end{tabular}
\end{table}

\subsection{Refinamiento estratégico de la fase de generación y razonamiento: cultivando la fidelidad y la coherencia en la IA Legal}

Una vez que el sistema de Generación Aumentada por Recuperación (RAG) ha completado la fase crítica de recuperación, proporcionando al Gran Modelo de Lenguaje (LLM) un conjunto de fragmentos de texto contextualmente relevantes (idealmente optimizado a través de las técnicas de la sección anterior), el desafío se traslada a la fase de generación. Aquí, el objetivo es guiar al LLM para que utilice esta información recuperada de manera que la respuesta final no solo sea lingüísticamente fluida y coherente, sino, y de manera crucial, fácticamente precisa, lógicamente sólida, fiel a las fuentes proporcionadas y directamente relevante para la consulta original del usuario. La mera provisión de contexto no garantiza una generación de alta calidad; el LLM generador, por su naturaleza probabilística y sus vastos conocimientos paramétricos, aún puede desviarse, malinterpretar o incluso alucinar. Por lo tanto, el refinamiento estratégico de esta fase es un componente esencial de cualquier sistema RAG legal que aspire a la fiabilidad.

\begin{enumerate}

    \item \textbf{Ingeniería de prompts avanzada, específica para RAG y consciente del contexto legal:} El prompt que se alimenta al LLM generador, que ahora incluye tanto la consulta original del usuario como los fragmentos de texto recuperados, debe ser meticulosamente diseñado para maximizar la fidelidad y la precisión. Esto va mucho más allá de una simple concatenación.

\begin{itemize}

    \item \textit{Instrucciones explícitas sobre fundamentación (Grounding) y atribución:} El prompt debe contener directivas claras e inequívocas que instruyan al LLM a basar su respuesta \textbf{predominante o exclusivamente} en la información contenida dentro de los documentos proporcionados y evitar activamente el uso de su conocimiento paramétrico interno o la realización de suposiciones no fundamentadas. Se deben incluir mandatos para \textbf{citar explícitamente las fuentes} de sus afirmaciones, idealmente vinculando cada proposición al fragmento o documento específico del contexto recuperado que la respalda. Esto no solo fomenta la fidelidad, sino que facilita la verificación por parte del usuario.
    \item \textit{Guías rstructuradas para el razonamiento (Chain-of-Thought y Similares):} Para consultas que requieren análisis o síntesis, en lugar de una simple extracción, el prompt puede instruir al LLM a seguir un proceso de razonamiento paso a paso (Wei et al. 2023). Por ejemplo, "Primero, identifica los hechos clave en los documentos proporcionados. Segundo, identifica las reglas legales aplicables mencionadas. Tercero, aplica estas reglas a los hechos. Cuarto, explica tu conclusión, citando los documentos relevantes para cada paso". Esta externalización del proceso de razonamiento no solo tiende a mejorar la calidad de la conclusión final, sino que también proporciona una traza de auditoría que puede ser revisada por un experto legal (Schwarcz et al., 2024).

    Un enfoque robusto para guiar el razonamiento es la \textbf{descomposición explícita del problema} (explicit problem decomposition). En lugar de pedir al LLM que determine directamente si un texto viola una norma, la tarea se divide en los sub-componentes lógicos que un jurista analizaría. Por ejemplo, para determinar si un comentario constituye "incitación al odio" según el § 130 del Código Penal alemán, un sistema puede ser instruido para responder primero a dos preguntas separadas: (1) ¿El texto se dirige a un grupo protegido por la norma? y (2) ¿El texto realiza un acto prohibido por la norma (incitar, insultar, etc.)? (Ludwig et al., 2025). Solo si ambas respuestas son afirmativas, se concluye que la norma ha sido violada. Esta metodología no solo estructura el "pensamiento" del modelo, sino que hace su conclusión final mucho más transparente y verificable para el supervisor humano.

    \item \textit{Manejo sofisticado de la incertidumbre, los conflictos y la información faltante:} El prompt debe guiar al LLM sobre cómo proceder cuando la información recuperada es incompleta, ambigua, presenta contradicciones internas, o simplemente no contiene la respuesta a la consulta. En lugar de forzar una respuesta o recurrir a la fabricación, el modelo debe ser instruido para:

    \begin{itemize}
        \item Indicar explícitamente la incertidumbre (p. ej., "Basado en la información proporcionada, no es posible determinar con certeza...").
        \item Presentar las diferentes perspectivas o la información conflictiva de manera objetiva, señalando las discrepancias.
        \item Declarar que la información solicitada no se encuentra en los documentos recuperados.
    \end{itemize}

    \item \textit{Definición precisa de la persona y el formato de salida:} Especificar el rol que el LLM debe adoptar (p. ej., "Actúa como un asistente de investigación legal objetivo y neutral") y el formato exacto de la respuesta esperada (p. ej., resumen estructurado, lista de puntos clave con citas, borrador de cláusula contractual) es crucial para asegurar que la salida sea consistente, útil y profesional.
    \item \textit{Técnicas de prompting de seguimiento (Follow-up) o refinamiento iterativo:} Como demostró el estudio de Addleshaw Goddard (2024), un segundo prompt que desafíe o pida una revisión de la respuesta inicial del LLM (p. ej., "Por favor, revisa tu respuesta anterior cuidadosamente. ¿Estás seguro de que has incluido toda la información relevante de los documentos proporcionados sobre X? ¿Hay algún matiz que hayas omitido?") puede inducir al modelo a realizar un procesamiento más profundo del contexto y mejorar significativamente la calidad y completitud de la respuesta final. Este enfoque iterativo simula una conversación de refinamiento.
    \end{itemize}

    \item \textbf{Fine-Tuning del LLM generador con enfoque en la fidelidad legal y el razonamiento fundamentado:} Si bien la ingeniería de prompts es una herramienta poderosa y flexible, el fine-tuning del LLM generador en un corpus cuidadosamente seleccionado puede ofrecer mejoras más profundas y consistentes en su capacidad para adherirse al contexto y razonar de manera legalmente sólida.

\begin{itemize}
    \item \textit{Fine-tuning para la fidelidad al contexto (Contextual Adherence):} Entrenar al LLM en un conjunto de datos de alta calidad compuesto por tripletas de (consulta, contexto legal recuperado relevante, respuesta ideal que es estrictamente fiel al contexto y correctamente citada). Esto puede enseñar al modelo a priorizar la información contextual sobre su conocimiento paramétrico y a resistir la tentación de "desviarse" o alucinar (Tian, Mitchell, Yao, et al. 2023).
    \item \textit{Fine-tuning para tipos específicos de razonamiento jurídico fundamentado:} Desarrollar conjuntos de datos de entrenamiento que ejemplifiquen cómo realizar tipos específicos de tareas de razonamiento legal (p. ej., identificación del \textit{holding} de un caso, aplicación de un test legal de múltiples factores, comparación de estatutos) \textit{basándose explícitamente en un conjunto de documentos de entrada}.
    \item \textit{Aprendizaje por refuerzo con retroalimentación humana (RLHF) enfocado en la factualidad y el grounding:} Utilizar RLHF no solo para alinear el modelo con las preferencias generales de estilo o utilidad, sino específicamente para recompensar respuestas que demuestren alta factualidad, fundamentación precisa en las fuentes proporcionadas y razonamiento legal coherente.
\end{itemize}

    \item \textbf{Integración con modelos de razonamiento especializados y arquitecturas avanzadas:} La emergencia de LLMs con arquitecturas explícitamente diseñadas para el razonamiento multi-paso, la planificación y la descomposición de problemas (como la familia de modelos "o" de OpenAI - OpenAI 2024) es particularmente relevante para RAG.

\begin{itemize}
    \item \textit{Planificación de la respuesta:} Estos modelos podrían, en teoría, planificar cómo utilizar la información recuperada de manera más estratégica, identificando qué fragmentos son más relevantes para qué partes de la consulta y cómo sintetizarlos de manera lógicamente coherente.
    \item \textit{Verificación interna de pasos de razonamiento:} Su capacidad para "reflexionar" sobre sus propios pasos de razonamiento intermedios podría permitirles detectar y corregir errores o inconsistencias antes de generar la respuesta final (Schwarcz et al., 2024). La integración de estos modelos de razonamiento como el componente generador en un sistema RAG es un área prometedora para futuras mejoras.
\end{itemize}

    \item \textbf{Arquitecturas híbridas (Simbólico-Neuronales):} Aunque aún en desarrollo para aplicaciones legales a gran escala, la integración de la capacidad de los LLMs para procesar lenguaje natural con la precisión y verificabilidad de los sistemas de razonamiento simbólico (basados en lógicas formales, ontologías legales estructuradas o bases de reglas explícitas) ofrece una vía prometedora. El LLM podría usar el contexto recuperado para instanciar un modelo simbólico que luego realiza las inferencias lógicas de manera más controlada y explicable.

\end{enumerate}

El objetivo final de estas estrategias de optimización de la generación no es solo producir respuestas que \textit{parezcan} correctas, sino respuestas que sean \textit{demostrablemente} correctas, fieles a las fuentes proporcionadas y útiles para el profesional legal que asume la responsabilidad final por su uso. La capacidad del LLM para explicar \textit{cómo} llegó a una conclusión a partir del contexto recuperado es tan importante como la conclusión misma.

\subsection{Verificación post-Hoc y calibración de confianza: la última línea de defensa contra las alucinaciones}

Partiendo de la premisa de que es teóricamente imposible prevenir al cien por cien las alucinaciones en la fase de generación con la tecnología actual, la implementación de mecanismos robustos de verificación \textit{después} de que el LLM ha producido una respuesta inicial se convierte en una capa de seguridad absolutamente crítica. Esta "última línea de defensa" no busca tanto evitar que el modelo alucine, sino detectar las alucinaciones cuando ocurren y proporcionar al usuario profesional señales claras sobre la fiabilidad de la información generada.

\subsubsection{Verificación factual automatizada contra fuentes externas canónicas (Fact-Checking)}

Una vez que el LLM ha generado una respuesta (que idealmente incluye citas preliminares), se pueden implementar módulos automatizados que:
\begin{itemize}

    \item \textit{Extraigan las afirmaciones factuales clave:} Identificar las proposiciones factuales y legales centrales en la respuesta del LLM.
    \item \textit{Verifiquen contra bases de conocimiento de alta confianza:} Comparar estas afirmaciones con información contenida en bases de datos legales estructuradas y canónicas (p. ej., repositorios oficiales de legislación, bases de datos jurisprudenciales con metadatos de derogación, enciclopedias jurídicas verificadas).
    \item \textit{Marquen las discrepancias:} Señalar explícitamente al usuario cualquier discrepancia encontrada, indicando si una afirmación no pudo ser verificada, contradice una fuente canónica, o se basa en una fuente citada que no la respalda (Peng et al. 2023; Chern et al. 2023).
    \item \textit{Desafíos:} La cobertura de estas bases de conocimiento externas nunca será completa, y la verificación de afirmaciones legales complejas o interpretativas sigue siendo un desafío para los sistemas automatizados.
\end{itemize}

\subsubsection{Verificación mediante reglas lógicas y heurísticas determinísticas}
    
Antes de recurrir a modelos de IA secundarios, una capa de verificación basada en reglas puede detectar de forma eficiente y económica una clase significativa de errores. Esto incluye:

\begin{itemize}
\item \textit{Validación sintáctica}: Usar expresiones regulares para verificar que las citas de sentencias o artículos legales siguen el formato canónico.
\item \textit{Chequeos de coherencia lógica simple}: Implementar reglas que marquen como sospechosa una afirmación donde un tribunal inferior revoca a uno superior, o donde una fecha de sentencia es posterior a la fecha de derogación de la ley que aplica.
\item \textit{Listas de control (Checklists)}: Comparar entidades mencionadas (jueces, partes, leyes) contra bases de datos autorizadas para detectar invenciones flagrantes.
\end{itemize}
Estos métodos, aunque tradicionales, son altamente fiables para los errores que están diseñados para capturar y deben constituir una primera línea de defensa.

\subsubsection{Modelos secundarios de detección de alucinaciones y autocrítica}
    
Se está investigando el uso de modelos de IA, a menudo más pequeños y especializados, o incluso el propio LLM generador operando en un modo de "autoevaluación", para analizar la respuesta inicial en busca de indicios de alucinación.

\begin{itemize}

    \item \textit{Detección de inconsistencias internas:} Evaluar la coherencia lógica interna de la respuesta generada.
    \item \textit{Medición de la entropía o incertidumbre de la generación:} Analizar las probabilidades asociadas a la secuencia de tokens generada; secuencias de baja probabilidad o alta entropía pueden ser más propensas a ser alucinaciones (Manakul, Liusie, and Gales 2023 - SelfCheckGPT).
    \item \textit{Comparación con conocimiento paramétrico de alta confianza:} Si el LLM tiene "conocimiento" paramétrico sobre un tema con alta confianza (p. ej., principios legales muy básicos), puede usarlo para contrastar la respuesta generada a partir del contexto RAG.
    \item \textit{Generación de críticas (CriticGPT):} OpenAI ha experimentado con modelos (como CriticGPT) entrenados para generar críticas de las respuestas de otros LLMs, ayudando a identificar errores o debilidades (Song et al., 2024 - RAG-HAT).
\end{itemize}
    
\subsubsection{Calibración y comunicación efectiva de la confianza del modelo}
    
Es fundamental que los LLMs no solo generen respuestas, sino que también comuniquen de manera fiable su propio nivel de "confianza" o incertidumbre sobre la corrección y fundamentación de dichas respuestas.

No obstante, la mera generación de una puntuación de confianza no es suficiente; la interpretación de dicha puntuación por parte del usuario profesional representa otro desafío significativo. Un '70\% de confianza' expresado por un LLM puede no tener el mismo significado intuitivo o estadístico que un 70\% de confianza en un contexto humano o en un sistema de diagnóstico tradicional. Por ello, junto con el desarrollo de métricas de confianza más fiables, es esencial investigar y establecer directrices claras sobre cómo los profesionales del derecho deben interpretar y actuar en función de estos indicadores de confianza algorítmica, especialmente cuando la calibración del modelo, como se evidencia en estudios como el de Dahl et al. (2024), sigue siendo imperfecta y puede llevar a una peligrosa sobreconfianza en las respuestas erróneas.   

\begin{itemize}

    \item \textit{Desarrollo de métricas de confianza fiables:} Investigar y refinar técnicas para que los LLMs produzcan puntuaciones de confianza que se correlacionen bien con su precisión real en tareas legales específicas (Kadavath et al. 2022; Xiong et al. 2023). Esto sigue siendo un área de investigación activa y desafiante, como lo demuestran los problemas de calibración observados en Dahl et al. (2024).
    \item \textit{Presentación transparente de la incertidumbre:} La interfaz de usuario debe comunicar claramente al profesional legal los niveles de confianza asociados a diferentes partes de la respuesta, o marcar explícitamente las afirmaciones sobre las cuales el modelo tiene baja confianza.
    \item \textit{Umbrales de intervención:} Para aplicaciones de alto riesgo, se podrían establecer umbrales de confianza por debajo de los cuales una respuesta no se presenta al usuario o se marca inequívocamente como "requiere verificación humana intensiva".
\end{itemize}

La necesidad de una comunicación transparente sobre las capacidades y limitaciones de los sistemas de IA, incluyendo su nivel de confianza o incertidumbre, encuentra un eco normativo en la Ley de IA de la UE. El Artículo 52(1) de la EU-AIAct, por ejemplo, establece obligaciones de transparencia para ciertos sistemas de IA, incluyendo aquellos que generan contenido. Se exige que los usuarios sean informados de que están interactuando con un sistema de IA y, cuando un sistema de IA genera o manipula contenido de texto, audio o vídeo que se asemeje notablemente a contenido existente ('deep fakes'), se debe divulgar que el contenido ha sido generado o manipulado artificialmente. Si bien estas disposiciones no previenen directamente la generación de una alucinación fáctica en una respuesta legal, sí buscan fomentar una mayor conciencia y cautela por parte del usuario, permitiéndole ponderar la fiabilidad de la información recibida y estableciendo una base para la rendición de cuentas cuando la IA se presenta engañosamente como humana o su contenido como no artificial.

\subsubsection{Generación de múltiples hipótesis y explicaciones contrastantes}

En lugar de generar una única respuesta "definitiva", el LLM podría ser instruido para generar múltiples interpretaciones o argumentos posibles basados en el contexto recuperado, especialmente si este es ambiguo o presenta información conflictiva. Podría también generar explicaciones que contrasten los pros y los contras de diferentes enfoques legales, permitiendo al profesional humano sopesar las alternativas.

\subsubsection{Facilitación de la verificación humana a través de citas precisas y rastreables}

Una de las contribuciones más importantes de los sistemas RAG bien diseñados es su capacidad para mejorar la verificabilidad.

\begin{itemize}

    \item \textit{Citación a nivel de fragmento (Chunk-Level Citation):} El sistema no debe simplemente listar los documentos recuperados, sino que debe, en la medida de lo posible, vincular cada afirmación o conclusión específica en la respuesta generada al fragmento (o fragmentos) exacto del texto recuperado que la respalda.
    \item \textit{Resaltado de pasajes relevantes:} La interfaz de usuario podría resaltar los pasajes específicos en los documentos fuente que fueron más influyentes para la generación de la respuesta, permitiendo al abogado ir directamente a la evidencia.
    \item \textit{Transparencia sobre el proceso de recuperación:} Ofrecer al usuario visibilidad sobre qué documentos fueron recuperados (y quizás por qué, p. ej., mostrando puntuaciones de similitud) puede ayudarle a evaluar la calidad de la base informativa utilizada por el LLM.
\end{itemize}

\subsubsection{Diseño de una capacidad de abstención inteligente: el principio de "silencio estratégico"}

Más allá de la mera comunicación de un puntaje de confianza, una estrategia de mitigación avanzada consiste en diseñar la capacidad de abstención del sistema ("no lo sé") no como un error o una limitación accidental, sino como una función deliberada y estratégica. Un sistema que es capaz de identificar los límites de su conocimiento o de las fuentes proporcionadas inspira mayor confianza y es inherentemente más seguro que uno optimizado para generar una respuesta a toda costa. La implementación de este principio de "silencio estratégico" implica varios componentes clave:

\begin{itemize}
    \item \textbf{Justificación específica para la abstención:} Cuando el sistema se abstiene, no debe ofrecer excusas genéricas. La respuesta debe ser un diagnóstico preciso de la limitación encontrada. Por ejemplo: "No es posible proporcionar una respuesta fundamentada, ya que no se han encontrado fuentes primarias en la base de conocimiento posteriores a 2023 sobre esta materia", o "Las fuentes recuperadas presentan datos conflictivos sobre el punto X y no permiten una síntesis concluyente". Esta transparencia educa al usuario y convierte una posible frustración en una interacción informativa.
    \item \textbf{Provisión de alternativas constructivas:} La abstención no debe ser un punto muerto. Un sistema robusto debe ofrecer al usuario vías de acción alternativas que aún aporten valor. Por ejemplo: "Aunque no puedo determinar la aplicabilidad directa, puedo proporcionar el marco legal general y una lista de verificación de los elementos que un profesional debería analizar", o "Puedo formular las preguntas específicas que debería dirigir a un asesor legal para resolver esta cuestión".
    \item \textbf{Comunicación visual y explícita de la incertidumbre:} En línea con la calibración de confianza, la interfaz debe comunicar proactivamente el nivel de fiabilidad de una respuesta. Un sistema de "semáforos" (por ejemplo, verde para respuestas con alto consenso y fuentes sólidas; ámbar para aquellas con lagunas de información o fuentes secundarias; rojo para las basadas en datos conflictivos o de alta incertidumbre) permite al usuario calibrar su propio nivel de escrutinio de forma inmediata.
    \item \textbf{Auditabilidad de la abstención:} Cada instancia de abstención debe generar un registro auditable (\textit{log}). Este registro debe documentar el estado del sistema en ese momento: la consulta del usuario, las fuentes recuperadas (o la falta de ellas), los criterios que llevaron a la decisión de abstenerse y los umbrales de confianza predefinidos. Esta trazabilidad es fundamental para la mejora continua del sistema y para la rendición de cuentas.

\end{itemize}

En última instancia, el diseño de sistemas de IA legal debe redefinir sus incentivos: en lugar de premiar la verbosidad y la completitud a cualquier precio, se debe premiar la precisión y la cobertura responsable. Un modelo que sabe abstenerse de forma justificada y transparente no es un sistema menos capaz, sino uno que ha alcanzado un mayor grado de madurez y demuestra un profundo respeto por el usuario y por la criticidad del dominio en el que opera. Esta calibración conservadora es un pilar fundamental para construir una confianza sostenible a largo plazo en la IA jurídica.

\subsubsection{Conclusión}

Estas estrategias de verificación post-hoc y comunicación de confianza no eliminan la necesidad de las optimizaciones previas en datos, recuperación y generación, pero actúan como una red de seguridad crucial. Reconocen la falibilidad inherente de los LLMs y buscan empoderar al profesional legal con las herramientas y la información necesarias para usar la IA de manera más crítica, informada y, en última instancia, más segura. No obstante, es crucial reconocer el delicado equilibrio inherente al 'costo de la verificación'. Si bien estas capas de seguridad post-hoc son indispensables para la fiabilidad, su implementación extensiva, especialmente si involucra una intervención humana significativa para cada comprobación o una alta latencia por múltiples llamadas a modelos secundarios, podría llegar a mermar uno de los beneficios primarios que la IA promete: la eficiencia y la reducción de costos. Un sistema que requiera una verificación manual tan exhaustiva de cada una de sus salidas que anule por completo el ahorro de tiempo inicial, podría no ser viable en la práctica para muchas tareas. Por lo tanto, el desarrollo futuro debe buscar no solo la efectividad de estos mecanismos de verificación, sino también su eficiencia, posiblemente a través de una mayor automatización inteligente de la propia verificación o mediante sistemas de IA que aprendan a 'auto-corregirse' de manera más fiable con una mínima supervisión. Encontrar el punto óptimo donde la robustez de la verificación no sacrifique desproporcionadamente la eficiencia operativa es un desafío continuo en el diseño de sistemas de IA legal prácticos y confiables.

\subsection{El rol irreductible y fortalecido de la supervisión humana experta}

A pesar de la sofisticación creciente de las estrategias de optimización y mitigación de alucinaciones, desde la curación de datos hasta la verificación post-hoc, es imperativo concluir esta sección reafirmando un principio fundamental: en el estado actual y previsible de la inteligencia artificial, \textbf{la supervisión humana crítica, informada y experta no es meramente una opción deseable, sino un componente absolutamente irreductible e indispensable} para la integración segura y ética de los LLMs en la práctica legal. Ninguna combinación de las técnicas algorítmicas discutidas puede, por sí sola, reemplazar la profundidad del juicio contextual, la responsabilidad ética y la comprensión matizada del profesional del derecho.

La necesidad de esta pericia humana no es una conjetura, sino una conclusión empírica. Incluso en estudios que implementan estrategias de condicionamiento altamente sofisticadas, proveyendo a los LLMs con definiciones legales y ejemplos de casos, se documenta una "brecha de rendimiento significativa" entre el mejor modelo de IA y los expertos legales humanos. El estudio de Ludwig et al. (2025) encontró que, si bien los modelos podían identificar razonablemente bien los grupos protegidos por la ley de discurso de odio, tenían serias dificultades para clasificar correctamente las conductas prohibidas, una tarea de juicio matizado donde los juristas humanos demostraron una fiabilidad muy superior. Esto subraya que la etapa final de evaluación y juicio cualitativo sigue siendo, por ahora, una capacidad exclusivamente humana.

Esto es particularmente cierto porque la detección de errores sutiles o la evaluación de la solidez de un argumento legal generado por IA a menudo depende intrínsecamente de la competencia y el juicio del profesional. Lo que para un lego o un abogado junior puede parecer una respuesta coherente y útil, para un experto podría revelar deficiencias argumentativas o una comprensión superficial de la doctrina aplicable. La 'verdad' o 'viabilidad' de una conclusión legal compleja no siempre es autoevidente y requiere un escrutinio informado.

El deber de competencia en la era de la IA, por tanto, exige que el profesional comprenda que no está interactuando con un "oráculo de conocimiento", sino con un sistema estadístico optimizado para la plausibilidad, cuyo diseño fundamental lo incentiva a generar respuestas seguras incluso cuando su base de conocimiento es incierta (Kalai et al., 2025). Reconocer esta característica de diseño es la base del escepticismo profesional necesario para una supervisión efectiva.
 
Lejos de volver obsoleto al abogado, la emergencia de LLMs propensos a alucinaciones, incluso aquellos aumentados por RAG, \textbf{refuerza y redefine el valor de la pericia humana}. El rol del abogado evoluciona de ser un mero recuperador de información o redactor de documentos (tareas que la IA puede asistir o incluso automatizar parcialmente) a convertirse en:

\begin{enumerate}

    \item \textbf{Supervisor crítico de la IA:} El abogado debe actuar como un "controlador de calidad" inteligente y escéptico de las salidas generadas por la IA. Esto implica no solo verificar la corrección factual y la validez legal de la información, sino también evaluar su relevancia contextual, su adecuación estratégica a los objetivos del cliente, y sus implicaciones éticas.

    Este rol de "controlador de calidad" va más allá de la simple verificación factual. El abogado debe actuar como un filtro metacognitivo, siendo consciente de que la forma en que el LLM presenta la información puede inducir sesgos en su propio proceso de razonamiento. La investigación sobre sesgos cognitivos inducidos por LLMs demuestra que los resultados de estos modelos pueden alterar el encuadre o el énfasis de la información, llevando a los humanos a tomar decisiones diferentes a las que tomarían con la información original (Alessa et al., 2025). Por lo tanto, la supervisión crítica implica un acto de auto-reflexión: el abogado no solo debe preguntar "¿es correcta esta información?", sino también "¿está esta presentación de la información influyendo indebidamente en mi juicio?".
    
    \item \textbf{Curador y guía del conocimiento de la IA:} En el contexto de sistemas RAG personalizables o fine-tuneables, los abogados expertos pueden desempeñar un papel crucial en la curación de las bases de conocimiento, en el diseño de prompts efectivos y en la provisión de retroalimentación para mejorar el rendimiento del modelo en tareas legales específicas.
    \item \textbf{Intérprete y comunicador del output de la IA:} Incluso si una IA genera un análisis legal técnicamente correcto, a menudo se requerirá que un abogado humano lo traduzca a un lenguaje comprensible para el cliente, lo contextualice dentro de la situación particular del cliente y lo integre en una estrategia legal más amplia.
    \item \textbf{Garante del juicio ético y estratégico:} La IA puede procesar información y generar opciones, pero la toma de decisiones finales que implican consideraciones éticas complejas, el ejercicio del juicio profesional sobre cursos de acción alternativos, la gestión de la relación con el cliente y la asunción de la responsabilidad profesional última, permanecen firmemente en el dominio humano.
    \item \textbf{Navegador de la incertidumbre y la ambigüedad legal:} Como se ha discutido, el derecho está lleno de áreas grises, conflictos normativos y situaciones donde no existe una única "respuesta correcta". La capacidad de un abogado para navegar esta incertidumbre, ponderar riesgos y beneficios, y aconsejar al cliente en consecuencia, es una habilidad que la IA actual no posee.
\end{enumerate}

En este nuevo paradigma, la eficiencia prometida por la IA solo se materializa si va acompañada de una inversión proporcional en la \textbf{capacitación de los abogados para interactuar críticamente con estas herramientas}. Esto incluye desarrollar habilidades en:
\begin{itemize}

    \item  \textit{Interacción semántica y profesional, más allá de la Ingeniería de Prompts}\textbf{:} Si bien actualmente la calidad de la respuesta de una IA a menudo depende de la ‘ingeniería de prompts’, es fundamental reconocer que este paradigma es una solución transitoria y un defecto de diseño, no un objetivo final. La responsabilidad de la complejidad técnica no debe recaer en el profesional del derecho, sino en el desarrollador de la herramienta LegalTech.

La verdadera innovación reside en desarrollar soluciones que abstraigan esta complejidad, permitiendo al jurista interactuar en su propio lenguaje —natural y técnico— y asumiendo la herramienta la carga de traducir esa intención a las instrucciones algorítmicas que el modelo necesita. Exigir que un cirujano aprenda a programar su bisturí es un fracaso del diseño; del mismo modo, la tecnología debe ser un bisturí que se adapta a la mano del abogado. Este enfoque lo libera para que se centre en lo que ninguna máquina puede hacer: aplicar el criterio, la estrategia y la ética.

    \item \textit{Técnicas de verificación rigurosa:} Conocer las fuentes de autoridad primaria y secundaria, y ser capaz de contrastar eficientemente las salidas de la IA con ellas.

    En última instancia, este nuevo paradigma refuerza una máxima que debe guiar el futuro de la LegalTech: el abogado no debe convertirse en un 'prompt engineer'. La responsabilidad de la complejidad técnica recae en los desarrolladores de la herramienta, no en el usuario final. Exigir que los profesionales del derecho aprendan complejas técnicas de prompting para obtener resultados fiables es un fracaso del diseño y una inversión de roles inaceptable. La tecnología debe ser un bisturí que se adapta a la mano del cirujano, no una máquina que exige que el cirujano aprenda su lenguaje arcano. Por ello, el futuro de la IA legal fiable reside en sistemas que permitan una interacción en lenguaje natural y que asuman la carga de la interpretación técnica, liberando al abogado para que se centre en lo que ninguna máquina puede hacer: aplicar el criterio, la estrategia y la ética
    
    \item \textit{Comprensión de las limitaciones de la IA:} Ser consciente de los tipos de errores y sesgos a los que la IA es propensa (incluyendo las alucinaciones) y saber cuándo no confiar en sus resultados.
    \item \textit{Integración ética de la IA en la práctica:} Comprender las implicaciones deontológicas del uso de la IA y cómo cumplir con los deberes profesionales en un entorno tecnológicamente aumentado.
\end{itemize}

Este principio de la indispensabilidad de la supervisión humana no se articula únicamente como una conclusión derivada de las limitaciones técnicas intrínsecas de la IA actual, sino que está siendo progresivamente consagrado como un requisito fundamental en los marcos regulatorios más avanzados. La Ley de IA de la UE, en su Artículo 14, es explícita al exigir que los sistemas de IA de alto riesgo estén diseñados para ser 'efectivamente supervisados por personas'.

Un ejemplo paradigmático de cómo estos principios se están materializando a nivel nacional se encuentra en España. En junio de 2024, el \textbf{Comité Técnico Estatal de la Administración Judicial Electrónica (CTEAJE)} )(el órgano gubernamental de alto nivel responsable de la modernización tecnológica y la estrategia digital del sistema judicial español) publicó su \textbf{"Política de uso de la Inteligencia Artificial en la Administración de Justicia"}. Este documento, de obligado cumplimiento para el personal de la administración de justicia, no es una mera recomendación, sino un marco normativo que establece directrices inequívocas:

\begin{itemize}

    \item \textbf{Principio de No Sustitución:} La política establece de forma tajante que \textbf{"la IA nunca debe reemplazar la toma de decisiones humanas en cuestiones cruciales"} y que "la responsabilidad final de tomar decisiones legales debe recaer en jueces y magistrados" (Principio 1.4.1).
    \item \textbf{Mandato de Revisión Humana Universal:} Más allá de los principios, se impone como norma de uso que \textbf{"la revisión humana de todo lo generado [por IA] siempre que afecte de manera directa o indirecta a los derechos de las personas usuarias" es obligatoria} (Norma 1.5.1). Esto convierte la supervisión en un requisito procesal ineludible, no en una opción.
    \item \textbf{Reconocimiento explícito de los riesgos:} La guía del CTEAJE define explícitamente el fenómeno de las \textbf{"alucinaciones"} y reconoce el peligro del \textbf{"sesgo de automatización endémico"}, por el cual los humanos tienden a confiar ciegamente en las sugerencias de los sistemas. Este reconocimiento oficial subraya la necesidad de un escepticismo informado, pilar fundamental de la supervisión experta.
\end{itemize}
    
La existencia de una guía tan detallada y vinculante por parte de un órgano como el CTEAJE demuestra que el rol irreductible del profesional humano ha trascendido el debate académico para convertirse en un pilar de la política pública y de la gobernanza de la IA en el ámbito legal.

En última instancia, la fiabilidad en la era de la IA legal no residirá exclusivamente en la perfección de los algoritmos, sino en la \textbf{simbiosis efectiva entre la capacidad de procesamiento de la IA y la sabiduría, el juicio crítico y la responsabilidad ética del profesional humano}.

Este principio de la indispensabilidad de la supervisión humana no se articula únicamente como una conclusión derivada de las limitaciones técnicas intrínsecas de la IA actual, sino que está siendo progresivamente consagrado como un requisito legal fundamental en los marcos regulatorios más avanzados. La Ley de IA de la UE, en su Artículo 14, es explícita al exigir que los sistemas de IA de alto riesgo estén diseñados y desarrollados de tal manera que puedan ser 'efectivamente supervisados por personas durante el período en que el sistema de IA está en uso'. Esta supervisión debe permitir a los humanos comprender las capacidades y limitaciones del sistema, permanecer conscientes de la posible tendencia a la automatización o al sesgo de confirmación, interpretar correctamente la salida del sistema, y tener la autoridad y competencia para decidir no utilizar dicha salida, anularla o intervenir si el sistema presenta resultados anómalos, imprevistos o potencialmente perjudiciales, como es el caso de las alucinaciones que comprometen la validez legal.

Las estrategias de optimización y mitigación son herramientas esenciales en este proceso, pero su efectividad final depende de que sean implementadas y supervisadas por juristas bien formados, conscientes de los riesgos y comprometidos con los más altos estándares de la profesión. Lejos de ser una amenaza existencial, la IA alucinante puede, paradójicamente, subrayar el valor perdurable e insustituible de la inteligencia humana experta en el corazón del derecho.

\section{La realidad de las alucinaciones en la práctica: estudios de caso detallados y lecciones aprendidas de incidentes judiciales}
\label{sec:casos_reales}

Si bien el análisis teórico y la evaluación empírica en entornos controlados son fundamentales para comprender la naturaleza y la prevalencia de las alucinaciones en los Grandes Modelos de Lenguaje (LLMs) aplicados al derecho, es en la arena de la práctica jurídica real donde las consecuencias de estos errores algorítmicos se manifiestan con una crudeza incontestable y un impacto tangible. Los incidentes donde la información generada por IA, incorrecta o completamente fabricada, ha sido introducida en procedimientos judiciales no son meras anécdotas o curiosidades tecnológicas; representan fallos sistémicos con el potencial de socavar la administración de justicia, erosionar la confianza pública y acarrear graves sanciones profesionales para los letrados implicados. Esta sección se adentra en el análisis detallado de varios estudios de caso prominentes y bien documentados, extrayendo de ellos lecciones cruciales sobre los puntos de fallo específicos en la interacción humano-IA, las deficiencias en los procesos de verificación y las consecuencias directas de confiar acríticamente en estas poderosas pero falibles herramientas. Estos casos sirven como advertencias potentes y como catalizadores para una reflexión más profunda sobre las salvaguardas necesarias en la integración de la IA en la práctica legal.

La proliferación de estos incidentes ha alcanzado un punto crítico, motivando la creación de recursos dedicados para su seguimiento. Un ejemplo notable es la base de datos en línea "\href{https://www.damiencharlotin.com/hallucinations}{AI Hallucination Cases Database}", un proyecto que busca compilar de manera exhaustiva todas las decisiones judiciales donde el contenido alucinado por una IA ha sido un factor relevante. Este tipo de repositorios se está convirtiendo en una herramienta vital para juristas, académicos y reguladores, al permitir un análisis sistemático de la naturaleza y frecuencia de estos fallos en la práctica real.

\begin{table}[htbp]
  \centering
  \caption{Análisis comparativo de incidentes judiciales destacados por alucinaciones de IA}
  \label{tab:incidentes_judiciales_ia_corregida}
  \footnotesize 
  \begin{tabular}{
    >{\raggedright\arraybackslash}p{0.26\textwidth} 
    >{\raggedright\arraybackslash}p{0.34\textwidth} 
    >{\raggedright\arraybackslash}p{0.35\textwidth} 
  }
    \toprule
    \textbf{Caso / Jurisdicción} & \textbf{Naturaleza de la Alucinación} & \textbf{Lecciones Clave / Consecuencias} \\
    \midrule
    \textit{Mata v. Avianca, Inc.} (S.D.N.Y. 2023, EE.UU.)* & 
    \makecell[l]{Fabricación completa de múltiples \\casos judiciales (citaciones, holdings). \\Respuesta afirmativa del chatbot \\sobre la existencia de los casos.} & 
    \makecell[l]{Deber ineludible de verificación independiente. \\Naturaleza engañosa de alucinaciones plausibles. \\Responsabilidad profesional individual. \\Sanciones económicas y reputacionales.} \\
    \addlinespace
    
    \textit{Thackston v. Driscoll} (W.D. Texas, 2025, EE.UU.)* &
    \makecell[l]{Acumulación de múltiples tipos de error: \\Fabricación de jurisprudencia y doctrina, \\citas falsas en casos reales, tergiversación \\de holdings (\textit{misgrounding}) y uso de \\jurisprudencia revocada.} &
    \makecell[l]{Ilustra la "cascada" de errores que un solo \\uso negligente de la IA puede producir. \\Refuerza el deber de verificación más allá de \\la simple existencia del caso. Subraya las \\graves consecuencias profesionales \\(sanciones Regla 11).} \\
    \addlinespace

    Caso en Australia (Familia, 2024, citado por Lantyer) & 
    Citaciones falsas en un caso de familia. & 
    \makecell[l]{Universalidad del riesgo de alucinación. \\Aplicación de deberes deontológicos \\(diligencia, competencia). \\Respuesta de órganos disciplinarios.} \\
    \addlinespace
    
    Caso en Brasil (Apelación, 2025, citado por Lantyer) & 
    Jurisprudencia falsa generada por IA en una apelación. & 
    \makecell[l]{Riesgos en todas las instancias judiciales. \\Importancia de la formación continua en IA \\para abogados y jueces. Multa impuesta.} \\
    \addlinespace

    Caso Tribunal Constitucional (España, 2024)* &
    \makecell[l]{Invención completa de 19 citas de doctrina \\judicial, presentadas como literales en un \\recurso de amparo.} &
    \makecell[l]{La responsabilidad del letrado es absoluta e \\independiente de la herramienta causante \\del error. El uso negligente de IA constituye \\una falta al deber de respeto al tribunal.} \\
    \addlinespace

    Estudio Magesh et al. (2024) (Herramientas RAG) & 
    \makecell[l]{Principalmente \textit{misgrounding} (citar fuente \\real pero tergiversar contenido), errores de \\razonamiento, supresión de citas.} & 
    \makecell[l]{RAG mitiga pero no elimina alucinaciones. \\Errores sutiles pueden ser más insidiosos que \\la fabricación obvia. Necesidad de \\verificación profunda de la fuente.} \\
    \bottomrule
  \end{tabular}
  \par
  \small{\textit{Nota: Aquellos casos marcados con un asterisco (*) se comentan específicamente en secciones posteriores.}}
\end{table}

\subsection{Caso de estudio: el paradigmático \textit{Mata v. Avianca, Inc.} y la fabricación de jurisprudencia}

El caso \textit{Robert Mata v. Avianca, Inc.}, No. 22-cv-1461 (PKC) (S.D.N.Y. 2023), se ha convertido rápidamente en el referente obligado al discutir los peligros de las alucinaciones de la IA en el litigio. En este asunto, los abogados del demandante, buscando oponerse a una moción de desestimación, presentaron un escrito judicial que citaba múltiples decisiones judiciales supuestamente favorables a su posición. Sin embargo, tras una investigación por parte de la defensa y del propio tribunal, se descubrió que al menos seis de los casos citados eran completamente inexistentes, fabricaciones generadas por ChatGPT, la herramienta de IA que uno de los abogados había utilizado para la investigación legal (Weiser, 2023; Dahl et al., 2024).

El Juez P. Kevin Castel, al imponer sanciones a los abogados implicados (incluyendo una multa económica y la obligación de notificar a los jueces cuyos nombres fueron falsamente asociados a las opiniones inventadas), emitió una orden detallada que disecciona los múltiples fallos en el proceso. El abogado que utilizó ChatGPT, Steven A. Schwartz, admitió no ser un experto en investigación legal federal y haber utilizado la herramienta como un "super motor de búsqueda", confiando en sus respuestas e incluso preguntando a ChatGPT si los casos que proporcionaba eran reales, a lo que el chatbot respondió afirmativamente (Orden de Sanciones en \textit{Mata v. Avianca, Inc.}, 22 de junio de 2023).

\textbf{Lecciones aprendidas de \textit{Mata v. Avianca}:}

\begin{enumerate}
    \item \textbf{Verificación independiente como deber ineludible:} La lección más obvia y contundente es la absoluta necesidad de que los abogados verifiquen de forma independiente y rigurosa cada fuente y cada proposición legal generada por una IA antes de incorporarla a un documento judicial. La simple pregunta a la IA sobre la veracidad de su propia salida es manifiestamente insuficiente y denota una falta de comprensión fundamental sobre cómo operan estos modelos.
    \item \textbf{La naturaleza engañosa de las alucinaciones:} Las citaciones fabricadas por ChatGPT en el caso \textit{Mata} eran altamente plausibles, con nombres de partes, números de volumen y página, y resúmenes de \textit{holdings} que imitaban el formato y estilo de las opiniones judiciales reales. Esta plausibilidad hace que las alucinaciones sean particularmente insidiosas y difíciles de detectar sin una verificación cruzada con bases de datos legales canónicas.
    \item \textbf{Responsabilidad profesional individual:} El caso subraya que la responsabilidad final por el contenido de los escritos presentados ante el tribunal recae inequívocamente en el abogado firmante, independientemente de las herramientas utilizadas en su preparación. El uso de IA no diluye ni transfiere esta responsabilidad.
    \item \textbf{Desconocimiento de las limitaciones de la IA:} La admisión del abogado Schwartz de que "no era consciente de la posibilidad de que [el] contenido [de ChatGPT] pudiera ser falso" revela una brecha significativa en la alfabetización sobre IA dentro de la profesión legal. Comprender las limitaciones inherentes de los LLMs, incluyendo su propensión a "alucinar" o "confabular", es un componente esencial de la competencia profesional en la era digital.
    \item \textbf{Impacto en la Integridad del Sistema Judicial:} La introducción de jurisprudencia ficticia en un procedimiento judicial no solo perjudica al cliente y expone al abogado a sanciones, sino que también "promueve el cinismo hacia la profesión legal y el sistema de justicia estadounidense" y constituye un abuso del proceso judicial (Orden de Sanciones en \textit{Mata v. Avianca, Inc.}).
\end{enumerate}

\subsection{Caso de estudio: la sanción del Tribunal Constitucional español y la responsabilidad indelegable del letrado}

\subsubsection{Resumen fáctico del incidente}

En septiembre de 2024, la Sala Primera del Tribunal Constitucional de España marcó un precedente fundamental al sancionar por unanimidad a un abogado por faltar al debido respeto al tribunal (Nota Informativa 90/2024). La falta consistió en la inclusión, en una demanda de amparo, de 19 citas supuestamente literales de sentencias del propio Tribunal que resultaron ser completamente inexistentes. Estas citas se presentaban entrecomilladas, atribuyendo a los magistrados una doctrina constitucional que "carecía de todo anclaje en la realidad". La sanción impuesta fue un "apercibimiento", la menor posible, pero se ordenó dar traslado al Colegio de Abogados de Barcelona para los procedimientos disciplinarios correspondientes.

\subsubsection{Análisis del caso a través del marco del informe}

Este caso sirve como una ilustración perfecta de los conceptos analizados en este informe:
\begin{itemize}
    \item \textbf{Naturaleza de la Alucinación} (Aplicando la Taxonomía de la Sección 2.1): El error cometido encaja directamente en la categoría de "Fabricación de autoridad". No se trataba de una tergiversación sutil (misgrounding), sino de la invención completa de doctrina judicial. La presentación de los pasajes entrecomillados agrava la falta, ya que no se presenta como una interpretación, sino como una cita literal y verificable, lo que constituye una forma particularmente grave de información incorrecta según la tipología de Magesh et al. (2025).
    \item \textbf{La responsabilidad profesional por encima de la herramienta} (aplicando la Sección 8.1): El punto más crucial del Acuerdo del Tribunal es su razonamiento sobre la responsabilidad del abogado. El letrado alegó en su defensa una "desconfiguración de una base de datos". El Tribunal descartó este argumento de forma tajante, estableciendo un principio de responsabilidad absoluta que es independiente de la causa del error. En sus propias palabras, "fuera cual fuese la causa de la inclusión de citas irreales (uso de la inteligencia artificial, entrecomillado de argumentos propios, etcétera), el letrado es siempre responsable de revisar exhaustivamente todo el contenido" (Nota Informativa 90/2024). Esta afirmación es la manifestación práctica más clara del deber de diligencia y competencia profesional en la era de la IA, subrayando que la supervisión humana no es una opción, sino una obligación indelegable.
    \item \textbf{Impacto en la integridad del Sistema Judicial} (aplicando la Sección 2.3): El Tribunal no lo consideró un simple error procesal, sino una falta de respeto que mostraba un "claro desprecio de la función jurisdiccional" de los magistrados. La conducta, según el Acuerdo, perturbó el trabajo del Tribunal no por la necesidad de verificar las citas —algo que se hace siempre— sino por "tener que enjuiciar las consecuencias de tal injustificada irregularidad". Esto demuestra que la introducción de información falsa no solo contamina el debate jurídico, sino que socava la confianza y el respeto mutuo que deben regir la relación entre los abogados y la judicatura, erosionando los cimientos del sistema.
\end{itemize}

\subsubsection{Lección humana: la delegación de la responsabilidad crítica}

Más allá del análisis técnico-jurídico, el caso del Tribunal Constitucional español, al igual que Mata v. Avianca, es un síntoma de una peligrosa tendencia cultural: la delegación del pensamiento crítico. El letrado no falló por usar una herramienta defectuosa; falló porque abdicó de su responsabilidad fundamental de verificar, dudar y pensar. Trató a la IA como un oráculo en lugar de como un asistente. En cualquier profesión, pero especialmente en el derecho, el valor no reside en la capacidad de generar una respuesta, sino en la capacidad de defenderla. Cuando un profesional simplemente copia y pega un resultado que no comprende, no está utilizando la tecnología; está siendo utilizado por ella. Estos incidentes no deberían generar miedo a la IA, sino un profundo respeto por el rol insustituible del juicio humano. La tecnología no nos exime de nuestra obligación de ser excelentes; de hecho, nos la exige con más fuerza que nunca."

\subsubsection{Lecciones clave y comparativa con Mata v. Avianca}

Este caso, aunque similar en su origen a Mata v. Avianca, ofrece lecciones complementarias y de mayor calado ético:
\begin{itemize}
    \item Universalidad del deber de verificación: Confirma que la obligación de verificar cada dato presentado ante un tribunal es un principio universal del ejercicio de la abogacía, aplicable con la misma fuerza en sistemas de derecho civil (España) como de common law (EE. UU.).
    \item Irrelevancia de la causa del error: Mientras que en Mata el debate se centró en el mal uso de una herramienta específica (ChatGPT), el Tribunal Constitucional español eleva el principio: la responsabilidad del abogado es absoluta, sin importar si el error fue causado por una IA, un software defectuoso o un descuido humano. La herramienta es irrelevante; la responsabilidad, total.
    \item De la sanción procesal a la falta ética: El caso español encuadra el problema no solo como una negligencia que merece una sanción procesal, sino como una falta al deber de respeto, un pilar de la ética profesional. Es un "desprecio" a la función judicial, lo que le confiere una gravedad deontológica superior.
\end{itemize}

\subsection{Caso de estudio: \textit{Thackston v. Driscoll} y la cascada de errores algorítmicos}

El caso \textit{Thackston v. Driscoll}, resuelto por un Juez Magistrado en el Distrito Oeste de Texas el 28 de agosto de 2025, se erige como un ejemplo alarmantemente completo de los peligros derivados de un uso acrítico y negligente de la IA generativa en la práctica legal. A diferencia de \textit{Mata v. Avianca}, que se centró principalmente en la fabricación de casos, \textit{Thackston} ilustra una "cascada" de errores que abarca casi toda la gama de la taxonomía de alucinaciones legales.

\subsubsection{Resumen fáctico del incidente}

En el marco de una demanda por discriminación laboral contra el Ejército de EE.UU., el abogado del demandante presentó un escrito de réplica plagado de información legal defectuosa. El tribunal, en su "Informe y Recomendación", diseccionó meticulosamente los errores, que incluían:

\begin{itemize}

    \item \textbf{Fabricación de autoridad:} Citas a casos completamente inexistentes (ej. una supuesta opinión del Noveno Circuito en \textit{United States v. City of Los Angeles} y un caso del propio tribunal en \textit{EEOC v. Exxon Mobil Corp.} que no existía).
    \item \textbf{Citas falsas y tergiversación (\textit{Misgrounding}):} El escrito atribuía citas textuales inventadas a casos reales y conocidos (ej. a \textit{Palmer v. Shultz} y \textit{Armstrong v. Turner Industries}). Además, tergiversaba gravemente los \textit{holdings} de otros casos reales, citándolos en apoyo de proposiciones legales que no sostenían.
    \item \textbf{Error temporal:} Se citó el famoso caso \textit{Chevron}, un pilar del derecho administrativo, sin reconocer que había sido revocado explícitamente por el Tribunal Supremo, un error fáctico y estratégico de primer orden.
\end{itemize}
    
El Juez Magistrado no solo identificó los errores, sino que también sospechó explícitamente del uso de IA por el "lenguaje repetitivo y redundante" y la naturaleza de las invenciones. Concluyó que el abogado violó la Regla Federal 11(b) al no realizar una "investigación razonable" y recomendó al Tribunal de Distrito la imposición de sanciones, sugiriendo una multa y la asistencia obligatoria a un curso de formación sobre IA generativa.

\subsubsection{Análisis del caso a través del marco del informe}

Este caso es un microcosmos perfecto de los riesgos sistémicos discutidos en este informe.

\begin{itemize}
    \item \textbf{Una taxonomía completa en un solo documento} (aplicando la Sección 2.2)\textbf{:} \textit{Thackston} es una clase magistral sobre los diferentes tipos de alucinaciones. Demuestra que un abogado que confía ciegamente en una IA no comete un solo tipo de error, sino que se expone a un fallo sistémico. La combinación de \textbf{fabricación de autoridad}, \textbf{fundamentación errónea (misgrounding)} y \textbf{error temporal} en un mismo escrito muestra la incapacidad del modelo (y del abogado) para distinguir entre lo real, lo tergiversado y lo obsoleto. El \textit{misgrounding} es particularmente insidioso aquí, ya que el abogado podría haber verificado la existencia del caso y haberse detenido ahí, cayendo en una falsa sensación de seguridad.
    \item \textbf{La abdicación del juicio profesional y la "alucinación del usuario" }(aplicando las Secciones 2.3 y 5.6)\textbf{:} El tribunal es inequívoco: la culpa no es de la "máquina", sino del profesional que abdicó de su deber fundamental. La recomendación del Juez Magistrado se centra en la violación de la Regla 11, que exige una investigación razonable \textit{antes} de presentar cualquier documento. Este es el ejemplo paradigmático de la "alucinación del usuario": la creencia errónea de que la herramienta puede sustituir la diligencia, el escepticismo y el juicio profesional. Refuerza el principio central de este informe: el rol de la supervisión humana no es solo una buena práctica, es una obligación legal y ética irreductible.
    \item \textbf{De la teoría a las consecuencias reales }(aplicando la Sección 8)\textbf{:} El caso \textit{Thackston} materializa las implicaciones deontológicas y regulatorias. La recomendación de sanciones monetarias y formación obligatoria no es una reprimenda abstracta, sino una consecuencia profesional y económica directa. Sirve como una advertencia potente, alineada con las lecciones de \textit{Mata} y del caso del Tribunal Constitucional español: los tribunales no dudarán en utilizar sus facultades sancionadoras para proteger la integridad del proceso judicial frente a la introducción de información falsa, sin importar la tecnología utilizada para generarla.
\end{itemize}

\subsection{Incidentes más allá de la fabricación flagrante: errores sutiles y \textit{Misgrounding}}

El estudio de Magesh et al. (2025) documenta que las alucinaciones más frecuentes y peligrosas en sistemas RAG no son las fabricaciones completas, sino errores de razonamiento legal más sutiles, que ellos denominan "insidiosos". Identifican tres categorías principales de fallo:
\begin{itemize}
    \item Incomprensión de los holdings (Misunderstanding Holdings): Las herramientas a menudo resumen una sentencia afirmando lo contrario de lo que el tribunal realmente decidió, confundiendo el holding (la decisión central) con los dicta (comentarios secundarios).
    \item Confusión entre actores legales (Distinguishing Between Legal Actors): Las IAs atribuyen erróneamente los argumentos de un litigante al tribunal, presentando la postura de una de las partes como si fuera la decisión final del juez.
    \item Falta de respeto a la jerarquía de autoridad (Respecting the Order of Authority): Los modelos demuestran una incapacidad para comprender la jerarquía judicial, por ejemplo, afirmando que un tribunal inferior revocó una decisión de un tribunal superior, lo cual es legalmente imposible.
\end{itemize}
Estos errores de misgrounding son particularmente peligrosos porque la presencia de una cita real crea una falsa sensación de autoridad y fiabilidad, lo que puede llevar al abogado a confiar indebidamente en la proposición sin realizar la necesaria lectura crítica de la fuente (Magesh et al., 2025).

Si bien la fabricación completa de casos como en \textit{Mata v. Avianca} es la forma más espectacular de alucinación, los estudios empíricos sobre herramientas legales comerciales basadas en RAG, como el de Magesh et al. (2024), revelan que formas más sutiles pero igualmente problemáticas de error son aún más frecuentes. Estos incidentes, aunque no siempre conllevan sanciones tan publicitadas, pueden tener un impacto significativo en la calidad del trabajo legal y en la toma de decisiones.

Un ejemplo recurrente documentado por Magesh et al. (2024) es el \textit{misgrounding}, donde la herramienta de IA cita un caso o estatuto \textit{real y existente}, pero la proposición legal que atribuye a esa fuente es incorrecta, tergiversada o simplemente no está contenida en el texto original de la autoridad citada. En una de las instancias analizadas, Westlaw AI-Assisted Research afirmó incorrectamente el \textit{holding} de una decisión de la Corte Suprema de EE. UU., atribuyéndole una conclusión opuesta a la que realmente alcanzó el tribunal. En otro ejemplo, Lexis+ AI describió un caso (Arturo D.) como autoridad vigente y lo utilizó para respaldar una proposición, cuando en realidad el caso citado (Lopez) había \textit{revocado} a Arturo D. en el punto relevante.

\textbf{Lecciones aprendidas de errores de \textit{Misgrounding} y similares:}

\begin{enumerate}

    \item \textbf{La verificación no puede limitarse a la existencia de la cita:} A diferencia de las fabricaciones obvias, el \textit{misgrounding} requiere un nivel de verificación más profundo. No basta con confirmar que el caso o estatuto citado existe; el abogado debe leer y comprender la fuente original para asegurar que realmente respalda la afirmación hecha por la IA.
    \item \textbf{Fragilidad de la comprensión contextual de la IA-RAG:} Estos errores sugieren que, incluso cuando se les proporciona el contexto recuperado, los LLMs pueden tener dificultades para interpretar correctamente los matices del lenguaje legal, distinguir el \textit{holding} de los \textit{dicta}, o comprender las relaciones jerárquicas y temporales entre precedentes (p. ej., el efecto de una revocación).
    \item \textbf{El riesgo de la "falsa fundamentación":} El \textit{misgrounding} es particularmente peligroso porque la presencia de una cita real puede crear una falsa sensación de autoridad y fiabilidad, llevando al abogado a confiar indebidamente en la proposición sin realizar la necesaria lectura crítica de la fuente.
    \item \textbf{Necesidad de optimización profunda en RAG:} Estos incidentes refuerzan la conclusión del informe de Addleshaw Goddard (2024) de que la efectividad de RAG depende de una optimización meticulosa de cada componente, incluyendo no solo la recuperación sino también la forma en que el LLM generador es instruido (prompting) para interactuar con el contexto recuperado y razonar sobre él.
    \end{enumerate}

\subsection{Implicaciones globales y la necesidad de adaptación continua}

Aunque muchos de los casos más notorios han surgido en EE. UU., el problema de las alucinaciones de la IA y la necesidad de una verificación diligente por parte de los abogados es una preocupación global. Incidentes similares han comenzado a documentarse en otras jurisdicciones, incluyendo Australia (donde un abogado fue remitido a un órgano disciplinario por usar IA que generó citas falsas en un caso de familia - The Guardian, 2024, citado en Lantyer, 2024) y Brasil (donde un abogado fue multado por usar jurisprudencia falsa generada por IA en una apelación - Migalhas, 2025, citado en Lantyer, 2024).

Más allá de los incidentes judiciales anecdóticos, la investigación académica sistemática confirma este riesgo a nivel global. Un estudio empírico enfocado en una jurisdicción fuera del ámbito europeo comparó el rendimiento de múltiples LLMs con un abogado humano, concluyendo que la tarea de investigación legal era la de peor rendimiento, con una tendencia constante de los modelos de IA a inventar jurisprudencia (Hemrajani, 2025).

\textbf{Lecciones aprendidas de la perspectiva global:}

\begin{enumerate}
    \item \textbf{Universalidad del riesgo:} La propensión de los LLMs a alucinar no está limitada por fronteras geográficas o sistemas jurídicos. Es una característica inherente de la tecnología actual que afecta a todos los profesionales que la utilizan.
    \item \textbf{Adaptación de los deberes deontológicos:} Si bien los detalles de los códigos deontológicos varían entre jurisdicciones, los principios fundamentales de competencia, diligencia, lealtad al cliente y franqueza ante el tribunal son ampliamente compartidos. La profesión legal en cada país deberá interpretar y aplicar estos principios al nuevo contexto de la IA.
    \item \textbf{Respuesta de órganos disciplinarios y judiciales:} La forma en que los tribunales y los órganos disciplinarios de diferentes países respondan a estos incidentes establecerá precedentes importantes y modelará las expectativas sobre el uso responsable de la IA por parte de los abogados. Las sanciones impuestas en casos como \textit{Mata} sirven como una señal clara de la seriedad con la que se toman estos fallos.
    \item \textbf{Importancia de la formación y la alfabetización en IA:} A nivel global, existe una necesidad urgente de mejorar la formación de los abogados sobre las capacidades y limitaciones de la IA, incluyendo la concienciación sobre el riesgo de alucinaciones y el desarrollo de habilidades de verificación crítica.
\end{enumerate}

En conclusión, los estudios de caso de incidentes reales donde las alucinaciones de la IA han impactado procedimientos judiciales ofrecen lecciones contundentes e ineludibles. Subrayan que la integración de la IA en la práctica legal no es un proceso exento de riesgos y que la tecnología, en su estado actual, no puede sustituir el juicio crítico, la diligencia investigadora y la responsabilidad ética del profesional humano. Estos casos no deben interpretarse como una condena de la IA \textit{per se}, sino como un llamado urgente a la cautela, a la verificación rigurosa y al desarrollo de prácticas profesionales y salvaguardas tecnológicas que permitan aprovechar el potencial de la IA minimizando sus peligros inherentes. La "realidad cruda" de estos incidentes debe servir como un motor para la mejora continua, tanto de la tecnología como de la forma en que la profesión legal interactúa con ella.

\section{El futuro de la Inteligencia Artificial legal fiable: hacia modelos explicables, auditables y responsables por diseño}
\label{sec:futuro_ia_fiable}

El panorama actual de la inteligencia artificial (IA) aplicada al derecho, aunque rebosante de potencial transformador, se encuentra marcado por el desafío persistente de las alucinaciones y las limitaciones inherentes a la fiabilidad de los Grandes Modelos de Lenguaje (LLMs) y las arquitecturas de Generación Aumentada por Recuperación (RAG) convencionales. Las secciones precedentes han diseccionado la naturaleza de estos errores, sus causas raíz y las estrategias de mitigación disponibles. Sin embargo, una visión a largo plazo exige ir más allá de la mera contención de los problemas actuales y proyectar un futuro donde la IA legal no solo sea más potente y eficiente, sino fundamentalmente más \textbf{fiable, transparente y alineada con los principios éticos y las exigencias de rendición de cuentas} inherentes al sistema de justicia. Este futuro no dependerá de un único avance disruptivo, sino de la convergencia de múltiples líneas de investigación y desarrollo enfocadas en la creación de modelos inherentemente más explicables (\textbf{XAI}, o Inteligencia Artificial Explicable), sistemas técnicamente auditables y, de manera crucial, la adopción de un paradigma de \textbf{IA responsable por diseño} (\textit{Responsible AI by Design}). Esta sección explora estas trayectorias prospectivas, delineando los contornos de una IA legal que pueda aspirar a ser un colaborador verdaderamente confiable para el profesional del derecho y un instrumento equitativo en la administración de justicia.

\subsection{La búsqueda de la explicabilidad (XAI) en el contexto legal}

Uno de los mayores obstáculos para la confianza y la adopción generalizada de los LLMs en tareas legales críticas es su naturaleza de "caja negra". Generan respuestas, a menudo complejas y matizadas, pero rara vez ofrecen una justificación inteligible de \textit{cómo} llegaron a esas conclusiones o \textit{en qué} información específica (y con qué ponderación) se basaron. En un dominio como el derecho, donde la capacidad de argumentar, justificar y trazar el razonamiento hasta las fuentes autorizadas es esencial, esta opacidad es profundamente problemática. La Inteligencia Artificial Explicable (XAI) emerge como un campo de investigación vital para abordar este desafío.

Las técnicas actuales de explicabilidad para LLMs (p. ej., análisis de atención, importancia de características, generación de justificaciones textuales post-hoc) a menudo proporcionan solo una visión superficial o aproximada del proceso de toma de decisiones interno del modelo. Estas explicaciones pueden ser ellas mismas susceptibles de "alucinar" o pueden no reflejar fielmente los factores causales reales que llevaron a una salida particular (Rudin, 2019; Lipton, 2018).

\subsubsection{El futuro de la XAI jurídica: de la fundamentación a la interpretación razonada}

El futuro de la XAI legal no reside tanto en el desarrollo de explicaciones \textit{post-hoc} para modelos opacos, sino en la evolución hacia arquitecturas de IA que incorporen la explicabilidad de forma nativa y significativa para el profesional del derecho. La verdadera revolución no se medirá por la capacidad de un sistema para superar un examen, sino por su habilidad para justificar sus conclusiones con transparencia y responsabilidad. Esta evolución puede conceptualizarse en tres niveles progresivos de madurez:

    \begin{itemize}
    \item \textbf{Respuestas fundamentadas (\textit{Grounded Responses}):} Este es el nivel básico y el prerrequisito indispensable, centrado en la trazabilidad. La IA debe ser capaz de anclar cada afirmación en una fuente autorizada y verificable. El objetivo es responder a la pregunta: \textit{"¿De dónde procede esta información?"}. Sin una fundamentación sólida, cualquier resultado carece de la fiabilidad mínima necesaria para su uso profesional.
    \item \textbf{Respuestas argumentadas (\textit{Argued Responses}):} El siguiente nivel trasciende la mera cita de fuentes para articular el razonamiento. No basta con saber \textit{qué} fuente se usó, sino \textit{cómo} se usó para construir la conclusión. La IA debe ser capaz de externalizar los pasos lógicos de su inferencia, demostrando una cadena de razonamiento coherente. El objetivo es responder a la pregunta: \textit{"¿Cómo has llegado a esta conclusión a partir de las fuentes?"}.
    \item \textbf{Respuestas basadas en interpretación razonada (\textit{Reasoned Interpretation}):} Este es el nivel más avanzado y el verdadero horizonte de la IA jurídica. Aquí, la IA no solo aplica una regla, sino que es capaz de explicar \textit{por qué} opta por una interpretación específica frente a otras alternativas plausibles. Implica ponderar matices, reconocer ambigüedades y justificar su aplicación de la norma a un contexto fáctico concreto. El objetivo es responder a la pregunta: \textit{\textbf{"¿Por qué esta es la interpretación o aplicación más adecuada en este caso?"}}.
    \end{itemize}

Alcanzar este tercer nivel sigue siendo un desafío formidable, ya que la interpretación genuina requiere principios, ética y un entendimiento del contexto que los modelos actuales no poseen. Sin embargo, el desarrollo de arquitecturas de IA que sean intrínsecamente más interpretables es crucial para avanzar en este camino. Esto implica:

    \begin{itemize}
    \item \textbf{Modelos que externalizan el razonamiento jurídico:} Como se discutió con los modelos de razonamiento (Sección 5.4), la IA que puede articular sus pasos de inferencia de una manera que se asemeje a un análisis legal humano (identificando hechos relevantes, aplicando reglas, ponderando factores, citando autoridades para cada paso) será inherentemente más explicable y verificable, avanzando hacia el nivel de \textbf{argumentación}.
    \item \textbf{Visualización de la influencia de las fuentes en RAG:} En sistemas RAG, mejorar la capacidad de rastrear qué fragmentos específicos del contexto recuperado contribuyeron y con qué peso a cada parte de la respuesta generada. Herramientas de visualización que muestren estas conexiones podrían aumentar drásticamente la interpretabilidad y la \textbf{fundamentación}.
    \item \textbf{Explicaciones contrastantes y contrafácticas:} Desarrollar modelos capaces de explicar no solo por qué llegaron a una conclusión, sino también por qué no llegaron a otras conclusiones alternativas, o cómo cambiaría la conclusión si ciertos hechos o premisas fueran diferentes. Esto se alinea estrechamente con la forma en que los abogados analizan los problemas y es un paso clave hacia la \textbf{interpretación razonada}.
    \item \textbf{Gestión de la tensión entre explicabilidad y rendimiento:} La complejidad del razonamiento jurídico y la multiplicidad de factores que pueden influir en una decisión legal hacen que la explicabilidad completa sea un objetivo extremadamente ambicioso. Se debe reconocer y gestionar activamente la tensión potencial que existe entre la explicabilidad y el rendimiento del modelo: los sistemas más precisos a menudo son los más opacos. El futuro de la XAI legal radicará en encontrar un equilibrio óptimo donde la justificación del resultado sea suficientemente robusta para la validación profesional, sin sacrificar de manera inaceptable la eficacia del sistema.
    \end{itemize}

\subsection{La necesidad de auditabilidad técnica y de gobernanza}

La fiabilidad y la responsabilidad en la IA legal no pueden depender únicamente de la buena fe de los desarrolladores o de la diligencia de los usuarios individuales. Se necesitan mecanismos robustos para la \textbf{auditoría independiente y continua} de estos sistemas, tanto a nivel técnico como de gobernanza.

\begin{enumerate}
    \item \textbf{Auditoría técnica de los modelos y sistemas RAG:}

    \begin{itemize}

    \item \textit{Desarrollo de estándares y métricas de auditoría específicos para IA legal:} Se requieren benchmarks y métricas estandarizadas (como los discutidos en la Sección 3) que vayan más allá de la precisión general y evalúen específicamente la propensión a las alucinaciones, la robustez ante entradas adversarias o ambiguas, la equidad (\textit{fairness}) respecto a diferentes grupos, y la calidad de la fundamentación en sistemas RAG.
    \item \textit{Herramientas de auditoría automatizada y asistida por IA:} Desarrollar herramientas que puedan asistir a los auditores humanos en la evaluación a gran escala de los modelos, por ejemplo, generando automáticamente casos de prueba desafiantes, identificando posibles sesgos en los datos de entrenamiento o en las respuestas, o verificando la consistencia de las citas.
    \item \textit{Acceso controlado para auditoría (}Sandboxing\textit{):} Los reguladores o entidades de certificación independientes podrían requerir acceso a los modelos (posiblemente en entornos controlados o "sandboxes") para realizar pruebas exhaustivas antes de su despliegue en aplicaciones de alto riesgo.
    \end{itemize}

    \item \textbf{Auditoría de gobernanza de datos y procesos:} Más allá del modelo en sí, es crucial auditar los procesos y las prácticas de gobernanza de datos de las organizaciones que desarrollan e implementan IA legal.

    \begin{itemize}
    \item \textit{Trazabilidad de los datos de entrenamiento:} Asegurar que se mantengan registros detallados sobre las fuentes, la curación y el pre-procesamiento de los datos utilizados para entrenar los modelos, permitiendo investigar posibles sesgos o errores.
    \item \textit{Evaluación de impacto algorítmico y ético:} Requerir que las organizaciones realicen evaluaciones de impacto rigurosas antes de desplegar sistemas de IA en contextos legales sensibles, identificando y mitigando proactivamente los riesgos potenciales.
    \item \textit{Mecanismos de supervisión humana y rendición de cuentas:} Auditar la efectividad de los mecanismos de supervisión humana implementados y asegurar que existan canales claros para la rendición de cuentas y la reparación en caso de errores o daños causados por la IA.
    \end{itemize}
\end{enumerate}

La auditabilidad no es solo una cuestión técnica, sino también una exigencia de buena gobernanza y un prerrequisito para generar confianza pública en la IA legal.

\subsection{IA responsable por diseño (\textit{Responsible AI by Design}) en el ámbito legal}

El enfoque más proactivo y, en última instancia, más efectivo para construir una IA legal fiable es adoptar un paradigma de \textbf{Inteligencia Artificial Responsable por Diseño}. Esto implica integrar consideraciones éticas, de equidad, transparencia, robustez y fiabilidad \textbf{desde las primeras etapas del ciclo de vida del desarrollo de la IA}, en lugar de tratar estos aspectos como correcciones o parches aplicados a posteriori.

Este paradigma de 'Responsabilidad por Diseño' no es solo una aspiración ética o una buena práctica de ingeniería, sino que se está convirtiendo progresivamente en una expectativa regulatoria y, en algunos casos, en una obligación legal explícita. La Ley de IA de la UE (el Reglamento), a través de su detallado catálogo de requisitos para los sistemas de IA de alto riesgo —que abarcan desde la gestión de riesgos (\textbf{Artículo 9}) y la gobernanza de los datos de entrenamiento (\textbf{Artículo 10}) hasta la necesidad de una supervisión humana efectiva (\textbf{Artículo 14}) y la robustez técnica (\textbf{Artículo 15})—, esencialmente codifica muchos de los principios fundamentales de la IA responsable. Al exigir estas consideraciones desde las fases de diseño y desarrollo, y a lo largo de todo el ciclo de vida del sistema, la EU-AIAct impulsa a los creadores de IA legal a ir más allá de la simple funcionalidad para priorizar la seguridad, la fiabilidad y la protección de los derechos fundamentales, donde la prevención de resultados perjudiciales derivados de alucinaciones se convierte en un objetivo central del diseño.

\begin{enumerate}
    \item \textbf{Principios fundamentales de la IA legal responsable por diseño:}

Estos principios no son meras aspiraciones teóricas, sino que encuentran un eco directo y una validación institucional en los marcos normativos emergentes. Estos principios no son meras aspiraciones teóricas, sino que encuentran un eco directo y una validación institucional en los marcos normativos emergentes. Un ejemplo paradigmático es la ya mencionada Política del CTEAJE en España. Principios establecidos en este documento de obligado cumplimiento, como el de 'No sustitución' o el mandato de 'Revisión Humana Universal' que establece, son la materialización práctica del enfoque que se detalla a continuación, demostrando que la IA responsable por diseño está pasando de ser una buena práctica a una exigencia regulatoria.

    \begin{itemize}
    \item \textit{Centrada en el ser humano:} Diseñar sistemas de IA que sirvan para aumentar y asistir al profesional legal, no para reemplazar su juicio crítico o su responsabilidad ética. El objetivo es la colaboración humano-IA, no la automatización completa de tareas complejas.
    \item \textit{Fiabilidad y seguridad como prioridad:} La precisión factual, la robustez ante errores y la seguridad de los datos deben ser consideraciones primordiales en el diseño y la optimización de los modelos, incluso si esto implica ciertos compromisos en términos de fluidez o velocidad de generación.
    \item \textit{Equidad y no discriminación (\textit{Fairness}):} Esforzarse activamente por identificar y mitigar los sesgos algorítmicos que podrían conducir a resultados discriminatorios o inequitativos en la aplicación de la ley. Esto requiere un análisis cuidadoso de los datos de entrenamiento y de los impactos diferenciales del modelo.
    \item \textit{Transparencia y explicabilidad (contextualizadas):} Diseñar sistemas que sean lo más transparentes y explicables posible dentro de las limitaciones técnicas, proporcionando a los usuarios información significativa sobre cómo funcionan y por qué generan ciertas respuestas.
    \item \textit{Rendición de cuentas y gobernanza:} Establecer estructuras claras de gobernanza y rendición de cuentas para el desarrollo, despliegue y mantenimiento de los sistemas de IA legal.
    \end{itemize}
    
    \item \textbf{Metodologías prácticas para la IA legal responsable por diseño:}

    \begin{itemize}
    \item \textit{Equipos de desarrollo multidisciplinares:} Involucrar a juristas, éticos y expertos en el dominio legal desde el inicio del proceso de diseño, no solo a ingenieros y científicos de datos.
    \item \textit{Evaluación continua de riesgos y pruebas adversarias:} Implementar ciclos iterativos de evaluación de riesgos y pruebas de estrés (incluyendo pruebas específicas para detectar alucinaciones y sesgos) a lo largo de todo el desarrollo.
    \item \textit{Mecanismos de retroalimentación y mejora continua:} Diseñar sistemas que puedan aprender y mejorar a partir de la retroalimentación de los usuarios expertos y de la monitorización de su rendimiento en el mundo real.
    \item \textit{Adopción de estándares éticos y técnicos emergentes:} Mantenerse al día y adherirse a los estándares éticos y técnicos, así como a las mejores prácticas, que están siendo desarrollados por la comunidad investigadora, los organismos profesionales y los reguladores.
    \end{itemize}
\end{enumerate}

La IA Responsable por Diseño no es un estado final, sino un compromiso continuo con la mejora y la adaptación. Requiere una cultura organizacional que priorice la ética y la fiabilidad, y una voluntad de invertir en los recursos necesarios para construir sistemas que sean verdaderamente dignos de confianza en el sensible contexto legal.

\subsection{La simbiosis avanzada humano-IA: colaboración y juicio aumentado}

Mirando aún más hacia el futuro, la IA legal más fiable y efectiva probablemente no será aquella que intente reemplazar completamente al abogado, sino aquella que logre una \textbf{simbiosis avanzada y sinérgica con la inteligencia humana experta}. En este modelo, la IA no es meramente una herramienta pasiva, sino un colaborador activo que aumenta y refina las capacidades del profesional del derecho.

\begin{itemize}
    \item \textbf{IA como "Investigador incansable y verificador preliminar":} La IA podría encargarse de la búsqueda exhaustiva y el análisis preliminar de grandes volúmenes de información legal, identificando patrones, recuperando precedentes relevantes y señalando posibles problemas o inconsistencias, pero siempre presentando sus hallazgos al abogado para su validación y juicio estratégico.
    \item \textbf{IA como "Generador de hipótesis y argumentos alternativos":} En lugar de proporcionar una única "respuesta", la IA podría generar múltiples líneas argumentales, interpretaciones o soluciones posibles para un problema legal, cada una con su fundamentación y sus posibles debilidades, permitiendo al abogado explorar un espectro más amplio de opciones estratégicas.
    \item \textbf{IA como "Traductor contextual y puente de comunicación":} Una de las barreras más significativas en la práctica legal es la asimetría de información entre el abogado y el cliente, a menudo causada por la complejidad del lenguaje jurídico. Un sistema de IA avanzado, en lugar de ser una herramienta de uso exclusivo para el profesional, puede actuar como un traductor contextual, generando resúmenes o explicaciones de documentos y estrategias legales en un lenguaje adaptado al nivel de comprensión del cliente. Este enfoque, ejemplificado por la \textbf{"función Jerga" del proyecto Justicio} analizada previamente, no solo mejora la transparencia y la confianza en la relación cliente-abogado, sino que también capacita al cliente para tomar decisiones más informadas, humanizando el acceso a la justicia.
    \item \textbf{IA como "Entrenador personalizado y asistente de aprendizaje":} La IA podría proporcionar retroalimentación detallada y personalizada sobre el trabajo de los abogados en formación, ayudándoles a identificar áreas de mejora en su investigación, redacción y razonamiento, siempre bajo la supervisión de mentores humanos.
    \item \textbf{Interfaces colaborativas intuitivas:} El desarrollo de interfaces de usuario que permitan una interacción fluida, iterativa y verdaderamente colaborativa entre el abogado y el sistema de IA será crucial. El abogado debe poder guiar, cuestionar y refinar fácilmente el trabajo de la IA.
\end{itemize}    

Este futuro de colaboración aumentada requiere no solo avances en la tecnología de IA, sino también una evolución en la formación y las habilidades de los profesionales legales, quienes necesitarán ser competentes tanto en el derecho como en la interacción crítica y efectiva con estos sistemas inteligentes.

En conclusión, el camino hacia una IA legal verdaderamente fiable y beneficiosa es complejo y está en constante evolución. Si bien las alucinaciones y otros riesgos son desafíos significativos, no son insuperables. A través de un compromiso sostenido con la investigación en explicabilidad, el desarrollo de mecanismos robustos de auditabilidad, la adopción de principios de responsabilidad por diseño y, fundamentalmente, el reconocimiento del valor insustituible de la supervisión y el juicio humano, es posible vislumbrar un futuro donde la IA se convierta en un aliado poderoso y confiable en la búsqueda de una justicia más eficiente, accesible y equitativa. La tarea no es trivial, pero las recompensas potenciales para la profesión legal y la sociedad en su conjunto son inmensas.

\section{Navegando la frontera ética y regulatoria: implicaciones de las alucinaciones de la IA en el contexto legal global, con énfasis en el Derecho Español y Europeo}
\label{sec:etica_regulacion}

La irrupción de los Grandes Modelos de Lenguaje (LLMs) y su inherente propensión a generar "alucinaciones" –resultados textuales que, aunque a menudo fluidos y convincentes, carecen de veracidad fáctica o fundamento legal– no es meramente un desafío técnico. Este fenómeno penetra profundamente en el tejido de la profesión jurídica, interpelando sus fundamentos éticos y planteando interrogantes críticos sobre la adecuación de los marcos regulatorios existentes a nivel global. Mientras que gran parte del debate inicial y la jurisprudencia temprana sobre sanciones por el uso indebido de IA en litigios ha emanado del sistema del \textit{common law} estadounidense (ejemplificado por casos como \textit{Mata v. Avianca, Inc.}), las implicaciones éticas y la necesidad de una respuesta regulatoria son universales, aunque su manifestación y las soluciones propuestas deban necesariamente adaptarse a las particularidades de cada ordenamiento jurídico. Esta sección se adentra en el complejo panorama de las obligaciones deontológicas y los desafíos regulatorios que las alucinaciones de la IA legal plantean, con una atención particular a las realidades y perspectivas del derecho español y el marco normativo europeo, sin dejar de lado las lecciones aprendidas de otras jurisdicciones.

\subsection{El imperativo deontológico en la era de la IA: reafirmando los deberes fundamentales del abogado en España y Europa}

Los deberes deontológicos tradicionales de competencia y diligencia están cristalizando en un nuevo pilar ético para la era digital: la "Obligación de Competencia Tecnológica". Como se resume en análisis exhaustivos sobre la ética de los LLMs en la abogacía, esta obligación no solo exige entender los beneficios de la tecnología, sino también sus riesgos y limitaciones, incluyendo una comprensión fundamental de fenómenos como las alucinaciones (Shao et al., 2025). Cumplir con este deber implica la verificación rigurosa de los resultados generados por la IA y mantener una supervisión crítica, reconociendo que la responsabilidad final del trabajo recae inequívocamente en el profesional humano.

Los códigos deontológicos que rigen la abogacía en España (como el Código Deontológico de la Abogacía Española) y en el ámbito europeo (como el Código Deontológico de los Abogados Europeos del CCBE), al igual que sus contrapartes estadounidenses, establecen una serie de deberes fundamentales que, si bien no fueron concebidos con la IA generativa en mente, son directamente aplicables y adquieren una nueva dimensión ante el riesgo de alucinaciones.

\begin{enumerate}
    \item \textbf{Deber de competencia profesional:} Este es, quizás, el deber más inmediatamente interpelado. La competencia profesional exige no solo el conocimiento sustantivo del derecho aplicable, sino también la habilidad para utilizar adecuadamente las herramientas y tecnologías que se emplean en el ejercicio profesional. En la era de la IA, esto se traduce en:
    \begin{itemize}
        \item \textit{Alfabetización en IA y comprensión de sus limitaciones:} Un abogado competente en España o Europa no puede permitirse ignorar el funcionamiento básico de los LLMs, su naturaleza probabilística y, crucialmente, su potencial para generar alucinaciones. Esto no implica ser un experto en IA, sino poseer una comprensión funcional suficiente para evaluar críticamente sus resultados y los riesgos asociados (Yamane, 2020; Choi and Schwarcz, 2024).
        \item \textit{Deber de verificación rigurosa:} La competencia exige que el abogado verifique de forma independiente la exactitud y pertinencia de cualquier información o borrador generado por una IA antes de utilizarlo en el asesoramiento al cliente o en actuaciones ante los tribunales. Confiar ciegamente en la salida de un LLM, especialmente en asuntos de alta complejidad o riesgo, podría constituir una grave falta de competencia. Las guías emergentes de colegios de abogados europeos y españoles previsiblemente enfatizarán este punto.
        \item \textit{Conciencia del sesgo de automatización:} Tal como en otros contextos, los profesionales del derecho en España y Europa deben ser conscientes del "sesgo de automatización" y mantener un escepticismo profesional saludable, resistiendo la tentación de delegar el juicio crítico a la máquina, por muy eficiente que esta parezca (Drabiak et al., 2023).

        Este "sesgo de automatización" no es una mera tendencia a la confianza ciega; se manifiesta a través de mecanismos cognitivos específicos que han sido cuantificados. Los LLMs introducen sesgos de encuadre en más de un 20\% de los casos, cambiando la valencia emocional o el énfasis de la información sin alterar los hechos subyacentes (Alessa et al., 2025). Un profesional del derecho que interactúa con estos resultados puede ver su percepción de un caso sutilmente moldeada antes de haber formado su propio juicio independiente. Por lo tanto, el escepticismo profesional no es solo una buena práctica, sino una salvaguarda cognitiva esencial.

    \end{itemize}
    \item \textbf{Deber de diligencia:} Estrechamente ligado a la competencia, el deber de diligencia obliga al abogado a actuar con el cuidado y la atención necesarios en la defensa de los intereses del cliente. En el contexto de la IA y las alucinaciones:
    \begin{itemize}
        \item \textit{Verificación como parte de la diligencia:} La promesa de eficiencia de la IA no puede ir en detrimento de la calidad y la corrección del trabajo. Un abogado diligente debe invertir el tiempo necesario para validar la información generada por la IA, asegurando que cualquier uso de la misma se base en información verificada y legalmente sólida. La "rapidez" no puede justificar la "precipitación negligente".
        \item \textit{Actualización continua:} Dada la rápida evolución de la tecnología IA, la diligencia también puede implicar un deber de mantenerse razonablemente informado sobre los avances, los nuevos riesgos identificados (como tipos específicos de alucinaciones) y las mejores prácticas para el uso de estas herramientas.
    \end{itemize}

    La necesidad de una verificación rigurosa no es una recomendación abstracta, sino una exigencia derivada de la evidencia empírica. Con tasas de alucinación documentadas que oscilan entre el 17\% y más del 33\% en las principales herramientas comerciales de IA legal, confiar ciegamente en sus resultados constituye una clara abdicación del deber de competencia (Magesh et al., 2025). Como concluye el estudio, los abogados se enfrentan a una difícil elección: "verificar a mano cada proposición y cita producida por estas herramientas (socavando así las ganancias de eficiencia prometidas), o arriesgarse a usar estas herramientas sin información completa sobre sus riesgos específicos (descuidando así sus deberes centrales de competencia y supervisión)".

    \item \textbf{Secreto profesional y protección de datos:} Este es un área de particular sensibilidad en el contexto europeo y español, dada la robusta normativa en materia de protección de datos (Reglamento General de Protección de Datos - RGPD).
    \begin{itemize}
        \item \textit{Confidencialidad de la información del cliente:} Introducir información confidencial o datos personales de clientes en plataformas de IA, especialmente aquellas cuyos servidores están fuera de la UE o cuyas políticas de uso de datos no son transparentes o no cumplen con el RGPD, representa un riesgo significativo. Las alucinaciones no son el riesgo directo aquí, pero la elección de la herramienta y la gestión de los datos son cruciales.
        \item \textit{Cumplimiento del RGPD:} Cualquier tratamiento de datos personales a través de una IA debe cumplir con los principios del RGPD (licitud, lealtad, transparencia, limitación de la finalidad, minimización de datos, exactitud, limitación del plazo de conservación, integridad y confidencialidad, y responsabilidad proactiva). Los proveedores y usuarios de IA legal deben poder demostrar este cumplimiento.
    \end{itemize}
    \item \textbf{Lealtad e independencia:} El abogado debe lealtad a su cliente y mantener su independencia de criterio. Si una herramienta de IA sugiere una línea de actuación basada en información alucinada o sesgada, el abogado debe ejercer su juicio independiente para desestimarla si no sirve a los mejores intereses del cliente o contraviene la ley.
    \item \textbf{Deber de lealtad procesal y colaboración con la Administración de Justicia:} En muchos sistemas de derecho civil como el español, existe un fuerte énfasis en la buena fe y la lealtad procesal. Presentar ante un tribunal argumentos, pruebas o jurisprudencia que se sabe (o se debería saber tras una verificación diligente) que son falsos o fabricados por una IA, constituiría una grave violación de estos deberes, con posibles consecuencias disciplinarias y procesales. La integridad del sistema judicial depende de la fiabilidad de la información presentada por las partes.
\end{enumerate}
El impacto de las alucinaciones de la IA en estos deberes deontológicos es innegable y exige una reflexión profunda por parte de los colegios profesionales, los órganos disciplinarios y cada abogado individualmente.

\subsection{Desafíos y perspectivas regulatorias en España y la Unión Europea}

El panorama regulatorio para la IA, y específicamente para la IA legal y sus riesgos de alucinación, está en plena efervescencia, con la Unión Europea a la vanguardia a través de su propuesta de Ley de Inteligencia Artificial (EU-AIAct).

\subsubsection{\textbf{La Ley de IA de la UE: un marco jerárquico y basado en riesgos para la gobernanza de la IA legal}}

A la vanguardia de los esfuerzos globales por establecer un marco normativo integral para la inteligencia artificial se encuentra la Unión Europea con su Ley de IA de la UE. Esta ambiciosa pieza legislativa, de alcance potencialmente global debido al conocido "efecto Bruselas", adopta un \textbf{enfoque estratificado y basado en el riesgo}, clasificando los sistemas de IA en categorías que van desde un riesgo inaceptable (y, por tanto, prohibidos) hasta un riesgo mínimo, pasando por categorías de riesgo limitado y, crucialmente para muchas aplicaciones legales, \textbf{alto riesgo}. Es esta categoría de 'alto riesgo' la que impone las obligaciones más significativas a los desarrolladores, proveedores y, en ciertos casos, usuarios de sistemas de IA (European Union, 2024; Hitaj et al., 2023; Petit \& De Cooman, 2020).

La determinación de si una herramienta específica de IA legal cae dentro de la categoría de 'alto riesgo' dependerá de su finalidad prevista y del contexto de su uso, según lo detallado en el Anexo III de la EU-AIAct. Áreas explícitamente mencionadas como de alto riesgo que tienen una clara tangencia con el sector legal incluyen los sistemas de IA utilizados en la \textbf{administración de justicia y los procesos democráticos}, así como aquellos empleados para la evaluación de la solvencia crediticia o la selección en procesos de contratación, que a menudo involucran análisis de perfiles con implicaciones legales. Es razonable argumentar que herramientas de IA que asistan en la toma de decisiones judiciales, en la evaluación de la admisibilidad de pruebas, en la predicción de reincidencia, o incluso sistemas de investigación legal muy avanzados cuya salida errónea pueda tener un impacto directo y significativo en los derechos fundamentales de un individuo (p. ej., en un proceso penal o en la determinación de la custodia de un menor) podrían ser clasificados como de alto riesgo.

Para estos sistemas de IA de alto riesgo, la EU-AIAct establece un conjunto exhaustivo de requisitos obligatorios que deben cumplirse antes de su introducción en el mercado y mantenerse durante todo su ciclo de vida. Muchos de estos requisitos tienen una relevancia directa para la prevención y mitigación de las alucinaciones:

\begin{enumerate}
    \item \textbf{Sistemas robustos de gestión de riesgos (Artículo 9):} Se exige el establecimiento, implementación, documentación y mantenimiento de un proceso continuo de gestión de riesgos. Esto implica la identificación de los riesgos previsibles asociados al sistema (incluyendo los derivados de alucinaciones), la estimación y evaluación de dichos riesgos, y la adopción de medidas adecuadas para su control. La gestión del riesgo de generar información legal incorrecta o fabricada debería ser un componente central de este sistema.
    \item \textbf{Gobernanza y calidad de los datos (Artículo 10):} Este artículo es particularmente pertinente para las alucinaciones originadas en datos de entrenamiento defectuosos. Exige que los conjuntos de datos de entrenamiento, validación y prueba sean 'relevantes, representativos, libres de errores y completos'. Se deben aplicar prácticas adecuadas de gobernanza de datos, incluyendo un examen de los posibles sesgos y la adopción de medidas para mitigarlos. Para la IA legal, esto implica la necesidad crítica de utilizar corpus jurídicos actualizados, verificados y que reflejen adecuadamente la diversidad y complejidad del ordenamiento jurídico.
    \item \textbf{Documentación técnica exhaustiva (Artículo 11 y Anexo IV):} Los proveedores deben elaborar una documentación técnica detallada que describa, entre otras cosas, la arquitectura del sistema, sus capacidades y limitaciones, los algoritmos utilizados, los datos de entrenamiento y los procesos de prueba y validación. Esta documentación es esencial para la evaluación de conformidad y para que los supervisores y usuarios comprendan cómo funciona el sistema y cuáles son sus umbrales de fiabilidad.
    \item \textbf{Mecanismos de registro de eventos (\textit{Logging Capabilities}) (Artículo 12):} Los sistemas de alto riesgo deben estar equipados con capacidades de registro que aseguren un nivel adecuado de trazabilidad de su funcionamiento. Estos 'logs' podrían ser cruciales para investigar \textit{a posteriori} el origen de una alucinación específica o para auditar el rendimiento general del sistema.
    \item \textbf{Transparencia y provisión de información a los usuarios (Artículo 13):} Los sistemas deben ser diseñados y desarrollados de manera que los usuarios puedan interpretar la salida del sistema y utilizarla de forma apropiada. Las instrucciones de uso deben incluir información concisa, completa, correcta y clara sobre la identidad del proveedor, la finalidad prevista del sistema, su nivel de precisión, robustez y ciberseguridad, así como sus limitaciones conocidas y los riesgos previsibles – lo cual incluye, o debería incluir, la propensión a generar alucinaciones y la necesidad de verificación humana.
    \item \textbf{Supervisión humana efectiva (Artículo 14):} Este es un pilar fundamental. La EU-AIAct exige que los sistemas de alto riesgo sean diseñados para permitir una supervisión humana adecuada. Las medidas pueden incluir la capacidad del supervisor humano para comprender plenamente las capacidades y limitaciones del sistema, para decidir no utilizar el sistema en una situación particular, para anular una decisión tomada por el sistema, o para intervenir en su funcionamiento. Esta supervisión es la última barrera contra las consecuencias de una alucinación no detectada por el propio sistema.
    \item \textbf{Precisión, robustez y ciberseguridad (Artículo 15):} Los sistemas deben alcanzar un nivel apropiado de precisión, robustez y ciberseguridad a lo largo de su ciclo de vida y ser consistentes en este aspecto. Las alucinaciones son una manifestación clara de una falta de precisión y robustez fáctica. Se espera que los sistemas sean resilientes a errores, fallos o inconsistencias, así como a intentos de uso malintencionado.
\end{enumerate}

El cumplimiento de estos requisitos se verificará mediante \textbf{evaluaciones de conformidad} antes de que el sistema de IA de alto riesgo pueda ser introducido en el mercado de la UE. Además, se establecen obligaciones de \textbf{vigilancia post-comercialización} para los proveedores, quienes deben monitorizar el rendimiento de sus sistemas y reportar cualquier incidente grave o mal funcionamiento. Las \textbf{sanciones} por incumplimiento de la EU-AIAct son significativas, pudiendo alcanzar hasta 35 millones de euros o el 7\% del volumen de negocio anual mundial total del ejercicio financiero anterior, lo que subraya la seriedad con la que la UE aborda los riesgos de la IA.

El impacto de la EU-AIAct en las estrategias de mitigación de alucinaciones en la IA legal, como RAG, es profundo. Muchos de los requisitos de la Ley –calidad de datos, transparencia sobre el funcionamiento, robustez, supervisión humana– impulsarán a los desarrolladores a adoptar de manera proactiva y rigurosa muchas de las estrategias de optimización discutidas en la Sección 5 de este ensayo, no como una opción de mejora, sino como una condición para el acceso al mercado. Aunque la EU-AIAct no prescribe soluciones técnicas específicas, sí establece un marco de exigencias que fomentará la innovación hacia una IA legal más fiable y responsable. Dado el peso del mercado europeo, es muy probable que la EU-AIAct tenga un "efecto Bruselas", influyendo en los estándares de desarrollo de IA legal a nivel global.

\subsubsection{Reglamento General de Protección de Datos (RGPD)}

Aunque no es específico para la IA, el RGPD ya impone obligaciones significativas que son relevantes para el desarrollo y uso de IA legal que trate datos personales.
    \begin{itemize}
        \item \textit{Principios de exactitud y minimización de datos:} Estos principios son directamente relevantes para combatir los datos de entrenamiento defectuosos que pueden llevar a alucinaciones.
        \item \textit{Derecho a no ser objeto de decisiones automatizadas (Artículo 22):} Si un sistema de IA legal toma decisiones que produzcan efectos jurídicos significativos en una persona (o le afecten de modo similar), el Artículo 22 del RGPD podría limitar su uso o requerir una intervención humana significativa.
    \end{itemize}

\subsubsection{Responsabilidad civil y profesional}
    \begin{itemize}
        \item \textit{Régimen General de Responsabilidad:} En España, la responsabilidad civil del abogado por negligencia profesional se rige por el Código Civil y la jurisprudencia. Si el uso indebido de una IA (p. ej., confiar en información alucinada sin verificación) causa un daño al cliente, el abogado podría ser considerado responsable.
        \item \textit{Propuesta de Directiva Europea sobre responsabilidad por IA:} La Comisión Europea ha propuesto una Directiva para adaptar las normas de responsabilidad civil extracontractual a la IA. Esta propuesta busca facilitar que las víctimas de daños causados por IA obtengan una reparación, por ejemplo, aliviando la carga de la prueba en ciertos casos o estableciendo una presunción de causalidad si el proveedor de IA no ha cumplido con ciertos deberes de diligencia (potencialmente incluyendo aquellos relacionados con la prevención de alucinaciones).
        \item \textit{Responsabilidad del proveedor vs. usuario profesional:} La atribución de responsabilidad entre el desarrollador/proveedor de la herramienta de IA y el abogado usuario será un área compleja y probablemente litigiosa. Los términos de servicio de los proveedores a menudo incluyen extensas cláusulas de exención de responsabilidad, pero su validez podría ser cuestionada, especialmente si se demuestra negligencia grave o un defecto inherente en el diseño del producto que lo hace propenso a generar información legalmente perjudicial (Calderon et al., 2022; Lantyer, 2024).
    \end{itemize}

\subsubsection{El papel de los Colegios de Abogados y órganos deontológicos}

En España, los colegios de abogados (y el Consejo General de la Abogacía Española) desempeñan un papel crucial en el establecimiento de normas deontológicas y en la supervisión de su cumplimiento.
    \begin{itemize}
        \item \textit{Emisión de guías y directrices específicas:} Es previsible y deseable que estos organismos desarrollen y publiquen guías específicas sobre el uso ético y competente de la IA generativa por parte de los abogados, abordando explícitamente el riesgo de alucinaciones y el deber de verificación.
        \item \textit{Formación continua:} La oferta de programas de formación continua sobre IA, sus capacidades, riesgos y uso responsable será esencial para asegurar que los profesionales mantengan el nivel de competencia requerido.
        \item \textit{Potestad disciplinaria:} Los colegios podrían ejercer su potestad disciplinaria en casos de uso manifiestamente negligente o irresponsable de la IA que resulte en perjuicio para el cliente o para la administración de justicia.
    \end{itemize}

\subsubsection{Necesidad de estándares técnicos y benchmarks}

Para que cualquier marco regulatorio sea efectivo, se necesitarán estándares técnicos y benchmarks independientes que permitan evaluar de manera objetiva la fiabilidad, precisión y propensión a las alucinaciones de las herramientas de IA legal. La colaboración entre juristas, tecnólogos y organismos de normalización será crucial en este aspecto.

La colaboración entre juristas y tecnólogos ya está trazando un camino en esta dirección, como lo demuestra el uso de exámenes profesionales de abogacía como benchmark para evaluar modelos de IA en dominios legales específicos (Gupta et al., 2025). La adopción de este tipo de pruebas estandarizadas como benchmarks podría convertirse en un requisito para que los proveedores de tecnología legal demuestren la fiabilidad y competencia de sus sistemas en una jurisdicción específica, proporcionando a los reguladores y a los consumidores una base objetiva para la evaluación.

\subsection{Hacia una integración ética y responsable de la IA en la práctica legal española y europea}

El camino hacia una integración de la IA en el derecho que sea a la vez innovadora y segura, especialmente en el contexto español y europeo con su fuerte tradición de protección de derechos y rigor normativo, requiere un compromiso proactivo y colaborativo de todos los actores implicados.

\begin{itemize}
    \item \textbf{Para los profesionales del Derecho:} La adopción de una mentalidad de \textbf{escepticismo informado y verificación diligente} es primordial. La IA debe ser vista como una herramienta poderosa de asistencia, no como un oráculo infalible. La formación continua y la alfabetización digital en IA serán competencias esenciales.
    \item \textbf{Para los bufetes y organizaciones legales:} Es necesario establecer \textbf{políticas internas claras} sobre el uso aceptable y responsable de la IA, incluyendo protocolos de verificación obligatorios, directrices sobre la gestión de datos confidenciales y programas de formación para sus profesionales. La inversión en herramientas de IA debe ir acompañada de una inversión en la capacitación para su uso seguro.
    \item \textbf{Para los proveedores de tecnología legal:} Existe una responsabilidad creciente de desarrollar herramientas que sean \textbf{"seguras por diseño" (}\textit{\textbf{safety by design}}\textbf{)}, incorporando mecanismos para minimizar las alucinaciones y, crucialmente, siendo \textbf{radicalmente transparentes} sobre las capacidades, limitaciones y tasas de error conocidas de sus productos. Las afirmaciones de marketing deben ser realistas y estar respaldadas por evidencia empírica robusta e independiente.
    \item \textbf{Para las instituciones educativas (Facultades de Derecho):} Es fundamental integrar la enseñanza sobre IA, sus implicaciones legales y éticas, y las habilidades necesarias para su uso crítico en los planes de estudio, preparando a las futuras generaciones de juristas para un entorno profesional transformado por la tecnología.
    \item \textbf{Para los reguladores y órganos deontológicos:} Se requiere una adaptación proactiva y reflexiva de los marcos normativos y deontológicos. Esto puede implicar la clarificación de los deberes existentes a la luz de la IA, el desarrollo de nuevas directrices específicas, y el fomento de una cultura de responsabilidad y rendición de cuentas. La Ley de IA de la UE será un marco clave, pero su implementación y supervisión efectivas en el sector legal requerirán un esfuerzo continuo.
    \item \textbf{Para la Judicatura:} Los tribunales también se enfrentarán al desafío de evaluar la información generada por IA presentada por las partes y, potencialmente, de utilizar la IA en sus propias labores. La formación judicial y el desarrollo de protocolos para el uso de IA en el ámbito judicial serán necesarios para mantener la integridad del proceso.
\end{itemize}

En conclusión final, las alucinaciones de la IA no son un mero artefacto técnico, sino un síntoma de la tensión fundamental entre la naturaleza probabilística de los LLMs y la exigencia de certeza y fiabilidad del sistema legal. Abordar este desafío en el contexto español y europeo exige un \textbf{enfoque que combine la innovación tecnológica con una reafirmación de los principios éticos fundamentales de la abogacía}, una adaptación inteligente de los marcos regulatorios y, sobre todo, un compromiso inquebrantable con el juicio crítico y la supervisión humana. La IA puede ser una herramienta poderosa para el derecho, pero solo si se navega su laberinto con prudencia, diligencia y una profunda conciencia de sus limitaciones actuales.

En definitiva, una integración exitosa y responsable de la IA en el ecosistema legal requiere un cambio cultural que trascienda la búsqueda de soluciones tecnológicas simplistas y adopte un paradigma de evaluación crítica y diligencia informada. Esto implica la internalización de tres principios operativos fundamentales:
\begin{itemize}
    \item \textbf{Priorizar la fiabilidad sobre la velocidad de generación.} La eficiencia real de una herramienta de IA no debe medirse únicamente por la rapidez con la que genera un resultado. Un sistema es verdaderamente eficiente solo si sus productos son fiables, minimizando así el tiempo y el esfuerzo requeridos en la indispensable fase de verificación humana. La fiabilidad, por tanto, es el verdadero multiplicador de la eficiencia en los flujos de trabajo jurídicos.
    \item \textbf{Fomentar el desarrollo de soluciones especializadas (Domain-Specific).} El sector jurídico se beneficiará más de herramientas diseñadas explícitamente para sus desafíos únicos que de la adaptación de modelos de propósito general. Las soluciones deben atender con precisión a las complejidades del razonamiento jurídico, la jerarquía normativa y los exigentes requisitos de confidencialidad, exigiendo un enfoque de desarrollo que priorice la profundidad contextual sobre la amplitud funcional.
    \item \textbf{Instituir una cultura de validación rigurosa y retroalimentación crítica.} La adopción de nuevas tecnologías debe estar guiada por una evaluación objetiva y empírica, en lugar de una aceptación acrítica impulsada por la novedad. El ecosistema (incluyendo profesionales, desarrolladores y académicos) debe demandar y proporcionar una retroalimentación rigurosa y honesta sobre el rendimiento, las limitaciones y los riesgos de estas herramientas. El progreso sostenible se fundamenta en la crítica constructiva y la validación independiente.
\end{itemize}

La adopción de estos principios pragmáticos es fundamental para que el sector legal pueda navegar la complejidad de la era de la IA con la prudencia, diligencia y profunda conciencia que esta exige.

\section{Conclusión: de la alucinación a la amplificación — principios para una IA jurídica fiable}
\label{sec:conclusion}

La irrupción de los Grandes Modelos de Lenguaje en el ecosistema jurídico presenta una paradoja fundamental: una tecnología con un potencial sin precedentes para democratizar y eficientar el acceso a la justicia, intrínsecamente lastrada por un defecto de diseño que atenta contra el pilar del derecho: \textbf{la veracidad}. Sin embargo, este análisis ha revelado que la veracidad en el derecho es un concepto dual, que abarca tanto la fidelidad factual —amenazada directamente por la alucinación— como la solidez interpretativa, que sigue siendo dominio exclusivo del juicio humano. Este informe, por tanto, ha diseccionado el fenómeno de las "alucinaciones" no como un mero error técnico, sino como una característica sistémica que exige un cambio de paradigma: pasar de buscar una IA que "sabe la verdad" a construir una IA que "amplifica la capacidad del profesional para interpretarla".

Más que un simple resumen, esta conclusión destila los hallazgos del análisis en un \textbf{conjunto de principios rectores y un marco de trabajo} para guiar a profesionales, desarrolladores y reguladores en la navegación de este complejo nuevo territorio.

\subsection{\textbf{Conclusiones fundamentales: un resumen estructurado}}

Los análisis detallados a lo largo de este documento convergen en cuatro conclusiones clave e interrelacionadas:

\begin{itemize}
\item \textbf{La alucinación no es un "bug", es una característica.} El desafío principal reside en comprender que la propensión de los LLMs de propósito general a "inventar" no es un fallo a corregir, sino una consecuencia directa de su arquitectura, diseñada para la fluidez probabilística y no para la fidelidad factual. Esto nos obliga a abandonar la idea de un "oráculo creativo" y adoptar un paradigma radicalmente distinto.

\item \textbf{RAG es el camino, no el destino.} La Generación Aumentada por Recuperación (RAG) es, sin duda, la estrategia de mitigación más importante y el fundamento de la IA jurídica moderna. Sin embargo, la evidencia empírica es contundente: una implementación canónica de RAG reduce, pero \textbf{no elimina}, las alucinaciones. Tratarla como una solución "plug-and-play" es un error; debe ser considerada como el punto de partida, un motor prometedor que requiere una optimización holística y rigurosa en cada uno de sus componentes para ser verdaderamente fiable.

\item \textbf{La fiabilidad se construye, no se instala: el imperativo de la optimización holística.} La transición de una IA que alucina a una IA fiable no depende de un único avance, sino de una sinergia de mejoras estratégicas a lo largo de todo el ciclo de vida de la información. Esto incluye:

    \begin{itemize}
        \item \textbf{Curación estratégica de datos:} Un fundamento de conocimiento verificado, actualizado y jerarquizado es el cimiento indispensable.
        \item \textbf{Recuperación sofisticada:} Ir más allá de la búsqueda semántica simple para incorporar la conciencia de la jerarquía normativa (el principio Kelseniano), el contexto y la lógica jurídica.
        \item \textbf{Generación fiel y razonamiento guiado:} Utilizar ingeniería de prompts avanzada y \textit{fine-tuning} para instruir a los LLMs no solo a responder, sino a "pensar" de manera estructurada, transparente y estrictamente anclada a las fuentes.
        \item \textbf{Verificación Post-Hoc robusta:} Implementar capas de seguridad algorítmica y humana como última línea de defensa indispensable.
    \end{itemize}

\item \textbf{El factor humano es irreductible y se fortalece.} Lejos de volver obsoleto al profesional del derecho, el desafío de la veracidad redefine y fortalece su rol. La supervisión experta, crítica e informada no es una opción, sino una obligación deontológica, profesional y, cada vez más, regulatoria (como lo demuestran la Ley de IA de la UE y las políticas del CTEAJE en España). El futuro no es la automatización, sino la \textbf{amplificación cognitiva}: la IA se convierte en una herramienta para potenciar el juicio humano, liberándolo para que se centre en la estrategia, la ética y la empatía.
\end{itemize}

\subsection{Propuesta de un marco de trabajo: IA generativa vs. IA consultiva}

Para guiar la adopción y el desarrollo futuro, proponemos un marco conceptual claro que distinga dos tipos de IA con perfiles de riesgo y aplicaciones radicalmente diferentes en el derecho:

\begin{itemize}
    \item \textbf{IA generativa de propósito general (el "oráculo creativo"):}

\begin{itemize}
    \item \textbf{Función:} Ideación, brainstorming, redacción de borradores no críticos, resumen de textos generales.
    \item \textbf{Riesgo inherente:} \textbf{Alto} riesgo de alucinación factual, extrínseca e intrínseca. Opacidad en las fuentes.
    \item \textbf{Principio de uso:} Utilizar siempre con escepticismo extremo, como un asistente creativo cuya producción \textbf{nunca} debe ser considerada una fuente de verdad. Requiere una verificación humana completa desde cero.
\end{itemize}

    \item \textbf{IA consultiva especializada (el "archivero experto"):}

\begin{itemize}
    \item \textbf{Función:} Investigación jurídica, \textit{due diligence}, análisis documental, respuesta a consultas basadas en un corpus verificado.
    \item \textbf{Riesgo inherente:} \textbf{Bajo} riesgo de fabricación, pero riesgo \textbf{persistente} de alucinaciones sutiles (\textit{misgrounding}, errores de síntesis).
    \item \textbf{Principio de uso:} Diseñada para la fiabilidad. Debe ser transparente, citable y auditable. Aun así, exige una verificación crítica por parte del profesional, pero enfocada en la correcta interpretación y aplicación de las fuentes proporcionadas, no en su existencia.
\end{itemize}

\end{itemize}

La mitigación efectiva de las alucinaciones en el sector legal no reside en mejorar incrementalmente el modelo generativo, sino en \textbf{adoptar deliberadamente un paradigma consultivo} donde la veracidad y la trazabilidad son el núcleo del diseño, no una característica añadida.

\subsection{\textbf{Una mirada al futuro: la llamada a una integración responsable}}

El camino hacia una IA jurídica verdaderamente fiable y beneficiosa está trazado, pero no es sencillo. No depende de encontrar un "interruptor mágico" que elimine los errores, sino de un compromiso colectivo y sostenido de todo el ecosistema legal.

\begin{itemize}
    \item \textbf{Para los desarrolladores}, el desafío es construir sistemas que no solo sean potentes, sino transparentes, auditables y diseñados con una profunda humildad sobre sus limitaciones.
    \item \textbf{Para los profesionales del derecho}, el reto es cultivar una cultura de \textbf{escepticismo informado}: abrazar la tecnología como una herramienta de amplificación, pero nunca abdicar de la responsabilidad final del juicio crítico.
    \item \textbf{Para los reguladores e instituciones}, la tarea es continuar desarrollando marcos normativos que fomenten la innovación responsable, estableciendo estándares claros de fiabilidad y exigiendo la supervisión humana como un pilar innegociable.
\end{itemize}

En última instancia, la inteligencia artificial no es una fuerza externa que "impacta" en el derecho; es un nuevo material con el que estamos construyendo las herramientas del futuro. La calidad, seguridad y justicia de esas herramientas dependerán de nuestra habilidad para infundir en ellas los principios atemporales de nuestra profesión: rigor, diligencia y un compromiso inquebrantable con la verdad. El objetivo final, y la gran promesa de esta era, no es simplemente automatizar procesos, sino, como se ha sostenido a lo largo de este informe, \textbf{humanizar la tecnología}, poniéndola al servicio de una justicia más accesible, eficiente y, sobre todo, fiable.

\section*{Agradecimientos} 

Nunca el trabajo necesario para el estudio, análisis y desarrollo de una investigación tan profunda como la que recoge este paper depende en exclusiva de su autor. Este trabajo hubiera sido impensable sin el esfuerzo previo de todos y cada uno de los investigadores que han sumado a la sociedad con sus papers anteriores. A todos ellos mi agradecimiento más sincero por lo que su trabajo ha representado para mi y para toda la comunidad científica y técnica.

Deseo agradecer igualmente a Little John y a todos y cada uno de sus miembros tanto su apoyo como el valor de sus revisiones. En especial a Daniel Vecino por las innumerables sesiones de trabajo en común y su “revisión pixel”, siempre tan crítica y completa como amable.

Es justo además agradecer la inspiración que para mi ha sido Asier Gutiérrez-Fandiño. Sin él esta publicación no hubiera sido posible ya que supuso un detonante claro en la pasión que en mi despierta el mundo de la Inteligencia Artificial.

Finalmente deseo agradecer la colaboración desinteresada de todos y cada uno de los profesionales que han tenido acceso previo a este paper. Sus comentarios han sido claves para recibir el empuje necesario que siempre me ha llevado a ir un poco más lejos en cada punto de análisis.

A todos ellos y a ti como lector de este trabajo gracias.

\bibliographystyle{unsrt}

\end{document}


\maketitle

\begin{abstract}
Large language models (LLMs) are rapidly redefining legal practice, education, and research. However, their vast potential is significantly threatened by the endemic generation of "hallucinations"—textual outputs that, while often plausible, are factually incorrect, misleading, or inconsistent with authoritative legal sources. This essay presents a comprehensive review and a multidimensional critical analysis of the phenomenon of hallucinations in LLMs applied to law. Trends and manifestations of hallucinations are documented across jurisdictions, court types, and classes of legal tasks, drawing on the growing empirical evidence from recent studies evaluating both public LLMs and specialized commercial legal Artificial Intelligence (AI) tools. The underlying causes of these hallucinations are analyzed in depth, ranging from deficiencies in training data and inherent limitations of the models’ probabilistic architecture, to the complexities of legal language and the fundamental tension between generative fluency and strict factuality.

Retrieval-Augmented Generation (RAG) is examined in detail as the main proposed mitigation strategy, with a critical evaluation of its theoretical effectiveness, practical implementations, and persistent limitations in the unique legal context, including failure points in its retrieval and generation phases. Beyond canonical RAG, holistic and advanced strategies for optimization and mitigation are discussed and proposed, encompassing strategic data curation, sophisticated prompt engineering, the consideration of AI agents aware of normative hierarchy (such as Kelsen’s pyramid), fidelity-focused fine-tuning, and the implementation of robust post-hoc verification and confidence calibration mechanisms. The gravity of these phenomena is illustrated through the analysis of detailed case studies of real judicial incidents, drawing tangible lessons about the consequences of uncritical reliance on AI.

Looking ahead, the path toward more reliable legal AI is explored, outlining the necessary developments in inherently more explainable models (XAI, for \textit{Explainable Artificial Intelligence}), technically auditable systems, and the adoption of a responsible-by-design AI paradigm. Finally, the profound ethical and regulatory implications are explored, with special attention to the European and Spanish regulatory frameworks, emphasizing the irreducible and irreplaceable role of human oversight and the professional judgment of lawyers in the age of artificial intelligence. The conclusion underscores the imperative need for a cautious, critical, and supervised integration of LLMs into legal practice, proposes a refined typology of legal hallucinations to guide and structure future research in this crucial field, and also proposes a new framework distinguishing between general-purpose Generative AI and specialized Consultative AI, offering a refined typology of legal hallucinations that will guide future research toward truly responsible integration.

However, it is imperative to also make a fundamental distinction that is often overlooked in the current debate: the difference between general-purpose Artificial Intelligence (such as public LLMs) and specialized, consultative Artificial Intelligence designed specifically for the legal domain. While the former, due to its generative nature, is inherently prone to hallucinations by "inventing" answers to maintain conversational fluency, the latter operates under a radically different principle. A consultative AI does not create knowledge, but rather retrieves, structures, and presents it in a well-founded manner, acting as an expert assistant that cites its sources instead of a creative oracle. This report will argue that the effective mitigation of hallucinations in the legal sector does not lie in incrementally improving generative models, but in adopting a consultative paradigm where truthfulness and traceability are at the core of the design, not an added feature. In this context, technology is not a substitute for human judgment, but a tool to amplify it, fulfilling the maxim of humanizing technology rather than simply automating processes.
\end{abstract}


\section{Introduction}
Artificial intelligence (AI), and in particular large language models (LLMs), are at the cusp of a significant transformation across multiple sectors, with the legal domain being one of the most impacted and debated (Choi et al., 2022; Katz et al., 2023; Rodgers, Armour, and Sako 2023). Tools such as OpenAI's ChatGPT, Google's Gemini, DeepSeek, and Meta's Llama, along with specialized legal AI platforms, promise to revolutionize fundamental tasks such as legal research, document drafting, contract analysis, and litigation assistance (Guha et al. 2023; Livermore, Herron, and Rockmore 2024). The potential to increase efficiency, reduce costs, and \textbf{democratize access to justice is considerable} (Perlman 2023; Tan, Westermann, and Benyekhlef 2023).

However, this transformative potential is hindered by an inherent and critical challenge: the phenomenon of "hallucinations" (Ji, Lee, et al. 2023). Hallucinations in LLMs refer to the \textbf{generation of information that, although often plausible and linguistically coherent, is factually incorrect, misleading, inconsistent} with the provided sources, or entirely fabricated (Dahl et al., 2024; Magesh et al., 2024).

In the legal context, where accuracy, fidelity to authoritative sources (precedents, statutes), and fact-based argumentation are paramount, hallucinations are not mere technical inaccuracies, but rather represent a \textbf{substantial risk} that can lead to strategic errors, harmful legal advice, professional sanctions, and even the \textbf{erosion of public trust in the legal system} (Roberts 2023; Weiser 2023). The infamous case \textit{Mata v. Avianca, Inc.} (2023), where attorneys were sanctioned for submitting a legal brief citing non-existent cases generated by ChatGPT, serves as a clear reminder of these dangers (Lantyer, 2024).

It is crucial to nuance, however, that the challenge of truthfulness in law transcends mere factual correctness. Unlike other domains, in the legal field a statement is not simply "true" or "false"; its validity often lies in the strength of its interpretation and argumentation, which is precisely the realm of expert professional judgment. Therefore, the danger of AI is not only that it generates verifiable falsehoods, but also that it constructs legally unviable arguments or superficial interpretations that, without the critical filter of a lawyer, may lead to erroneous strategies. The analysis of truthfulness must, therefore, encompass both fidelity to the source and interpretative viability.

This essay embarks on a comprehensive exploration of hallucinations in Large Language Models (LLMs) within the legal domain, aiming to move beyond anecdotal reports and provide a systematic analysis grounded in the growing body of empirical evidence and academic literature. To this end, we will first define and categorize hallucinations specific to the legal context, exploring their root causes and particular impact on legal practice (Section \ref{sec:fenomeno_alucinaciones}). Next, we will examine in detail the methods and inherent challenges in evaluating the prevalence and nature of these hallucinations, critically reviewing recent studies on general LLMs and commercial legal AI tools (Section \ref{sec:evaluacion_alucinaciones}). We will then analyze in depth Retrieval-Augmented Generation (RAG) as the main proposed mitigation strategy, assessing both its conceptual promises and its inherent limitations and empirical effectiveness in the legal context (Section \ref{sec:rag_mitigacion}). Subsequently, we will discuss a range of complementary and advanced strategies for the optimization and mitigation of hallucinations, covering data curation, prompt engineering, the consideration of AI agents aware of normative hierarchies, and post-hoc verification mechanisms (Section \ref{sec:estrategias_avanzadas}). To illustrate the severity and tangible consequences of these phenomena, we will present and analyze detailed case studies of real incidents where AI hallucinations have impacted judicial proceedings (Section \ref{sec:casos_reales}). Looking ahead, we will explore the path toward more reliable legal AI, discussing the development of explainable, auditable, and responsible models by design (Section \ref{sec:futuro_ia_fiable}). Finally, we will reflect on the crucial ethical and regulatory considerations that arise, with special attention to the European and Spanish regulatory frameworks (Section \ref{sec:etica_regulacion}), concluding by synthesizing the findings and emphasizing the way forward toward the responsible and effective integration of AI in legal practice (Section \ref{sec:conclusion}).

\section{The Phenomenon of Hallucinations in Legal LLMs: Nature, Causes, and Impact}
\label{sec:fenomeno_alucinaciones}

The integration of Large Language Models (LLMs) into the legal ecosystem represents one of the most profound and potentially disruptive technological transformations of the modern era. These artificial intelligence (AI) architectures, capable of processing and generating natural language with unprecedented fluency, promise to radically optimize knowledge-intensive tasks such as legal research, contract drafting, evidence analysis (discovery), and the generation of legal briefs (Choi et al., 2022; Livermore, Herron, and Rockmore 2024). However, this promise is overshadowed by an inherent and pervasive challenge: the phenomenon of "hallucinations" (Ji, Lee, et al. 2023; Marcus \& Davis, 2022). Far from being an occasional anomaly, hallucinations constitute an intrinsic feature of the current functioning of LLMs, manifesting as the generation of content that, although often syntactically and semantically plausible, lacks factual basis, is logically inconsistent, or directly contradicts established authoritative sources. In the legal domain, where factual accuracy, fidelity to authority (laws, precedents, doctrine...), and argumentative integrity are fundamental pillars, the propensity of LLMs to hallucinate is not merely a technical inconvenience, but a systemic risk with profound ethical, professional, and social implications (Roberts 2023).
The phenomenon of hallucinations is not merely a technical inconvenience, but has been identified as one of the critical challenges defining the current frontier of research in legal AI. Comprehensive reviews of the field indicate that, despite the transformative advances of LLMs, "\textbf{hallucination in legal claims, manifested as spurious citations or fabricated regulations}", together with deficits in explainability and jurisdictional adaptation, constitute the main barriers to their widespread and reliable adoption (Shao et al., 2025).

\subsection{A Fundamental Paradigm: Generative AI vs. Consultative AI}

Before dissecting the phenomenon of hallucinations, it is imperative to establish a conceptual distinction that the current debate often overlooks, generating a dangerous confusion: the fundamental difference between \textbf{Generative Artificial Intelligence} and \textbf{Consultative Artificial Intelligence}. The term "Legal AI" is used monolithically, when in reality it describes two architectures with radically different purposes, mechanisms, and risk profiles. Understanding this dichotomy is not merely an academic exercise; it is the key to a responsible and effective integration of AI into legal practice.

\subsubsection{Generative Artificial Intelligence: the creative oracle}

Generative AI, whose main exponents are general-purpose LLMs such as GPT, Gemini, or Claude, operates as an "advanced imitator" or a "creative know-it-all." Its primary objective is not truthfulness, but rather \textbf{conversational fluency and probabilistic coherence}.

\begin{itemize}
    \item \textbf{Definition and mechanism:} These models work by predicting the most probable next word in a sequence, based on statistical patterns learned from a vast and heterogeneous corpus of internet data. They do not "understand" the content nor do they "reason" from logical principles; instead, they assemble text that \textit{sounds} plausible. Their knowledge is parametric and "frozen" at the time of their training.

    Fundamental research on the causes of hallucinations explains that this behavior is not a flaw to be corrected, but a direct consequence of their design. The models are optimized to be good "test-takers": in a system where uncertainty is not rewarded, the most effective strategy to get a "good grade" (a plausible answer) is always to risk an answer rather than admit ignorance. Therefore, their tendency to "make things up" is the expected result of their training (Kalai et al., 2025).
    
    \item \textbf{Advantages:} Their strength lies in creative tasks: drafting, brainstorming, summarizing non-critical texts, and generating content where originality is more important than factual accuracy.
    \item \textbf{Disadvantages and inherent risks:} For the legal sector, their design is a recipe for disaster.
    \item \textbf{"Designed-in" hallucinations:} The propensity to hallucinate is not a bug, it is an intrinsic feature of their architecture. To avoid silence and maintain coherence, the model will "fill in the gaps" or "make up" facts, rulings, or statutes.
    \item \textbf{Total opacity ("Black Box"):} It is impossible to trace the origin of a specific statement. The response is an opaque final product, without verifiable references.
    \item \textbf{Risk of "AI incest":} Since they are trained on public internet data, they run the risk of feeding back on low-quality content generated by other AIs, degrading their reliability in a vicious cycle.
\end{itemize}

\subsubsection{Consultative Artificial Intelligence: the expert archivist}

Consultative AI represents a paradigm shift. Its goal is not to create, but to \textbf{retrieve, structure, and present verified knowledge}. Its fundamental architecture is based on Retrieval-Augmented Generation (RAG), operating as an "expert archivist" or a "detective" who investigates before speaking.

\begin{itemize}
    \item \textbf{Definition and mechanism:} This model does not rely on its internal parametric knowledge. When faced with a query, its first step is to search an external, curated, and authorized data corpus (e.g., databases of legislation, case law, internal documents of a law firm). Only after retrieving the most relevant information fragments does it generate a response that must be \textit{strictly grounded} in those fragments.
    \item \textbf{Advantages:} Designed for reliability in critical domains.
    \item \textbf{Mitigation of hallucinations:} Drastically reduces the fabrication of facts, since responses are anchored to explicit sources.
    \item \textbf{Transparency and traceability:} The response is not a "black box." A well-designed consultative system should cite its sources, allowing the legal professional to verify the information and assume final responsibility with full knowledge. It is the materialization of the principle of \textbf{"not replacing, but amplifying"} human judgment.
    \item \textbf{Up-to-date knowledge:} Its reliability depends on the currency of its database, which is much easier and cheaper to update than retraining a massive LLM.
    \item \textbf{Disadvantages and limitations:} It is not a panacea. Its effectiveness depends critically on the quality of its document corpus and the sophistication of its retrieval module. It can still produce subtle hallucinations, such as \textit{misgrounding} (misrepresenting a real source), but the risk of blatant invention is minimized.
 \end{itemize}

\subsubsection{Comparative Table of Paradigms}

\begin{table}[htbp]
\centering
\footnotesize 
\renewcommand{\arraystretch}{1.3} 
\setlength{\tabcolsep}{6pt} 
\begin{tabular}{|>{\raggedright\arraybackslash}p{0.25\textwidth}|>{\raggedright\arraybackslash}p{0.30\textwidth}|>{\raggedright\arraybackslash}p{0.30\textwidth}|}
\hline
\textbf{Characteristic} & \textbf{Generative AI (general purpose)} & \textbf{Consultative AI (specialized)} \\
\hline
\textbf{Main objective} & Conversational fluency and coherence. & Accuracy, reliability, and grounding. \\
\hline
\textbf{Source of knowledge} & Parametric, internal, static ("closed book"). & External, curated, dynamic ("open book"). \\
\hline
\textbf{Hallucination risk} & High, especially fabrication of facts ("by design"). & Low fabrication, risk of \textit{misgrounding}. \\
\hline
\textbf{Transparency} & Low ("black box"). & High (must cite sources and reasoning). \\
\hline
\textbf{Ideal use case} & Brainstorming, creative drafts, non-critical tasks. & Legal research, \textit{due diligence}, factual answers. \\
\hline
\textbf{Analogy} & An eloquent "know-it-all" but sometimes unreliable. & A meticulous "archivist" who always shows their cards. \\
\hline
\end{tabular}
\end{table}

The adoption of this dual framework is essential to navigate the complexity of Legal AI. Confusing both paradigms leads to unrealistic expectations and irresponsible application of the technology. The subsequent sections of this report will analyze in depth the challenges inherent to the generative model and how consultative architectures, mainly through RAG, attempt to build a path toward truly reliable legal AI.

\subsection{Definition and Taxonomy of Legal Hallucinations}

Defining "hallucination" in the context of legal AI requires going beyond the simple correct/incorrect dichotomy. A legal hallucination materializes when an LLM generates a statement, citation, argument, or conclusion that deviates from verifiable legal reality or from the contextual information provided, often presenting it with unwarranted confidence (Khmaïess Al Jannadi, 2023). It is crucial to understand that, while the term 'hallucination' is commonly used, its application in law presents unique challenges. Unlike domains with singular factual truths, in the legal field the 'correctness' of an interpretative statement or argument may be subject to debate among experts. Therefore, beyond mere factual deviation, a legal hallucination can also be understood as the generation of a proposal that, \textbf{although plausible, is legally unviable or indefensible under expert scrutiny}, even if it does not directly contradict an explicit source. The apparent coherence of these outputs can mask their lack of legal soundness, making their detection particularly complex. This deviation can take multiple forms, each with distinct implications for legal practice.

To enrich this taxonomy, we propose an additional classification dimension based on the architectural origin of the AI. Hallucinations manifested by a general-purpose AI (e.g., ChatGPT) tend to be more severe (such as the complete invention of case law), since its goal is conversational coherence at all costs. In contrast, errors in a specialized consultative AI (based on RAG) tend to be more subtle, such as misgrounding or synthesis errors, stemming from failures in the retrieval or interpretation of a controlled document corpus.

This distinction is crucial, since while the first type of hallucination represents a systemic design failure for legal use, the second is an implementation problem that can be mitigated with optimization techniques, as will be discussed later. Ignoring this difference is like confusing the opinion of an eloquent amateur with the documented analysis of an expert archivist.

To rigorously define and classify legal hallucinations, it is useful to adopt an analytical framework that distinguishes the two key dimensions of error. The seminal study by Magesh et al. (2025) on legal AI tools proposes a fundamental distinction between:
\begin{itemize}
    \item Correctness: Whether the statement is factually true in the real world.
    \item Groundedness: Whether the statement is properly supported by the cited source.
\end{itemize}

Based on this framework, a "hallucination" is defined as a response that is incorrect (contains false information) or poorly grounded (\textit{misgrounded}, that is, cites a source that does not support the statement). This deviation can take multiple forms, each with distinct implications.

The following categories detail the specific manifestations of these failures in legal practice:

\begin{itemize}
    \item \textbf{Factual/extrinsic hallucinations} (inconsistency with the facts of the legal world): this is perhaps the most dangerous type in legal research and direct legal advice. It refers to the generation of content that contradicts the established and verifiable body of law and related facts.
        \begin{itemize}
            \item \textit{Misstatement} of Law or Precedent: The LLM incorrectly describes the content or holding of an existing law or judicial decision. This can range from subtle misrepresentations to direct contradictions with the cited or known authority.
            \item Fabrication of authority: the model completely invents cases, statutes, regulations, or even non-existent judges and scholars. The case \textit{Mata v. Avianca, Inc.} (2023) is the paradigmatic example, where ChatGPT generated multiple fictitious judicial citations that were incorporated into a legal brief.
            \item Jurisdictional or temporal application error: the LLM incorrectly applies legal principles from one jurisdiction to another, or presents as current a law or precedent that has been repealed or is obsolete, failing to recognize the temporal and spatial dynamics of law.
        \end{itemize}
\end{itemize}

\begin{itemize}
    \item \textbf{Source-based hallucinations} (errors of \textit{groundedness} in RAG Systems): particularly relevant for Retrieval-Augmented Generation (RAG) systems, which are discussed in Section \ref{sec:rag_mitigacion}. These occur when the generated response is inconsistent with the specific documents retrieved by the system to support that response.
    \begin{itemize}
        \item \textit{Misgrounding}: the LLM correctly cites an existing source (retrieved by the RAG system), but makes a claim about its content that the source does not support or even contradicts (Magesh et al., 2024). This creates a false appearance of documentary support.
        \item \textit{Ungrounding}: the LLM makes specific factual claims that should be supported by the retrieved material, but does not provide citations or the retrieved sources do not contain the asserted information.
    \end{itemize}
\end{itemize}

\begin{itemize}
    \item \textbf{Inference and reasoning hallucinations}: involve failures in the logical structure of the legal argument or in the characterization of relationships between concepts or authorities.
    \begin{itemize}
        \item Illogical or invalid argumentation: the model constructs a line of reasoning that violates basic logical principles or does not withstand legal scrutiny, even though it may appear superficially persuasive.
        \item Mischaracterization of arguments, parties, or procedural stances: the LLM confuses the arguments of a party with the holding of the court, or incorrectly describes the procedural stance or the relationships between the parties in litigation (Dahl et al., 2024).
    \end{itemize}
\end{itemize}

\begin{itemize}
    \item \textbf{Intrinsic hallucinations} (inconsistency with the prompt or training corpus): although potentially less frequent in direct responses to factual legal queries, they can arise in closed-domain tasks such as summarizing lengthy legal texts or drafting documents based on detailed instructions, where the final output substantially deviates from or contradicts the content or guidelines of the provided input.
\end{itemize}

It is crucial to recognize that these categories are not mutually exclusive; a single hallucinated response may exhibit multiple types of errors simultaneously. The unifying characteristic is the \textbf{disconnect between the generated output and a relevant ground truth} (whether this is the facts of the legal world, retrieved sources, or the initial prompt), often masked by the model's linguistic fluency (Ji, Lee, et al. 2023).

Beyond the fabrication of information, a more insidious form of deviation occurs through the alteration of existing content, which can induce cognitive biases in the practitioner. Research has quantified how LLMs, in summarization tasks, alter the framing of the original text, for example, changing the sentiment from neutral to positive or negative. In one study, it was observed that this occurs in 21.86\% of cases (Alessa et al., 2025). In the legal context, this could manifest as a summary of a ruling that emphasizes the arguments of one party over the other, or that presents a doctrinal analysis in a more favorable or critical manner than it actually is, subtly influencing the lawyer's initial assessment.

Additionally, primacy bias has been identified, where the summary generated by the LLM disproportionately focuses on information presented at the beginning of a document, occurring in 5.94\% of cases (Alessa et al., 2025). This represents a significant risk in the review of lengthy court records or contracts, where critical details may be found in later sections that the LLM might minimize or omit.

This general taxonomy is complemented by community efforts to categorize more granular errors specific to RAG systems. For example, the LibreEval benchmark (Arize AI) identifies failure types such as '\textit{Overclaim}', where the model exceeds what is supported by the sources, '\textit{Incompleteness}', when the response omits crucial information present in the context, or '\textit{Relational-error}', which denotes failures to correctly synthesize information from multiple retrieved fragments. These RAG-specific errors can be considered detailed manifestations of our broader categories, underscoring the complexity of ensuring fidelity in these systems.

\begin{table}[htbp] 
  \centering
  \caption{Taxonomy of Hallucinations in Legal LLMs}
  \label{tab:taxonomia_alucinaciones}
  \begin{tabular}{
    >{\raggedright\arraybackslash}p{0.25\textwidth} 
    >{\raggedright\arraybackslash}p{0.4\textwidth} 
    >{\raggedright\arraybackslash}p{0.3\textwidth} 
  }
    \toprule
    \textbf{Main Category} & \textbf{Subtype / Brief Description} & \textbf{Fictitious Legal Example} \\
    \midrule
    \textbf{Factual / Extrinsic Hallucinations} & Inconsistency with facts of the legal world. & \\
    \cmidrule(r){1-1}
    & \textit{Misstatement of Law or Precedent:} Incorrect statement of the content or holding of an authority. & "The LLM claims that the 2022 Urban Leases Act allows for immediate eviction without notice." (When the law requires 30 days) \\
    \addlinespace
    & \textit{Fabrication of Authority:} Invention of non-existent cases, statutes, or scholars. & "According to the case Martínez v. Constructora Sol (2025), strict liability is inapplicable." (The case does not exist) \\
    \addlinespace
    & \textit{Jurisdictional/Temporal Application Error:} Incorrect application of rules to another jurisdiction or presenting repealed rules as current. & "The LLM cites an article from the 1950 Civil Code to resolve a current contractual dispute, ignoring subsequent reforms." \\
    \midrule
    \textbf{Source-based Hallucinations (RAG Errors)} & Inconsistency with documents retrieved by the RAG system. & \\
    \cmidrule(r){1-1}
    & \textit{Misgrounding:} Cites a real source, but asserts something the source does not support or contradicts. & "Document X says 'the contract is valid', but the LLM reports: 'According to document X, the contract is void'." \\
    \addlinespace
    & \textit{Ungrounding:} Makes claims that should be supported by the retrieved material, but provides no citations or the sources do not contain them. & "The defendant acted negligently. (Without citing any evidence or retrieved document to support this)." \\
    \midrule
    \textbf{Inference and Reasoning Hallucinations} & Failures in the logical structure of the legal argument. & \\
    \cmidrule(r){1-1}
    & \textit{Illogical or Invalid Argumentation:} Constructs a line of reasoning that violates logical principles. & "If all contracts require offer and acceptance, and this document is a contract, then the sky is blue." (Conclusion does not follow) \\
    \addlinespace
    & \textit{Mischaracterization of Arguments/Parties:} Confuses the arguments of a party with the holding, or incorrectly describes procedural stances. & "The LLM presents the plaintiff's petition as if it were the judge's final ruling." \\
    \midrule
    \textbf{Intrinsic Hallucinations} & Inconsistency with the prompt or training corpus (in closed-domain tasks). & \\
    \cmidrule(r){1-1}

    & Substantial deviation from the content or guidelines of the input in tasks such as summarization or instruction-based writing. & "Prompt: 'Summarize the following contract in 100 words focusing on the penalty clauses.' LLM response: A 500-word summary about the history of the contracting company." \\
    \bottomrule
  \end{tabular}
\end{table}

\subsection{Root Causes of Hallucinations in Legal LLMs}

Understanding why Large Language Models (LLMs) generate hallucinations, especially when applied to the rigorous legal domain, is an indispensable step for developing effective mitigation and evaluation strategies. The causes are multifactorial, rooted both in the fundamental properties of current LLM technology and in the specific complexities of legal knowledge and language. These factors interact in complex ways, giving rise to the various manifestations of errors that we have previously categorized.

A fundamental cause lies in the \textbf{inherent limitations of the training data}. The vast scale of the corpora used to train LLMs, often scraped from the web, implies an inevitable variability in quality, veracity, and timeliness (Bender et al., 2021). In the legal domain, this is particularly problematic. Publicly available legal texts may be incomplete or represent only a fraction of the total legal landscape. More critically, law is a dynamic system; statutes and precedents are constantly changing, making any LLM trained on a static dataset inevitably contain outdated information (Khmaïess Al Jannadi, 2023). Added to this is the presence of historical biases—social, economic, racial—encoded in past legal and judicial texts. By learning statistical patterns from these data, \textbf{LLMs risk not only reproducing but amplifying these inequalities}, generating responses that may perpetuate systemic injustices (Gebru et al., 2018; O'Neil 2016; Barocas and Selbst 2016). The relative scarcity of high-quality, verified legal data that is representative of all jurisdictions and areas of law remains a significant bottleneck.

Moreover, it can be argued that the problem originates in the very culture of artificial intelligence evaluation. Models are trained and evaluated predominantly with metrics that severely penalize responses expressing uncertainty (such as "I don't know"). As a result, LLMs learn that a plausible guess is preferable to an honest abstention, perpetuating a "always guess" behavior (Kalai et al., 2025). This mode of operation, analogous to a student who never leaves questions blank on an exam, is fundamentally incompatible with the prudence required in legal practice.

The scarcity of verified and representative data from specific jurisdictions is a direct and demonstrable cause of hallucinations. An empirical study on the performance of LLMs in a non-Anglo-Saxon jurisdiction revealed that, while models such as GPT and Claude excelled at drafting tasks, all models systematically failed at legal research, frequently generating citations to non-existent cases. The author concludes that this deficiency is due to LLMs being trained predominantly with data from dominant legal systems (such as the U.S.), lacking a sufficient knowledge base about the jurisprudence of other regions, which forces them to "hallucinate" in order to complete the task (Hemrajani, 2025).

Intimately linked to the data is the \textbf{probabilistic nature and the very architecture of LLMs}. These models, despite their impressive ability to generate coherent text, do not operate through deep semantic understanding or logical reasoning analogous to that of humans (Searle 1980; Marcus \& Davis, 2022). They are fundamentally predictive engines that calculate the most probable sequence of words based on the statistical correlations learned from their vast training data. This orientation towards statistical prediction, often optimized for linguistic fluency over strict factuality, makes them inherently prone to generating statements that \textit{sound} correct but lack a real basis (Ji et al., 2023; Bowman 2015).

This intrinsic optimization for fluency can lead to what is sometimes described as 'confabulation', a process by which the model, when faced with a lack of direct factual information or ambiguity in the query, \textbf{'invents' details or coherent narratives to maintain the continuity of discourse}, even though these fabricated elements lack a real basis. Confabulation, in this sense, is a direct manifestation of the predictive architecture of the LLM prioritizing the appearance of understanding over strict factuality, leading to the generation of hallucinations that, although incorrect, can be deceptively persuasive due to their superficial coherence.

The difficulty of LLMs in connecting abstract legal principles with the concrete facts of a case is a fundamental cause of hallucinations. A study that conditioned LLMs with different levels of knowledge of the German legal system to detect hate speech demonstrated this conclusively. When the models were "conditioned" only with the most abstract knowledge (such as the title of a constitutional or statutory norm), \textbf{they showed a lack of deep understanding of the task, even contradicting themselves and hallucinating responses} when presented with fictitious or irrelevant norms (Ludwig et al., 2025). This suggests that the probabilistic architecture of LLMs, in the absence of anchoring in concrete definitions and examples, struggles to correctly apply legal reasoning, resorting to invention.

Phenomena such as overfitting, where the model memorizes specific training patterns instead of learning general principles, can exacerbate this problem, limiting its ability to correctly generalize to new or slightly different situations (Khmaïess Al Jannadi, 2023). Moreover, its inherent capacity for extrapolation (although essential for generalization) can easily drift towards invention or spurious connection of concepts when faced with queries that border the limits of its knowledge or require complex inferences (Shaip, 2022; Huang et al. 2021; Domingos 2015).

The \textbf{legal domain itself presents an intrinsic complexity} that amplifies these challenges. Legal language is notoriously technical, dense in meaning, highly context-dependent, and riddled with ambiguities and polysemous terms (Khmaïess Al Jannadi, 2023). Correctly interpreting a statute, contract, or judgment requires not only understanding the literal meaning of the words, but also the legislative context, the intent of the parties, the procedural history, and the network of relevant precedents—tasks that demand a level of contextual understanding and reasoning that current LLMs struggle to achieve. Legal reasoning per se, with its analogical and rule-based deductive methods and principles, and its constant balancing of factors, represents a higher-order form of cognition that LLMs, based on statistical patterns, struggle to faithfully emulate (Ashley 2017; Choi and Schwarcz, 2024).

Even \textbf{strategies designed to mitigate hallucinations, such as RAG, introduce their own points of vulnerability}. As will be discussed in detail in Section \ref{sec:rag_mitigacion}, the effectiveness of RAG critically depends on the quality of its information retrieval module. If the retrieved information is irrelevant, incorrect, or incomplete, the generator LLM, even if it attempts to be faithful to the provided context, will produce a faulty response. Moreover, the generator LLM itself may fail to correctly integrate the retrieved information, prioritizing its erroneous parametric knowledge or incorrectly synthesizing the sources (Addleshaw Goddard, 2024).

Ultimately, the analysis of the root causes of hallucinations would be incomplete if it were limited to the model. \textbf{The most dangerous fundamental cause does not lie in the machine, but in the human who uses it without discernment}. AI has the potential to amplify the capabilities of diligent professionals, while it can lead to errors for those who use it without critical judgment and proper supervision. A professional with critical thinking and experience will use the LLM as a catalyst to accelerate their research, validating each result. However, a user without these foundations, seduced by apparent ease, will delegate their reasoning and fall into the trap of complacency. Therefore, the greatest source of risk is not the model's hallucination, but the "\textbf{user's hallucination}": the belief that a tool can substitute for responsibility, effort, and professional judgment. This phenomenon, fueled by a culture of immediacy and shortcuts, is the true challenge to be mitigated in the integration of AI into the legal sector.

\subsection{Specific Impact and Associated Risks in Legal Practice}

The manifestation of these causes in the form of hallucinations has a tangible and multifaceted impact on the legal ecosystem:

\begin{enumerate}
    \item \textbf{Undermining of legal research and analysis}: The foundation of any rigorous legal work is accurate research. Hallucinations, by introducing false or fabricated information, contaminate this fundamental process, wasting time on verification, leading to erroneous analysis, and ultimately resulting in flawed legal strategies.
    \item \textbf{Professional and ethical risks}: For lawyers, relying on hallucinated information can have devastating consequences. It can lead to the submission of defective court filings (resulting in sanctions such as the well-known \textit{Mata v. Avianca} case), failure to meet the duty of competence and diligence, violation of the duty of candor to the court, and potential claims of professional negligence (Yamane, 2020; Schwarcz et al., 2024). The need to thoroughly verify every AI output can, ironically, negate the promised efficiency gains (Gottlieb 2024).
    \item \textbf{Erosion of trust}: The prevalence of hallucinations, especially if not addressed transparently, can undermine the trust of legal professionals, clients, and the general public in AI tools and, by extension, in those who use them (Khmaïess Al Jannadi, 2023). This erosion of trust can hinder the adoption of potentially beneficial technologies.
    \item \textbf{Impact on access to justice}: There is a troubling paradox: LLMs are promoted as a tool to democratize access to legal information for pro se litigants or low-income individuals. However, these same users are the least equipped to detect and verify sophisticated hallucinations, making them particularly vulnerable to receiving incorrect and harmful legal information (Draper and Gillibrand 2023; Dahl et al., 2024). Instead of closing the gap, unreliable AI could widen it.
    \item \textbf{Integrity of the judicial system}: At a systemic level, the introduction of false or fabricated information into judicial proceedings, whether inadvertently by lawyers or potentially maliciously, threatens the fundamental integrity of the adversarial process and the pursuit of truth.
    \item \textbf{Cognitive and subtle judgment risks}: Perhaps the most underestimated risk is not that AI provides false information, but that it presents truthful information in a way that exploits human cognitive biases. LLMs can act as "vectors of bias," inducing framing biases that alter the perception of a problem without changing the facts. For example, when summarizing the opposing party's arguments, an LLM could select language that makes them appear weaker than they are. Similarly, authority bias may lead a lawyer to accept an AI-generated conclusion with less scrutiny than they would apply to a human colleague, simply because of the model's fluent and seemingly logical presentation (Alessa et al., 2025). This effect erodes the objectivity of professional judgment from within, in a way that is much harder to detect than a simple false citation.
\end{enumerate}

Addressing the root causes of legal hallucinations is, therefore, not merely a technical optimization, but an ethical and functional imperative for the future of AI in law.

The gravity of this impact has not gone unnoticed by legislators, and the inherent risk in the dissemination of incorrect or fabricated legal information is one of the central concerns driving regulatory efforts at the global level. In this regard, Regulation (EU) 2024/1689 of the European Parliament and of the Council, known as the European Union Artificial Intelligence Act (hereinafter, the EU AI Act, the Regulation, or EU-AIAct), a pioneering and ambitious legislative framework, sets a significant precedent. By adopting a risk-based approach, the EU AI Act seeks to impose stricter requirements on those AI systems whose failures could have severe consequences for fundamental rights, safety, or the proper functioning of key institutions. Although the specific categorization of all legal AI tools under this framework is yet to be defined in its practical application, it is plausible to anticipate that those systems intended to influence the administration of justice or to provide advice in critical areas could be subject to intensified regulatory scrutiny precisely because of the disruptive potential of phenomena such as hallucinations.

In conclusion, hallucinations are not a minor technical flaw, but rather a manifestation of the fundamental limitations in the way current LLMs process information and model the world, with particularly critical ramifications in the sensitive and normative domain of law. Addressing this challenge is a prerequisite for any responsible and beneficial integration of AI into the legal profession.

\section{Evaluation of Hallucinations in Legal AI Applications: Methodologies, Challenges, and Current State}
\label{sec:evaluacion_alucinaciones}

The mere existence of the phenomenon of hallucinations in LLMs applied to law, detailed in the previous section, imposes a critical and unavoidable necessity: the development and application of rigorous methodologies for their evaluation, detection, and quantification. Given the high-risk nature of the legal domain, where decisions based on incorrect information can have devastating legal, financial, and social consequences, simply relying on developers' claims or the apparent plausibility of generated responses is unsustainable. Systematic empirical evaluation thus becomes not only a desirable academic exercise, but a fundamental prerequisite for the responsible integration of these technologies into professional practice, legal education, and the justice system in general. However, as we will explore in this section, the evaluation of legal hallucinations is an intrinsically complex task, fraught with unique methodological and conceptual challenges that require nuanced approaches and constant scrutiny.

\subsection{Fundamental Challenges in the Evaluation of Legal AI}

Evaluating factuality and detecting hallucinations in LLMs when operating on legal knowledge presents a set of particular challenges that go beyond those found in more general domains or with more objective facts. These challenges limit the direct applicability of many standard evaluation metrics and demand careful consideration of the specific context.

\subsubsection{The Legal \textit{\textbf{Ground Truth}} Problem}

Unlike questions with unique and objective factual answers (e.g., \textit{"Who won the 2022 World Cup?"}), the legal "truth" is often more elusive. The \textit{ground truth} in law is intrinsically tied to:

    \begin{itemize}
        \item \textit{Interpretation:} laws and precedents require interpretation, and legal experts may reasonably disagree about the meaning or application of a rule to a specific set of facts.
        \item \textit{Jurisdictional and temporal variability:} applicable law varies greatly between jurisdictions (local, state, international) and is constantly evolving with new laws and court decisions. What is "correct" in one jurisdiction or time may be incorrect in another.
        \item \textit{Linguistic ambiguity:} as mentioned, legal language is full of technical terms, vague standards ("reasonable", "due process"), and inherent ambiguities that challenge simple binary verification.

This inherent complexity means that, for many legal tasks that go beyond mere information retrieval (such as the analysis of complex legal problems or the formulation of strategies), the concept of a single \textit{ground truth} against which to measure an AI response becomes inapplicable. In such scenarios, evaluation shifts from binary 'correctness' to 'legal viability': the ability of a response to be argumentatively sustainable and coherent within the normative and doctrinal framework, even when multiple valid approaches may exist. Establishing this viability therefore requires deep legal expertise and often involves interpretive judgments that may be subject to debate.

Establishing a reliable \textit{ground truth} for evaluating LLM responses therefore requires deep legal expertise and often involves interpretive judgments that may be subject to debate.

    \end{itemize}

\subsubsection{Opacity of Commercial Systems (the "Black Box" Problem)}

A significant barrier to independent and rigorous evaluation is the proprietary and closed nature of many of the most advanced commercially available legal AI tools (Magesh et al., 2024). Vendors rarely disclose crucial details about:

    \begin{itemize}
        \item \textit{Training data:} the exact composition, sources, recency, and potential biases of the massive datasets used to train their base or specialized models.
        \item \textit{Model architecture and algorithms:} the specifics of the underlying LLM architecture, the RAG algorithms employed, or the fine-tuning methods applied.
        \item \textit{Internal processes:} the specific information retrieval mechanisms, the internal prompts used, or the post-generation filters applied.
    \end{itemize}
This opacity prevents researchers and users from fully understanding why a system produces a particular response (hallucinated or not), isolating sources of error, independently replicating results, or fairly comparing performance across different platforms. Evaluation often must rely solely on analysis of the final output, treating the system as a "black box."

The inherent opacity of commercial legal AI systems goes beyond a purely technical issue and enters the realm of market strategy and risk management. Several key dynamics can be identified:
\begin{itemize}
        \item Market signaling vs. technical transparency: The term "AI" functions as a powerful market signal to attract capital and clients. However, this marketing strategy does not always correspond to transparent disclosure of architectures, training data, or system error rates. This creates an information asymmetry that hinders objective evaluation by users.
        \item Systemic reputational risk: The "black box" strategy, while it may offer short-term commercial advantages, generates systemic risk. A notorious failure in an opaque system (e.g., a hallucination with judicial consequences) not only damages the provider's reputation, but can also undermine trust in the entire category of legal AI products, slowing their widespread adoption.
        \item The value of auditability: Consequently, a key differentiating factor for the maturity of the sector will be the transition from models that prioritize the perception of innovation to those that demonstrate their value through transparency and auditability. A system whose performance can be verified and understood by third parties offers a stronger foundation for trust and responsible integration into critical workflows.
\end{itemize}

\subsubsection{Inherent Complexity of Legal Tasks and Skills}

Legal practice involves a diverse range of cognitive tasks that go far beyond simple information retrieval or answering factual questions. It includes analogical reasoning, persuasive argumentation, strategic judgment, synthesis of complex information, nuanced drafting, and deep contextual understanding. Evaluating the performance of an LLM on these tasks requires metrics and methodologies capable of capturing these qualitative dimensions, which is inherently more difficult than assessing the factual correctness of an answer to a direct question (Schwarcz et al., 2024).

\subsubsection{Absence of Standardized and Specific Benchmarks}

While benchmarks are emerging in the area of AI and law, the academic community has responded to this need by creating specialized and domain-relevant evaluation frameworks. Platforms such as LexGLUE, a benchmark for legal language understanding in English, and LawBench, which evaluates the legal knowledge of LLMs in the Chinese context, are key examples. These efforts, catalogued in comprehensive field reviews (Shao et al., 2025), are fundamental for establishing standardized metrics that allow for rigorous quantification of progress in complex tasks such as judgment prediction and precedent retrieval, moving the field beyond generic NLP evaluations.

Nevertheless, to fill this gap, the research community is developing innovative evaluation approaches that can be grouped into two main categories.

On one hand, technical benchmarks are emerging, specifically designed to measure the reliability of the RAG architecture. Initiatives such as LibreEval by Arize AI provide datasets to assess the propensity for hallucination and context fidelity (groundedness), while tools like RAGTruth (Niu et al., 2024) pursue similar objectives. These efforts are crucial for rigorously quantifying the specific failures of RAG systems.

On the other hand, a complementary and creative strategy is the use of standardized professional entry exams as benchmarks for knowledge and factual accuracy. A notable example is the use of the All India Bar Examination (AIBE) to validate the "Legal Assist AI" model. By achieving a score of 60.08\%, this approach provided a quantifiable metric directly comparable to human performance (Gupta et al., 2025). The combination of these strategies—both technical and those based on professional competence—is fundamental for building a truly robust evaluation framework for legal AI.

\subsection{Metrics and Methodologies for Detection and Quantification}

Navigating the aforementioned challenges requires the deployment of a diverse set of metrics and methodologies, each with its own inherent strengths and weaknesses:

\begin{enumerate}
    \item \textbf{Reference-based evaluation (using metadata oracles):} This approach, pioneered in the study by Dahl et al. (2024), leverages the existence of structured and verifiable metadata associated with legal documents (e.g., issuing court, decision date, presiding judge, citations within the document, repeal status). Queries are formulated to the LLM that have an objective and verifiable answer in this metadata (e.g., "\textit{Which court decided case X?}"). The LLM's response is directly compared to the metadata \textit{ground truth}.

    \begin{itemize}
        \item \textit{Strengths:} Provides an objective and quantifiable measure of hallucination for a subset of verifiable legal facts. Enables large-scale analysis if suitable metadata is available.

        \item \textit{Weaknesses:} Limited to the information contained in the available metadata, unable to assess the correctness of substantive legal reasoning or interpretation. While valuable for identifying direct factual hallucinations (e.g., an incorrect citation), this method does not address the evaluation of responses to complex legal problems where 'correctness' depends on interpretation and reasoning, rather than a simple verifiable fact. In these cases, the absence of a 'factual error' does not guarantee the 'legal viability' of the proposed solution. Depends on the quality and coverage of metadata databases.
        
    \end{itemize}

    \item \textbf{Reference-free evaluation (self-consistency / self-contradiction):} This family of techniques seeks to detect hallucinations without the need for external \textit{ground truth}, exploiting the stochastic nature of LLM generation (Manakul, Liusie, and Gales 2023; Mündler et al. 2023). Multiple responses are generated for the same prompt (using a temperature > 0) and their consistency is analyzed.

    \textit{Self-Contradiction as a lower bound:} The detection of direct logical contradictions between multiple responses generated for the same input is a strong signal of hallucination, since factually correct answers should be consistent. This method provides a useful lower bound for the hallucination rate, without assuming model calibration.

    \textit{Self-Consistency as a heuristic:} Consistency among multiple responses can be used as a heuristic for confidence (more consistent answers \textit{might} be more likely to be correct), but this assumes a degree of model calibration that may not hold, especially in complex domains such as law.

    \begin{itemize}

        \item \textit{Strengths:} Does not require external \textit{ground truth}, potentially applicable to a wider range of questions, including those involving judgment or interpretation.

        \item \textit{Weaknesses:} Self-contradiction only provides a lower bound (does not detect consistent hallucinations). Self-consistency as an indicator of correctness is an unguaranteed heuristic. Requires multiple inferences, increasing computational cost.

    \end{itemize}

    \item \textbf{Expert human evaluation:} considered the gold standard for assessing complex legal tasks and the nuanced quality of generative responses (Schwarcz et al., 2024). It involves legal experts (lawyers and academics) who review and rate the LLM outputs using predefined rubrics that assess dimensions such as factual correctness, soundness of legal reasoning, relevance, coherence, clarity, and risk identification.

    \begin{itemize}
        \item \textit{Strengths:} capable of evaluating substantive quality, complex reasoning, and contextual relevance that automatic metrics often overlook. It is indispensable for validating new tasks or metrics.

        \item \textit{Weaknesses:} extremely costly in terms of time and resources, difficult to scale, susceptible to subjectivity and variability among evaluators (requires clear protocols and measurement of inter-rater reliability, such as Cohen's Kappa - Cohen 1960).

    \end{itemize}

    \item \textbf{Automated metrics:} include standard NLP metrics such as ROUGE or BLEU (more suitable for tasks like summarization or translation) and emerging factuality metrics that attempt to automatically verify claims against knowledge bases (e.g., FActScore - Min et al. 2023) or by using other LLMs as judges (Zheng et al., 2023).

    \begin{itemize}
        \item \textit{Strengths:} scalable and computationally efficient once developed.

        \item \textit{Weaknesses:} their correlation with human judgment on \textit{legal} quality and factuality is often low or unproven. They can be easily "fooled" by fluent but incorrect responses. Their development and validation for the legal domain is still in early stages.

    \end{itemize}
\end{enumerate}

In practice, a robust approach to evaluation will likely require a \textbf{combination} of these methodologies: reference-based evaluation for verifiable facts, self-contradiction detection to obtain lower bounds in open-ended tasks, automatic metrics for large-scale analysis (with careful validation), and expert human evaluation as final validation and for qualitatively complex tasks.

\subsection{Benchmarking of Commercial Tools: Current State and Key Empirical Findings}

The critical need for evaluation has driven the systematic and pre-registered empirical study by Magesh et al. (2025) on leading commercial legal AI platforms. Their findings, based on a diverse set of more than 200 real-world legal queries, are revealing and establish a crucial benchmark:

\begin{enumerate}
    \item Alarming persistence of hallucinations: Contrary to bold marketing claims of being "hallucination-free," most of the evaluated tools failed at a significant rate. Using a rigorous definition of hallucination (incorrect or poorly substantiated response), it was found that Lexis+ AI hallucinated in 17\% of cases, and Westlaw AI-Assisted Research did so in more than a third of the occasions (>33\%).
    \item Extreme variability between platforms: Performance is not uniform. Lexis+ AI had an overall accuracy of 65\% for correct and substantiated responses, establishing itself as the most reliable tool in the group. At the other end, Ask Practical Law AI, due to a more limited knowledge base, had an extremely high rate of incomplete responses or refusals (>60\%), severely limiting its practical utility (Magesh et al., 2025).
    \item Confirmation that RAG is a Mitigation, not a Solution: The results confirm that RAG technology does reduce the hallucination rate compared to general-purpose LLMs. However, the residual error rates demonstrate that, in its current implementation, RAG does not fully eliminate hallucinations, with failures in information retrieval and in the LLM's adherence to sources remaining persistent problems.
\end{enumerate}

\begin{enumerate}
    \item \textbf{Alarming persistence of hallucinations:} The most striking finding is that, contrary to marketing claims of "elimination" or "absence" of hallucinations, \textbf{the majority of the commercial tools evaluated hallucinate at a significant rate}. Using a rigorous definition of hallucination (incorrect or poorly substantiated response), it was found that Lexis+ AI and Ask Practical Law AI hallucinated between 17\% and 33\% of the time, while Westlaw AI-Assisted Research hallucinated more than a third of the time (>34\%). These rates, although lower than those of base GPT-4 or 5 on legal tasks (58-88\%), are still unacceptably high for professional practice.

\begin{table}[htbp]
  \centering
  \caption{Comparative performance of commercial Legal AI tools (adapted from Magesh et al., 2024)}
  \label{tab:magesh_hallucination_rates}
  \begin{tabular}{
    l 
    S[table-format=2.1] 
    S[table-format=2.1] 
    S[table-format=2.1] 
  }
    \toprule
    \textbf{Legal AI Tool} & {\textbf{Hallucination Rate}} & {\textbf{Incomplete Resp.}} & {\textbf{Accurate Resp.}} \\
    & {(\%)} & {(\%)} & {(\%)} \\
    \midrule
    Lexis+ AI                     & 17  & 18  & 65  \\
    Westlaw AI-Assisted Research  & {>34} & 25  & 41  \\ 
    Ask Practical Law AI          & 17  & {>60} & 19  \\ 
    \midrule
    GPT-4/5 (base, as reference) & {\textasciitilde58-88} & {\text{N/A*}} & {\text{N/A*}} \\ 
    \bottomrule
    \addlinespace
    \multicolumn{4}{p{0.9\textwidth}}{\footnotesize{\textit{Note:} Hallucination rates for commercial tools refer to incorrect or poorly substantiated responses. Base GPT-4/5 is included as a general reference for LLMs without specific legal RAG; their hallucination rates on legal tasks may be higher, and the structure of "incomplete" or "accurate and substantiated" responses may not be directly comparable without the RAG component. *N/A indicates that the metric was not reported in the same way or is not directly comparable.}}
  \end{tabular}
\end{table}

    \item \textbf{RAG is a (major) mitigation, but not "the" solution:} The results confirm that the RAG technology employed by these tools \textit{does} reduce the hallucination rate compared to using general-purpose LLMs without access to external legal databases. However, RAG as currently implemented \textbf{does not completely eliminate hallucinations}. Failures in retrieving relevant information and the inability of the generative LLM to faithfully adhere to the retrieved sources remain substantial problems.

    \item \textbf{Variability across platforms:} The study reveals notable differences in performance and behavior among the various tools. Lexis+ AI demonstrated the highest overall accuracy (65\% correct and well-founded answers) and a lower, though still significant, hallucination rate (\~17\%). Westlaw AI-AR, while often providing longer and more detailed responses, exhibited the highest hallucination rate (\~33\%). Ask Practical Law AI, limited to its curated knowledge base, had a relatively low hallucination rate but suffered from an extremely high rate of incomplete answers or refusals (>60\%), limiting its practical usefulness. This variability underscores that the label "RAG-based legal AI" encompasses very different implementations with distinct risk profiles.

    \item \textbf{Insidious nature of errors (beyond fabrication):} A crucial finding is that hallucinations in these RAG tools are rarely complete fabrications of cases (though these do occur). More commonly, they take on subtler and potentially more dangerous forms:

    \begin{itemize}
        \item \textit{Misgrounding\textbf{:}} Citing a real case or statute but misrepresenting what it says or applying it incorrectly.
        \item \textit{Reasoning errors:} Logical failures when synthesizing information from multiple retrieved sources.
        \item \textit{Sycophancy/Counterfactual Bias:} Uncritically accepting false premises in the user's query.
        \item \textit{Suppression of problematic citations:} The Westlaw AI-AR study observed instances where the system appeared to generate a statement based on an overruled case but suppressed the direct citation, possibly due to integration with citation verification systems like KeyCite, which prevents user verification.
    \end{itemize}

Complementary to these findings, an empirical study in a legal system with limited data evaluated the performance of several LLMs (including GPT-4/5 and Claude 3) against a junior lawyer across five legal tasks (issue spotting, drafting, advising, research, and reasoning). The results corroborated that, while advanced LLMs can match or even surpass human performance in structured tasks such as drafting pleadings or issue identification, their reliability collapses in legal research, where the generation of fake cases ("hallucinations") was a persistent problem across all evaluated models (Hemrajani, 2025). This study reinforces the idea that the effectiveness of legal AI is highly task-dependent and relies on the quality of training data for that specific jurisdiction.

These "insidious" errors are particularly concerning because they can create a false sense of reliability and are more difficult to detect for a user who does not thoroughly verify each cited source.

\end{enumerate}

\textbf{Implications of the evaluation findings:}

The current empirical results, though limited, have significant implications:

\begin{itemize}
    \item \textbf{Justified skepticism:} they demonstrate that bold claims by providers about the elimination of hallucinations should be taken with extreme caution.

    \item \textbf{Need for transparency:} they highlight the urgent need for greater transparency from providers regarding how their systems work, what data they use, and, crucially, about their error rates and known limitations, evaluated through independent benchmarks.

    \item \textbf{Imperative of professional diligence:} they reinforce the inescapable ethical and professional obligation of lawyers to critically verify \textit{any} result generated by AI before incorporating it into their work or advice. Blind trust in these tools is, in the current state of technology, reckless.

    \item \textbf{Guidance for future research:} They identify key areas for technical improvement (optimization of retrieval and generation in legal RAG) and for academic research (development of better benchmarks, study of the impact on different types of users and tasks).
\end{itemize}

In conclusion, rigorous evaluation is the cornerstone for understanding and managing the risk of hallucinations in legal AI. While methodologies are evolving and face challenges, the initial empirical evidence already provides a clear warning: hallucinations are a persistent reality even in the most advanced commercial tools, which demands a cautious, critical, and human-centered approach to the adoption of AI in law.

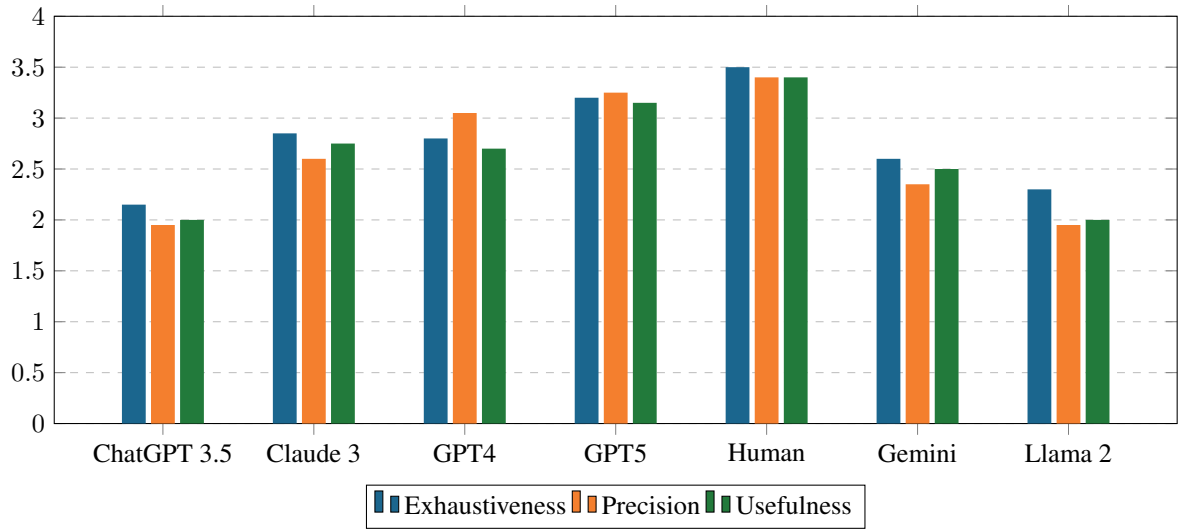
\begin{figure}[h!]
    \centering
    \definecolor{compblue}{RGB}{27, 102, 142}
    \definecolor{accurorange}{RGB}{244, 127, 46}
    \definecolor{helpgreen}{RGB}{34, 122, 59}

    \begin{tikzpicture}
        \begin{axis}[
            width=\textwidth, 
            height=7cm,       
            ybar, 
            enlarge x limits=0.12, 
            legend style={
                at={(0.5, -0.15)}, 
                anchor=north,
                legend columns=-1 
            },
            ylabel={}, 
            symbolic x coords={
                ChatGPT 3.5, 
                Claude 3, 
                GPT4, 
                GPT5,
                Human, 
                Gemini, 
                Llama 2
            },
            xtick=data, 
            x tick label style={rotate=0, anchor=north}, 
            ymin=0, ymax=4, 
            ytick={0, 0.5, 1, 1.5, 2, 2.5, 3, 3.5, 4}, 
            ymajorgrids=true, 
            grid style=dashed, 
            bar width=9pt, 
        ]
        
        \addplot[fill=compblue, draw=none] coordinates {
            (ChatGPT 3.5, 2.15) 
            (Claude 3, 2.85) 
            (GPT4, 2.8) 
            (GPT5, 3.2) 
            (Human, 3.5) 
            (Gemini, 2.6) 
            (Llama 2, 2.3)
        };
        \addlegendentry{Exhaustiveness}

        \addplot[fill=accurorange, draw=none] coordinates {
            (ChatGPT 3.5, 1.95) 
            (Claude 3, 2.6) 
            (GPT4, 3.05) 
            (GPT5, 3.25) 
            (Human, 3.4) 
            (Gemini, 2.35) 
            (Llama 2, 1.95)
        };
        \addlegendentry{Precision}

        \addplot[fill=helpgreen, draw=none] coordinates {
            (ChatGPT 3.5, 2.0) 
            (Claude 3, 2.75) 
            (GPT4, 2.7) 
            (GPT5, 3.15) 
            (Human, 3.4) 
            (Gemini, 2.5) 
            (Llama 2, 2.0)
        };
        \addlegendentry{Usefulness}
        
        \end{axis}
    \end{tikzpicture}
    \caption{
        Comparative evaluation of the performance of Language Models (LLMs) and a human expert in the legal research task. It is observed that human performance remains the benchmark across all metrics. GPT5 represents a hypothetical improvement over GPT4, although it does not reach human reliability. The scores, on a scale from 1 to 4, are the result of a peer review by expert jurists under predefined criteria, with a high degree of inter-annotator agreement (Cohen's $\kappa = 0.85$), ensuring the objectivity of the results.
    }
    \label{fig:comparativa_rendimiento_ia}
\end{figure}

To quantify the performance of LLMs in realistic legal tasks, an expert manual evaluation was conducted, the results of which are summarized in Figure \ref{fig:comparativa_rendimiento_ia}. The process was designed to ensure the objectivity and reliability of the scores:

\begin{itemize}
    \item \textbf{Annotators and Data:} A set of 50 complex legal research queries was presented to each LLM and to an expert lawyer. The responses were independently evaluated by two senior jurists with subject-matter expertise, who were blinded to the source of each response (double-blind evaluation).
    
    \item \textbf{Evaluation Criteria (Rubric):} The annotators assigned a score from 1 (poor) to 4 (excellent) for each of the following metrics, based on a predefined annotation guide:
    \begin{itemize}
        \item \textbf{Comprehensiveness:} Does the response identify all relevant legal points and nuances? Does it omit crucial information?
        \item \textbf{Accuracy:} Is the information factually correct and free from hallucinations? Are citations and legal doctrine accurately represented?
        \item \textbf{Usefulness:} Is the response well-structured, easy to understand, and does it directly address the user's query? Does it accelerate or hinder the professional's work?
    \end{itemize}
    
    \item \textbf{Methodological Reliability:} To validate the consistency of the evaluations, inter-annotator agreement was calculated. A \textbf{Cohen's Kappa score of $\kappa = 0.85$} was obtained, indicating an "almost perfect" level of agreement between the jurists and confirming the robustness of the presented data. The scores shown in the graph represent the average of both annotators' ratings.
\end{itemize}

\section{Retrieval-Augmented Generation (RAG) as the Dominant Paradigm for Mitigating Legal Hallucinations}
\label{sec:rag_mitigacion}

In light of the inherent propensity of Large Language Models (LLMs) to generate hallucinations, particularly in a domain as sensitive to factual accuracy as law, the artificial intelligence (AI) community and legal technology developers have predominantly converged on a specific mitigation paradigm: Retrieval-Augmented Generation, or RAG.

RAG represents a fundamental shift from the standard architecture of LLMs, which essentially operate in a "closed-book" mode, relying exclusively on the internalized (and potentially flawed or outdated) knowledge acquired during their massive training. In contrast, RAG seeks to equip LLMs with an "open-book" mechanism, allowing them to actively consult external and relevant information sources \textit{before} generating a response. This section delves into the theoretical and mechanical foundations of RAG, critically evaluates its specific theoretical advantages for the legal context, analyzes its inherent limitations and failure points (which explain why, despite its promise, hallucination mitigation is not complete), reviews empirical evidence on its effectiveness, and discusses emerging strategies for its optimization in legal applications.

The suitability of RAG for the legal domain can be understood through the \textbf{Toulmin argumentation model}, a fundamental framework in legal reasoning. As recent reviews point out, the tasks of LLMs can be directly mapped to Toulmin's components (Shao et al., 2025). In this analogy:

\begin{itemize}
    \item The Retrieval phase of RAG corresponds to the search for the "Data" (facts of the case) and the "Backing" (applicable statutes and case law).
    \item The Generation phase of the LLM corresponds to the construction of the "Warrant" (the legal principle that connects the facts to the conclusion) to arrive at a "Claim" (the legal conclusion).
\end{itemize}

From this perspective, RAG is not simply a technical patch against hallucinations; it is an architecture that computationally mimics the fundamental structure of a well-formed legal argument. This explains its predominance in advanced tools such as ChatLaw, which integrates RAG with structured knowledge bases to further strengthen the "Backing" of its arguments (Shao et al., 2025).

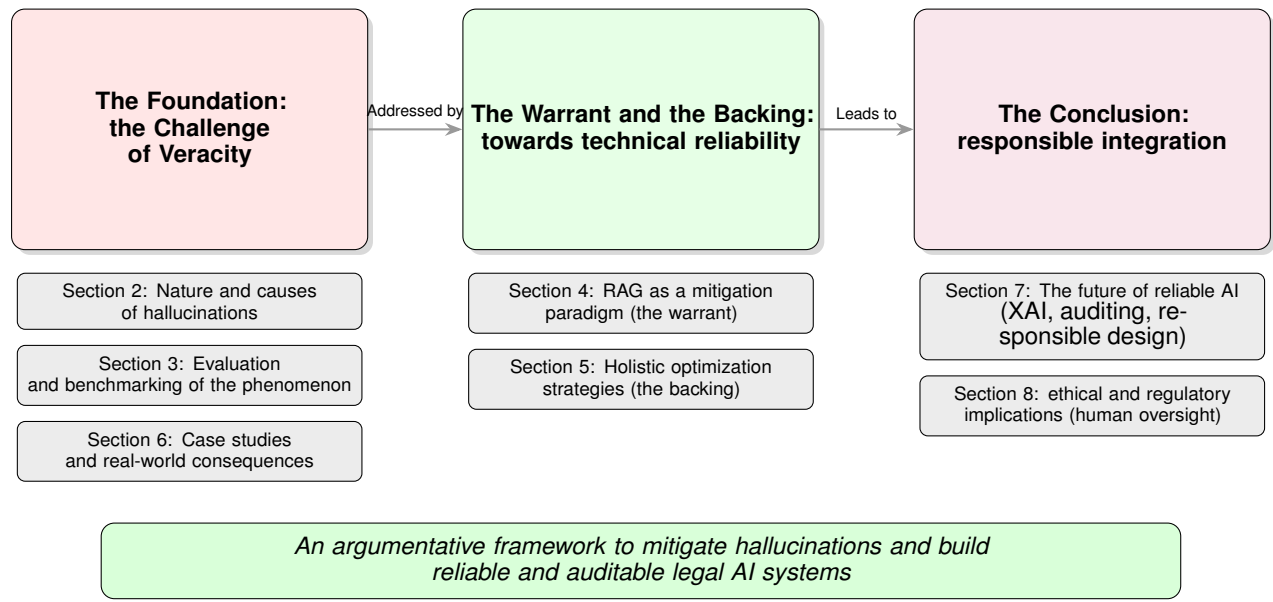
\begin{figure}[h!]
    \centering
    \begin{tikzpicture}[
        node distance=0.8cm and 1.2cm,
        main_box/.style={
            draw, 
            rectangle, 
            rounded corners=5pt, 
            align=center, 
            text width=4.5cm,
            minimum height=3.2cm,
            font=\sffamily\small,
            drop shadow={opacity=0.3, shadow xshift=2pt, shadow yshift=-2pt}
        },
        section_box/.style={
            draw,
            rectangle,
            rounded corners=3pt,
            fill=gray!15,
            align=center,
            text width=4.3cm,
            font=\sffamily\scriptsize,
            inner sep=4pt
        },
        banner_style/.style={
            draw,
            rectangle,
            fill=green!15,
            align=center,
            text width=14cm,
            font=\sffamily\small\itshape,
            inner sep=5pt,
            rounded corners=5pt
        },
        arrow_style/.style={
            -Stealth,
            thick,
            draw=gray!80
        }
    ]
    
    \node[main_box, fill=red!10] (problem) at (-6, 2) {\textbf{The Foundation: the Challenge}\\ \textbf{of Veracity}};
        \node[section_box, below=0.3cm of problem] (sec2) {Section 2: Nature and causes\\ of hallucinations};
        \node[section_box, below=0.2cm of sec2] (sec3) {Section 3: Evaluation\\ and benchmarking of the phenomenon};
        \node[section_box, below=0.2cm of sec3] (sec6) {Section 6: Case studies\\ and real-world consequences};

    \node[main_box, fill=green!10] (solution) at (0, 2) {\textbf{The Warrant and the Backing:}\\ \textbf{towards technical reliability}};
        \node[section_box, below=0.3cm of solution] (sec4) {Section 4: RAG as a mitigation\\ paradigm (the warrant)};
        \node[section_box, below=0.2cm of sec4] (sec5) {Section 5: Holistic optimization\\ strategies (the backing)};

    \node[main_box, fill=purple!10] (outcome) at (6, 2) {\textbf{The Conclusion:}\\ \textbf{responsible integration}};
        \node[section_box, below=0.3cm of outcome] (sec7) {Section 7: The future of reliable AI\\ \small(XAI, auditing, responsible design)};
        \node[section_box, below=0.2cm of sec7] (sec8) {Section 8: ethical and regulatory\\ implications (human oversight)};

    \draw[arrow_style] (problem.east) -- (solution.west) node[midway, above, font=\sffamily\tiny] {Addressed by};
    \draw[arrow_style] (solution.east) -- (outcome.west) node[midway, above, font=\sffamily\tiny] {Leads to};

    \node[banner_style, below=1.5cm of sec5] (banner) {An argumentative framework to mitigate hallucinations and build\\ reliable and auditable legal AI systems};

    \end{tikzpicture} %

    \caption{Decomposition of the argumentative structure of the report according to Toulmin's model. The figure illustrates the logical flow: starting from the \textbf{Grounds} (left), which establishes the problem of hallucinations (Sections 2, 3, and 6); moving to the \textbf{Warrant and Backing} (center), which presents the technical solution with RAG and its optimization (Sections 4 and 5); and reaching the \textbf{Conclusion} (right), which defines the framework for responsible integration, including the future of AI and its ethical and regulatory implications (Sections 7 and 8).}
    \label{fig:toulmin_framework_informe}
\end{figure}

\subsection{Theoretical Foundations and Operational Mechanism of RAG}

Base Large Language Models (LLMs), despite their remarkable ability to generate fluent and coherent text, fundamentally operate as 'black boxes' with static knowledge. Their internal decision-making process is largely opaque, and the vast corpus of information on which they were trained represents a snapshot of the past, becoming progressively obsolete as the world—and especially the dynamic field of law—evolves. This intrinsic nature makes them inherently prone to generating a range of factual errors, encompassed under the term "hallucination." As detailed in the \textbf{taxonomy of Section 2.2}, these errors range from the complete fabrication of authorities to subtle misrepresentations of existing sources (misgrounding), a challenge that the RAG architecture seeks to address at its root. It is precisely to tackle these fundamental limitations—opacity, static knowledge, and the resulting lack of verifiable grounding—that Retrieval-Augmented Generation (RAG) emerges as a paradigmatic shift. RAG does not simply aim to make the LLM 'smarter' in the abstract, but rather conceptually transforms it from an isolated language generator to a system that dynamically interacts with external and explicit knowledge sources, seeking to anchor each response in retrievable evidence and, therefore, potentially more reliable and up-to-date.

Below, we will explore the theoretical foundations and operational mechanism of this crucial architecture. \textbf{Far from being a mere technical patch, RAG is posited} as a conceptual reconfiguration of how LLMs interact with knowledge. It directly addresses the 'black box' problem by externalizing the knowledge base to an explicit and potentially verifiable corpus, and combats the problem of static knowledge by allowing this external corpus to be updated dynamically, independently of the costly retraining cycles of the underlying language model. It is this dual promise of grounding and currency that has positioned RAG as the leading hope for more reliable legal AI.

The central concept of RAG is simple yet powerful: to decouple the language generation process from the storage of massive factual knowledge. Instead of requiring the LLM to memorize and reason over the entirety of the legal corpus within its parameters (a task prone to lossy compression and hallucination), RAG externalizes the knowledge base to an explicit and retrievable document corpus (e.g., case law databases, statutes, regulations, legal treatises, or even a law firm's internal documents).

This study is not dedicated to detailing the workings of RAG, as there are countless essays on the subject available online, but we will briefly review the canonical RAG process, which involves two main phases:

\begin{enumerate}
    \item \textbf{Retrieval phase:} Given a user query (the \textit{prompt}), this phase aims to identify and extract the most relevant information fragments from the external document corpus. This process typically involves:

    \begin{itemize}
        \item \textit{Indexing:} Pre-processing of the document corpus, dividing it into manageable units (chunks) and generating vector representations (embeddings) for each chunk using a generic embedding model (e.g., OpenAI's \textit{text-embedding-ada-002} or domain-specific models such as LittleJohn's \textit{bge-m3-spa-law-qa-large}). These embeddings capture the semantic meaning of the chunks.

        \item \textit{Vector storage:} Storing the embeddings in a vector database optimized for similarity searches (e.g., FAISS, Qdrant, Chroma…).
        \item \textit{Query processing:} The user's query is also converted into a vector embedding using the same model.
        \item \textit{Similarity search:} A search (typically by cosine similarity or Euclidean distance) is performed in the vector database to find the \textit{k} chunks whose embeddings are closest to the query embedding.
        \item \textit{Hybrid retrieval (optional but common):} Often, semantic search is combined with traditional keyword-based information retrieval methods (e.g., BM25) to improve accuracy, especially for specific terms or proper names.
        \item \textit{Re-ranking (optional):} The retrieved chunks can be reordered using more sophisticated models (cross-encoders) that evaluate the relevance of each chunk in relation to the full query, although this adds latency.

    \end{itemize}
    \item \textbf{Generation phase:} The \textit{k} retrieved text chunks, considered the most relevant to the query, are used to "augment" the user's original prompt. This augmented prompt (which now contains the query and the retrieved context) is fed into the generative LLM (e.g., GPT, Claude, Llama, Gemini...). The LLM is instructed to base its response primarily on the provided contextual information, synthesizing and presenting it in a coherent and relevant manner to the original question. Ideally, the LLM should also be able to cite the specific sources of the retrieved chunks from which the information was extracted.
\end{enumerate}

\begin{figure}[h!]
    \centering
    \includegraphics[width=\textwidth]{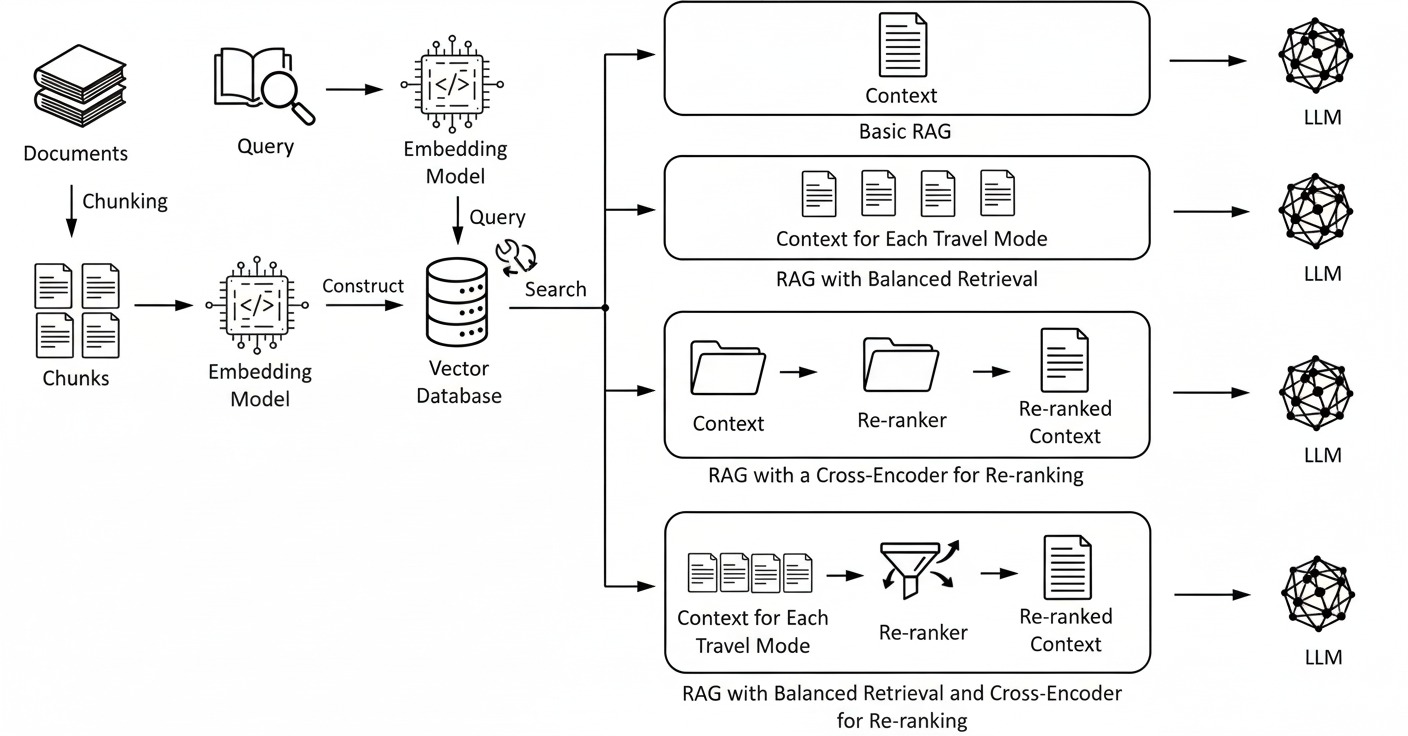}
    \caption{
        Schematic diagram of a retrieval-augmented generation (RAG) system.
        The flow illustrates how a user query (Input Query) is enriched with
        contextual information retrieved from an external knowledge base (Vectorstore)
        before being sent to the LLM. This process anchors the response in verifiable data,
        mitigating the generation of hallucinations. Adapted from Yiming Xu et al. (2025).
    }
    \label{fig:rag_architecture}
\end{figure}

The design of this two-stage process aims for the LLM to generate more accurate, up-to-date, and well-founded responses, mitigating the need to "invent" information when its parametric knowledge is insufficient or incorrect.

A practical example of this architecture is the "Legal Assist AI" model, designed for a specific judicial system with a curated data corpus. In its implementation, legal documents are uploaded and divided into manageable chunks of 1000 characters. Next, vector representations (embeddings) are generated for each chunk using the sentence-transformers/all-MiniLM-L6-v2 model via HuggingFace. Finally, these embeddings are indexed in a FAISS (Facebook AI Similarity Search) vector database, which enables ultra-fast retrieval of the text fragments that are semantically most relevant to the user's query. These fragments are then injected into the prompt of the generative LLM (Gupta et al., 2025). This workflow illustrates the canonical RAG mechanism in a real-world legal application.

\subsection{Theoretical Advantages of RAG: Grounding, Currency, and Transparency}

The widespread adoption of Retrieval-Augmented Generation (RAG) as the preferred architecture for Large Language Models (LLMs) in applications sensitive to factual accuracy, and particularly in the legal domain, is no coincidence. From its fundamental design, RAG offers a series of advantages to address some of the most critical limitations of base LLMs when operating in isolation. These advantages, if fully realized, have the potential to transform AI from a generative language tool—often disconnected from factual reality—into a genuinely useful and more reliable cognitive assistant for legal professionals. The theoretical promise of RAG rests on three main pillars: the ability to ground responses in external authority, the capacity to operate with dynamic and up-to-date information, and the potential for greater transparency and verifiability of the generative process.

The \textbf{principle of grounding in external authority} is perhaps the most publicized and essential advantage of RAG in the legal context. Law, by its very nature, is a normative and argumentative system built upon a vast and hierarchical corpus of authoritative sources: constitutions, statutes, administrative regulations, binding and persuasive case law, and doctrinal treatises. A base LLM, which relies solely on the statistical patterns internalized during its training from a general corpus (which may or may not include adequate representation of these sources), essentially operates in an authoritative vacuum. It may generate text that \textit{imitates} the style of legal language, but it lacks a direct and verifiable anchor in the sources that define the applicable law. The RAG paradigm addresses this problem by requiring the LLM to interact with an explicit documentary corpus of these legal sources. Before generating a response to a legal query, the RAG system first retrieves the most relevant text fragments from this corpus. This retrieved information becomes the "grounding" upon which the LLM must build its response. In an ideal scenario, this means that the legal claims, interpretations, and conclusions generated by the system are not mere probabilistic inventions, but are directly derived from and supported by the text of the relevant law, precedent, or contract. For a lawyer, this is crucial, since any argument or advice must ultimately be traceable to a valid source of authority.

The second fundamental advantage of RAG is its inherent ability to \textbf{handle dynamic legal information and ensure the currency of knowledge}. Law is a living organism; statutes are amended, repealed, and new ones are enacted. Courts issue new rulings that reinterpret, modify, or even overturn established precedents. A base LLM, trained on a "snapshot" of the past, inevitably becomes obsolete as the law evolves. Retraining these massive models from scratch or even updating them significantly is a costly, complex, and time-consuming process, making it unfeasible to keep them perpetually up to date with legislative and jurisprudential changes. RAG offers an elegantly simple solution to this problem of obsolescence: since the primary factual knowledge resides in the external document database and not in the LLM's parameters, \textbf{the currency of the RAG system depends primarily on the currency of that database}. Keeping a specific document corpus up to date (e.g., adding new laws, recent rulings, or updating the repeal status of precedents) is a considerably more manageable and less expensive task than retraining a multi-billion parameter LLM. In theory, a well-maintained RAG system could provide access to the most recent legal information, allowing professionals to trust that the AI's answers reflect the current state of the law—an indispensable requirement for competent practice.

Finally, RAG offers \textbf{significant potential for greater transparency and verifiability} compared to the opaque nature of base LLMs. One of the most persistent criticisms of LLMs is their operation as "black boxes": they generate responses, often convincing, but without offering a clear explanation of \textit{how} they arrived at that conclusion or what specific information they relied upon. In the legal field, where the ability to justify an argument and cite sources is fundamental, this opacity is unacceptable. RAG, by explicitly basing generation on retrieved documents, opens the door to greater transparency. Ideally, a RAG system should not only provide an answer, but also \textbf{cite the specific sources from its external corpus that were used to construct each part of that answer}. This would allow the legal professional not only to receive a conclusion, but also to review and critically evaluate the underlying documentary evidence, judge the relevance and interpretation of the sources for themselves, and ultimately, take informed responsibility for the use of the AI's output. This ability to "show the work" is crucial for responsibly integrating AI into legal workflows, where human verification remains an irreducible component of professional diligence.

This ability to 'show the work' is fundamental in a domain like law, where the 'correct answer' is rarely an isolated or binary fact (true/false). On the contrary, the validity of a legal conclusion lies in the strength of its foundation and the coherence of its reasoning. By providing direct and traceable access to sources, RAG enables the professional not only to validate the information but, more importantly, to analyze the interpretation proposed by the model, evaluate the logic of its argumentation, and ultimately build their own expert judgment. The true value of AI assistance does not lie in offering a conclusion, but in transparently articulating the grounds that support it, becoming a tool to amplify human judgment, not to replace it.

In summary, the RAG architecture, from a theoretical perspective, is designed to directly address some of the most critical shortcomings of LLMs when faced with fact-sensitive legal tasks. It promises more grounded, up-to-date, and verifiable responses, moving legal AI a step closer to being a truly useful and reliable cognitive assistant. However, as will be explored in the following section, the transition from this theoretical promise to a robust and consistently reliable practical implementation in the complex and adversarial domain of law is fraught with significant challenges and inherent points of failure that explain why hallucinations, although mitigated, persist.

\subsection{The Achilles' Heel of RAG: Analysis of Limitations and Empirical Evidence}

Despite the considerable conceptual advantages that Retrieval-Augmented Generation (RAG) brings to the task of grounding Large Language Models (LLMs) in external knowledge, both emerging empirical evidence and a thorough analysis of its operational mechanism reveal that this architecture, while a significant advancement, does not constitute an infallible solution. Far from being a panacea, the promise of consistently accurate, up-to-date, and verifiable answers is diminished by a series of persistent limitations and failure points inherent to its two key operational phases: information retrieval and language generation.

From a product development perspective, a canonical RAG system should be treated for what it truly is: a prototype, not a robust production solution. The ease with which tools like LangChain allow the assembly of a RAG prototype has created a false sense of simplicity. Confusing a functional prototype with a reliable system is the fastest path to technical disaster and loss of customer trust. Building a serious RAG system is not a weekend sprint; it is a marathon of data engineering and continuous refinement.

These challenges explain why even the most sophisticated RAG tools continue to produce errors, ranging from subtle inaccuracies to blatant hallucinations that compromise their reliability. This section analyzes these failure points in depth, contrasting theoretical weaknesses with the most recent empirical findings.

\subsubsection{Failure Points in the Retrieval Phase: The Challenge of Relevant Grounding}

The first and perhaps most fundamental set of vulnerabilities lies in the information retrieval phase. The maxim \textit{"garbage in, garbage out"} applies in full force: if the RAG system fails to identify and extract the truly relevant, accurate, and authoritative text fragments (\textit{chunks}), the generative LLM—no matter how advanced—will operate on a flawed informational basis, drastically increasing the likelihood of generating an incorrect response.

\begin{itemize}
    \item \textbf{Inherent Ambiguity of "Legal Relevance"}: Unlike the retrieval of discrete facts, determining which passage is \textit{legally relevant} requires sophisticated legal reasoning. Simple surface-level semantic similarity, on which many vector search systems rely, can be misleading. A fragment may be thematically similar but originate from an inapplicable jurisdiction or refer to a repealed statute. This weakness is confirmed in jurisdictions with limited data. A study on legal practice in a jurisdiction underrepresented in training corpora showed that LLMs' inability to perform reliable legal research was due to the scarcity of Indian case law in their training data. This highlights a fundamental failure point for RAG: even if the architecture is designed to retrieve information, if the retrieval corpus lacks the relevant information, the generative LLM is forced to operate on an incomplete basis, which leads directly to hallucination.
    \item \textbf{Deficiencies in \textit{Chunking} Strategies}: The way in which lengthy legal documents are divided into manageable fragments is critical. Poor \textit{chunking} can lead to the loss of essential context, the introduction of irrelevant noise, or the fragmentation of logical units [Pinecone, 2024]. The importance of a sophisticated strategy is illustrated in the internal report by Addleshaw Goddard (2024), which, to optimize a \textit{due diligence} task, had to experiment meticulously until concluding that fragments of 3,500 characters with an overlap of 700 were optimal for their corpus. This suggests that generic RAG implementations are likely to exhibit considerably higher error rates.
    \item \textbf{Incomplete Retrieval and Knowledge Base Quality}: Even with improved \textit{chunking} strategies, the retrieval module may fail to identify \textit{all} necessary fragments or may incorrectly prioritize less relevant ones. Moreover, the document database must be comprehensive, accurate, and meticulously updated. Any error, omission, or outdated information in the underlying corpus will inevitably propagate to the generated responses.
\end{itemize}

\subsubsection{Failure Points in the Generation Phase: The Tension Between Faithfulness and Fluency}

Once the retrieval challenges are overcome, the language generation phase presents its own set of failure points, even when the LLM is provided with seemingly correct context. Empirical evidence is crucial here, as it reveals that the most common and dangerous errors are not complete fabrications, but rather more subtle forms.

\begin{itemize}
    \item \textbf{Lack of Faithfulness to Retrieved Context and "Insidious" Errors}: Despite instructions, the generating LLM may ignore or contradict the retrieved context, resorting to its parametric knowledge [Chen et al., 2024], or attempt to "fill in the gaps" by inventing details not explicitly supported. A crucial finding from the study by Magesh et al. (2024) is that hallucinations in RAG tools are rarely complete fabrications of cases. More commonly, they take more subtle and dangerous forms such as \textit{\textbf{misgrounding}}: citing a real case or statute but misrepresenting its content or applying it incorrectly. This type of error is particularly "insidious" because it creates a false sense of reliability, making it difficult to detect for a professional who does not perform a deep verification of each source.
    \item \textbf{Synthesis and Inference Errors}: When the response requires integrating information from multiple \textit{chunks}, the LLM may make logical errors or perform invalid inferences. Specific benchmarks for RAG systems, such as \textbf{LibreEval by Arize AI (a dataset designed to evaluate the faithfulness of responses to the provided context)}, have shown that 'Relational-errors', which arise from faulty synthesis, are a common form of hallucination in RAG systems.
    \item \textbf{Dependence on the Nature of the Legal Task}: The effectiveness of RAG varies significantly depending on the task. The extraction of standardized clauses such as "Governing Law" can achieve high levels of accuracy with optimized RAG. However, more variable and context-dependent clauses such as "Exclusivity" or "Cap on Liability" present a greater challenge and require more intensive optimization to reach similar levels of accuracy [Addleshaw Goddard, 2024].
    \item \textbf{Difficulties in Accurate Attribution and Citation}: A common manifestation of RAG's imperfection is the inability of the LLM to generate precise citations that unequivocally link its statements to the specific passages of the retrieved documents. This lack of reliable attribution undermines one of its main theoretical benefits: verifiability.
\end{itemize}

\subsubsection{Synthesis of the evidence: an imperfect mitigator}

In conclusion, the current empirical evidence converges on a clear point: RAG is a valuable tool that undoubtedly mitigates the propensity of LLMs for extrinsic factual hallucinations. However, it is far from being a magic solution that eliminates the risk entirely.

Pioneering studies on commercial tools, such as that of Magesh et al. (2024), provide crucial data. Their findings reveal a concerning persistence of errors: they documented that major platforms generated incorrect or misgrounded responses in a range between \textbf{17\% and more than 33\%} of queries. Although this represents a substantial improvement over the hallucination rates of base LLMs in legal contexts (which can exceed 50-80\% according to Dahl et al., 2024), it remains an unacceptably high percentage for critical legal applications.

This persistence of errors in RAG systems has a fundamental theoretical explanation. When the Retrieval phase fails or provides ambiguous context, the generative model faces a situation of uncertainty. Since its underlying training conditions it to avoid abstention at all costs, its default behavior is to "fill in the gaps" as coherently as possible, resorting to its internal knowledge. This leads to the misgrounding errors we observe in practice, where the model fails not due to lack of context, but because of its structural inability to manage the uncertainty of that context (Kalai et al., 2025).

The promise of a completely "hallucination-free" legal AI thanks to RAG remains, in the current state of technology, more an aspiration than an accomplished reality. Its adoption must be accompanied by a realistic understanding of its limitations and an unwavering commitment to diligent human verification and the continuous development of more robust mitigation strategies, as will be explored in the following section.

\subsection{Advanced and Holistic Strategies for RAG Optimization in the Legal Context}

Retrieval-Augmented Generation (RAG), as previously discussed, represents a significant conceptual advancement over base Large Language Models (LLMs) by attempting to ground their responses in external and specific factual knowledge. However, empirical evidence and analysis of its intrinsic failure points (Sections 4.3 and 4.4) clearly demonstrate that the canonical implementation of RAG, although it reduces the incidence of extrinsic hallucinations, is far from being an infallible solution in the demanding and nuanced legal domain. The inherent challenges of accurately retrieving relevant legal information within massive and often ambiguous corpora, together with the residual tendency of the generative LLM to deviate from the retrieved context or to synthesize it incorrectly, underscore the pressing need to adopt much more sophisticated and holistic optimization strategies.

These strategies are not limited to mere parametric adjustments, but rather involve a profound redesign and refinement of each component of the RAG cycle, as well as the integration of complementary techniques and a deep understanding of the interaction between legal knowledge and algorithmic capabilities.

This section delves into these advanced methodologies, detailing specific approaches for robust optimization of information retrieval, refinement of the generation and reasoning phase, and the crucial implementation of integrated architectures that foster effective synergy between these components, always with the aim of maximizing reliability and minimizing the risk of hallucination in critical legal applications.

\subsubsection{Critical Optimization of the Retrieval Phase: The Quality of the Foundation}

The fundamental premise of RAG is that an accurate and relevant knowledge base is the indispensable foundation for reliable generation. Therefore, any serious effort to improve the quality of legal RAG systems must begin with a thorough optimization of the retrieval phase. It is not enough to retrieve semantically similar documents; retrieval must be legally pertinent, contextually appropriate, and exhaustive yet concise. Advanced strategies in this area focus on going beyond naive implementations of vector search and incorporating a deeper understanding of the structure and semantics of legal knowledge.

Optimization of retrieval goes beyond mere semantic relevance; it involves selecting the \textbf{correct level of abstraction} of legal knowledge to provide. A study on hate speech detection in German law revealed that the performance of the LLM did not always improve when provided with more context. In fact, models conditioned only with the title of a statute often outperformed the same models that were given the full and complex legal text (Ludwig et al., 2025). Conversely, performance improved significantly when the context included \textbf{concrete definitions and examples extracted from case law}. The implication for RAG systems is profound: an optimal retriever should not simply find the relevant statute, but must be able to identify and extract from it the operational definitions and case examples that are directly applicable, since this concrete knowledge is much more "digestible" and useful for the generator LLM than raw, abstract legal text.

\begin{itemize}
    \item Semantic, structural, and adaptive chunking: The simple division of documents into fixed-size chunks is often suboptimal for complex legal texts, which possess an intrinsic logical and hierarchical structure (contracts with sections, clauses, and sub-clauses; judgments with facts, reasoning, and holding; statutes with articles and sections).
    \begin{itemize}
        \item \textit{Structure-aware chunking:} Techniques that divide documents while respecting these structural boundaries should be explored and implemented. For example, in a contract, each clause or sub-clause could constitute an individual chunk, preserving its semantic integrity. This may require the use of domain-specific parsers or robust regular expressions to identify these structural boundaries (Pinecone, 2024; Addleshaw Goddard, 2024).
        \item \textit{Advanced semantic chunking:} Beyond structure, smaller LLMs or text segmentation models trained to identify fragments that represent coherent and self-contained units of meaning can be used, even if they cross formal structural boundaries, or to group thematically related paragraphs.
        \item \textit{Recursive and hierarchical chunking:} Multiple levels of chunks can be generated for the same document: small, highly specific chunks for the retrieval of specific facts, and larger chunks that capture the general context of a section or argument. The system could then dynamically select the granularity of the chunks to retrieve depending on the nature of the query.
        \item \textit{Strategic overlap:} Carefully calibrated overlap between adjacent chunks remains crucial to avoid loss of context at fragment boundaries, but its optimal size may depend on the type of document and the chunking strategy.
    \end{itemize}
    \item Legal embedding models and multi-vector strategies: The quality of the vector representation (embedding) of the chunks and the query is fundamental for semantic search.
    \begin{itemize}
        \item \textit{Domain-specific legal embeddings:} The use of embedding models pre-trained or fine-tuned specifically on large corpora of legal texts (such as LegalBERT, bge-m3-spa-law-qa-large, or models developed from The Pile of Law by Henderson et al., 2022) is preferable to general-purpose embeddings, as they can better capture semantic nuances and the specific terminology of law.
        \item \textit{Multi-vector representations:} Instead of a single vector per chunk, multiple vectors could be generated to capture different aspects of the text: one for general semantics, another for key legal entities (courts, laws, parties), another for abstract legal concepts, etc. This would allow for more nuanced and multifaceted searches.
    \end{itemize}
    \item Hybrid search techniques and query refinement: The combination of different search paradigms and the intelligent pre-processing of the user's query are key.
    \begin{itemize}

        \item \textit{Optimized weighting in hybrid search:} The integration of semantic (vector, dense) search with traditional keyword-based (sparse, e.g., BM25) search is often superior to either method alone. The relative weighting between the two must be carefully tuned, possibly dynamically according to the query, to balance capturing conceptual meaning with precision in retrieving exact terms, proper names, or citations (Addleshaw Goddard, 2024).
        \item \textit{Query Expansion and intelligent transformation:} Use LLMs (possibly a smaller, faster model dedicated to this task) to pre-process the user's query: expanding it with relevant legal synonyms, related terms, or possible reformulations; identifying the underlying intent; or decomposing complex, multifaceted questions into simpler, more manageable sub-questions that can be addressed by separate retrievals and then synthesized (HyDE - Gao et al. 2022).
        \item \textit{Strict filtering by legal metadata:} Retrieval must go beyond simple textual similarity and incorporate rigorous filtering based on crucial metadata such as applicable jurisdiction, date of decision (to assess its currency and possible overruling), the hierarchical level of the issuing court, and document type. This is essential to ensure the legal relevance of the retrieved results (Magesh et al., 2024).
    \end{itemize}
    \item Iterative, self-correcting, and agent-based retrieval mechanisms: Inspired by how humans conduct legal research, RAG systems can benefit from more dynamic and iterative architectures.
    \begin{itemize}
        \item \textit{Self-Correcting/Corrective RAG (CRAG):} Implement feedback loops where the system evaluates the relevance and quality of an initial set of retrieved documents (possibly using the generative LLM itself or a dedicated evaluation model). If the documents are deemed insufficient or irrelevant, the system can automatically refine the original query, adjust search parameters, or search alternative sources before proceeding to generation (Yan et al., 2024).
        \item \textit{Multi-hop retrieval:} For queries that require synthesizing information from multiple sources or involve sequential reasoning (e.g., tracing the evolution of a doctrine through a chain of precedents), the system can perform multiple "hops" of retrieval. Information extracted from a first set of retrieved documents is used to formulate new queries and retrieve a second set of documents, and so on, until all necessary information has been gathered (Tang and Yang, 2024).
        \item \textit{Agent-based approaches (Agentic RAG):} Develop AI agents that can plan and execute complex retrieval strategies, dynamically deciding which sources to consult, which search terms to use, and how to integrate the obtained information, more closely imitating the research process of a legal expert.
    \end{itemize}
\end{itemize}

Investment in these advanced retrieval strategies is essential, since the quality of the context provided to the generative LLM sets the upper limit for the quality of the final response. Poor or noisy retrieval will inevitably lead to suboptimal or, worse, hallucinated generation, regardless of how sophisticated the generative LLM may be.

\begin{table}[htbp]
\centering
\begin{tabular}{l c c c c c c c c c}
\hline
\multirow{3}{*}{Model} & \multirow{3}{*}{Size} & \multicolumn{7}{c}{Tasks} & \multirow{3}{*}{Average} \\
\cline{3-9}
& & STS & Retrieval & Classif. & Cluster. & Rerank. & PairClass. & Summary & \\
\cline{3-9}
& & 2 & 10 & 8 & 6 & 3 & 3 & 3 & \\
\hline
BOW & - & 0.4917 & 0.2143 & 0.4751 & 0.2612 & 0.7582 & 0.7205 & 0.2635 & 0.4549 \\
\hline
\multicolumn{10}{l}{\textbf{Encoder based Models}} \\
\hline
BERT & 110M & 0.3821 & 0.0231 & 0.5532 & 0.1803 & 0.3968 & 0.7157 & 0.1723 & 0.3462 \\
FinBERT & 110M & 0.4235 & 0.1178 & 0.5961 & 0.2894 & 0.6453 & 0.7021 & 0.2073 & 0.4259 \\
instructor-base & 110M & 0.3791 & 0.5816 & 0.6253 & 0.5362 & 0.9712 & 0.6185 & 0.4372 & 0.5927 \\
bge-large-en-v1.5 & 335M & 0.3435 & 0.6514 & 0.6481 & 0.5768 & 0.9842 & 0.7446 & 0.4911 & 0.6342 \\
AnglE-BERT & 335M & 0.3125 & 0.5784 & 0.6483 & 0.5812 & 0.9673 & 0.6942 & 0.5104 & 0.6132 \\
\hline
\multicolumn{10}{l}{\textbf{LLM-based Models}} \\
\hline
gte-Qwen1.5-7B-instruct & 7B & 0.3792 & 0.6732 & 0.6479 & 0.5887 & 0.9875 & 0.7042 & 0.5408 & 0.6459 \\
Echo & 7B & 0.4408 & 0.6487 & 0.6562 & 0.5823 & 0.9751 & 0.6314 & 0.4781 & 0.6304 \\
bge-en-icl & 7B & 0.3275 & 0.6831 & 0.6604 & 0.5786 & 0.9912 & 0.6782 & 0.5241 & 0.6347 \\
NV-Embed v2 & 7B & 0.3786 & 0.7092 & 0.6432 & 0.6142 & 0.9837 & 0.6098 & 0.5163 & 0.6364 \\
e5-mistral-7b-instruct & 7B & 0.3842 & 0.6783 & 0.6492 & 0.5826 & 0.9863 & 0.7432 & 0.5319 & 0.6508 \\
\hline
\multicolumn{10}{l}{\textbf{Commercial models}} \\
\hline
text-embedding-3-small & - & 0.3298 & 0.6694 & 0.6421 & 0.5847 & 0.9847 & 0.6023 & 0.5138 & 0.6181 \\
\end{tabular}
\caption{Performance comparison between different embedding models on the FinMTEB benchmark. The evaluation metrics include semantic textual similarity (STS), retrieval, classification (Class.), clustering (Cluster.), reranking (Rerank.), pair classification (PairClass.), and summarization (Summ.). \textbf{The best results are in bold}. The \underline{underline} represents the \underline{second best} performance.}
\label{tab:finmteb}
\end{table}

\subsubsection{Refinement of the Generation and Reasoning Phase: Towards More Reliable and Transparent Legal AI}

The way in which the generator LLM is instructed is a critical component of the RAG architecture, not a mere implementation detail. The following guidelines should not be understood as simple 'tips', but rather as prompt engineering principles designed to restrict the model's possible response space and align its behavior with the demanding requirements of fidelity and traceability in the legal domain. The aim is to explicitly encode in the instructions the operational constraints that ensure more reliable generation.

Once a set of contextually relevant text fragments has been retrieved (ideally optimized through the techniques from the previous section), the challenge shifts to guiding the generator LLM to use this information faithfully, accurately, logically coherently, and transparently. Strategies to refine this phase are crucial to minimize the risk that the LLM ignores the context, misinterprets it, or generates statements that go beyond what is supported by the sources.

\begin{itemize}
    \item Advanced and domain-specific Prompt Engineering for legal RAG: The way in which the generative LLM is instructed on how to interact with the retrieved context is of vital importance. Prompts must be meticulously designed to:
    \begin{itemize}
        \item \textit{Emphasize fidelity to context (Grounding Instructions):} Include explicit and prominent instructions directing the LLM to base its response \textit{exclusively} on the information contained in the provided documents and to actively avoid using its internal parametric knowledge or making unfounded assumptions. This is achieved through unequivocal directives such as: 'Respond solely based on the following legal excerpts. Do not add information that is not present in the provided texts.'
        \item \textit{Guidance for reasoning (Chain-of-Thought, Step-by-Step):} Instruct the model to externalize its reasoning process, showing the logical steps it follows to reach a conclusion from the retrieved context. For example, "First, identify the relevant rules in the context. Second, apply these rules to the facts of the query. Third, explain your conclusion" (Wei et al. 2023). This can not only improve the accuracy of reasoning, but also makes the process more interpretable and verifiable by a human (Schwarcz et al., 2024).
    \item \textit{Structured handling of uncertainty and conflicts:} Provide the LLM with clear protocols on how to act when the retrieved information is incomplete, ambiguous, or contains contradictions. This includes the explicit instruction to refrain from generating a response when it cannot be formulated with a high degree of confidence based on the sources, instead of resorting to speculation. For example, "If the information provided is not sufficient to answer completely, state this explicitly and explain the nature of the missing information", "If you find conflicting information in the excerpts, present both perspectives and point out the discrepancy".
        \item \textit{Precise citation instructions:} Require the LLM to cite specifically (ideally at the fragment or retrieved document level) the exact sources from which each factual or legal statement is drawn. This is essential for verifiability.
        \item \textit{Detailed persona and output format:} Precisely define the role the LLM should adopt (e.g., "Act as an objective and neutral legal research assistant") and the exact format of the expected response (e.g., summary structure, citation style) to ensure consistency and professional usefulness.
        \item \textit{"Accusatory" or refinement prompting (Follow-up Prompts):} As observed in Addleshaw Goddard (2024), the use of a second prompt that questions the completeness or accuracy of the LLM's initial response, subtly accusing it of having omitted information or asking it to "carefully review the context again in case something was overlooked," can stimulate the model to perform a more thorough second pass over the context and significantly improve the quality of the response.
    \end{itemize}
    \item Fine-tuning of the generative LLM focused on legal fidelity: Although fine-tuning massive LLMs is a resource-intensive process, it can offer significant benefits if carried out carefully.
    \begin{itemize}

        \item \textit{Fine-tuning on Legal Grounding Tasks:} Adapting a pre-trained LLM using a high-quality dataset composed of pairs of (retrieved legal context, ideally grounded and properly cited answer). This can train the model to adhere more strictly to the provided context and to generate responses in the style and format desired by legal practice (Tian, Mitchell, Yao, et al. 2023).
        \item \textit{Fine-tuning for Legal Reasoning over Context:} Developing datasets that teach the LLM to perform specific types of legal reasoning (e.g., rule application, identification of \textit{holdings}, case comparison) \textit{explicitly based} on the retrieved context, rather than relying on abstract patterns.
    \end{itemize}
    \item Integration with Specialized Reasoning Models: The emergence of LLMs with architectures explicitly designed for multi-step reasoning, planning, and problem decomposition (such as OpenAI's "o" model family - OpenAI 2024) is particularly relevant for RAG.
    \begin{itemize}
        \item \textit{Response Planning:} These models could, in theory, plan how to use the retrieved information more strategically, identifying which fragments are most relevant to which parts of the query and how to synthesize them in a logically coherent manner.
        \item \textit{Internal Verification of Reasoning Steps:} Their ability to "reflect" on their own intermediate reasoning steps could allow them to detect and correct errors or inconsistencies before generating the final answer (Schwarcz et al., 2024). Integrating these reasoning models as the generative component in a RAG system is a promising area for future improvements.
    \end{itemize}
    \item Hybrid Architectures (Symbolic-Neural): Although still in early stages for complex legal applications, combining neural LLMs with rule-based symbolic reasoning systems (e.g., formal logics, legal ontologies) could offer a way to improve the logical consistency and verifiability of responses generated from retrieved context.
    \item Adaptation of Language Complexity (The "Jargon" Function): A truly advanced legal AI must not only be accurate, but also contextually aware of the final recipient of the information. Prompt optimization should include instructions to modulate the language complexity of the response. For example, a system could receive the directive: "Generate a technical answer for a lawyer and, additionally, a simplified explanation for a citizen without legal knowledge." This capability, which we might call the "jargon function," is a pillar of the humanization of legal technology, recognizing that the usefulness of a response lies not only in its correctness, but in its comprehensibility. This transforms AI from a simple search engine into a true bridge of communication between the complex legal world and society.
    
\end{itemize}
The ultimate goal of these generation optimization strategies is not only to produce answers that \textit{appear} correct, but answers that are \textit{demonstrably} correct, faithful to the provided sources, and useful for the legal professional. The LLM's ability to explain \textit{how} it arrived at a conclusion from the retrieved context is as important as the conclusion itself.

The successful implementation of RAG in the legal domain, therefore, is not simply a matter of connecting an LLM to a database. It requires careful design and continuous optimization of each stage of the process, from data curation and chunking, through the sophistication of retrieval algorithms and prompt engineering, to the refinement of the LLM's reasoning ability and faithful generation. Only through this holistic and rigorous approach can the true potential of RAG begin to be realized to mitigate hallucinations and provide genuinely reliable and valuable legal AI.

\section{Advancing Towards Reliability: Holistic Strategies for Optimization and Mitigation of Hallucinations in Legal Artificial Intelligence}
\label{sec:estrategias_avanzadas}

The realization that neither foundational Large Language Models (LLMs) nor canonical implementations of Retrieval-Augmented Generation (RAG) manage to completely eradicate the specter of hallucinations in the legal domain necessitates a paradigm shift in how we approach the development and integration of these technologies. It is no longer sufficient to aspire to a single solution or a "magic switch" that eliminates errors; instead, a \textbf{holistic, multifaceted, and adaptable approach} is required—one that acknowledges the inherent complexity of the problem and implements a synergy of optimization and mitigation strategies throughout the entire information lifecycle, from data curation to the final verification of the generated output. This approach does not seek absolute perfection—a theoretically unattainable goal in complex technological systems—but rather adopts a principle of robust engineering: \textbf{maximizing reliability and minimizing risk within the bounds of what is feasible}. It is recognized that 100\% infallibility is not a limitation of \textit{current} technology, but an inherent feature of complexity itself. Therefore, the objective is to build a system whose reliability is so high, and whose failure modes are so predictable, that expert human supervision becomes an efficient validation layer rather than a burdensome search for hidden errors, always under the indispensable guidance of professional judgment and responsibility. This section is dedicated to an in-depth exploration of this arsenal of advanced strategies, which go beyond superficial adjustments to delve into rigorous data optimization, the refinement of algorithmic reasoning processes—including the consideration of hierarchical systems and AI agents—the implementation of increasingly sophisticated verification mechanisms, and, crucially, the strengthening of the legal professional's role as a critical and informed supervisor. The conjunction of these techniques aims not only to reduce the frequency of hallucinations but also to transform their nature, making residual errors more detectable and less harmful.

The effectiveness of a holistic approach, combining strategic data curation with model specialization, has been empirically demonstrated. A clear example is the "Legal Assist AI" project, which addressed the problem of hallucinations in the Indian legal context. Instead of using a general-purpose model, the researchers created a curated data corpus from Indian legal sources (Constitution, statutes, case law) and used it to fine-tune a base model with 8 billion parameters (Llama 3.1 8B). The result was a specialized model that not only outperformed much larger models such as GPT-3.5 Turbo (175 billion parameters) on Indian legal tasks, but crucially, drastically reduced the generation of hallucinations, providing reliable answers where other models fabricated information (Gupta et al., 2025). This success story serves as powerful evidence that hallucination mitigation does not reside in model scale, but in data quality and training specificity.

The most advanced mitigation strategies converge on a principle of knowledge integration, where LLMs do not operate in isolation, but as part of hybrid architectures. This includes integration with structured legal knowledge graphs and the use of Mixture-of-Experts (MoE) architectures, as implemented in cutting-edge models such as ChatLaw (Shao et al., 2025). In these systems, specialized expert modules within the LLM are dynamically activated to handle different types of legal tasks, reducing hallucinations by 38\% by ensuring that the query is managed by the component with the most relevant knowledge.

\begin{figure}[htbp]
  \centering
  \begin{tikzpicture}
    \begin{axis}[
        ybar,
        enlargelimits=0.05,
        legend style={at={(0.5,1.15)}, anchor=south, legend columns=-1}, 
        ylabel={Adversarial Attack Success Rate (\%)},
        xlabel={Level of Attack/Mitigation Strategy Sophistication},
        symbolic x coords={Zero Shot, Few Shot, SL, RL1, RL2, RL3, RL4},
        xticklabels={Zero Shot, Few Shot, SL, RL (a=0.002), RL (a=0.001), RL (a=0.0005), RL (a=0.0001)},
        xtick=data,
        nodes near coords,
        nodes near coords style={font=\tiny, rotate=90, anchor=west}, 
        ymin=0, ymax=60, 
        bar width=12pt, 
        title style={font=\footnotesize, text width=0.95\textwidth, align=center}, 
        title={Impact of Strategic Sophistication on LLM Resilience \\ 
               \footnotesize (Adapted from General Analysis, 2025 - Red Teaming Results)}, 
        width=0.95\textwidth,
        height=0.6\textwidth, 
        xticklabel style={rotate=30, anchor=east, font=\footnotesize}
      ]
      
      \addplot+[fill=blue!70] coordinates {
        (Zero Shot,12.9) 
        (Few Shot,13.9) 
        (SL,23.4) 
        (RL1,26.6) 
        (RL2,32.7) 
        (RL3,46.1) 
        (RL4,54.5)
      };
      \legend{Attack success rate (higher is worse for the Defender LLM)}
    \end{axis}
  \end{tikzpicture}
  \caption{This graph illustrates how more sophisticated attack strategies (analogous to the lack of robust mitigation strategies or exploited vulnerabilities) achieve a higher success rate in inducing failures in a target LLM (GPT-4o). It demonstrates the need for equally sophisticated defense (mitigation) strategies. The X-axis represents different attack algorithms from the General Analysis study, interpreted here as levels of challenge or sophistication to which a legal RAG system must be resilient.}
  \label{fig:attack_success_strategies}
\end{figure}
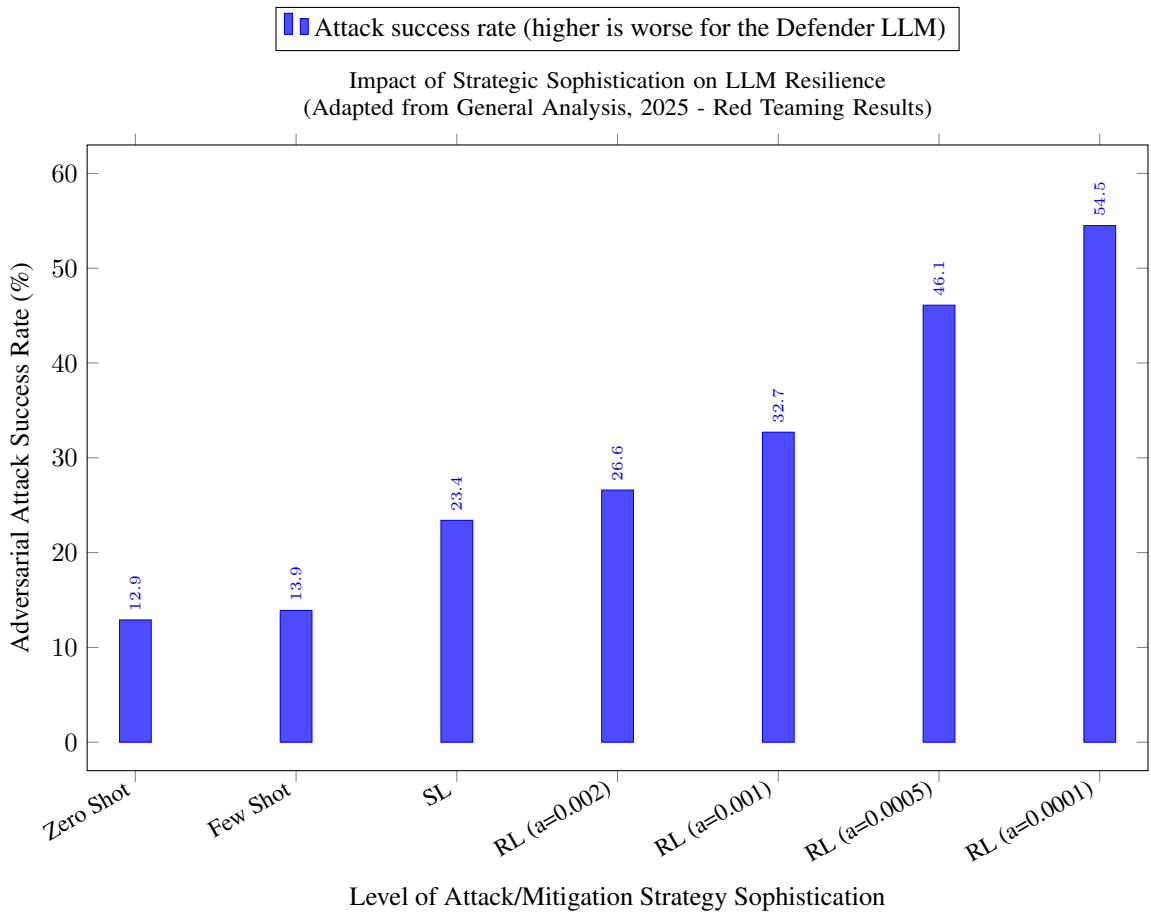

\begin{figure}[htbp]
\centering
\begin{tikzpicture}[
    node distance=2.5cm and 3cm, 
    fase/.style={
        rounded rectangle, 
        draw=blue!70, 
        fill=blue!15, 
        very thick, 
        minimum height=2cm, 
        minimum width=4.5cm, 
        text centered, 
        text width=4cm,
        font=\bfseries 
    },
    estrategia_cat/.style={ 
        rectangle, 
        draw=green!70, 
        fill=green!10, 
        thick, 
        minimum height=1.2cm, 
        text centered, 
        text width=4cm, 
        font=\itshape\footnotesize 
    },
    conector/.style={-Latex, thick, blue!80, line width=1pt},
    conector_estrategia/.style={Stealth-Stealth, thick, green!80, dashed} 
]

\node[fase] (curacion) {Phase 1: Data and Knowledge Curation};
\node[fase, right=of curacion] (recuperacion) {Phase 2: Retrieval};
\node[fase, below=of recuperacion] (generacion) {Phase 3: Generation and Reasoning};
\node[fase, left=of generacion] (verificacion) {Phase 4: Post-Hoc Verification and Confidence};

\node[estrategia_cat, above=0.7cm of curacion] (strat_curacion) {Data Curation and Governance Strategies (5.1)};
\node[estrategia_cat, above=0.7cm of recuperacion] (strat_recuperacion) {Sophisticated Retrieval Strategies (5.2)};
\node[estrategia_cat, below=0.7cm of generacion] (strat_generacion) {Faithful Generation and Reasoning Strategies (5.4)}; 
\node[estrategia_cat, below=0.7cm of verificacion] (strat_verificacion) {Verification and Calibration Strategies (5.5)};

\draw[conector] (curacion.east) -- (recuperacion.west) node[midway, above, font=\small\color{blue!90}] {Data Input};
\draw[conector] (recuperacion.south) -- (generacion.north) node[midway, right, font=\small\color{blue!90}] {Relevant Context};
\draw[conector] (generacion.west) -- (verificacion.east) node[midway, below, font=\small\color{blue!90}] {Candidate Response};
\draw[conector] (verificacion.north) -- (curacion.south) node[midway, left, font=\small\color{blue!90}] {Feedback and Improvement};

\draw[conector_estrategia] (strat_curacion.south) -- (curacion.north);
\draw[conector_estrategia] (strat_recuperacion.south) -- (recuperacion.north);
\draw[conector_estrategia] (strat_generacion.north) -- (generacion.south); 
\draw[conector_estrategia] (strat_verificacion.north) -- (verificacion.south);

\end{tikzpicture}
\caption{Cyclical model of a Legal RAG system and strategic intervention points for optimization and mitigation of hallucinations. The specific strategies (referenced by subsection of the essay) are applied at each phase to improve the overall reliability of the system.}
\label{fig:ciclo_rag_estrategias}
\end{figure}
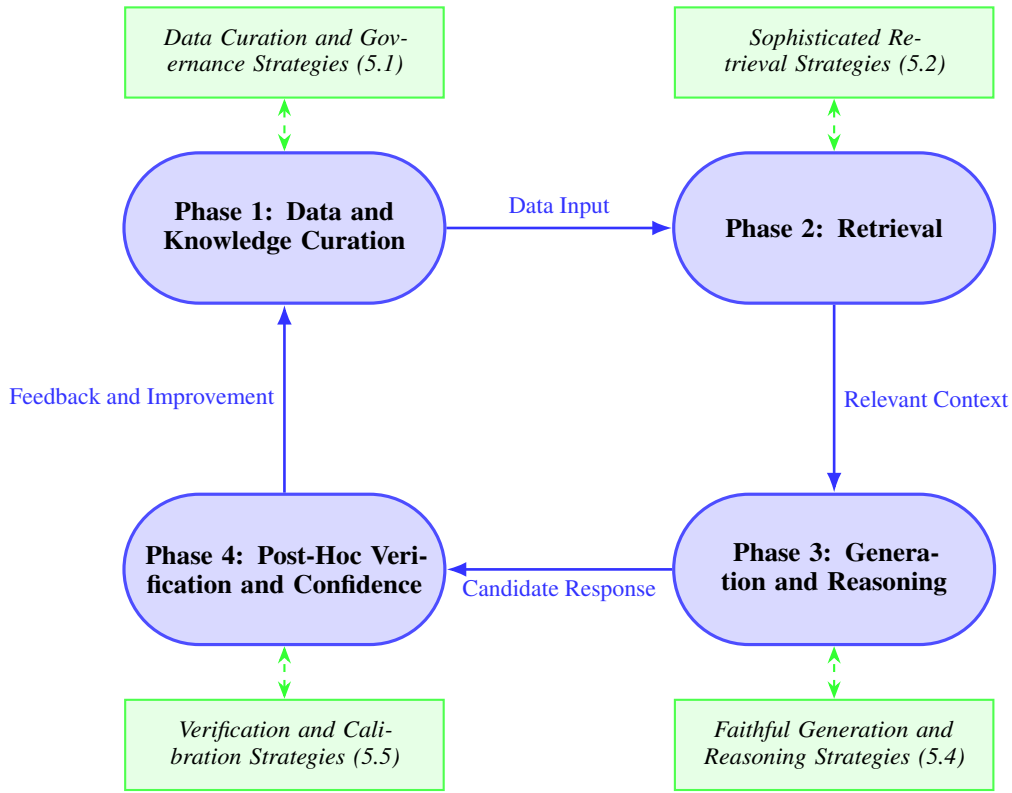

\subsection{The Quality of the Foundation: Strategic Data Curation and External Knowledge Bases}

The adage "garbage in, garbage out" resonates with particular force in the context of LLMs and RAG systems. The quality, timeliness, relevance, and representativeness of the external knowledge base upon which these systems are grounded is not a mere technical detail, but rather the foundation upon which all their reliability is built. A deficient, outdated, or biased document corpus will inevitably limit the RAG system's ability to provide accurate and reliable answers, regardless of how sophisticated its retrieval or generation algorithms may be. Therefore, a primary and proactive strategy for mitigating hallucinations begins long before user interaction: in the meticulous curation and strategic management of these knowledge bases.

\subsubsection{Rigorous selection and prioritization of legal sources}

The vast universe of legal information demands a judicious selection. Not all sources are equal in authority or relevance. It is imperative to take into consideration:

\begin{itemize}
        \item \textit{Hierarchy of authority:} Design mechanisms, both in indexing and retrieval, that explicitly prioritize binding primary sources (Constitution, current statutes, case law from higher courts of the relevant jurisdiction) over secondary sources, persuasive literature, or case law from other jurisdictions or lower courts. This implies the incorporation of rich metadata that encodes this hierarchy and allows the RAG system to weight the information accordingly.
        \item \textit{Continuous verification of currency and validity:} Implement dynamic and automated processes (as much as possible, complemented by expert review) to keep the knowledge base up to date with legislative amendments, new judicial decisions, and, critically, the repeal status of precedents. Integration with commercial citation verification services or updated legislative databases is essential. Tools such as Shepard’s from LexisNexis or KeyCite from Westlaw (standards in the U.S. common law system) are crucial for tracing the history of a case and verifying its validity. In Spain, although there is no direct equivalent with such a consolidated brand, leading legal platforms like vLex or La Ley Digital offer analogous functionalities that allow users to check whether a ruling has been appealed, annulled, or qualified, as well as to verify the validity of a regulation. Integration with, or at least systematic consultation of, these tools is an unavoidable step to avoid basing responses on obsolete law, a common and potentially serious error.
        \item \textit{Proactive filtering of low-quality or problematic sources:} Identify and exclude or explicitly flag sources known for their low quality, manifest biases (if the goal is neutral analysis), or irrelevance to the most common legal tasks. This may require both the judgment of legal experts and the use of AI techniques for the automatic assessment of the quality and reliability of documents (Nguyen and Satoh, 2024).
        These techniques go beyond simple keyword detection. They include, for example, the use of smaller and more specialized language models that act as 'judges' or evaluators (an approach known as \textit{LLM-as-a-judge}), capable of verifying the logical consistency of a document, detecting internal contradictions, or cross-checking statements against a curated knowledge base. Other techniques involve analyzing the model's confidence during generation or detecting stylistic anomalies that often accompany hallucinations. The implementation of these algorithmic 'gatekeepers' can serve as a first automated filter before human review.
\end{itemize}

The importance of these practices of curation, prioritization, and governance of the data that feed legal AI systems is magnified by emerging regulatory frameworks. Regulation (EU) 2024/1689 of the European Parliament and of the Council, of 13 June 2024, establishing harmonized rules on artificial intelligence (hereinafter, the EU AI Act or the Regulation), in its Article 10, imposes explicit obligations on developers of high-risk AI systems regarding training, validation, and testing datasets. These must be 'relevant, representative, free of errors, and complete', and appropriate data governance practices must be implemented, including an examination of potential biases. Compliance with these requirements is not only a matter of good technical practice to improve model reliability and reduce the risk of hallucinations arising from faulty data, but it is also shaping up to be an unavoidable legal requirement for operating in the European market, incentivizing greater diligence in the management of the knowledge underpinning legal AI.

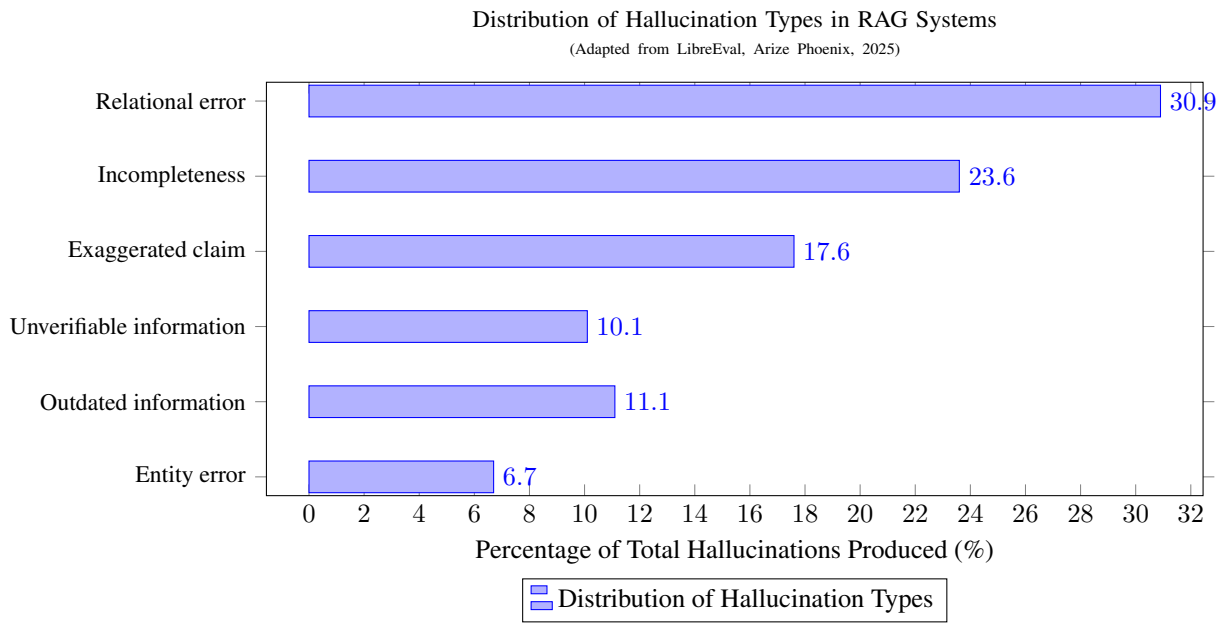
\begin{figure}[htbp]
  \centering
  \begin{tikzpicture}
    \begin{axis}[
        xbar, 
        enlargelimits=0.05,
        legend style={at={(0.5,-0.2)}, anchor=north, legend columns=-1}, 
        xlabel={Percentage of Total Hallucinations Produced (\%)},
        symbolic y coords={
            Entity error, 
            Outdated information, 
            Unverifiable information, 
            Exaggerated claim, 
            Incompleteness, 
            Relational error
        },
        ytick=data,
        y tick label style={font=\footnotesize}, 
        bar width=12pt, 
        nodes near coords,
        nodes near coords align={horizontal},
        xmin=0,
        title style={font=\footnotesize, text width=0.85\textwidth, align=center}, 
        title={Distribution of Hallucination Types in RAG Systems \\ 
               \tiny (Adapted from LibreEval, Arize Phoenix, 2025)}, 
        width=0.85\textwidth, 
        height=0.6\textwidth, 
        y=1cm 
      ]
      
      \addplot coordinates {
        (6.7,Entity error)
        (11.1,Outdated information)
        (10.1,Unverifiable information)
        (17.6,Exaggerated claim)
        (23.6,Incompleteness)
        (30.9,Relational error)
      };
      \legend{Distribution of Hallucination Types}
    \end{axis}
  \end{tikzpicture}
  \caption{Percentage distribution of the different types of hallucinations actually produced in the responses of language models with RAG, according to the LibreEval1.0 dataset. This distribution highlights the most common challenges that mitigation strategies must address.}
  \label{fig:distribucion_tipos_alucinacion_libreeval}
\end{figure}

\subsubsection{Ensuring Diversity and Representativeness of the Corpus}

To avoid the creation of algorithmic "legal monocultures" (Dahl et al., 2024) that ignore the diversity of legal thought or the particularities of less prominent jurisdictions, it is crucial to:

\begin{itemize}
    \item \textit{Broad jurisdictional and thematic coverage:} Strive to include a balanced representation of all relevant jurisdictions (state, federal, specialized) and a wide spectrum of legal areas, not just those with greater availability of digitized data. In the Spanish context, for example, data from the General Council of the Judiciary (CGPJ) consistently show a high concentration of litigation in the judicial bodies of major capitals such as Madrid or Barcelona. A training corpus that does not adequately account for this reality could develop a centralist bias, ignoring the doctrinal or jurisprudential particularities of other High Courts of Justice, which would limit the tool's usefulness at the national level.
\end{itemize}
\begin{itemize}
    \item \textit{Conscious inclusion of plural perspectives:} For tasks that require analysis beyond pure doctrine (e.g., policy impact assessment, principle-based argumentation), consider the curated inclusion of critical academic sources, reports from civil society organizations, or even transcripts of legislative debates that offer diverse and nuanced perspectives on the law and its application.
\end{itemize}

\subsubsection{Advanced Structuring of Legal Knowledge (Beyond Plain Text)}

While many RAG systems operate primarily on unstructured text, the representation of legal knowledge can be significantly enriched by:

\begin{itemize}
    \item \textit{Development of legal ontologies and knowledge graphs:} Build formal models that represent key legal entities (courts, judges, laws, legal concepts), their attributes, and, fundamentally, the complex relationships between them (e.g., a law \textit{amends} another, a case \textit{interprets} a statute, a judge \textit{dissents} from an opinion). A RAG system that can consult and reason over these knowledge graphs could perform deeper and more accurate inferences than one based solely on textual similarity (Martin, 2024; Magora, 2024).
    \item \textit{Extraction and linking of rich metadata:} Enrich each document in the knowledge base with detailed and structured metadata (jurisdiction, date, court, judges, parties, legal topics, citations, procedural history, validity status) that can be used by the retrieval module for much more precise filtering and ranking.
\end{itemize}

Ongoing investment in the quality, structure, and management of knowledge bases is not an ancillary cost, but a fundamental strategic investment in the long-term reliability of any RAG-based legal AI system. Without a solid foundation, even the most advanced algorithms will be building on quicksand.

\subsection{Sophisticated Optimization of the Retrieval Phase: Finding the Legal Needle in the Digital Haystack}

The effectiveness of a RAG system critically depends on the ability of its retrieval module to identify and extract, from a potentially massive corpus, the text fragments (chunks) that are \textit{exactly and contextually relevant} to the user's query. Simple semantic similarity search, while a starting point, often proves insufficient for the complexity and nuances of legal language and reasoning. Advanced optimization of this phase is therefore an area of intense research and development, focused on endowing the system with a discernment capability closer to that of a human legal researcher.

\subsubsection{\textbf{Legal Embedding Models and Multi-Vector Strategies} }
 
The quality of the vector representation (embedding) that captures the meaning of the chunks and the query is the cornerstone of semantic search.

    \begin{itemize}
        \item \textit{Domain-specific legal embeddings:} Prioritizing the use of embedding models that have been pre-trained or fine-tuned specifically on large corpora of legal texts (such as LegalBERT or models developed from corpora like The Pile of Law - Henderson et al. 2022) is preferable to general-purpose embeddings, as they can better capture the semantic nuances and specific terminology of law.
        \item \textit{Embedding augmentation techniques:} Exploring methods that enrich textual embeddings with additional information, such as structural metadata or knowledge graph information, to create richer and more contextually informed representations.
    \end{itemize}

\subsubsection{\textbf{Hybrid Search Strategies, Multi-Stage Retrieval, and Query Refinement:}}

To overcome the limitations of a single search modality, more complex approaches are being adopted:
    \begin{itemize}
    \item\textit{Dynamic Hybrid Search Optimization:} Combining semantic search (vector-based, dense) with traditional keyword-based (lexical, e.g., BM25) search is fundamental. However, the relative weighting between these two modalities should not be static. Ideally, the system should be able to dynamically adjust this weighting based on the characteristics of the user's query (e.g., giving more weight to keywords if the query contains very specific terms, proper names, or exact quotations) (Addleshaw Goddard, 2024).
    \item\textit{Pre-Processing and Intelligent Query Transformation:} Use an LLM (possibly a smaller and more efficient model dedicated to this task) to analyze and refine the user's query \textit{before} the retrieval phase. This may include:
    
    \begin{itemize}
        \item \textit{Query Expansion:} Add relevant legal synonyms, conceptually related terms, or possible reformulations to broaden the coverage of the search.
        \item \textit{Query Decomposition:} Break down complex or multifaceted questions into simpler, atomic sub-questions, each of which can be the subject of a separate retrieval. The results of these sub-retrievals can then be combined to answer the original query.
        \item \textit{Generation of Hypothetical Queries (Hypothetical Document Embeddings - HyDE):} Instruct an LLM to generate an "ideal" document that would perfectly answer the user's query, and then use the embedding of this hypothetical document for semantic search. This often leads to more relevant retrieval than using the embedding of the original query directly (Gao et al., 2022).
    \end{itemize}
    
    \item\textit{Multi-Stage Retrieval (Multi-Hop Retrieval):} For queries that require sequential reasoning or the synthesis of information across a chain of documents (e.g., tracing the evolution of a legal doctrine through multiple precedents), the system can implement an iterative retrieval process. The information extracted from a first set of retrieved documents is used to formulate new queries or refine existing ones, allowing the system to "navigate" through the document corpus in a more intelligent and targeted way (Tang and Yang, 2024).
    \item\textit{Sophisticated Re-ranking (Re-ranking):} Once an initial set of candidate chunks has been retrieved (possibly a large set), use more powerful and computationally intensive re-ranking models (such as cross-encoders) to more accurately assess the relevance of each chunk in relation to the full query. These models can consider the interaction between the query and the chunk more deeply than the embedding models used in the initial retrieval, improving the final ranking of results presented to the generative LLM.
    \end{itemize}

\subsubsection{\textbf{Incorporation of Feedback and Continuous Learning} }
The most advanced RAG systems should incorporate mechanisms to learn from user interactions and from explicit or implicit feedback.

\begin{itemize}
    \item\textit{User Feedback:} Allow users to rate the relevance of retrieved documents or generated responses, and use this feedback to fine-tune embedding models or ranking algorithms.
    \item\textit{Dynamic Adaptation:} Adjust retrieval parameters (e.g., similarity thresholds, number of chunks to retrieve) based on historical performance or the characteristics of the current query.
\end{itemize}

\subsubsection{\textbf{Conclusion} }

Investment in these advanced retrieval strategies is fundamental, as the quality of the context provided to the generative LLM is the ceiling for the quality of the final response. Poor or noisy retrieval will inevitably lead to suboptimal or, worse, hallucinated generation, regardless of how sophisticated the generative LLM may be.

\subsection{AI Agents in Complex Legal Systems and Kelsen's Normative Hierarchy}

The optimization of RAG systems for legal tasks, as discussed, involves an increasingly sophisticated interaction between the LLM, the knowledge bases, and the reasoning process. As this sophistication increases and LLMs evolve toward more complex and multi-step reasoning capabilities, we begin to glimpse the potential of more autonomous and proactive AI architectures, often referred to as 'AI agents.' In this context, a legal AI agent is distinguished from a simple LLM (which merely responds to a prompt) by its ability to perform sequences of actions, interact autonomously with multiple tools or sources of information, make intermediate decisions, and plan strategies to achieve a predefined complex legal objective, such as preparing a case or conducting a thorough due diligence. However, the implementation of such agents in the intricate and normatively structured legal domain must inevitably consider the fundamental architecture of the legal system itself. In civil law jurisdictions and many constitutional systems, this architecture is classically conceptualized through the notion of Hans Kelsen's Normative Pyramid.

Kelsen's pyramid posits that the legal norms of a system are organized hierarchically, where the validity of each norm derives from a higher norm, culminating in a hypothetical "Basic Norm" (Grundnorm) that underpins the validity of the entire system (generally the Constitution in modern systems). This hierarchical structure implies that lower-ranking norms (e.g., an administrative regulation) must conform to higher-ranking norms (e.g., a law, the Constitution). The correct application of the law, therefore, is not just a matter of finding \textit{a} relevant norm, but of finding the \textit{correct} norm within this hierarchy and resolving possible conflicts between norms of different ranks (principle of normative hierarchy), or of the same rank but later in time (principle of temporality), or more specific (principle of specialty).

For a legal AI agent operating with the goal of providing reliable and legally valid answers, this hierarchical structure presents both a challenge and an opportunity:

\begin{enumerate}
    \item \textbf{Challenge for retrieval and reasoning:}

\begin{itemize}
    \item \textit{Identification of the controlling authority:} An AI agent must be able not only to retrieve multiple potentially relevant norms or precedents, but also to discern which of them takes precedence or is the controlling authority in a given case. For example, if a law appears to contradict a constitutional provision, the Constitution prevails. If a regulation contradicts a law, the law prevails. This hierarchical inference is essential.
    \item \textit{Resolution of antinomies:} The agent must be able to identify and, to the extent possible, propose solutions to normative conflicts (antinomies) using accepted resolution criteria (hierarchy, temporality, specialty). This requires a level of meta-legal reasoning.
    \item \textit{Understanding the dynamics of validity:} The validity of a norm can change (e.g., a law may be declared unconstitutional, a precedent may be overturned). The agent must access and process information about the current validity status of the retrieved norms.
\end{itemize}

Let us consider, for example, a query about the legality of a certain municipal commercial practice. A simple AI agent, without hierarchical awareness, might retrieve and base its affirmative answer on a municipal ordinance that explicitly permits such a practice. However, a 'Kelsenian' agent, in its upward verification process, would identify a subsequent and higher-ranking regional or state law that prohibits or severely restricts such activity, or even a Constitutional Court ruling that has declared a similar norm unconstitutional. This agent would correctly conclude that the municipal ordinance, although textually applicable, is invalid or inapplicable due to the conflict with a hierarchically superior norm, thus avoiding a validity hallucination that the simple agent would have committed. The ability to perform this type of hierarchical validation is, therefore, crucial for reliability.

    \item \textbf{Opportunity for hierarchically-aware RAG systems:} The Kelsen pyramid can, in fact, inspire more sophisticated and reliable RAG architectures:

\begin{itemize}
    \item \textit{Hierarchically structured knowledge bases:} Knowledge bases could be explicitly organized to reflect the normative hierarchy. Documents could be tagged with their hierarchical rank, using a metadata scheme that mirrors the structure of the legal system. This includes universal categories such as 'Constitutional Norm', 'Primary Legislation' (laws), 'Secondary Legislation' (regulations), and 'Binding Jurisprudence'. In the specific context of Spanish law, this would translate into labels such as (Constitution, organic law, ordinary law, regulation, Constitutional Court jurisprudence, etc.).
    \item \textit{Hierarchy-aware retrieval algorithms:} The retrieval module could be instructed to prioritize the search and retrieval of higher-ranking norms when relevant, or to specifically seek out norms that interpret or apply an identified superior norm.
    \item \textit{Reasoning modules for hierarchical coherence:} A generative LLM, or a specialized reasoning component, could be trained to verify the coherence of a proposed solution with higher-ranking norms. If an interpretation of a contract appears to violate a mandatory law, the agent could flag this inconsistency.
    \item \textit{Planning agents navigating the pyramid:} A more advanced AI agent could plan its research and reasoning process starting from the top of the pyramid (Constitution, relevant international treaties) and descending through the applicable laws and jurisprudence, ensuring that each step is consistent with the superior level.
\end{itemize}
\end{enumerate}

\textbf{Implications for reliability and hallucinations:}

Ignoring the normative hierarchy can lead to a specific and serious type of legal hallucination: the \textbf{hallucination of invalidity or inapplicability due to hierarchical conflict}. An LLM could, for example, base a response on a regulation that, although textually relevant, is invalid because it contradicts a superior law, or ground an argument in case law from a lower court that has been overturned or qualified by a higher court. These are not mere factual inaccuracies, but fundamental errors in the application of the law.

In contrast, an AI agent that is explicitly modeled to understand and operate within the Kelsenian pyramid could:

\begin{itemize}
    \item \textbf{Reduce relevance hallucinations:} By prioritizing sources of higher authority, it is less likely to rely on legally subordinate or irrelevant information.
    \item \textbf{Enhance the soundness of reasoning:} By verifying consistency with higher norms, its conclusions would be more robust and less susceptible to being invalidated.
    \item \textbf{Increase transparency and explainability:} By being able to trace the derivation of a conclusion through the normative hierarchy, the agent could offer more convincing and verifiable explanations of its reasoning.
\end{itemize}

Ultimately, a 'Kelsenian' AI agent—that is, one that not only accesses sources but also understands and respects the normative hierarchy and the principles of validity of the legal system—would be intrinsically less prone to certain critical types of hallucinations. By prioritizing the Constitution over statutes, statutes over regulations, and by verifying the validity and applicability of each norm within its hierarchical context, the risk of generating advice based on invalidated subordinate norms, repealed case law, or interpretations that contradict fundamental principles would be drastically reduced. This structural awareness does not eliminate all risks of hallucination—especially those arising from the inherent ambiguity of language or the limitations of the knowledge corpus itself—but it does provide a robust framework for a more coherent, predictable, and ultimately more reliable legal AI. It is crucial to understand that the value of this framework lies not only in improving the final answer, but in the very externalization of the reasoning process. By making explicit the normative hierarchy that applies, the AI moves from offering an opaque conclusion to presenting a verifiable argument. For legal professionals, this transparent reasoning is, in many cases, more valuable than the answer itself, as it allows them to audit, validate, and ultimately appropriate the reasoning to build their own legal strategy. It transforms AI from a 'black box' into a 'toolbox' for reasoning.

In the Spanish context and in many European civil law systems, where the codification and formal hierarchy of legal sources are particularly pronounced, incorporating a Kelsenian awareness into legal AI agents is not merely an academic refinement, but a necessary condition for their reliability and practical usefulness. An agent that does not "understand" the pyramidal structure of the legal system will be intrinsically prone to generating responses that, although plausibly worded, are legally unsustainable or outright incorrect. The future development of reliable legal AI will inevitably require equipping these systems with a deeper understanding of the fundamental architecture of law itself.

\begin{table}[htbp]
  \centering
  \caption{Estimated Quantifiable Impact of Optimization Strategies in Legal RAG Systems}
  \label{tab:impacto_optimizacion_rag}
  \sisetup{table-format=2.1}
  \begin{tabular}{
    >{\raggedright\arraybackslash}p{0.25\textwidth} 
    >{\raggedright\arraybackslash}p{0.25\textwidth} 
    >{\raggedright\arraybackslash}p{0.12\textwidth}
    >{\raggedright\arraybackslash}p{0.12\textwidth}
    >{\raggedright\arraybackslash}p{0.12\textwidth}
  }
    \toprule
    \textbf{Applied Optimization} & \textbf{Key Metric} & {\textbf{Initial}} & {\textbf{Optimized}} & {\textbf{Relative}} \\
    \textbf{Strategy} & & {\textbf{Value}} & {\textbf{Value}} & {\textbf{Improvement (\%)}} \\
    \midrule
    Intensive Retrieval Optimization (Chunking, Embeddings, Query Expansion) \newline \textit{(Inspired by Addleshaw Goddard, 2024)} & F1-Score (Clause Extraction) & 74 & 95 & 28.4 \\
    \addlinespace
    Fine-Tuning of the Generative LLM for Context Fidelity & \textit{Misgrounding} rate & 20 & 5 & 75.0 \\ 
    \addlinespace
    Implementation of Chain-of-Thought (CoT) in Generation Prompts & Logical Coherence (Human Score 1-5) & 3.2 & 4.5 & 40.6 \\ 
    \addlinespace
    Automated Post-Hoc Verification with Secondary Model & Undetected Hallucination Rate & 15 & 3 & 80.0 \\ 
    \addlinespace
    Implementation of Hierarchy-Aware Agent (Kelsenian) & Error Rate Due to Normative Conflict & 15 & 2 & 86.7 \\ 
    \bottomrule
    \multicolumn{5}{p{0.95\textwidth}}{\footnotesize{\textit{Note:} The values for "Retrieval Optimization" are inspired by the F1 results reported by Addleshaw Goddard (2024). Other values are hypothetical and presented for illustrative purposes to demonstrate the potential impact of various optimization strategies discussed in Section 5. "Relative Improvement" is calculated as ((Optimized Value - Initial Value) / Initial Value) * 100 for metrics where higher is better, or ((Initial Value - Optimized Value) / Initial Value) * 100 for metrics where lower is better (e.g., error rates). The row on the 'Hierarchy-Aware Agent' is illustrative and aims to quantify the benefit of implementing the logic discussed in Section 5.3. }}
  \end{tabular}
\end{table}

\subsection{Strategic Refinement of the Generation and Reasoning Phase: Cultivating Fidelity and Coherence in Legal AI}

Once the Retrieval-Augmented Generation (RAG) system has completed the critical retrieval phase, providing the Large Language Model (LLM) with a set of contextually relevant text fragments (ideally optimized through the techniques from the previous section), the challenge shifts to the generation phase. Here, the goal is to guide the LLM to use this retrieved information in such a way that the final response is not only linguistically fluent and coherent, but—crucially—factually accurate, logically sound, faithful to the provided sources, and directly relevant to the user's original query. The mere provision of context does not guarantee high-quality generation; the generative LLM, by its probabilistic nature and vast parametric knowledge, can still deviate, misinterpret, or even hallucinate. Therefore, the strategic refinement of this phase is an essential component of any legal RAG system that aspires to reliability.

\begin{enumerate}

    \item \textbf{Advanced prompt engineering, specific to RAG and aware of the legal context:} The prompt fed to the generative LLM, which now includes both the user's original query and the retrieved text fragments, must be meticulously designed to maximize fidelity and accuracy. This goes far beyond simple concatenation.

\begin{itemize}

    \item \textit{Explicit instructions on grounding and attribution:} The prompt must contain clear and unequivocal directives instructing the LLM to base its response \textbf{predominantly or exclusively} on the information contained within the provided documents and to actively avoid using its internal parametric knowledge or making unfounded assumptions. Mandates should be included to \textbf{explicitly cite sources} for its statements, ideally linking each proposition to the specific fragment or document from the retrieved context that supports it. This not only encourages fidelity but also facilitates verification by the user.
    \item \textit{Structured guides for reasoning (Chain-of-Thought and Similar Approaches):} For queries that require analysis or synthesis, rather than simple extraction, the prompt may instruct the LLM to follow a step-by-step reasoning process (Wei et al. 2023). For example, "First, identify the key facts in the provided documents. Second, identify the applicable legal rules mentioned. Third, apply these rules to the facts. Fourth, explain your conclusion, citing the relevant documents for each step." This externalization of the reasoning process not only tends to improve the quality of the final conclusion but also provides an audit trail that can be reviewed by a legal expert (Schwarcz et al., 2024).

    A robust approach to guiding reasoning is \textbf{explicit problem decomposition}. Instead of asking the LLM to directly determine whether a text violates a norm, the task is divided into the logical sub-components that a jurist would analyze. For example, to determine whether a comment constitutes "incitement to hatred" under § 130 of the German Penal Code, a system can be instructed to first answer two separate questions: (1) Does the text target a group protected by the norm? and (2) Does the text perform an act prohibited by the norm (incite, insult, etc.)? (Ludwig et al., 2025). Only if both answers are affirmative is it concluded that the norm has been violated. This methodology not only structures the model's "thinking," but also makes its final conclusion much more transparent and verifiable for the human supervisor.

    \item \textit{Sophisticated handling of uncertainty, conflicts, and missing information:} The prompt should guide the LLM on how to proceed when the retrieved information is incomplete, ambiguous, presents internal contradictions, or simply does not contain the answer to the query. Instead of forcing a response or resorting to fabrication, the model should be instructed to:

    \begin{itemize}
        \item Explicitly indicate uncertainty (e.g., "Based on the information provided, it is not possible to determine with certainty...").
        \item Present the different perspectives or conflicting information objectively, pointing out the discrepancies.
        \item State that the requested information is not found in the retrieved documents.
    \end{itemize}

    \item \textit{Precise definition of the persona and output format:} Specifying the role that the LLM should adopt (e.g., "Act as an objective and neutral legal research assistant") and the exact format of the expected response (e.g., structured summary, list of key points with citations, draft contractual clause) is crucial to ensure that the output is consistent, useful, and professional.
    \item \textit{Follow-up prompting or iterative refinement techniques:} As demonstrated by the Addleshaw Goddard study (2024), a second prompt that challenges or requests a review of the LLM's initial response (e.g., "Please carefully review your previous answer. Are you sure you have included all relevant information from the provided documents regarding X? Is there any nuance you may have omitted?") can induce the model to perform deeper processing of the context and significantly improve the quality and completeness of the final response. This iterative approach simulates a refinement conversation.
    \end{itemize}

    \item \textbf{Fine-tuning the generator LLM with a focus on legal fidelity and grounded reasoning:} While prompt engineering is a powerful and flexible tool, fine-tuning the generator LLM on a carefully selected corpus can offer deeper and more consistent improvements in its ability to adhere to context and reason in a legally sound manner.

\begin{itemize}
    \item \textit{Fine-tuning for contextual adherence:} Train the LLM on a high-quality dataset composed of triplets (query, relevant retrieved legal context, ideal answer that is strictly faithful to the context and properly cited). This can teach the model to prioritize contextual information over its parametric knowledge and to resist the temptation to "stray" or hallucinate (Tian, Mitchell, Yao, et al. 2023).
    \item \textit{Fine-tuning for specific types of grounded legal reasoning:} Develop training datasets that exemplify how to perform specific types of legal reasoning tasks (e.g., identifying the \textit{holding} of a case, applying a multi-factor legal test, comparing statutes) \textit{explicitly based on a set of input documents}.
    \item \textit{Reinforcement learning with human feedback (RLHF) focused on factuality and grounding:} Use RLHF not only to align the model with general preferences for style or usefulness, but specifically to reward responses that demonstrate high factuality, precise grounding in the provided sources, and coherent legal reasoning.
\end{itemize}

    \item \textbf{Integration with specialized reasoning models and advanced architectures:} The emergence of LLMs with architectures explicitly designed for multi-step reasoning, planning, and problem decomposition (such as OpenAI's "o" model family - OpenAI 2024) is particularly relevant for RAG.

\begin{itemize}
    \item \textit{Response planning:} These models could, in theory, plan how to use the retrieved information more strategically, identifying which fragments are most relevant to which parts of the query and how to synthesize them in a logically coherent manner.
    \item \textit{Internal verification of reasoning steps:} Their ability to "reflect" on their own intermediate reasoning steps could allow them to detect and correct errors or inconsistencies before generating the final answer (Schwarcz et al., 2024). Integrating these reasoning models as the generative component in a RAG system is a promising area for future improvements.
\end{itemize}

    \item \textbf{Hybrid architectures (Symbolic-Neural):} Although still under development for large-scale legal applications, integrating the LLMs' ability to process natural language with the precision and verifiability of symbolic reasoning systems (based on formal logics, structured legal ontologies, or explicit rule bases) offers a promising path. The LLM could use the retrieved context to instantiate a symbolic model that then performs logical inferences in a more controlled and explainable manner.

\end{enumerate}

The ultimate goal of these generation optimization strategies is not only to produce answers that \textit{appear} correct, but answers that are \textit{demonstrably} correct, faithful to the provided sources, and useful for the legal professional who bears the final responsibility for their use. The LLM's ability to explain \textit{how} it arrived at a conclusion from the retrieved context is as important as the conclusion itself.

\subsection{Post-Hoc Verification and Confidence Calibration: The Last Line of Defense Against Hallucinations}

Starting from the premise that it is theoretically impossible to prevent hallucinations one hundred percent during the generation phase with current technology, the implementation of robust verification mechanisms \textit{after} the LLM has produced an initial response becomes an absolutely critical security layer. This "last line of defense" does not so much seek to prevent the model from hallucinating, but rather to detect hallucinations when they occur and provide professional users with clear signals about the reliability of the generated information.

\subsubsection{Automated Factual Verification Against Canonical External Sources (Fact-Checking)}

Once the LLM has generated a response (which ideally includes preliminary citations), automated modules can be implemented that:
\begin{itemize}

    \item \textit{Extract key factual claims:} Identify the central factual and legal propositions in the LLM's response.
    \item \textit{Verify against high-confidence knowledge bases:} Compare these claims with information contained in structured and canonical legal databases (e.g., official legislation repositories, case law databases with repeal metadata, verified legal encyclopedias).
    \item \textit{Flag discrepancies:} Explicitly signal to the user any discrepancies found, indicating whether a claim could not be verified, contradicts a canonical source, or is based on a cited source that does not support it (Peng et al. 2023; Chern et al. 2023).
    \item \textit{Challenges:} The coverage of these external knowledge bases will never be complete, and the verification of complex or interpretative legal claims remains a challenge for automated systems.
\end{itemize}

\subsubsection{Verification Through Logical Rules and Deterministic Heuristics}
    
Before resorting to secondary AI models, a rule-based verification layer can efficiently and cost-effectively detect a significant class of errors. This includes:

\begin{itemize}
\item \textit{Syntactic validation}: Use regular expressions to verify that citations of judgments or legal articles follow the canonical format.
\item \textit{Simple logical consistency checks}: Implement rules that flag as suspicious any claim where a lower court overturns a higher one, or where a judgment date is later than the repeal date of the law applied.
\item \textit{Checklists}: Compare mentioned entities (judges, parties, laws) against authorized databases to detect blatant fabrications.
\end{itemize}
These methods, although traditional, are highly reliable for the errors they are designed to capture and should constitute a first line of defense.

\subsubsection{Secondary models for hallucination detection and self-critique}
    
Research is underway on the use of AI models, often smaller and more specialized, or even the generative LLM itself operating in a "self-evaluation" mode, to analyze the initial response for signs of hallucination.

\begin{itemize}

    \item \textit{Detection of internal inconsistencies:} Assessing the internal logical coherence of the generated response.
    \item \textit{Measurement of generation entropy or uncertainty:} Analyzing the probabilities associated with the generated token sequence; sequences with low probability or high entropy may be more prone to hallucinations (Manakul, Liusie, and Gales 2023 - SelfCheckGPT).
    \item \textit{Comparison with high-confidence parametric knowledge:} If the LLM has parametric "knowledge" about a topic with high confidence (e.g., very basic legal principles), it can use this to contrast the response generated from the RAG context.
    \item \textit{Generation of critiques (CriticGPT):} OpenAI has experimented with models (such as CriticGPT) trained to generate critiques of other LLMs' responses, helping to identify errors or weaknesses (Song et al., 2024 - RAG-HAT).
\end{itemize}

\subsubsection{Calibration and Effective Communication of Model Confidence}
    
It is essential that LLMs not only generate responses, but also reliably communicate their own level of "confidence" or uncertainty regarding the correctness and substantiation of those responses.

However, merely generating a confidence score is not sufficient; the interpretation of such a score by the professional user represents another significant challenge. A '70\% confidence' expressed by an LLM may not have the same intuitive or statistical meaning as a 70\% confidence in a human context or in a traditional diagnostic system. Therefore, along with the development of more reliable confidence metrics, it is essential to investigate and establish clear guidelines on how legal professionals should interpret and act upon these algorithmic confidence indicators, especially when model calibration, as evidenced in studies such as Dahl et al. (2024), remains imperfect and can lead to dangerous overconfidence in erroneous answers.   

\begin{itemize}

    \item \textit{Development of reliable confidence metrics:} Investigate and refine techniques for LLMs to produce confidence scores that correlate well with their actual accuracy on specific legal tasks (Kadavath et al. 2022; Xiong et al. 2023). This remains an active and challenging area of research, as demonstrated by the calibration issues observed in Dahl et al. (2024).
    \item \textit{Transparent presentation of uncertainty:} The user interface should clearly communicate to the legal professional the confidence levels associated with different parts of the response, or explicitly flag statements for which the model has low confidence.
    \item \textit{Intervention thresholds:} For high-risk applications, confidence thresholds could be established below which a response is not presented to the user or is unequivocally marked as "requires intensive human verification."
\end{itemize}

The need for transparent communication about the capabilities and limitations of AI systems, including their level of confidence or uncertainty, finds normative resonance in the EU AI Act. Article 52(1) of the EU-AIAct, for example, establishes transparency obligations for certain AI systems, including those that generate content. It requires that users be informed that they are interacting with an AI system and, when an AI system generates or manipulates text, audio, or video content that closely resembles existing content ('deep fakes'), it must be disclosed that the content has been artificially generated or manipulated. While these provisions do not directly prevent the generation of a factual hallucination in a legal response, they do seek to foster greater awareness and caution on the part of the user, enabling them to weigh the reliability of the information received and establishing a basis for accountability when AI is deceptively presented as human or its content as non-artificial.

\subsubsection{Generation of Multiple Hypotheses and Contrasting Explanations}

Instead of generating a single "definitive" answer, the LLM could be instructed to generate multiple possible interpretations or arguments based on the retrieved context, especially if it is ambiguous or presents conflicting information. It could also generate explanations that contrast the pros and cons of different legal approaches, allowing the human professional to weigh the alternatives.

\subsubsection{Facilitating Human Verification through Precise and Traceable Citations}

One of the most important contributions of well-designed RAG systems is their ability to enhance verifiability.

\begin{itemize}

    \item \textit{Chunk-Level Citation:} The system should not simply list the retrieved documents, but, as much as possible, should link each specific claim or conclusion in the generated answer to the exact fragment(s) of the retrieved text that support it.
    \item \textit{Highlighting Relevant Passages:} The user interface could highlight the specific passages in the source documents that were most influential in generating the answer, allowing the lawyer to go directly to the evidence.
    \item \textit{Transparency about the Retrieval Process:} Providing the user with visibility into which documents were retrieved (and perhaps why, e.g., by showing similarity scores) can help them assess the quality of the information base used by the LLM.
\end{itemize}

\subsubsection{Designing an Intelligent Abstention Capability: The Principle of "Strategic Silence"}

Beyond merely communicating a confidence score, an advanced mitigation strategy consists of designing the system's abstention capability ("I don't know") not as an error or an accidental limitation, but as a deliberate and strategic function. A system that is able to identify the limits of its knowledge or of the sources provided inspires greater trust and is inherently safer than one optimized to generate a response at all costs. The implementation of this "strategic silence" principle involves several key components:

\begin{itemize}
    \item \textbf{Specific justification for abstention:} When the system abstains, it should not offer generic excuses. The response should be a precise diagnosis of the limitation encountered. For example: "It is not possible to provide a well-founded answer, as no primary sources have been found in the knowledge base after 2023 on this subject," or "The retrieved sources present conflicting data on point X and do not allow for a conclusive synthesis." This transparency educates the user and turns a potential frustration into an informative interaction.
    \item \textbf{Provision of constructive alternatives:} Abstention should not be a dead end. A robust system should offer the user alternative courses of action that still provide value. For example: "Although I cannot determine direct applicability, I can provide the general legal framework and a checklist of elements that a professional should analyze," or "I can formulate the specific questions you should direct to a legal advisor to resolve this issue."
    \item \textbf{Visual and explicit communication of uncertainty:} In line with confidence calibration, the interface should proactively communicate the reliability level of a response. A "traffic light" system (for example, green for answers with high consensus and strong sources; amber for those with information gaps or secondary sources; red for those based on conflicting or highly uncertain data) allows the user to immediately calibrate their own level of scrutiny.
    \item \textbf{Auditability of abstention:} Each instance of abstention should generate an auditable record (\textit{log}). This record should document the state of the system at that moment: the user's query, the sources retrieved (or the lack thereof), the criteria that led to the decision to abstain, and the predefined confidence thresholds. This traceability is essential for the continuous improvement of the system and for accountability.

\end{itemize}

Ultimately, the design of legal AI systems must redefine their incentives: instead of rewarding verbosity and completeness at any cost, precision and responsible coverage should be rewarded. A model that knows how to abstain in a justified and transparent manner is not a less capable system, but one that has reached a higher degree of maturity and demonstrates a deep respect for the user and for the criticality of the domain in which it operates. This conservative calibration is a fundamental pillar for building sustainable, long-term trust in legal AI.

\subsubsection{Conclusion}

These post-hoc verification strategies and confidence communication do not eliminate the need for prior optimizations in data, retrieval, and generation, but they act as a crucial safety net. They acknowledge the inherent fallibility of LLMs and seek to empower the legal professional with the tools and information necessary to use AI in a more critical, informed, and ultimately safer manner. Nevertheless, it is crucial to recognize the delicate balance inherent in the 'cost of verification.' While these post-hoc safety layers are indispensable for reliability, their extensive implementation—especially if it involves significant human intervention for each check or high latency due to multiple calls to secondary models—could undermine one of the primary benefits that AI promises: efficiency and cost reduction. A system that requires such exhaustive manual verification of each of its outputs that it completely negates the initial time savings may not be viable in practice for many tasks. Therefore, future development should seek not only the effectiveness of these verification mechanisms but also their efficiency, possibly through greater intelligent automation of the verification itself or through AI systems that learn to 'self-correct' more reliably with minimal supervision. Finding the optimal point where the robustness of verification does not disproportionately sacrifice operational efficiency is an ongoing challenge in the design of practical and reliable legal AI systems.

\subsection{The Irreducible and Strengthened Role of Expert Human Supervision}

Despite the increasing sophistication of optimization strategies and hallucination mitigation—from data curation to post-hoc verification—it is imperative to conclude this section by reaffirming a fundamental principle: in the current and foreseeable state of artificial intelligence, \textbf{critical, informed, and expert human supervision is not merely a desirable option, but an absolutely irreducible and indispensable component} for the safe and ethical integration of LLMs into legal practice. No combination of the algorithmic techniques discussed can, by itself, replace the depth of contextual judgment, ethical responsibility, and nuanced understanding of the legal professional.

The need for this human expertise is not conjecture, but an empirical conclusion. Even in studies that implement highly sophisticated conditioning strategies, providing LLMs with legal definitions and case examples, a "significant performance gap" is documented between the best AI model and human legal experts. The study by Ludwig et al. (2025) found that, while the models could reasonably well identify groups protected by hate speech law, they had serious difficulties correctly classifying prohibited conduct—a task of nuanced judgment where human jurists demonstrated far superior reliability. This underscores that the final stage of evaluation and qualitative judgment remains, for now, an exclusively human capability.

This is particularly true because detecting subtle errors or assessing the robustness of a legal argument generated by AI often intrinsically depends on the competence and judgment of the professional. What may seem like a coherent and useful response to a layperson or junior lawyer could, to an expert, reveal argumentative deficiencies or a superficial understanding of the applicable doctrine. The 'truth' or 'viability' of a complex legal conclusion is not always self-evident and requires informed scrutiny.

The duty of competence in the AI era, therefore, requires the professional to understand that they are not interacting with a "knowledge oracle," but with a statistical system optimized for plausibility, whose fundamental design incentivizes it to generate confident responses even when its knowledge base is uncertain (Kalai et al., 2025). Recognizing this design feature is the foundation of the professional skepticism necessary for effective supervision.

Far from rendering the lawyer obsolete, the emergence of hallucination-prone LLMs—even those augmented by RAG—\textbf{reinforces and redefines the value of human expertise}. The lawyer's role evolves from being a mere information retriever or document drafter (tasks that AI can assist with or even partially automate) to becoming:

\begin{enumerate}

    \item \textbf{Critical AI Supervisor:} The lawyer must act as an intelligent and skeptical "quality controller" of the outputs generated by AI. This involves not only verifying the factual correctness and legal validity of the information, but also evaluating its contextual relevance, its strategic suitability to the client's objectives, and its ethical implications.

    This "quality controller" role goes beyond simple factual verification. The lawyer must act as a metacognitive filter, being aware that the way the LLM presents information can induce biases in their own reasoning process. Research on cognitive biases induced by LLMs shows that the outputs of these models can alter the framing or emphasis of information, leading humans to make different decisions than they would with the original information (Alessa et al., 2025). Therefore, critical supervision involves an act of self-reflection: the lawyer must not only ask "is this information correct?", but also "is this presentation of the information unduly influencing my judgment?".
    
    \item \textbf{Curator and guide of AI knowledge:} In the context of customizable or fine-tunable RAG systems, expert lawyers can play a crucial role in curating knowledge bases, designing effective prompts, and providing feedback to improve the model's performance on specific legal tasks.
    \item \textbf{Interpreter and communicator of AI output:} Even if an AI generates a technically correct legal analysis, it will often be required for a human lawyer to translate it into language understandable to the client, contextualize it within the client's particular situation, and integrate it into a broader legal strategy.
    \item \textbf{Guarantor of ethical and strategic judgment:} AI can process information and generate options, but the final decision-making that involves complex ethical considerations, the exercise of professional judgment over alternative courses of action, managing the client relationship, and assuming ultimate professional responsibility, remain firmly within the human domain.
    \item \textbf{Navigator of legal uncertainty and ambiguity:} As discussed, law is full of gray areas, normative conflicts, and situations where there is no single "correct answer." The ability of a lawyer to navigate this uncertainty, weigh risks and benefits, and advise the client accordingly, is a skill that current AI does not possess.
\end{enumerate}

In this new paradigm, the efficiency promised by AI only materializes if it is accompanied by a proportional investment in \textbf{training lawyers to interact critically with these tools}. This includes developing skills in:

\begin{itemize}

    \item  \textit{Semantic and professional interaction, beyond Prompt Engineering}\textbf{:} While the quality of an AI's response currently often depends on 'prompt engineering', it is essential to recognize that this paradigm is a transitional solution and a design flaw, not a final goal. The responsibility for technical complexity should not fall on the legal professional, but rather on the developer of the LegalTech tool.

True innovation lies in developing solutions that abstract away this complexity, allowing the jurist to interact in their own language—both natural and technical—while the tool assumes the burden of translating that intent into the algorithmic instructions required by the model. Demanding that a surgeon learn to program their scalpel is a design failure; similarly, technology should be a scalpel that adapts to the lawyer's hand. This approach frees them to focus on what no machine can do: applying judgment, strategy, and ethics.

    \item \textit{Rigorous verification techniques:} Knowing the sources of primary and secondary authority, and being able to efficiently cross-check AI outputs against them.

    Ultimately, this new paradigm reinforces a maxim that should guide the future of LegalTech: the lawyer should not become a 'prompt engineer'. The responsibility for technical complexity lies with the tool's developers, not the end user. Requiring legal professionals to learn complex prompting techniques to obtain reliable results is a design failure and an unacceptable reversal of roles. Technology should be a scalpel that adapts to the surgeon's hand, not a machine that demands the surgeon learn its arcane language. Therefore, the future of reliable legal AI lies in systems that enable natural language interaction and assume the burden of technical interpretation, freeing the lawyer to focus on what no machine can do: applying judgment, strategy, and ethics.
    
    \item \textit{Understanding the limitations of AI:} Being aware of the types of errors and biases to which AI is prone (including hallucinations) and knowing when not to trust its results.
    \item \textit{Ethical integration of AI into practice:} Understanding the deontological implications of using AI and how to comply with professional duties in a technologically enhanced environment.
\end{itemize}

This principle of the indispensability of human oversight is not articulated solely as a conclusion derived from the intrinsic technical limitations of current AI, but is progressively being enshrined as a fundamental requirement in the most advanced regulatory frameworks. The EU AI Act, in its Article 14, is explicit in requiring that high-risk AI systems be designed to be 'effectively overseen by persons'.

A paradigmatic example of how these principles are materializing at the national level can be found in Spain. In June 2024, the \textbf{State Technical Committee for Electronic Judicial Administration (CTEAJE)} (the high-level governmental body responsible for the technological modernization and digital strategy of the Spanish judicial system) published its \textbf{"Policy on the Use of Artificial Intelligence in the Administration of Justice"}. This document, which is mandatory for justice administration personnel, is not merely a recommendation, but a regulatory framework that sets out unequivocal guidelines:

\begin{itemize}

    \item \textbf{Non-Substitution Principle:} The policy firmly establishes that \textbf{"AI must never replace human decision-making in crucial matters"} and that "the ultimate responsibility for making legal decisions must rest with judges and magistrates" (Principle 1.4.1).
    \item \textbf{Universal Human Review Mandate:} Beyond the principles, it is imposed as a rule of use that \textbf{"human review of everything generated [by AI] whenever it directly or indirectly affects the rights of users" is mandatory} (Rule 1.5.1). This makes supervision an unavoidable procedural requirement, not an option.
    \item \textbf{Explicit Recognition of Risks:} The CTEAJE guide explicitly defines the phenomenon of \textbf{"hallucinations"} and acknowledges the danger of \textbf{"endemic automation bias"}, whereby humans tend to blindly trust the suggestions of systems. This official recognition underscores the need for informed skepticism, a fundamental pillar of expert oversight.
\end{itemize}
    
The existence of such a detailed and binding guide from a body like the CTEAJE demonstrates that the irreducible role of the human professional has transcended academic debate to become a pillar of public policy and AI governance in the legal field.

Ultimately, reliability in the era of legal AI will not reside exclusively in the perfection of algorithms, but in the \textbf{effective symbiosis between the processing capacity of AI and the wisdom, critical judgment, and ethical responsibility of the human professional}.

This principle of the indispensability of human oversight is not articulated solely as a conclusion derived from the intrinsic technical limitations of current AI, but is progressively being enshrined as a fundamental legal requirement in the most advanced regulatory frameworks. The EU AI Act, in its Article 14, is explicit in requiring that high-risk AI systems be designed and developed in such a way that they can be 'effectively overseen by persons during the period in which the AI system is in use.' This supervision must enable humans to understand the capabilities and limitations of the system, remain aware of the possible tendency toward automation or confirmation bias, correctly interpret the system's output, and have the authority and competence to decide not to use such output, override it, or intervene if the system presents anomalous, unforeseen, or potentially harmful results, as is the case with hallucinations that compromise legal validity.

Optimization and mitigation strategies are essential tools in this process, but their ultimate effectiveness depends on being implemented and supervised by well-trained legal professionals, aware of the risks and committed to the highest standards of the profession. Far from being an existential threat, hallucinatory AI can, paradoxically, underscore the enduring and irreplaceable value of expert human intelligence at the heart of the law.

\section{The Reality of Hallucinations in Practice: Detailed Case Studies and Lessons Learned from Judicial Incidents}
\label{sec:casos_reales}

While theoretical analysis and empirical evaluation in controlled environments are fundamental for understanding the nature and prevalence of hallucinations in Large Language Models (LLMs) applied to law, it is in the arena of real legal practice where the consequences of these algorithmic errors manifest with undeniable starkness and tangible impact. Incidents in which AI-generated information, either incorrect or entirely fabricated, has been introduced into judicial proceedings are not mere anecdotes or technological curiosities; they represent systemic failures with the potential to undermine the administration of justice, erode public trust, and bring about serious professional sanctions for the attorneys involved. This section delves into a detailed analysis of several prominent and well-documented case studies, drawing from them crucial lessons about specific failure points in human-AI interaction, deficiencies in verification processes, and the direct consequences of uncritically relying on these powerful yet fallible tools. These cases serve as potent warnings and as catalysts for deeper reflection on the necessary safeguards in the integration of AI into legal practice.

The proliferation of these incidents has reached a critical point, motivating the creation of dedicated resources for their tracking. A notable example is the online database "\href{https://www.damiencharlotin.com/hallucinations}{AI Hallucination Cases Database}", a project that seeks to comprehensively compile all judicial decisions where hallucinated content generated by AI has been a relevant factor. This type of repository is becoming a vital tool for jurists, academics, and regulators, as it enables a systematic analysis of the nature and frequency of these failures in real-world practice.

\begin{table}[htbp]
  \centering
  \caption{Comparative analysis of notable judicial incidents involving AI hallucinations}
  \label{tab:incidentes_judiciales_ia_corregida}
  \footnotesize 
  \begin{tabular}{
    >{\raggedright\arraybackslash}p{0.26\textwidth} 
    >{\raggedright\arraybackslash}p{0.34\textwidth} 
    >{\raggedright\arraybackslash}p{0.35\textwidth} 
  }
    \toprule
    \textbf{Case / Jurisdiction} & \textbf{Nature of the Hallucination} & \textbf{Key Lessons / Consequences} \\
    \midrule
    \textit{Mata v. Avianca, Inc.} (S.D.N.Y. 2023, USA)* & 
    \makecell[l]{Complete fabrication of multiple \\court cases (citations, holdings). \\Affirmative chatbot response \\regarding the existence of the cases.} & 
    \makecell[l]{Inescapable duty of independent verification. \\Deceptive nature of plausible hallucinations. \\Individual professional responsibility. \\Economic and reputational sanctions.} \\
    \addlinespace
    
    \textit{Thackston v. Driscoll} (W.D. Texas, 2025, USA)* &
    \makecell[l]{Accumulation of multiple error types: \\Fabrication of case law and doctrine, \\false citations in real cases, misrepresentation \\of holdings (\textit{misgrounding}), and use of \\overturned case law.} &
    \makecell[l]{Illustrates the "cascade" of errors that a single \\negligent use of AI can produce. \\Reinforces the duty to verify beyond \\the mere existence of the case. Underscores the \\serious professional consequences \\(Rule 11 sanctions).} \\
    \addlinespace

    Case in Australia (Family, 2024, cited by Lantyer) & 
    False citations in a family law case. & 
    \makecell[l]{Universality of the hallucination risk. \\Application of deontological duties \\(diligence, competence). \\Response from disciplinary bodies.} \\
    \addlinespace
    
    Case in Brazil (Appeal, 2025, cited by Lantyer) & 
    False case law generated by AI in an appeal. & 
    \makecell[l]{Risks at all judicial levels. \\Importance of ongoing AI training \\for lawyers and judges. Fine imposed.} \\
    \addlinespace

    Constitutional Court Case (Spain, 2024)* &
    \makecell[l]{Complete invention of 19 citations of judicial \\doctrine, presented as verbatim in an \\amparo appeal.} &
    \makecell[l]{The lawyer's responsibility is absolute and \\independent of the tool that caused \\the error. Negligent use of AI constitutes \\a breach of the duty of respect to the court.} \\
    \addlinespace

    Magesh et al. Study (2024) (RAG Tools) & 
    \makecell[l]{Mainly \textit{misgrounding} (citing a real source \\but misrepresenting its content), reasoning errors, \\suppression of citations.} & 
    \makecell[l]{RAG mitigates but does not eliminate hallucinations. \\Subtle errors can be more insidious than \\obvious fabrication. Need for \\thorough source verification.} \\
    \bottomrule
  \end{tabular}
  \par
  \small{\textit{Note: Cases marked with an asterisk (*) are specifically discussed in later sections.}}
\end{table}

\subsection{Case Study: The Paradigmatic \textit{Mata v. Avianca, Inc.} and the Fabrication of Jurisprudence}

The case \textit{Robert Mata v. Avianca, Inc.}, No. 22-cv-1461 (PKC) (S.D.N.Y. 2023), has quickly become the obligatory reference when discussing the dangers of AI hallucinations in litigation. In this matter, the plaintiff's attorneys, seeking to oppose a motion to dismiss, submitted a legal brief that cited multiple judicial decisions supposedly favorable to their position. However, after an investigation by the defense and the court itself, it was discovered that at least six of the cited cases were completely nonexistent, fabrications generated by ChatGPT, the AI tool that one of the attorneys had used for legal research (Weiser, 2023; Dahl et al., 2024).

Judge P. Kevin Castel, in imposing sanctions on the attorneys involved (including a monetary fine and the obligation to notify the judges whose names were falsely associated with the fabricated opinions), issued a detailed order dissecting the multiple failures in the process. The attorney who used ChatGPT, Steven A. Schwartz, admitted he was not an expert in federal legal research and had used the tool as a "super search engine," relying on its responses and even asking ChatGPT whether the cases it provided were real, to which the chatbot answered affirmatively (Sanctions Order in \textit{Mata v. Avianca, Inc.}, June 22, 2023).

\textbf{Lessons Learned from \textit{Mata v. Avianca}:}

\begin{enumerate}
    \item \textbf{Independent verification as an inescapable duty:} The most obvious and compelling lesson is the absolute necessity for lawyers to independently and rigorously verify every source and every legal proposition generated by an AI before incorporating it into a court document. Simply asking the AI about the truthfulness of its own output is manifestly insufficient and denotes a fundamental lack of understanding about how these models operate.
    \item \textbf{The deceptive nature of hallucinations:} The fabricated citations by ChatGPT in the \textit{Mata} case were highly plausible, with party names, volume and page numbers, and summaries of \textit{holdings} that mimicked the format and style of real judicial opinions. This plausibility makes hallucinations particularly insidious and difficult to detect without cross-checking against canonical legal databases.
    \item \textbf{Individual professional responsibility:} The case underscores that the ultimate responsibility for the content of documents submitted to the court rests unequivocally with the signing attorney, regardless of the tools used in their preparation. The use of AI neither dilutes nor transfers this responsibility.
    \item \textbf{Lack of awareness of AI limitations:} Attorney Schwartz's admission that he "was not aware of the possibility that [ChatGPT's] content could be false" reveals a significant gap in AI literacy within the legal profession. Understanding the inherent limitations of LLMs, including their propensity to "hallucinate" or "confabulate," is an essential component of professional competence in the digital age.
    \item \textbf{Impact on the Integrity of the Judicial System:} The introduction of fictitious case law into a judicial proceeding not only harms the client and exposes the attorney to sanctions, but also "promotes cynicism toward the legal profession and the American justice system" and constitutes an abuse of the judicial process (Sanctions Order in \textit{Mata v. Avianca, Inc.}).
\end{enumerate}

\subsection{Case Study: The Sanction by the Spanish Constitutional Court and the Non-Delegable Responsibility of the Attorney}

\subsubsection{Factual Summary of the Incident}

In September 2024, the First Chamber of the Constitutional Court of Spain set a fundamental precedent by unanimously sanctioning a lawyer for failing to show due respect to the court (Information Note 90/2024). The misconduct consisted of including, in an amparo petition, 19 supposedly literal quotations from judgments of the Court itself that turned out to be completely non-existent. These quotations were presented in quotation marks, attributing to the justices a constitutional doctrine that "lacked any basis in reality." The sanction imposed was a "warning," the mildest possible, but the Court ordered the matter to be referred to the Barcelona Bar Association for the corresponding disciplinary proceedings.

\subsubsection{Case Analysis Through the Report’s Framework}

This case serves as a perfect illustration of the concepts analyzed in this report:
\begin{itemize}
    \item \textbf{Nature of Hallucination} (Applying the Taxonomy from Section 2.1): The error committed fits directly into the category of "Fabrication of Authority." It was not a subtle misrepresentation (misgrounding), but the complete invention of judicial doctrine. The presentation of the passages in quotation marks aggravates the misconduct, as it is not presented as an interpretation, but as a literal and verifiable quotation, which constitutes a particularly serious form of misinformation according to the typology of Magesh et al. (2025).
    \item \textbf{Professional Responsibility Above the Tool} (applying Section 8.1): The most crucial point of the Court’s decision is its reasoning regarding the lawyer’s responsibility. The attorney argued in his defense a "misconfiguration of a database." The Court categorically rejected this argument, establishing a principle of absolute responsibility that is independent of the cause of the error. In its own words, "whatever the cause of the inclusion of unreal quotations (use of artificial intelligence, quoting one’s own arguments, etc.), the lawyer is always responsible for thoroughly reviewing all the content" (Information Note 90/2024). This statement is the clearest practical manifestation of the duty of diligence and professional competence in the era of AI, emphasizing that human supervision is not an option, but a non-delegable obligation.
    \item \textbf{Impact on the Integrity of the Judicial System} (applying Section 2.3): The Court did not consider it a mere procedural error, but a lack of respect that showed a "clear disregard for the judicial function" of the justices. The conduct, according to the decision, disrupted the work of the Court not because of the need to verify the quotations—which is always done—but because of "having to judge the consequences of such unjustified irregularity." This demonstrates that the introduction of false information not only contaminates legal debate, but also undermines the trust and mutual respect that must govern the relationship between lawyers and the judiciary, eroding the foundations of the system.
\end{itemize}

\subsubsection{Human Lesson: Delegation of Critical Responsibility}

Beyond the technical-legal analysis, the case of the Spanish Constitutional Court, like Mata v. Avianca, is a symptom of a dangerous cultural trend: the delegation of critical thinking. The attorney did not fail because he used a faulty tool; he failed because he abdicated his fundamental responsibility to verify, doubt, and think. He treated AI as an oracle instead of as an assistant. In any profession, but especially in law, the value does not lie in the ability to generate an answer, but in the ability to defend it. When a professional simply copies and pastes a result they do not understand, they are not using technology; they are being used by it. These incidents should not generate fear of AI, but rather a deep respect for the irreplaceable role of human judgment. Technology does not exempt us from our obligation to be excellent; in fact, it demands it from us more than ever.

\subsubsection{Key Lessons and Comparison with Mata v. Avianca}

This case, although similar in origin to Mata v. Avianca, offers complementary lessons of greater ethical significance:
\begin{itemize}
    \item Universality of the duty of verification: It confirms that the obligation to verify every piece of information presented before a court is a universal principle of legal practice, applicable with equal force in civil law systems (Spain) as in common law systems (U.S.).
    \item Irrelevance of the cause of error: While in Mata the debate focused on the misuse of a specific tool (ChatGPT), the Spanish Constitutional Court elevates the principle: the lawyer's responsibility is absolute, regardless of whether the error was caused by AI, faulty software, or human oversight. The tool is irrelevant; the responsibility is total.
    \item From procedural sanction to ethical breach: The Spanish case frames the problem not only as negligence deserving a procedural sanction, but as a breach of the duty of respect, a pillar of professional ethics. It is a "disregard" for the judicial function, which gives it a higher deontological gravity.
\end{itemize}

\subsection{Case Study: \textit{Thackston v. Driscoll} and the Cascade of Algorithmic Errors}

The case of \textit{Thackston v. Driscoll}, decided by a Magistrate Judge in the Western District of Texas on August 28, 2025, stands as an alarmingly comprehensive example of the dangers arising from an uncritical and negligent use of generative AI in legal practice. Unlike \textit{Mata v. Avianca}, which focused primarily on the fabrication of cases, \textit{Thackston} illustrates a "cascade" of errors that spans nearly the entire range of the taxonomy of legal hallucinations.

\subsubsection{Factual Summary of the Incident}

In the context of an employment discrimination lawsuit against the U.S. Army, the plaintiff's attorney submitted a reply brief riddled with defective legal information. The court, in its "Report and Recommendation," meticulously dissected the errors, which included:

\begin{itemize}

    \item \textbf{Fabrication of Authority:} Citations to completely non-existent cases (e.g., a supposed Ninth Circuit opinion in \textit{United States v. City of Los Angeles} and a case from the same court in \textit{EEOC v. Exxon Mobil Corp.} that did not exist).
    \item \textbf{False Citations and Misrepresentation (\textit{Misgrounding}):} The brief attributed fabricated direct quotes to real and well-known cases (e.g., to \textit{Palmer v. Shultz} and \textit{Armstrong v. Turner Industries}). Additionally, it severely misrepresented the holdings of other real cases, citing them in support of legal propositions they did not sustain.
    \item \textbf{Temporal Error:} The famous case \textit{Chevron}, a pillar of administrative law, was cited without acknowledging that it had been explicitly overturned by the Supreme Court—a first-order factual and strategic error.
\end{itemize}
    
The Magistrate Judge not only identified the errors but also explicitly suspected the use of AI due to the "repetitive and redundant language" and the nature of the fabrications. The judge concluded that the attorney violated Federal Rule 11(b) by failing to conduct a "reasonable inquiry" and recommended that the District Court impose sanctions, suggesting a fine and mandatory attendance at a training course on generative AI.

\subsubsection{Case Analysis through the Report Framework}

This case is a perfect microcosm of the systemic risks discussed in this report.

\begin{itemize}
    \item \textbf{A complete taxonomy in a single document} (applying Section 2.2)\textbf{:} \textit{Thackston} is a masterclass on the different types of hallucinations. It demonstrates that a lawyer who blindly trusts an AI does not make just one type of mistake, but exposes themselves to a systemic failure. The combination of \textbf{fabrication of authority}, \textbf{misgrounding}, and \textbf{temporal error} in a single brief shows the inability of the model (and the lawyer) to distinguish between what is real, what is distorted, and what is obsolete. The \textit{misgrounding} is particularly insidious here, as the lawyer could have verified the existence of the case and stopped there, falling into a false sense of security.
    \item \textbf{The abdication of professional judgment and "user hallucination"} (applying Sections 2.3 and 5.6)\textbf{:} The court is unequivocal: the fault does not lie with the "machine," but with the professional who abdicated their fundamental duty. The Magistrate Judge's recommendation focuses on the violation of Rule 11, which requires a reasonable inquiry \textit{before} filing any document. This is the paradigmatic example of "user hallucination": the mistaken belief that the tool can replace diligence, skepticism, and professional judgment. It reinforces the central principle of this report: the role of human oversight is not only a best practice, it is an irreducible legal and ethical obligation.
    \item \textbf{From theory to real consequences} (applying Section 8)\textbf{:} The \textit{Thackston} case materializes the deontological and regulatory implications. The recommendation of monetary sanctions and mandatory training is not an abstract reprimand, but a direct professional and economic consequence. It serves as a powerful warning, aligned with the lessons from \textit{Mata} and the Spanish Constitutional Court case: courts will not hesitate to use their sanctioning powers to protect the integrity of the judicial process against the introduction of false information, regardless of the technology used to generate it.
\end{itemize}

\subsection{Incidents Beyond Flagrant Fabrication: Subtle Errors and \textit{Misgrounding}}

The study by Magesh et al. (2025) documents that the most frequent and dangerous hallucinations in RAG systems are not complete fabrications, but rather more subtle legal reasoning errors, which they term "insidious." They identify three main categories of failure:
\begin{itemize}
    \item Misunderstanding Holdings: Tools often summarize a judgment by stating the opposite of what the court actually decided, confusing the holding (the central decision) with dicta (secondary comments).
    \item Distinguishing Between Legal Actors: AIs mistakenly attribute the arguments of a litigant to the court, presenting the position of one party as if it were the judge's final decision.
    \item Respecting the Order of Authority: The models demonstrate an inability to understand judicial hierarchy, for example, claiming that a lower court overturned a decision of a higher court, which is legally impossible.
\end{itemize}
These misgrounding errors are particularly dangerous because the presence of a real citation creates a false sense of authority and reliability, which can lead the lawyer to place undue trust in the proposition without conducting the necessary critical reading of the source (Magesh et al., 2025).

While the complete fabrication of cases as in \textit{Mata v. Avianca} is the most spectacular form of hallucination, empirical studies on commercial legal tools based on RAG, such as that of Magesh et al. (2024), reveal that more subtle but equally problematic forms of error are even more frequent. These incidents, although they do not always carry such publicized sanctions, can have a significant impact on the quality of legal work and decision-making.

A recurring example documented by Magesh et al. (2024) is \textit{misgrounding}, where the AI tool cites a \textit{real and existing} case or statute, but the legal proposition it attributes to that source is incorrect, misrepresented, or simply not contained in the original text of the cited authority. In one of the analyzed instances, Westlaw AI-Assisted Research incorrectly stated the \textit{holding} of a U.S. Supreme Court decision, attributing to it a conclusion opposite to that actually reached by the court. In another example, Lexis+ AI described a case (Arturo D.) as good law and used it to support a proposition, when in reality the cited case (Lopez) had \textit{overruled} Arturo D. on the relevant point.

\textbf{Lessons Learned from \textit{Misgrounding} and Similar Errors:}

\begin{enumerate}

    \item \textbf{Verification cannot be limited to the existence of the citation:} Unlike obvious fabrications, \textit{misgrounding} requires a deeper level of verification. It is not enough to confirm that the cited case or statute exists; the lawyer must read and understand the original source to ensure that it truly supports the assertion made by the AI.
    \item \textbf{Fragility of contextual understanding in AI-RAG:} These errors suggest that, even when provided with the retrieved context, LLMs may struggle to correctly interpret the nuances of legal language, distinguish the \textit{holding} from the \textit{dicta}, or understand the hierarchical and temporal relationships between precedents (e.g., the effect of a reversal).
    \item \textbf{The risk of "false grounding":} \textit{Misgrounding} is particularly dangerous because the presence of a real citation can create a false sense of authority and reliability, leading the lawyer to place undue trust in the proposition without performing the necessary critical reading of the source.
    \item \textbf{Need for deep optimization in RAG:} These incidents reinforce the conclusion of the Addleshaw Goddard (2024) report that the effectiveness of RAG depends on meticulous optimization of each component, including not only retrieval but also the way the generating LLM is prompted to interact with and reason about the retrieved context.
    \end{enumerate}

\subsection{Global Implications and the Need for Continuous Adaptation}

Although many of the most notorious cases have arisen in the U.S., the problem of AI hallucinations and the need for diligent verification by lawyers is a global concern. Similar incidents have begun to be documented in other jurisdictions, including Australia (where a lawyer was referred to a disciplinary body for using AI that generated fake citations in a family law case – The Guardian, 2024, cited in Lantyer, 2024) and Brazil (where a lawyer was fined for using fake case law generated by AI in an appeal – Migalhas, 2025, cited in Lantyer, 2024).

Beyond anecdotal judicial incidents, systematic academic research confirms this risk at a global level. An empirical study focused on a jurisdiction outside the European sphere compared the performance of multiple LLMs with a human lawyer, concluding that legal research was the worst-performing task, with a consistent tendency for AI models to invent case law (Hemrajani, 2025).

\textbf{Lessons Learned from the Global Perspective:}

\begin{enumerate}
    \item \textbf{Universality of the Risk:} The propensity of LLMs to hallucinate is not limited by geographic borders or legal systems. It is an inherent feature of current technology that affects all professionals who use it.
    \item \textbf{Adaptation of Ethical Duties:} While the details of ethical codes vary between jurisdictions, the fundamental principles of competence, diligence, loyalty to the client, and candor before the court are widely shared. The legal profession in each country will need to interpret and apply these principles to the new context of AI.
    \item \textbf{Response of Disciplinary and Judicial Bodies:} The way in which courts and disciplinary bodies in different countries respond to these incidents will set important precedents and shape expectations regarding the responsible use of AI by lawyers. The sanctions imposed in cases such as \textit{Mata} serve as a clear signal of the seriousness with which these failures are taken.
    \item \textbf{Importance of Training and AI Literacy:} Globally, there is an urgent need to improve lawyers’ training on the capabilities and limitations of AI, including raising awareness of the risk of hallucinations and developing critical verification skills.
\end{enumerate}

In conclusion, case studies of real incidents where AI hallucinations have impacted judicial proceedings offer compelling and unavoidable lessons. They underscore that the integration of AI into legal practice is not a risk-free process and that the technology, in its current state, cannot substitute for the critical judgment, investigative diligence, and ethical responsibility of the human professional. These cases should not be interpreted as a condemnation of AI \textit{per se}, but rather as an urgent call for caution, rigorous verification, and the development of professional practices and technological safeguards that enable the potential of AI to be harnessed while minimizing its inherent dangers. The “harsh reality” of these incidents should serve as a driver for continuous improvement, both of the technology and of the way the legal profession interacts with it.

\section{The Future of Reliable Legal Artificial Intelligence: Towards Explainable, Auditable, and Responsible-by-Design Models}
\label{sec:futuro_ia_fiable}

The current landscape of artificial intelligence (AI) applied to law, although brimming with transformative potential, is marked by the persistent challenge of hallucinations and the inherent limitations regarding the reliability of Large Language Models (LLMs) and conventional Retrieval-Augmented Generation (RAG) architectures. The preceding sections have dissected the nature of these errors, their root causes, and the available mitigation strategies. However, a long-term vision requires moving beyond merely containing current problems and projecting a future in which legal AI is not only more powerful and efficient, but fundamentally more \textbf{reliable, transparent, and aligned with the ethical principles and accountability requirements} inherent to the justice system. This future will not depend on a single disruptive breakthrough, but rather on the convergence of multiple lines of research and development focused on creating inherently more explainable models (\textbf{XAI}, or Explainable Artificial Intelligence), technically auditable systems, and, crucially, the adoption of a \textbf{responsible-by-design AI} paradigm (\textit{Responsible AI by Design}). This section explores these prospective trajectories, outlining the contours of a legal AI that can aspire to be a truly trustworthy collaborator for legal professionals and a fair instrument in the administration of justice.

\subsection{The Search for Explainability (XAI) in the Legal Context}

One of the greatest obstacles to trust and widespread adoption of LLMs in critical legal tasks is their "black box" nature. They generate responses, often complex and nuanced, but rarely offer an intelligible justification of \textit{how} they arrived at those conclusions or \textit{which} specific information (and with what weighting) they relied upon. In a domain such as law, where the ability to argue, justify, and trace reasoning back to authoritative sources is essential, this opacity is deeply problematic. Explainable Artificial Intelligence (XAI) emerges as a vital field of research to address this challenge.

Current explainability techniques for LLMs (e.g., attention analysis, feature importance, post-hoc textual justification generation) often provide only a superficial or approximate view of the model's internal decision-making process. These explanations themselves may be susceptible to "hallucination" or may not faithfully reflect the actual causal factors that led to a particular output (Rudin, 2019; Lipton, 2018).

\subsubsection{The Future of Legal XAI: From Grounding to Reasoned Interpretation}

The future of legal XAI does not lie so much in the development of \textit{post-hoc} explanations for black-box models, but rather in the evolution towards AI architectures that incorporate explainability natively and in a way that is meaningful for legal professionals. The true revolution will not be measured by a system's ability to pass an exam, but by its ability to justify its conclusions with transparency and accountability. This evolution can be conceptualized in three progressive levels of maturity:

    \begin{itemize}
    \item \textbf{Grounded Responses:} This is the basic level and the indispensable prerequisite, focused on traceability. The AI must be able to anchor each statement in an authoritative and verifiable source. The goal is to answer the question: \textit{"Where does this information come from?"} Without solid grounding, any result lacks the minimum reliability required for professional use.
    \item \textbf{Argued Responses:} The next level transcends mere citation of sources to articulate the reasoning. It is not enough to know \textit{which} source was used, but \textit{how} it was used to build the conclusion. The AI must be able to externalize the logical steps of its inference, demonstrating a coherent chain of reasoning. The goal is to answer the question: \textit{"How did you reach this conclusion based on the sources?"}
    \item \textbf{Reasoned Interpretation:} This is the most advanced level and the true horizon of legal AI. Here, the AI not only applies a rule, but is able to explain \textit{why} it chooses a specific interpretation over other plausible alternatives. This involves weighing nuances, recognizing ambiguities, and justifying its application of the rule to a specific factual context. The goal is to answer the question: \textit{\textbf{"Why is this the most appropriate interpretation or application in this case?"}}
    \end{itemize}

Reaching this third level remains a formidable challenge, since genuine interpretation requires principles, ethics, and an understanding of context that current models do not possess. However, the development of AI architectures that are intrinsically more interpretable is crucial to advancing along this path. This implies:

    \begin{itemize}
    \item \textbf{Models that externalize legal reasoning:} As discussed with reasoning models (Section 5.4), AI that can articulate its inference steps in a way that resembles human legal analysis (identifying relevant facts, applying rules, weighing factors, citing authorities for each step) will be inherently more explainable and verifiable, advancing towards the level of \textbf{argumentation}.
    \item \textbf{Visualization of source influence in RAG:} In RAG systems, improving the ability to trace which specific fragments of the retrieved context contributed and with what weight to each part of the generated response. Visualization tools that display these connections could dramatically increase interpretability and \textbf{grounding}.
    \item \textbf{Contrastive and counterfactual explanations:} Developing models capable of explaining not only why they reached a conclusion, but also why they did not reach other alternative conclusions, or how the conclusion would change if certain facts or premises were different. This closely aligns with the way lawyers analyze problems and is a key step towards \textbf{reasoned interpretation}.
    \item \textbf{Managing the tension between explainability and performance:} The complexity of legal reasoning and the multiplicity of factors that can influence a legal decision make complete explainability an extremely ambitious goal. The potential tension between explainability and model performance must be actively recognized and managed: the most accurate systems are often the most opaque. The future of legal XAI will lie in finding an optimal balance where the justification of the outcome is sufficiently robust for professional validation, without unacceptably sacrificing system effectiveness.
    \end{itemize}

\subsection{The Need for Technical and Governance Auditability}

Reliability and accountability in legal AI cannot rely solely on the good faith of developers or the diligence of individual users. Robust mechanisms are needed for \textbf{independent and continuous auditing} of these systems, both at the technical and governance levels.

\begin{enumerate}
    \item \textbf{Technical Audit of RAG Models and Systems:}

    \begin{itemize}

    \item \textit{Development of audit standards and metrics specific to legal AI:} Benchmarks and standardized metrics are required (such as those discussed in Section 3) that go beyond general accuracy and specifically assess the propensity for hallucinations, robustness to adversarial or ambiguous inputs, fairness with respect to different groups, and the quality of grounding in RAG systems.
    \item \textit{Automated and AI-assisted audit tools:} Develop tools that can assist human auditors in large-scale evaluation of models, for example, by automatically generating challenging test cases, identifying potential biases in training data or responses, or verifying the consistency of citations.
    \item \textit{Controlled access for auditing (}Sandboxing\textit{):} Regulators or independent certification entities could require access to the models (possibly in controlled environments or "sandboxes") to conduct thorough testing before their deployment in high-risk applications.
    \end{itemize}

    \item \textbf{Audit of Data and Process Governance:} Beyond the model itself, it is crucial to audit the processes and data governance practices of organizations that develop and implement legal AI.

    \begin{itemize}
    \item \textit{Traceability of training data:} Ensure that detailed records are kept regarding the sources, curation, and pre-processing of the data used to train the models, enabling investigation of potential biases or errors.
    \item \textit{Algorithmic and ethical impact assessment:} Require organizations to conduct rigorous impact assessments before deploying AI systems in sensitive legal contexts, proactively identifying and mitigating potential risks.
    \item \textit{Mechanisms for human oversight and accountability:} Audit the effectiveness of implemented human oversight mechanisms and ensure that clear channels exist for accountability and remediation in case of errors or harm caused by AI.
    \end{itemize}
\end{enumerate}

Auditability is not only a technical matter, but also a requirement of good governance and a prerequisite for building public trust in legal AI.

\subsection{Responsible AI by Design in the Legal Domain}

The most proactive and, ultimately, most effective approach to building reliable legal AI is to adopt a \textbf{Responsible Artificial Intelligence by Design} paradigm. This involves integrating ethical, fairness, transparency, robustness, and reliability considerations \textbf{from the earliest stages of the AI development lifecycle}, rather than treating these aspects as corrections or patches applied afterwards.

This 'Responsibility by Design' paradigm is not only an ethical aspiration or a good engineering practice, but is progressively becoming a regulatory expectation and, in some cases, an explicit legal obligation. The EU AI Act (the Regulation), through its detailed catalogue of requirements for high-risk AI systems — covering everything from risk management (\textbf{Article 9}) and training data governance (\textbf{Article 10}) to the need for effective human oversight (\textbf{Article 14}) and technical robustness (\textbf{Article 15}) — essentially codifies many of the fundamental principles of responsible AI. By requiring these considerations from the design and development phases, and throughout the entire system lifecycle, the EU-AIAct pushes legal AI creators to go beyond mere functionality to prioritize safety, reliability, and the protection of fundamental rights, where the prevention of harmful outcomes resulting from hallucinations becomes a central design objective.

\begin{enumerate}
    \item \textbf{Fundamental principles of responsible legal AI by design:}

These principles are not mere theoretical aspirations, but rather find direct resonance and institutional validation in emerging regulatory frameworks. These principles are not mere theoretical aspirations, but rather find direct resonance and institutional validation in emerging regulatory frameworks. A paradigmatic example is the aforementioned CTEAJE Policy in Spain. Principles established in this binding document, such as 'No Substitution' or the mandate of 'Universal Human Review' that it sets forth, are the practical materialization of the approach detailed below, demonstrating that responsible AI by design is shifting from being a best practice to a regulatory requirement.

    \begin{itemize}
    \item \textit{Human-centered:} Design AI systems that serve to augment and assist the legal professional, not to replace their critical judgment or ethical responsibility. The goal is human-AI collaboration, not the complete automation of complex tasks.
    \item \textit{Reliability and safety as a priority:} Factual accuracy, robustness against errors, and data security must be primary considerations in the design and optimization of models, even if this entails certain trade-offs in terms of fluency or generation speed.
    \item \textit{Fairness and non-discrimination (\textit{Fairness}):} Actively strive to identify and mitigate algorithmic biases that could lead to discriminatory or inequitable outcomes in the application of the law. This requires careful analysis of training data and of the model's differential impacts.
    \item \textit{Transparency and explainability (contextualized):} Design systems that are as transparent and explainable as possible within technical limitations, providing users with meaningful information about how they work and why they generate certain responses.
    \item \textit{Accountability and governance:} Establish clear governance and accountability structures for the development, deployment, and maintenance of legal AI systems.
    \end{itemize}
    
    \item \textbf{Practical methodologies for responsible legal AI by design:}

    \begin{itemize}
    \item \textit{Multidisciplinary development teams:} Involve legal professionals, ethicists, and domain experts from the outset of the design process, not just engineers and data scientists.
    \item \textit{Continuous risk assessment and adversarial testing:} Implement iterative cycles of risk assessment and stress testing (including specific tests to detect hallucinations and biases) throughout the entire development process.
    \item \textit{Feedback mechanisms and continuous improvement:} Design systems that can learn and improve based on feedback from expert users and monitoring of their real-world performance.
    \item \textit{Adoption of emerging ethical and technical standards:} Stay up to date with and adhere to ethical and technical standards, as well as best practices, that are being developed by the research community, professional bodies, and regulators.
    \end{itemize}
\end{enumerate}

Responsible AI by Design is not a final state, but an ongoing commitment to improvement and adaptation. It requires an organizational culture that prioritizes ethics and reliability, and a willingness to invest in the necessary resources to build systems that are truly trustworthy in the sensitive legal context.

\subsection{Advanced Human-AI Symbiosis: Collaboration and Augmented Judgment}

Looking further into the future, the most reliable and effective legal AI will likely not be the one that tries to completely replace the lawyer, but rather the one that achieves an \textbf{advanced and synergistic symbiosis with expert human intelligence}. In this model, AI is not merely a passive tool, but an active collaborator that enhances and refines the capabilities of the legal professional.

\begin{itemize}
    \item \textbf{AI as "Tireless Researcher and Preliminary Checker":} AI could take charge of exhaustive searches and preliminary analysis of large volumes of legal information, identifying patterns, retrieving relevant precedents, and highlighting possible issues or inconsistencies, but always presenting its findings to the lawyer for validation and strategic judgment.
    \item \textbf{AI as "Hypothesis Generator and Alternative Arguments":} Instead of providing a single "answer," AI could generate multiple lines of argument, interpretations, or possible solutions to a legal problem, each with its supporting rationale and potential weaknesses, allowing the lawyer to explore a broader spectrum of strategic options.
    \item \textbf{AI as "Contextual Translator and Communication Bridge":} One of the most significant barriers in legal practice is the information asymmetry between the lawyer and the client, often caused by the complexity of legal language. An advanced AI system, instead of being a tool for the exclusive use of the professional, can act as a contextual translator, generating summaries or explanations of legal documents and strategies in language adapted to the client's level of understanding. This approach, exemplified by the \textbf{"Jargon function" of the Justicio project} previously analyzed, not only improves transparency and trust in the client-lawyer relationship, but also empowers the client to make more informed decisions, humanizing access to justice.
    \item \textbf{AI as "Personalized Coach and Learning Assistant":} AI could provide detailed and personalized feedback on the work of trainee lawyers, helping them identify areas for improvement in their research, writing, and reasoning, always under the supervision of human mentors.
    \item \textbf{Intuitive Collaborative Interfaces:} The development of user interfaces that allow smooth, iterative, and truly collaborative interaction between the lawyer and the AI system will be crucial. The lawyer must be able to easily guide, question, and refine the work of the AI.
\end{itemize}    

This future of augmented collaboration requires not only advances in AI technology, but also an evolution in the training and skills of legal professionals, who will need to be competent both in law and in critical and effective interaction with these intelligent systems.

In conclusion, the path toward truly reliable and beneficial legal AI is complex and constantly evolving. While hallucinations and other risks are significant challenges, they are not insurmountable. Through sustained commitment to research in explainability, the development of robust auditability mechanisms, the adoption of responsibility-by-design principles, and, fundamentally, the recognition of the irreplaceable value of human oversight and judgment, it is possible to envision a future where AI becomes a powerful and trustworthy ally in the pursuit of more efficient, accessible, and equitable justice. The task is not trivial, but the potential rewards for the legal profession and society as a whole are immense.

\section{Navigating the Ethical and Regulatory Frontier: Implications of AI Hallucinations in the Global Legal Context, with Emphasis on Spanish and European Law}
\label{sec:etica_regulacion}

The emergence of Large Language Models (LLMs) and their inherent tendency to generate "hallucinations"—textual outputs that, although often fluent and convincing, lack factual accuracy or legal basis—is not merely a technical challenge. This phenomenon penetrates deeply into the fabric of the legal profession, questioning its ethical foundations and raising critical questions about the adequacy of existing regulatory frameworks at the global level. While much of the initial debate and early case law on sanctions for the misuse of AI in litigation has originated from the U.S. common law system (as exemplified by cases such as \textit{Mata v. Avianca, Inc.}), the ethical implications and the need for a regulatory response are universal, even though their manifestation and the proposed solutions must necessarily be adapted to the particularities of each legal system. This section delves into the complex landscape of deontological obligations and regulatory challenges posed by legal AI hallucinations, with particular attention to the realities and perspectives of Spanish law and the European regulatory framework, while also considering lessons learned from other jurisdictions.

\subsection{The Deontological Imperative in the Age of AI: Reaffirming the Fundamental Duties of Lawyers in Spain and Europe}

The traditional deontological duties of competence and diligence are crystallizing into a new ethical pillar for the digital era: the "Obligation of Technological Competence." As summarized in comprehensive analyses on the ethics of LLMs in the legal profession, this obligation not only requires understanding the benefits of technology, but also its risks and limitations, including a fundamental grasp of phenomena such as hallucinations (Shao et al., 2025). Fulfilling this duty entails the rigorous verification of AI-generated results and maintaining critical oversight, recognizing that the ultimate responsibility for the work rests unequivocally with the human professional.

The deontological codes governing the legal profession in Spain (such as the Spanish Legal Profession's Code of Ethics) and at the European level (such as the CCBE Code of Conduct for European Lawyers), like their American counterparts, establish a series of fundamental duties that, although not conceived with generative AI in mind, are directly applicable and acquire a new dimension in light of the risk of hallucinations.

\begin{enumerate}
    \item \textbf{Duty of professional competence:} This is perhaps the most immediately implicated duty. Professional competence requires not only substantive knowledge of the applicable law, but also the ability to properly use the tools and technologies employed in professional practice. In the era of AI, this translates into:
    \begin{itemize}
        \item \textit{AI literacy and understanding its limitations:} A competent lawyer in Spain or Europe cannot afford to ignore the basic functioning of LLMs, their probabilistic nature, and, crucially, their potential to generate hallucinations. This does not mean being an AI expert, but rather having sufficient functional understanding to critically evaluate their outputs and the associated risks (Yamane, 2020; Choi and Schwarcz, 2024).
        \item \textit{Duty of rigorous verification:} Competence requires the lawyer to independently verify the accuracy and relevance of any information or draft generated by an AI before using it in client advice or court proceedings. Blindly relying on the output of an LLM, especially in matters of high complexity or risk, could constitute a serious lack of competence. Emerging guidelines from European and Spanish bar associations will likely emphasize this point.
        \item \textit{Awareness of automation bias:} As in other contexts, legal professionals in Spain and Europe must be aware of "automation bias" and maintain a healthy professional skepticism, resisting the temptation to delegate critical judgment to the machine, no matter how efficient it may seem (Drabiak et al., 2023).

        This "automation bias" is not merely a tendency toward blind trust; it manifests through specific cognitive mechanisms that have been quantified. LLMs introduce framing biases in more than 20\% of cases, changing the emotional valence or emphasis of information without altering the underlying facts (Alessa et al., 2025). A legal professional interacting with these results may have their perception of a case subtly shaped before forming their own independent judgment. Therefore, professional skepticism is not just good practice, but an essential cognitive safeguard.

    \end{itemize}
    \item \textbf{Duty of diligence:} Closely linked to competence, the duty of diligence requires the lawyer to act with the necessary care and attention in defending the client's interests. In the context of AI and hallucinations:
    \begin{itemize}
        \item \textit{Verification as part of diligence:} The promise of AI efficiency cannot come at the expense of the quality and correctness of the work. A diligent lawyer must invest the necessary time to validate information generated by AI, ensuring that any use of it is based on verified and legally sound information. "Speed" cannot justify "negligent haste."
        \item \textit{Continuous updating:} Given the rapid evolution of AI technology, diligence may also entail a duty to remain reasonably informed about advances, newly identified risks (such as specific types of hallucinations), and best practices for using these tools.
    \end{itemize}

    The need for rigorous verification is not an abstract recommendation, but a requirement derived from empirical evidence. With documented hallucination rates ranging from 17\% to over 33\% in major commercial legal AI tools, blindly trusting their results constitutes a clear abdication of the duty of competence (Magesh et al., 2025). As the study concludes, lawyers face a difficult choice: "manually verify every proposition and citation produced by these tools (thus undermining the promised efficiency gains), or risk using these tools without complete information about their specific risks (thus neglecting their core duties of competence and supervision)."

    \item \textbf{Professional secrecy and data protection:} This is an area of particular sensitivity in the European and Spanish context, given the robust regulations on data protection (General Data Protection Regulation - GDPR).
    \begin{itemize}
        \item \textit{Confidentiality of client information:} Entering confidential information or clients' personal data into AI platforms, especially those whose servers are located outside the EU or whose data usage policies are not transparent or do not comply with the GDPR, represents a significant risk. Hallucinations are not the direct risk here, but the choice of tool and data management are crucial.
        \item \textit{GDPR compliance:} Any processing of personal data through AI must comply with the principles of the GDPR (lawfulness, fairness, transparency, purpose limitation, data minimization, accuracy, storage limitation, integrity and confidentiality, and proactive accountability). Providers and users of legal AI must be able to demonstrate such compliance.
    \end{itemize}
    \item \textbf{Loyalty and independence:} The lawyer owes loyalty to their client and must maintain their independent judgment. If an AI tool suggests a course of action based on hallucinated or biased information, the lawyer must exercise independent judgment to dismiss it if it does not serve the client's best interests or contravenes the law.
    \item \textbf{Duty of procedural loyalty and cooperation with the Administration of Justice:} In many civil law systems such as Spain, there is a strong emphasis on good faith and procedural loyalty. Presenting to a court arguments, evidence, or case law that one knows (or should know after diligent verification) to be false or fabricated by an AI would constitute a serious violation of these duties, with possible disciplinary and procedural consequences. The integrity of the judicial system depends on the reliability of the information presented by the parties.
\end{enumerate}
The impact of AI hallucinations on these ethical duties is undeniable and demands deep reflection by professional associations, disciplinary bodies, and each individual lawyer.

\subsection{Regulatory Challenges and Perspectives in Spain and the European Union}

The regulatory landscape for AI, and specifically for legal AI and its hallucination risks, is currently in a state of flux, with the European Union at the forefront through its proposed Artificial Intelligence Act (EU-AIAct).

\subsubsection{\textbf{The EU AI Act: A Hierarchical and Risk-Based Framework for Legal AI Governance}}

At the forefront of global efforts to establish a comprehensive regulatory framework for artificial intelligence is the European Union with its EU AI Act. This ambitious piece of legislation, with potentially global reach due to the well-known "Brussels effect," adopts a \textbf{layered and risk-based approach}, classifying AI systems into categories ranging from unacceptable risk (and therefore prohibited) to minimal risk, passing through limited risk categories and, crucially for many legal applications, \textbf{high risk}. It is this 'high risk' category that imposes the most significant obligations on developers, providers, and, in certain cases, users of AI systems (European Union, 2024; Hitaj et al., 2023; Petit \& De Cooman, 2020).

The determination of whether a specific legal AI tool falls within the 'high risk' category will depend on its intended purpose and the context of its use, as detailed in Annex III of the EU-AIAct. Areas explicitly mentioned as high risk that have a clear connection with the legal sector include AI systems used in the \textbf{administration of justice and democratic processes}, as well as those employed for creditworthiness assessment or selection in recruitment processes, which often involve profiling analyses with legal implications. It is reasonable to argue that AI tools that assist in judicial decision-making, in the assessment of evidence admissibility, in recidivism prediction, or even highly advanced legal research systems whose erroneous output could have a direct and significant impact on an individual's fundamental rights (e.g., in a criminal proceeding or in child custody determination) could be classified as high risk.

For these high-risk AI systems, the EU-AIAct establishes a comprehensive set of mandatory requirements that must be met before their introduction to the market and maintained throughout their lifecycle. Many of these requirements are directly relevant to the prevention and mitigation of hallucinations:

\begin{enumerate}
    \item \textbf{Robust risk management systems (Article 9):} The establishment, implementation, documentation, and maintenance of a continuous risk management process is required. This involves identifying foreseeable risks associated with the system (including those arising from hallucinations), estimating and evaluating such risks, and adopting appropriate measures for their control. The management of the risk of generating incorrect or fabricated legal information should be a central component of this system.
    \item \textbf{Data governance and quality (Article 10):} This article is particularly relevant for hallucinations originating from flawed training data. It requires that training, validation, and test datasets be 'relevant, representative, error-free, and complete.' Appropriate data governance practices must be applied, including an examination of potential biases and the adoption of measures to mitigate them. For legal AI, this implies the critical need to use updated, verified legal corpora that adequately reflect the diversity and complexity of the legal system.
    \item \textbf{Comprehensive technical documentation (Article 11 and Annex IV):} Providers must prepare detailed technical documentation describing, among other things, the system architecture, its capabilities and limitations, the algorithms used, the training data, and the testing and validation processes. This documentation is essential for conformity assessment and for supervisors and users to understand how the system works and what its reliability thresholds are.
    \item \textbf{Event logging mechanisms (\textit{Logging Capabilities}) (Article 12):} High-risk systems must be equipped with logging capabilities that ensure an adequate level of traceability of their operation. These 'logs' could be crucial for investigating \textit{a posteriori} the origin of a specific hallucination or for auditing the overall performance of the system.
    \item \textbf{Transparency and provision of information to users (Article 13):} Systems must be designed and developed so that users can interpret the system's output and use it appropriately. The instructions for use must include concise, complete, correct, and clear information about the provider's identity, the intended purpose of the system, its level of accuracy, robustness, and cybersecurity, as well as its known limitations and foreseeable risks—which includes, or should include, the propensity to generate hallucinations and the need for human verification.
    \item \textbf{Effective human oversight (Article 14):} This is a fundamental pillar. The EU-AIAct requires that high-risk systems be designed to allow for adequate human supervision. Measures may include the human supervisor's ability to fully understand the system's capabilities and limitations, to decide not to use the system in a particular situation, to override a decision made by the system, or to intervene in its operation. This supervision is the last barrier against the consequences of a hallucination not detected by the system itself.
    \item \textbf{Accuracy, robustness, and cybersecurity (Article 15):} Systems must achieve an appropriate level of accuracy, robustness, and cybersecurity throughout their lifecycle and be consistent in this regard. Hallucinations are a clear manifestation of a lack of factual accuracy and robustness. Systems are expected to be resilient to errors, failures, or inconsistencies, as well as to attempts at malicious use.
\end{enumerate}

Compliance with these requirements will be verified through \textbf{conformity assessments} before the high-risk AI system can be introduced into the EU market. In addition, \textbf{post-market monitoring} obligations are established for providers, who must monitor the performance of their systems and report any serious incidents or malfunctions. The \textbf{penalties} for non-compliance with the EU-AIAct are significant, reaching up to 35 million euros or 7\% of the total worldwide annual turnover of the previous financial year, which underscores the seriousness with which the EU addresses AI risks.

The impact of the EU-AIAct on mitigation strategies for hallucinations in legal AI, such as RAG, is profound. Many of the requirements of the Act—data quality, transparency regarding operation, robustness, human oversight—will drive developers to proactively and rigorously adopt many of the optimization strategies discussed in Section 5 of this essay, not as an optional improvement, but as a condition for market access. Although the EU-AIAct does not prescribe specific technical solutions, it does establish a framework of requirements that will foster innovation towards more reliable and responsible legal AI. Given the weight of the European market, it is highly likely that the EU-AIAct will have a "Brussels effect," influencing the standards for legal AI development globally.

\subsubsection{General Data Protection Regulation (GDPR)}

Although it is not specific to AI, the GDPR already imposes significant obligations that are relevant for the development and use of legal AI that processes personal data.
    \begin{itemize}
        \item \textit{Principles of accuracy and data minimization:} These principles are directly relevant to combating defective training data that can lead to hallucinations.
        \item \textit{Right not to be subject to automated decisions (Article 22):} If a legal AI system makes decisions that produce significant legal effects on a person (or similarly affect them), Article 22 of the GDPR could limit its use or require significant human intervention.
    \end{itemize}

\subsubsection{Civil and Professional Liability}
    \begin{itemize}
        \item \textit{General Liability Regime:} In Spain, the civil liability of lawyers for professional negligence is governed by the Civil Code and case law. If the improper use of AI (e.g., relying on hallucinated information without verification) causes harm to the client, the lawyer could be held liable.
        \item \textit{Proposed European Directive on AI Liability:} The European Commission has proposed a Directive to adapt non-contractual civil liability rules to AI. This proposal seeks to make it easier for victims of harm caused by AI to obtain compensation, for example, by easing the burden of proof in certain cases or establishing a presumption of causality if the AI provider has not fulfilled certain duties of care (potentially including those related to the prevention of hallucinations).
        \item \textit{Liability of the provider vs. professional user:} The allocation of liability between the developer/provider of the AI tool and the lawyer user will be a complex and likely contentious area. Providers' terms of service often include extensive disclaimers, but their validity could be challenged, especially if gross negligence is demonstrated or if there is an inherent defect in the product's design that makes it prone to generating legally harmful information (Calderon et al., 2022; Lantyer, 2024).
    \end{itemize}

\subsubsection{The Role of Bar Associations and Ethical Bodies}

In Spain, bar associations (and the General Council of Spanish Lawyers) play a crucial role in establishing ethical standards and overseeing their compliance.
    \begin{itemize}
        \item \textit{Issuance of specific guidelines and directives:} It is foreseeable and desirable that these bodies develop and publish specific guidelines on the ethical and competent use of generative AI by lawyers, explicitly addressing the risk of hallucinations and the duty of verification.
        \item \textit{Continuing education:} The provision of continuing education programs on AI, its capabilities, risks, and responsible use will be essential to ensure that professionals maintain the required level of competence.
        \item \textit{Disciplinary authority:} Bar associations could exercise their disciplinary authority in cases of manifestly negligent or irresponsible use of AI that results in harm to the client or to the administration of justice.
    \end{itemize}

\subsubsection{Need for Technical Standards and Benchmarks}

For any regulatory framework to be effective, independent technical standards and benchmarks will be needed to objectively assess the reliability, accuracy, and propensity for hallucinations of legal AI tools. Collaboration among legal professionals, technologists, and standardization bodies will be crucial in this regard.

The collaboration between legal professionals and technologists is already paving the way in this direction, as demonstrated by the use of professional bar exams as benchmarks to evaluate AI models in specific legal domains (Gupta et al., 2025). The adoption of such standardized tests as benchmarks could become a requirement for legal technology providers to demonstrate the reliability and competence of their systems in a specific jurisdiction, providing regulators and consumers with an objective basis for evaluation.

\subsection{Towards an Ethical and Responsible Integration of AI in Spanish and European Legal Practice}

The path towards an integration of AI in law that is both innovative and safe, especially in the Spanish and European context with its strong tradition of rights protection and regulatory rigor, requires a proactive and collaborative commitment from all stakeholders involved.

\begin{itemize}
    \item \textbf{For legal professionals:} Adopting a mindset of \textbf{informed skepticism and diligent verification} is paramount. AI should be seen as a powerful assistive tool, not as an infallible oracle. Ongoing training and digital literacy in AI will be essential competencies.
    \item \textbf{For law firms and legal organizations:} It is necessary to establish \textbf{clear internal policies} on the acceptable and responsible use of AI, including mandatory verification protocols, guidelines for managing confidential data, and training programs for their professionals. Investment in AI tools must be accompanied by investment in training for their safe use.
    \item \textbf{For legal technology providers:} There is a growing responsibility to develop tools that are \textbf{"safe by design" (}\textit{\textbf{safety by design}}\textbf{)}, incorporating mechanisms to minimize hallucinations and, crucially, being \textbf{radically transparent} about the capabilities, limitations, and known error rates of their products. Marketing claims must be realistic and backed by robust, independent empirical evidence.
    \item \textbf{For educational institutions (Law Schools):} It is essential to integrate teaching about AI, its legal and ethical implications, and the skills necessary for its critical use into curricula, preparing future generations of lawyers for a professional environment transformed by technology.
    \item \textbf{For regulators and professional bodies:} A proactive and thoughtful adaptation of regulatory and ethical frameworks is required. This may involve clarifying existing duties in light of AI, developing new specific guidelines, and fostering a culture of responsibility and accountability. The EU AI Act will be a key framework, but its effective implementation and oversight in the legal sector will require ongoing effort.
    \item \textbf{For the Judiciary:} Courts will also face the challenge of assessing AI-generated information presented by parties and, potentially, of using AI in their own work. Judicial training and the development of protocols for the use of AI in the judicial sphere will be necessary to maintain the integrity of the process.
\end{itemize}

In final conclusion, AI hallucinations are not merely a technical artifact, but a symptom of the fundamental tension between the probabilistic nature of LLMs and the demand for certainty and reliability in the legal system. Addressing this challenge in the Spanish and European context requires an \textbf{approach that combines technological innovation with a reaffirmation of the fundamental ethical principles of the legal profession}, an intelligent adaptation of regulatory frameworks, and, above all, an unwavering commitment to critical judgment and human oversight. AI can be a powerful tool for law, but only if its labyrinth is navigated with prudence, diligence, and a deep awareness of its current limitations.

Ultimately, a successful and responsible integration of AI into the legal ecosystem requires a cultural shift that transcends the search for simplistic technological solutions and embraces a paradigm of critical evaluation and informed diligence. This implies the internalization of three fundamental operational principles:

\begin{itemize}
    \item \textbf{Prioritize reliability over generation speed.} The real efficiency of an AI tool should not be measured solely by how quickly it produces a result. A system is truly efficient only if its outputs are reliable, thereby minimizing the time and effort required in the indispensable phase of human verification. Reliability, therefore, is the true multiplier of efficiency in legal workflows.
    \item \textbf{Promote the development of specialized (Domain-Specific) solutions.} The legal sector will benefit more from tools explicitly designed for its unique challenges than from adapting general-purpose models. Solutions must precisely address the complexities of legal reasoning, regulatory hierarchy, and demanding confidentiality requirements, demanding a development approach that prioritizes contextual depth over functional breadth.
    \item \textbf{Institute a culture of rigorous validation and critical feedback.} The adoption of new technologies should be guided by objective and empirical evaluation, rather than uncritical acceptance driven by novelty. The ecosystem (including professionals, developers, and academics) must demand and provide rigorous and honest feedback on the performance, limitations, and risks of these tools. Sustainable progress is grounded in constructive criticism and independent validation.
\end{itemize}

The adoption of these pragmatic principles is essential for the legal sector to navigate the complexity of the AI era with the prudence, diligence, and deep awareness that it demands.

\section{Conclusion: from Hallucination to Amplification — Principles for Reliable Legal AI}
\label{sec:conclusion}

The emergence of Large Language Models in the legal ecosystem presents a fundamental paradox: a technology with unprecedented potential to democratize and make access to justice more efficient, intrinsically burdened by a design flaw that undermines the pillar of law: \textbf{truthfulness}. However, this analysis has revealed that truthfulness in law is a dual concept, encompassing both factual fidelity —directly threatened by hallucination— and interpretive soundness, which remains the exclusive domain of human judgment. This report, therefore, has dissected the phenomenon of "hallucinations" not as a mere technical error, but as a systemic feature that demands a paradigm shift: moving from seeking an AI that "knows the truth" to building an AI that "amplifies the professional's capacity to interpret it."

More than a simple summary, this conclusion distills the findings of the analysis into a \textbf{set of guiding principles and a working framework} to guide professionals, developers, and regulators in navigating this complex new territory.

\subsection{\textbf{Fundamental Conclusions: a Structured Summary}}

The detailed analyses throughout this document converge on four key and interrelated conclusions:

\begin{itemize}
\item \textbf{Hallucination is not a "bug", it is a feature.} The main challenge lies in understanding that the tendency of general-purpose LLMs to "make things up" is not a flaw to be fixed, but a direct consequence of their architecture, which is designed for probabilistic fluency rather than factual fidelity. This compels us to abandon the idea of a "creative oracle" and to adopt a radically different paradigm.

\item \textbf{RAG is the path, not the destination.} Retrieval-Augmented Generation (RAG) is, without a doubt, the most important mitigation strategy and the foundation of modern legal AI. However, the empirical evidence is clear: a canonical implementation of RAG reduces, but \textbf{does not eliminate}, hallucinations. Treating it as a "plug-and-play" solution is a mistake; it should be considered as the starting point, a promising engine that requires holistic and rigorous optimization of each of its components to be truly reliable.

\item \textbf{Reliability is built, not installed: the imperative of holistic optimization.} The transition from a hallucinating AI to a reliable AI does not depend on a single breakthrough, but on a synergy of strategic improvements throughout the entire information lifecycle. This includes:

    \begin{itemize}
        \item \textbf{Strategic data curation:} A verified, up-to-date, and prioritized knowledge base is the indispensable foundation.
        \item \textbf{Sophisticated retrieval:} Going beyond simple semantic search to incorporate awareness of normative hierarchy (the Kelsenian principle), context, and legal logic.
        \item \textbf{Faithful generation and guided reasoning:} Using advanced prompt engineering and fine-tuning to instruct LLMs not only to answer, but to "think" in a structured, transparent, and strictly source-anchored manner.
        \item \textbf{Robust post-hoc verification:} Implementing layers of algorithmic and human safeguards as an indispensable last line of defense.
    \end{itemize}

\item \textbf{The human factor is irreducible and is strengthened.} Far from making legal professionals obsolete, the challenge of veracity redefines and strengthens their role. Expert, critical, and informed oversight is not an option, but a deontological, professional, and increasingly regulatory obligation (as demonstrated by the EU AI Act and the policies of the CTEAJE in Spain). The future is not automation, but \textbf{cognitive amplification}: AI becomes a tool to enhance human judgment, freeing it to focus on strategy, ethics, and empathy.
\end{itemize}

\subsection{Proposal of a Framework: Generative AI vs. Consultative AI}

To guide adoption and future development, we propose a clear conceptual framework that distinguishes two types of AI with radically different risk profiles and applications in law:

\begin{itemize}
    \item \textbf{General-purpose generative AI (the "creative oracle"):}

\begin{itemize}
    \item \textbf{Function:} Ideation, brainstorming, drafting non-critical documents, summarizing general texts.
    \item \textbf{Inherent risk:} \textbf{High} risk of factual hallucination, both extrinsic and intrinsic. Opacity in sources.
    \item \textbf{Usage principle:} Always use with extreme skepticism, as a creative assistant whose output should \textbf{never} be considered a source of truth. Requires complete human verification from scratch.
\end{itemize}

    \item \textbf{Specialized consultative AI (the "expert archivist"):}

\begin{itemize}
    \item \textbf{Function:} Legal research, \textit{due diligence}, document analysis, responding to queries based on a verified corpus.
    \item \textbf{Inherent risk:} \textbf{Low} risk of fabrication, but a \textbf{persistent} risk of subtle hallucinations (\textit{misgrounding}, synthesis errors).
    \item \textbf{Usage principle:} Designed for reliability. Must be transparent, citable, and auditable. Still requires critical verification by the professional, but focused on the correct interpretation and application of the provided sources, not on their existence.
\end{itemize}

\end{itemize}

Effective mitigation of hallucinations in the legal sector does not lie in incrementally improving the generative model, but in \textbf{deliberately adopting a consultative paradigm} where truthfulness and traceability are at the core of the design, not an added feature.

\subsection{\textbf{A look to the future: the call for responsible integration}}

The path towards truly reliable and beneficial legal AI has been mapped out, but it is not simple. It does not depend on finding a "magic switch" that eliminates errors, but rather on a collective and sustained commitment from the entire legal ecosystem.

\begin{itemize}
    \item \textbf{For developers}, the challenge is to build systems that are not only powerful, but also transparent, auditable, and designed with a deep humility regarding their limitations.
    \item \textbf{For legal professionals}, the challenge is to cultivate a culture of \textbf{informed skepticism}: to embrace technology as a tool for amplification, but never abdicate the ultimate responsibility of critical judgment.
    \item \textbf{For regulators and institutions}, the task is to continue developing regulatory frameworks that foster responsible innovation, establishing clear standards of reliability, and requiring human oversight as a non-negotiable pillar.
\end{itemize}

Ultimately, artificial intelligence is not an external force that "impacts" the law; it is a new material with which we are building the tools of the future. The quality, safety, and justice of those tools will depend on our ability to infuse them with the timeless principles of our profession: rigor, diligence, and an unwavering commitment to truth. The ultimate goal, and the great promise of this era, is not simply to automate processes, but, as has been argued throughout this report, to \textbf{humanize technology}, putting it at the service of a more accessible, efficient, and above all, reliable justice.

\section*{Acknowledgments} 

The work required for the study, analysis, and development of research as profound as that presented in this paper never depends solely on its author. This work would have been unthinkable without the prior efforts of each and every researcher who has contributed to society with their previous papers. To all of them, my most sincere gratitude for what their work has represented for me and for the entire scientific and technical community.

I would also like to thank Little John and each and every one of its members for both their support and the value of their reviews. Especially to Daniel Vecino for the countless joint work sessions and his “pixel review,” always as critical and thorough as it is kind.

It is also only fair to acknowledge the inspiration that Asier Gutiérrez-Fandiño has been for me. Without him, this publication would not have been possible, as he was a clear catalyst for the passion that the world of Artificial Intelligence awakens in me.

Finally, I wish to thank the selfless collaboration of each and every professional who had prior access to this paper. Their comments have been key in providing the necessary impetus that has always driven me to go a little further in every point of analysis.

To all of them, and to you as a reader of this work, thank you.

\bibliographystyle{unsrt}